%% file: archive_cmfd.tex
\documentclass[10pt,twocolumn,twoside]{IEEEtran}

\usepackage{cite}

\ifCLASSINFOpdf
  \usepackage[pdftex]{graphicx}
  \graphicspath{{images/}}
  \DeclareGraphicsExtensions{.pdf,.jpg,.jpeg,.png}
\else
  \usepackage[dvips]{graphicx}
\fi
\usepackage{array}
\usepackage[tight,footnotesize]{subfigure}
\usepackage{xspace}
\usepackage{color}
\usepackage{rotating}
\usepackage{multirow}

\newcommand{\momA}{\textsc{Blur}\xspace}
\newcommand{\blur}{\momA}
\newcommand{\momB}{\textsc{Hu}\xspace}
\newcommand{\hu}{\momB}
\newcommand{\momC}{\textsc{Zernike}\xspace}
\newcommand{\zernike}{\momC}
\newcommand{\pca}{\textsc{PCA}\xspace}
\newcommand{\svd}{\textsc{SVD}\xspace}
\newcommand{\kpca}{\textsc{KPCA}\xspace}
\newcommand{\intA}{\textsc{Luo}\xspace}
\newcommand{\luo}{\intA}
\newcommand{\intB}{\textsc{Bravo}\xspace}
\newcommand{\bravo}{\intB}
\newcommand{\intC}{\textsc{Lin}\xspace}
\newcommand{\lin}{\intC}
\newcommand{\intD}{\textsc{Circle}\xspace}
\newcommand{\cir}{\intD}
\newcommand{\dct}{\textsc{DCT}\xspace}
\newcommand{\dwt}{\textsc{DWT}\xspace}
\newcommand{\fmt}{\textsc{FMT}\xspace} 
\newcommand{\sift}{\textsc{Sift}\xspace} 
\newcommand{\surf}{\textsc{Surf}\xspace}

\newcommand{\secRef}[1]{Sec.~\ref{#1}\xspace}
\newcommand{\subsecRef}[1]{Sec.~\ref{#1}\xspace}
\newcommand{\Figurename}{Fig.\xspace}

\newcommand{\Tablename}{Tab.\xspace}

\newcommand{\figRef}[1]{\Figurename~\ref{#1}\xspace}
\newcommand{\tabRef}[1]{\Tablename~\ref{#1}\xspace}

\newcommand{\etal}[1]{#1~\emph{et~al.}}

\newcommand{\FNm}{F_\mathrm{N}}
\newcommand{\FN}{$\FNm$\xspace}
\newcommand{\FPm}{F_\mathrm{P}}
\newcommand{\FP}{$\FPm$\xspace}

\newcommand{\TPm}{T_\mathrm{P}}
\newcommand{\TP}{$\TPm$\xspace}
\newcommand{\Precision}{p}
\newcommand{\Recall}{r}
\newcommand{\FMm}{F_{\mathrm{1}}}
\newcommand{\FM}{$\FMm$\xspace}

\newcommand{\libJPEG}{\texttt{libjpeg}}
\newcommand{\openCV}{\texttt{OpenCV}}
\newcommand{\eg}{e.\,g.~}
\newcommand{\ie}{i.\,e.~}

\newcommand{\minEuclidean}{\tau_1}

\newcommand{\minCluster}{\tau_2}
\newcommand{\areaThreshold}{\tau_3}

\newcommand{\thisVariantSupp}[1]{}
\newcommand{\thisVariantDiff}[1]{#1}
\newcommand{\thisVariantArch}[2]{#2}

\begin{document}
\title{An Evaluation of Popular Copy-Move Forgery Detection Approaches}

\author{Vincent~Christlein,~\IEEEmembership{Student~Member,~IEEE,} %
        Christian~Riess,~\IEEEmembership{Student~Member,~IEEE,} %
        Johannes~Jordan,~\IEEEmembership{Student~Member,~IEEE,} %
        Corinna~Riess, %
        and~Elli~Angelopoulou,~\IEEEmembership{Member,~IEEE}
\thanks{V. Christlein, C. Riess, J. Jordan and E. Angelopoulou are with the Pattern Recognition Lab,
University of Erlangen-Nuremberg, Germany,
e-mail see http://www5.cs.fau.de/our-team. Contact: riess{@}i5.cs.fau.de}}

\markboth{IEEE Transactions on Information Forensics and Security}
{Christlein \MakeLowercase{\textit{et al.}}: An Evaluation of Popular Copy-Move Forgery Detection Approaches}

\maketitle

\begin{abstract}
A copy-move forgery is created by copying and pasting content within the same
image, and potentially post-processing it. In recent years, the detection
of copy-move forgeries has become one of the most actively researched topics in
blind image forensics. A considerable number of different algorithms have been
proposed focusing on different types of postprocessed copies. In this paper, we
aim to answer which copy-move forgery detection algorithms and processing steps
(\eg, matching, filtering, outlier detection, affine transformation estimation)
perform best in various postprocessing scenarios. The focus of our analysis is
to evaluate the
performance of previously proposed feature sets. We achieve this by casting
existing algorithms in a common pipeline. In this paper, we examined the $15$
most prominent feature sets. We analyzed the detection performance on a
per-image basis and on a per-pixel basis. We created a challenging real-world
copy-move dataset, and a software framework for systematic image manipulation.
Experiments show, that the keypoint-based features \sift and \surf, as well as
the block-based \dct, \dwt, \kpca, \pca and \zernike features perform very
well. These feature sets exhibit the best robustness against various noise
sources and downsampling, while reliably identifying the copied regions.
\end{abstract}


\IEEEpeerreviewmaketitle

\section{Introduction}

\begin{figure}[!t]
	\centering
	\ifCLASSOPTIONdraftcls
		\includegraphics[width=0.28\linewidth]{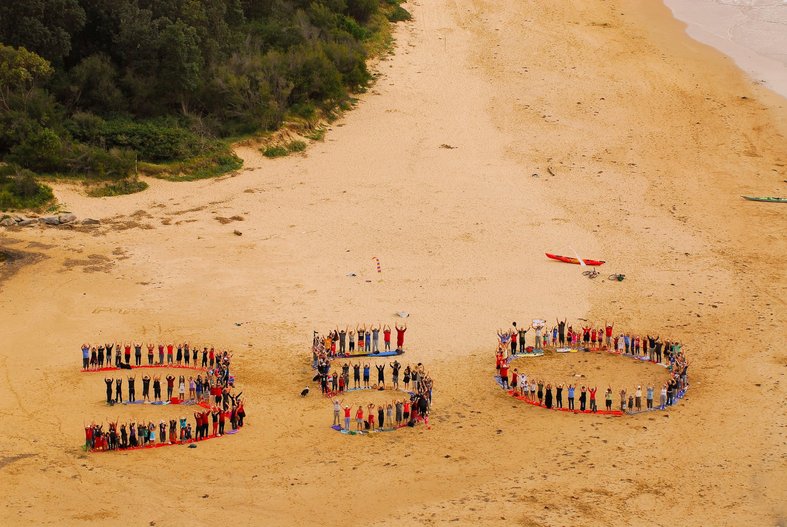}
		\includegraphics[width=0.28\linewidth]{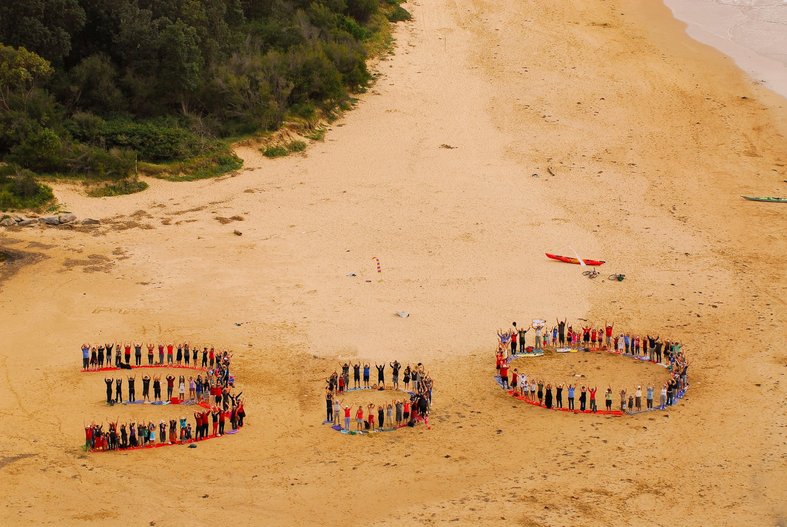}
	\else
		\includegraphics[width=0.49\linewidth]{db_examples/small_threehundred}
		\includegraphics[width=0.49\linewidth]{db_examples/small_threehundred_copy}
	\fi
	\caption{Example image of a typical copy-move forgery. Left: the original image. 
			Right: the tampered image. An example output of a CMFD detector for this image is shown
			in \figRef{fig:qualitative_results}.}
\label{fig:motivation}
\end{figure}

\IEEEPARstart{T}{he} goal of blind image forensics is to determine the authenticity
and origin of digital images without the support of an embedded security
scheme (see \eg\cite{Redi11:DIF,Farid09:ASO}). Within this field, copy-move
forgery detection (CMFD) is probably the most actively investigated subtopic. A
copy-move forgery denotes an image where
part of its content has been copied and pasted within the same image. Typical
motivations are either to hide an element in the image, or to emphasize
particular objects, \eg a crowd of demonstrators. A copy-move forgery is
straightforward to create. Additionally, both the source and the target regions stem from the
same image, thus properties like the color temperature, illumination conditions
and noise are expected to be well-matched between the tampered region and the image.
The fact that both the source and the target regions are contained in
the same image is directly exploited by many CMFD
algorithms,
\eg~\cite{Amerini11:SFM,Bashar10:EDR,Bayram05:IMD,Bayram09:AEA,Bravo11:EDR,%
Christlein10:ORI,Dybala07:DFC,Fridrich03:DOC,Huang08:DOC,Ju07:AAM,%
Kang08:ITR,Ke04:AEP,Li07:ASN,Langille06:AEM,Lin01:RSA,Lin09:FCM,Luo06:RDO,%
Mahdian07:DOC,Myrna07:DOR,Pan10:RDD,Popescu04:EDFDDIR,Ryu10:DCR,Shieh06:ASB,%
Wang09:FAR,Wang09:DOI,Zhang08:ANA}. We will briefly review existing methods in
\secRef{sec:pipeline}. 
An example of a typical copy-move forgery is shown in \figRef{fig:motivation}.

The goal of the paper is to examine which CMFD method to use under different
image attributes, like different image sizes and quantities of JPEG compression. Some limited prior work already exists on this topic,
\eg \cite{Christlein10:ORI,Bayram09:ASO}. However, to our knowledge, the
scope of such prior work is typically limited to the particular algorithm
under examination.
In this paper, we adopt a practitioner's view to copy-move forgery detection.
If we need to build a system to perform CMFD independent
of image attributes, which may be unknown, which method should we use? For that
purpose, we created a realistic database of
forgeries, accompanied by a software that generates copy-move forgeries of
varying complexity.
We defined a set of what we believe are ``common CMFD scenarios''
and did exhaustive testing over their parameters. A competitive
CMFD method should be able to cope with all these scenarios, as it is not known
beforehand how the forger applies the forgery. We implemented $15$
feature sets that have been proposed in the literature, and integrated them in
a joint pipeline with different pre- and postprocessing methods.
Results show, that keypoint-based methods have a clear advantage in terms of
computational complexity, while the most precise detection results can be
achieved using Zernike moments~\cite{Ryu10:DCR}.

The paper is organized as follows. In \secRef{sec:pipeline}, we present
existing CMFD algorithms within a unified workflow. In \secRef{sec:benchmark},
we introduce our benchmark data and the software framework for evaluation. In
\secRef{sec:error} we describe the employed error metrics. The experiments
are presented in
\secRef{sec:experiments}. We discuss our observations in
\secRef{sec:discussion}. \secRef{sec:conclusion} contains a brief summary and
closing remarks.

\section{Typical Workflow for Copy-Move Forgery Detection}
\label{sec:pipeline}

\begin{figure}[!t]
	\centering
\ifCLASSOPTIONdraftcls
		\includegraphics[width=.8\linewidth]{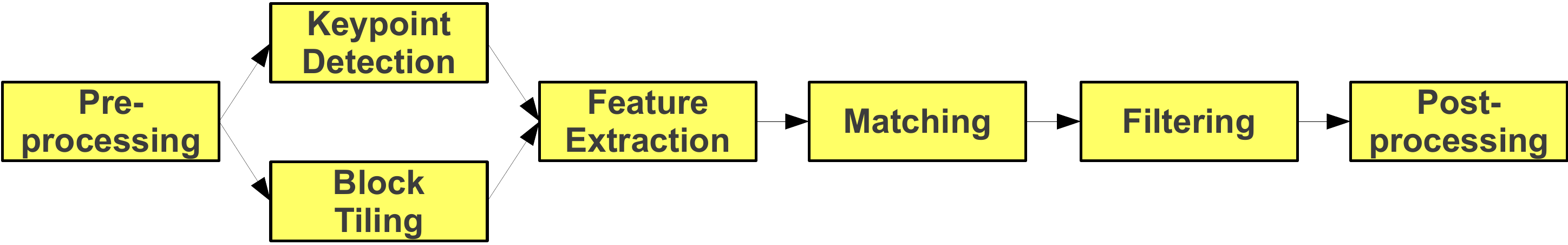}
\else
		\includegraphics[width=\linewidth]{cmfd_pipeline_alternate_new}
\fi
	\caption{Common processing pipeline for the detection of copy-move
	forgeries. The feature extraction differs for keypoint-based features (top)
	and block-based features (bottom). Except of method-specific threshold
	values, all remaining steps are the same.}
	\label{fig:pipeline}
\end{figure}

Although a large number of CMFD methods have been proposed, most techniques
follow a common
pipeline, as shown in \figRef{fig:pipeline}. Given an original image, there
exist two processing alternatives. CMFD methods are either
keypoint-based methods (\eg
\cite{Amerini11:SFM,Huang08:DOC,Pan10:RDD}) or block-based methods (\eg
\cite{Bashar10:EDR,Bayram05:IMD,Bayram09:AEA,Bravo11:EDR,%
Dybala07:DFC,Fridrich03:DOC,Ju07:AAM,%
Kang08:ITR,Ke04:AEP,Li07:ASN,Langille06:AEM,Lin01:RSA,Lin09:FCM,Luo06:RDO,%
Mahdian07:DOC,Myrna07:DOR,Popescu04:EDFDDIR,Ryu10:DCR,Shieh06:ASB,%
Wang09:FAR,Wang09:DOI,Zhang08:ANA}).
In both cases, preprocessing of the images is
possible. For instance, most methods operate on grayscale
images, and as such require that the color channels be first merged. For feature
extraction, block-based methods subdivide the image in rectangular regions. For
every such region, a feature vector is computed. Similar feature vectors are subsequently
matched. By contrast, keypoint-based methods compute 
their features only on image regions with high
entropy, without any image subdivision. Similar features within an image are afterwards matched. A forgery shall be
reported if regions of such matches cluster into larger areas. Both, keypoint-
and block-based methods include further filtering for removing spurious
matches. An optional postprocessing step of the detected regions may also be
performed, in order to group matches that jointly follow a transformation pattern.

Due to differences in the computational cost, as well as the detected detail,
we consider the difference between block- and keypoint-based methods
very important. Thus, we separately describe these two variants for feature
vector computation in the next two subsections, \subsecRef{subsec:block_algorithms} and
\subsecRef{subsec:keypoint_algorithms}, respectively.
Additional relevant details to the remaining steps in the pipeline are
presented below.  
\paragraph{Matching}
High similarity between two feature descriptors is interpreted as a cue for a
duplicated region. For block-based methods,
most authors propose the use of \emph{lexicographic sorting} in identifying
similar feature vectors (see \eg \cite{Bashar10:EDR,%
Bayram05:IMD,Bayram09:AEA,Bravo11:EDR,Dybala07:DFC,Fridrich03:DOC,Ju07:AAM,%
Kang08:ITR,Ke04:AEP,Li07:ASN,Langille06:AEM,Lin01:RSA,Lin09:FCM,Luo06:RDO,%
Myrna07:DOR,Popescu04:EDFDDIR,Ryu10:DCR,Shieh06:ASB,%
Wang09:FAR,Wang09:DOI,Zhang08:ANA}).
In lexicographic sorting a matrix of feature vectors is built so that
every feature vector becomes a row in the matrix. This matrix is then row-wise
sorted. Thus, the most similar features appear in consecutive rows.

Other authors use the Best-Bin-First search method derived from the kd-tree 
algorithm~\cite{Beis97:SIU} to get approximate nearest neighbors~\cite{Huang08:DOC,%
Pan10:RDD,Mahdian07:DOC}. In particular, keypoint-based methods often use this
approach.
Matching with a kd-tree yields a relatively efficient nearest
neighbor search. Typically, the Euclidean distance is used as a similarity
measure. In prior work~\cite{Christlein10:SFD}, it has been shown that the use of kd-tree matching
leads, in general, to better results than lexicographic sorting, but the memory
requirements are significantly higher. Note, however,
that the dimensions of some feature sets are ordered by importance (for
instance, in \cite{Bashar07:WBM,Lin09:FCM,Kang08:ITR}). For these features, the
performance gain over lexicographic sorting is minimal.
In our setup we matched feature vectors using the approximate
nearest neighbor method of \etal{Muja}~\cite{Muja09:FAN}. It uses multiple 
randomized kd-trees for a fast neighbor search.

\paragraph{Filtering}
Filtering schemes have been proposed in order to reduce the probability of false
matches. For instance, a common noise suppression measure involves the removal of
matches between spatially close regions. Neighboring pixels often have similar
intensities, which can lead to false forgery detection.
Different distance criteria were also proposed in order to filter out weak
matches. For example, several authors proposed the Euclidean distance between
matched feature vectors~\cite{Mahdian07:DOC,Wang09:DOI,Ryu10:DCR}.
In contrast, Bravo-Solorio and Nandi~\cite{Bravo11:EDR} proposed the correlation coefficient between two
feature vectors as a similarity criterion.

\paragraph{Postprocessing}
The goal of this last step is to only preserve matches that exhibit a
common behavior. Consider a set of matches that belongs to a copied region.
These matches are expected to be spatially close to each other in both the
source and the target blocks (or keypoints).
Furthermore, matches that originate from the same copy-move action should
exhibit similar amounts of translation, scaling and rotation.

The most widely used postprocessing
variant handles outliers by imposing a minimum number of similar shift vectors
between matches. A shift vector contains the translation (in image coordinates)
between two matched feature vectors. Consider, for example, a number of blocks which are
simple copies,
without rotation or scaling. Then, the histogram of shift vectors exhibits a
peak at the translation parameters of the copy operation.

Mahdian and Saic~\cite{Mahdian07:DOC} consider a pair of matched feature vectors as forged if: a)  they
are sufficiently similar, \ie their Euclidean distance is below a threshold,
and b) the neighborhood around their spatial locations contains
similar features.
Other authors use morphological operations to connect
matched pairs and remove outliers~\cite{Zhang08:ANA,Langille06:AEM,Pan10:RDD,Wang09:FAR}. 
An area threshold can also be applied, so that the detected region has
at least a minimum number of points~\cite{Luo06:RDO,Pan10:RDD,Wang09:FAR,Wang09:DOI}.
To handle rotation and scaling, Pan and Lyu~\cite{Pan10:RDD} proposed to use
RANSAC.  For a certain number of iterations, a random subset
of the matches is selected, and the transformations of the matches
are computed. The transformation which is
satisfied by most matches (\ie which yields most inliers) is chosen.
Recently, \etal{Amerini}~\cite{Amerini11:SFM} proposed a scheme which first builds clusters from the
locations of detected features and then uses RANSAC to estimate the geometric
transformation between the original area and its copy-moved
version. 
Alternatively, the \emph{Same Affine
Transformation Selection (SATS)}~\cite{Christlein10:ORI} groups locations 
of feature vectors to clusters. In principle, it performs region growing on
areas that can be mapped onto each other by an affine transformation.
\thisVariantDiff{More precisely, if the features computed on three spatially close blocks
match to three feature vectors whose blocks are also spatially close, then
these groups of blocks might be part of a copied region. The affine
transformation to map both groups of blocks onto each other is computed.
Further blocks from the neighborhood are added if the matched pairs satisfy the
same affine transformation. For details, see~\cite{Christlein10:ORI}.}
Although not explicitly reported in this paper, we evaluated the impact of
each of these methods. Ultimately, we adopted two strategies. For block-based approaches, we used a
threshold $\tau_2$ based on the SATS-connected area to filter out spurious
detections, as SATS provided the most reliable results in early experiments.
To obtain pixel-wise results for keypoint-based
methods, we combined the methods of \etal{Amerini} and Pan and Lyu. We built
the clusters described by \etal{Amerini}, but avoided the search for the
reportedly hard to calibrate \emph{inconsistency threshold}~\cite{Amerini11:SFM}.
Instead, clusters 
stop merging when the distance to their nearest neighbors are too high,
then the affine transformation between
clusters is computed using 
RANSAC and afterwards refined by applying
the gold standard algorithm for affine homography
matrices~\cite[pp. 130]{Hartley03:MVG}. For each such estimated transform, we
computed the correlation map according to Pan and Lyu~\cite{Pan10:RDD}. For
full details on our implementation, and a more elaborate discussion on our
choice of postprocessing, please refer to the supplemental material\thisVariantSupp{{} in~\cite{SuppMat}}.

To summarize, we present the joint CMFD algorithm below. 
Given an $M\times N$ image, the detected regions are computed as follows:
\begin{enumerate}
\item Convert the image to grayscale when applicable (exceptions: the features of
\etal{Bravo-Solorio}~\cite{Luo06:RDO} and \etal{Luo}~\cite{Bravo11:EDR} require
all color channels for the feature calculation)
\item For block-based methods:
	\begin{enumerate}
	\item Tile the image in $B_i$ overlapping blocks of size $b \times b$, where $0
	\leq i < \left((M-b+1) \cdot (N-b+1)\right)$
	\item Compute a feature vector $\vec{f}_i$ for every block $B_i$.
	\end{enumerate}

	For keypoint-based methods:
	\begin{enumerate}
	\item Scan the image for keypoints (\ie high entropy landmarks).
	\item Compute for every keypoint a feature vector $\vec{f}_i$. These two
	steps are typically integrated in a keypoint extraction algorithm like SIFT
	or SURF.
	\end{enumerate}
\item Match every feature vector by searching its approximate nearest neighbor. Let
$F_{ij}$ be a matched pair consisting of features $\vec{f}_i$ and $\vec{f}_j$,
where $i$, $j$ denote feature indices, and $i \neq j$. Let $c(\vec{f}_i)$
denote the image coordinates of the block or keypoint from which $\vec{f}_i$
was extracted. Then, $\vec{v}_{ij}$ denotes the translational difference
(``shift vector'') between positions $c(\vec{f}_i)$ and $c(\vec{f}_j)$. 
\item Remove pairs $F_{ij}$
where $\|\vec{v}_{ij}\|_2 < \minEuclidean$, where $\|\cdot\|_2$ denotes the
Euclidean norm. 
\item \thisVariantDiff{Clustering of the remaining matches that adhere to a joint pattern.}
	\begin{itemize}
	\item For block-based methods: Let $H(A)$ be the number of pairs satisfying the same affine transformation
		$A$. Remove all matched pairs where $H(A) < \minCluster$.
	\item \thisVariantDiff{For keypoint-based methods: Apply homography-based clustering as described in the paragraph above.}
	\end{itemize}
\item If an image contains connected regions of more than $\areaThreshold$
connected pixels, it is denoted as tampered.
\end{enumerate}
Please note that it is quite common to set the thresholds $\tau_2$ and $\tau_3$
to the same value.

\subsection{Block-based Algorithms}\label{subsec:block_algorithms}

We investigated $13$ block-based features, which we considered
representative of the entire field. They can be grouped in four categories:
moment-based, dimensionality reduction-based, intensity-based, and frequency
domain-based features
(see \tabRef{tab:methods}).

\begin{table}[tb]
\centering
	\ifCLASSOPTIONdraftcls
		\renewcommand{\arraystretch}{0.7}
	\fi
\caption{Grouping of evaluated feature sets for copy-move forgery detection.}
\label{tab:methods}
\begin{tabular}{|c|l|l|}
\hline
Group & Methods & Feature-length \footnotemark \\
\hline
\multirow{3}{*}{Moments} & \momA~\cite{Mahdian07:DOC} & $24$ \\
& \momB~\cite{Wang09:FAR} & $5$ \\
& \momC~\cite{Ryu10:DCR}  & $12$ \\
\hline
\multirow{3}{2cm}{\centering Dimensionality reduction} & \pca~\cite{Popescu04:EDFDDIR} & --\\
& \svd~\cite{Kang08:ITR} & --\\
& \kpca~\cite{Bashar10:EDR} & $192$\\
\hline
\multirow{4}{*}{Intensity} & \intA~\cite{Luo06:RDO} & $7$ \\
& \intB~\cite{Bravo11:EDR} & $4$ \\
& \intC~\cite{Lin09:FCM} & $9$ \\
& \intD~\cite{Wang09:DOI} & $8$ \\
\hline
\multirow{3}{*}{Frequency} & \dct~\cite{Fridrich03:DOC} & $256$ \\
& \dwt~\cite{Bashar10:EDR} & $256$ \\
& \fmt~\cite{Bayram09:AEA} & $45$ \\
\hline
\multirow{2}{*}{Keypoint} & \sift~\cite{Huang08:DOC},\cite{Pan10:RDD},\cite{Amerini11:SFM} & $128$ \\
& \surf~\cite{Shivakumar11:DOR},\cite{Bo10:ICP} & $64$ \\
\hline
\end{tabular}
\end{table}

\footnotetext{Some feature-sizes depend on the block size, which we fixed to
	$16\times16$. Also note that the feature-sizes of \pca and \svd depend
		on the image or block content, respectively.}

\noindent
\textbf{Moment-based:}
We evaluated $3$ distinct approaches within this class. Mahdian and
Saic~\cite{Mahdian07:DOC} proposed the use of $24$ blur-invariant moments as
features (\blur).  \etal{Wang}~\cite{Wang09:FAR} used the first four Hu moments
(\hu) as features.  Finally, \etal{Ryu}~\cite{Ryu10:DCR} recently proposed the
use of Zernike moments (\zernike). 

\noindent
\textbf{Dimensionality reduction-based:} In \cite{Popescu04:EDFDDIR}, the
feature matching space was reduced via principal component analysis
(\pca). \etal{Bashar}~\cite{Bashar10:EDR} proposed the Kernel-PCA~(\kpca) variant of PCA.
\etal{Kang}~\cite{Kang08:ITR} computed the singular values of a
reduced-rank approximation (\svd). A fourth approach using a combination of the discrete
wavelet transform and Singular Value Decomposition
\cite{Li07:ASN} did not yield reliable results in our setup and was, thus,
excluded from the evaluation.

\noindent
\textbf{Intensity-based:} The first three features used in~\cite{Luo06:RDO}
and~\cite{Bravo11:EDR} are the average red, green and blue
components. Additionally, \etal{Luo}~\cite{Luo06:RDO} used directional
information of blocks (\intA) while
\etal{Bravo-Solorio}~\cite{Bravo11:EDR} consider the entropy of a block as a
discriminating feature (\intB). \etal{Lin}~\cite{Lin09:FCM} (\lin)
computed the average grayscale intensities of a block and its
sub-blocks.
\etal{Wang}~\cite{Wang09:DOI} used the mean intensities of circles with
different radii around the block center (\cir). 

\noindent
\textbf{Frequency-based:} \etal{Fridrich}~\cite{Fridrich03:DOC} proposed the use of $256$
coefficients of the discrete cosine transform as features
(\dct). The coefficients
of a discrete wavelet transform (\dwt) using Haar-Wavelets were proposed as
features by \etal{Bashar}~\cite{Bashar10:EDR}.
\etal{Bayram}~\cite{Bayram09:AEA} recommended the use of the Fourier-Mellin
Transform (\fmt) for generating feature vectors.

\subsection{Keypoint-based Algorithms}\label{subsec:keypoint_algorithms}

Unlike block-based algorithms,
keypoint-based methods rely on the identification and selection of high-entropy
image regions (\ie the ``keypoints''). A feature
vector is then extracted per keypoint. Consequently, fewer feature vectors are estimated,
resulting in reduced computational complexity of
feature matching and post-processing. The lower number of
feature vectors dictates that postprocessing thresholds are also to be lower than
that of 
block-based methods. A drawback of keypoint methods is that copied regions are
often only sparsely covered by matched keypoints. If the
copied regions exhibit little structure, it may happen that the region is
completely missed. We examined two different versions of 
keypoint-based feature vectors. One uses the SIFT features while the other uses
the SURF features (see~\cite{Shivakumar11:DOR}). They are
denoted as \sift and \surf, respectively. The feature extraction is implemented
in standard libraries. However, particular differences of keypoint-based
algorithms lie in the postprocessing of the matched features, as stated in the
previous section (confer
also~\cite{Huang08:DOC,Pan10:RDD,Amerini11:SFM,Shivakumar11:DOR,Bo10:ICP}).

\section{Benchmark Data}\label{sec:benchmark}

Since image forensics is a relatively new field, there exist
only a few benchmarking datasets for algorithm
evaluation. \etal{Ng}~\cite{Ng04:ADS} developed a dataset that consists of
automatically spliced images. In their dataset, portions of an
image are quasi-randomly copied and inserted in a different image, without
post-processing. Thus, the seams of the spliced regions often exhibit
sharp edges. Furthermore, these splices are frequently not semantically meaningful. The CASIA
dataset~\cite{CASIA} addresses these issues.  However, the majority of the images are
$384\times
256$ pixels, and thus unrealistically small. \etal{Battiato}~\cite{Battiato10:DFE} presented a tampered image
database which focuses on the evaluation of detection methods based on JPEG
artifacts. These images are also of low resolution.
Almost all are $384\times 512$ pixels.
Other related databases have slightly different goals. For instance, the
Dresden Image Database~\cite{Gloe10:DID} is targeting methods for camera
identification.
Similarly, \etal{Goljan}~\cite{Goljan09:LST} presented a large-scale database for the
identification of sensor fingerprints.

To our knowledge, none of the existing databases is suited for an in-depth
evaluation of copy-move forgery techniques. Concurrent to our work on this
paper, \etal{Amerini} published two ground truth databases for CMFD algorithms,
called MICC F220 and MICC F2000~\cite{Amerini11:SFM}. They consist of $220$ and
$2000$ images, respectively. In each of these datasets, half of the images are
tampered.
The image size is $2048\times1536$ pixels.
The type of processing on the copy-move forgeries is limited to
rotation and scaling. Additionally, the source files are not available. Thus,
adding noise or other artifacts to the copied region is not feasible.
To
address these issues,
we built a new benchmark database \thisVariantDiff{aiming at the analysis of consumer photographs}.
Our images were created in a two-step process.
\begin{enumerate}
\item We selected $48$ source images and manually prepared per image
semantically meaningful regions that should be copied. We
call these regions \emph{snippets}. Three persons of varying artistic skill
manually created the snippets. When creating the snippets, we asked the artists
to vary the snippets in their size. Additionally, the snippet content should be
either \emph{smooth} (\eg, sky), \emph{rough} (\eg, rocks) or \emph{structured}
(typically man-made buildings). These groups can be used as categories for CMFD
images. More details on the categorization are provided in the supplemental
material\thisVariantSupp{{} in~\cite{SuppMat}}. In total $87$ snippets were constructed.
\item To create copy-move forgeries in a
controlled setup (\ie as the result of a parameterized algorithm), we developed
a software framework to generate copy-move forgeries using the snippets. Common
noise sources, such as JPEG artifacts, noise,
additional scaling or rotation, are automatically included. Additionally, a pixelwise ground truth map is
computed as part of this process.
\end{enumerate}
When creating the forgeries, we aimed to create realistic
copy-move forgeries in high-resolution pictures from consumer cameras.
Here, ``realistic'' refers to the number of copied pixels, the
treatment of the boundary pixels and the content of the snippet. The average size of
an image is about
$3000\times{}2300$ pixels.
In total, around $10\%$ of the pixels belong to tampered image
regions.

In the ground truth images, pixels are identified as either background or
copy-move pixels, or as belonging to a third class of otherwise processed
pixels (\eg painted, smeared, etc.). The latter group of pixels includes border
pixels of snippets where the copied data is semi-transparent, in order to form
a smoother transition between the neighboring original pixels and the copied
one. This is a typical process in real copy-move forgeries.
\ifCLASSOPTIONdraftcls
Pixels that belong to this third class are not fully opaque.
\figRef{fig:beachwood} shows a typical ground truth setup. The source image is
shown on the left. The snippet is the second image, the tampered image is on
the middle right. The ground truth map is shown on the right.
\else
Pixels that belong to this third class are not fully opaque.
\figRef{fig:beachwood} shows a typical ground truth setup. The source image is
shown on the top left, the tampered image on the top right. The snippet is
shown on the bottom left, the ground truth map is shown on the bottom right. 
\fi
Here, white denotes copied pixels, black denotes background. The boundary
pixels of the copied regions are marked gray for visualization purposes.
Please note that accurate ground truth is difficult to define in such mixed
data regions.
\ifCLASSOPTIONdraftcls
	\begin{figure}[!t]
		\centering
				\includegraphics[width=0.22\linewidth]{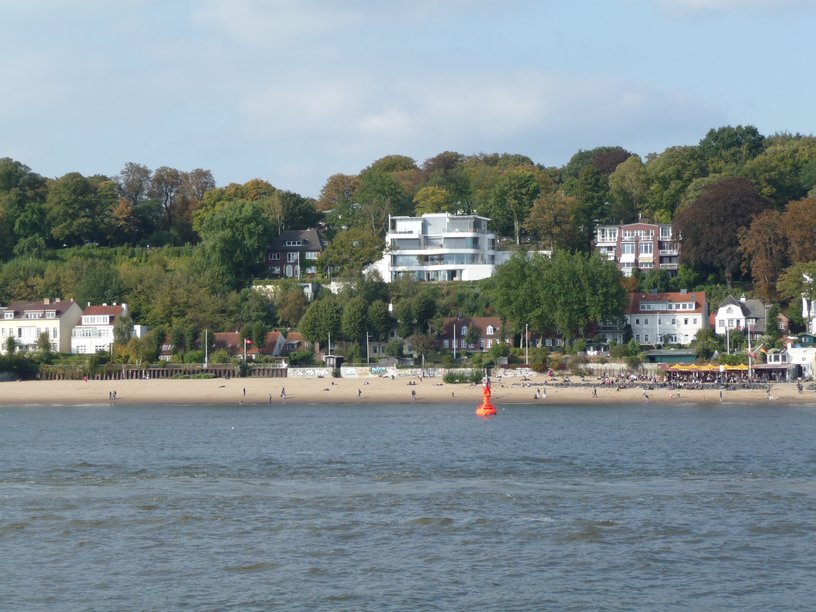}
				\includegraphics[width=0.22\linewidth]{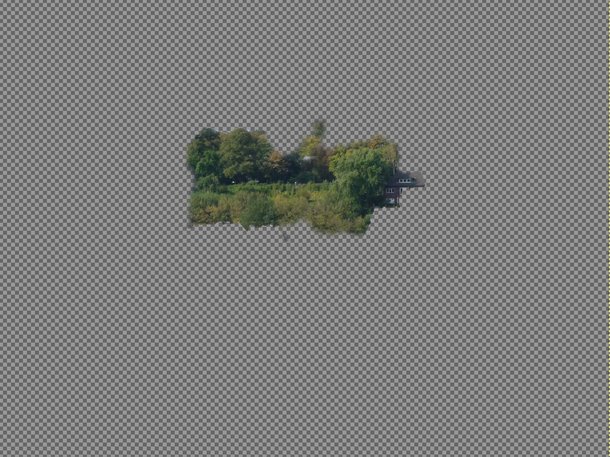}
				\includegraphics[width=0.22\linewidth]{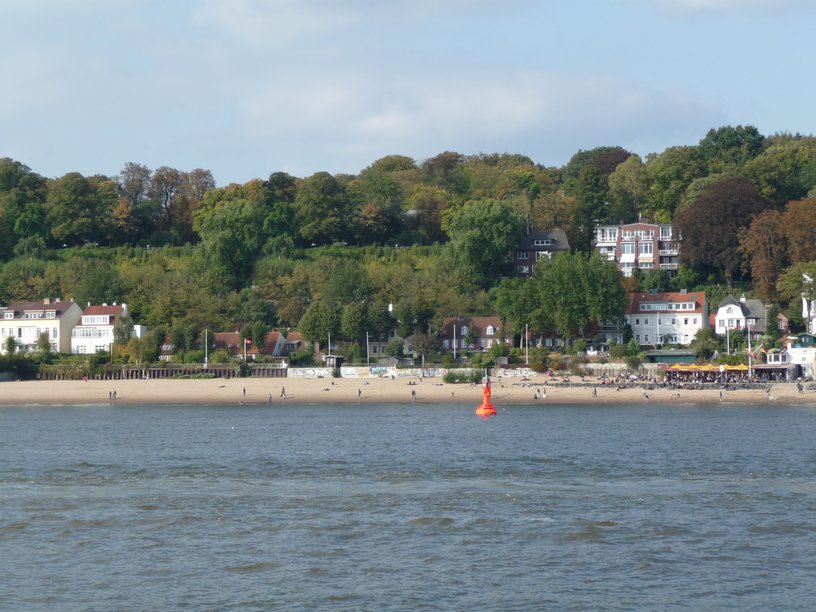}
				\includegraphics[width=0.22\linewidth]{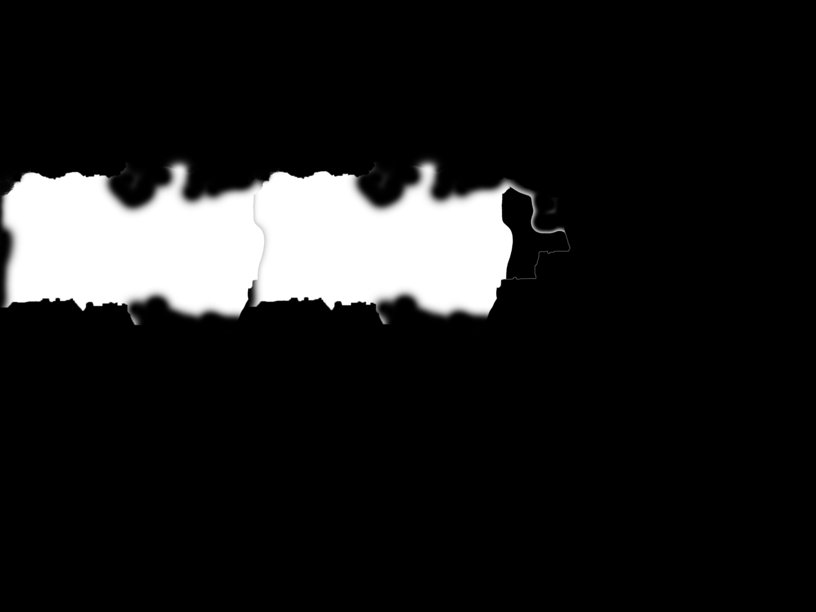}
		\caption{Ground truth example. From left to right: the image \emph{beachwood} is forged with
				a green patch (middle left) to conceal a building (middle right). A
				ground truth map (right) is generated where copy-moved pixels
				are white, unaltered pixels are black and boundary pixels are
				gray.}
		\label{fig:beachwood}
	\end{figure}
\else
	\begin{figure}[!t]
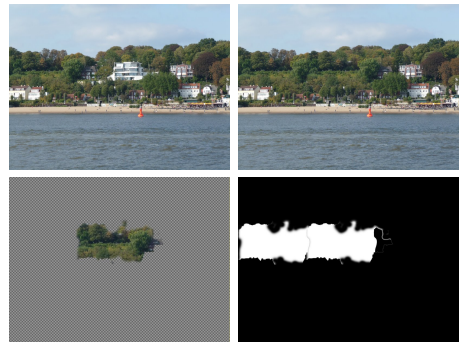

		\centering
				\includegraphics[width=0.33\linewidth]{beachwood_o}
				\includegraphics[width=0.33\linewidth]{beachwood}
			\\ \vspace{0.1cm}
				\includegraphics[width=0.33\linewidth]{beachwood_patch_fullsize}
				\includegraphics[width=0.33\linewidth]{beachwood_map}
		\caption{The image \emph{beachwood} (upper left) is forged with
				a green patch (bottom left) to conceal a building (upper right). A
				ground truth map (bottom right) is generated where copy-moved
				pixels are white, unaltered pixels are black and boundary pixels
				are gray.}
		\label{fig:beachwood}
	\end{figure}

\fi

A software that can be downloaded together with the images allows the
flexible combination of the original image and the tampered regions.
Whenever a tampered image is created, the corresponding ground truth is
automatically generated.
\ifCLASSOPTIONdraftcls
	\begin{figure}[!t]
		\centering
			\includegraphics[width=0.4\linewidth]{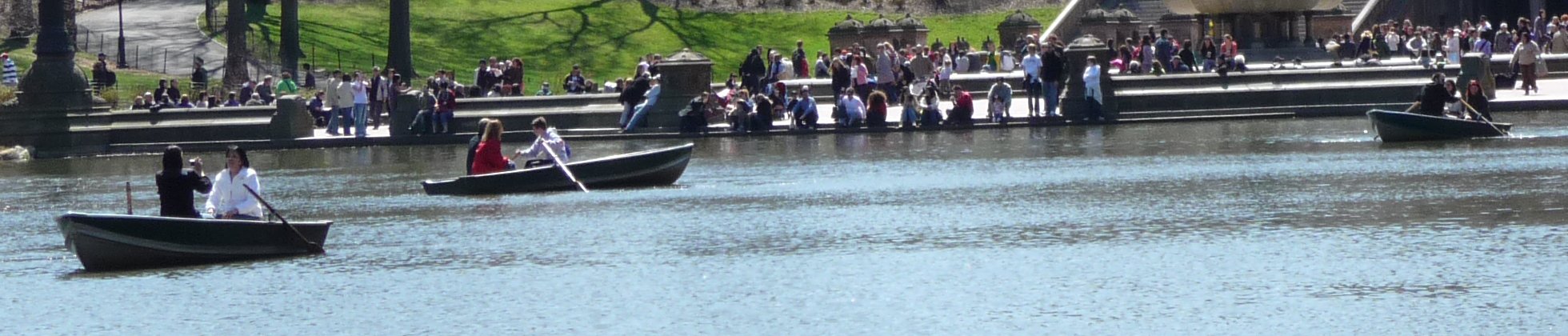}
			\includegraphics[width=0.4\linewidth]{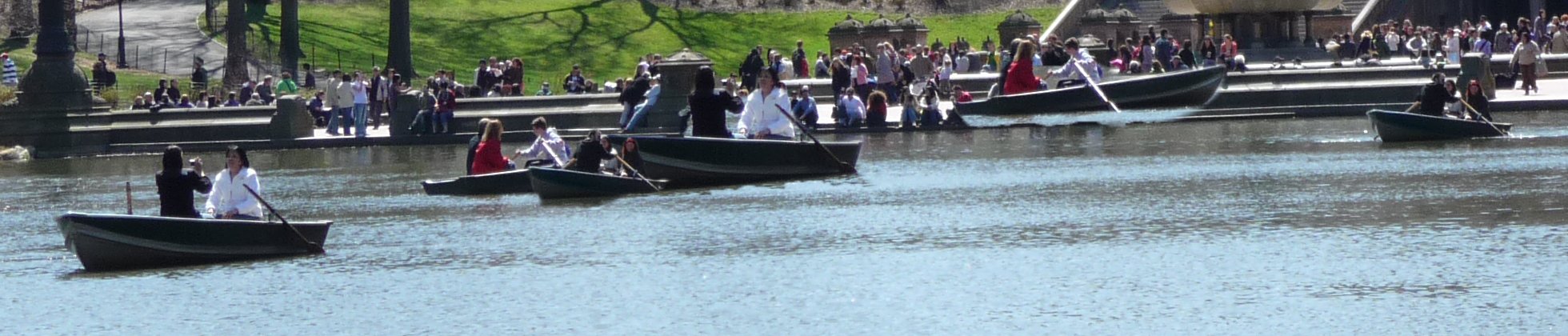}

			\vspace{1mm}
			\includegraphics[width=0.4\linewidth]{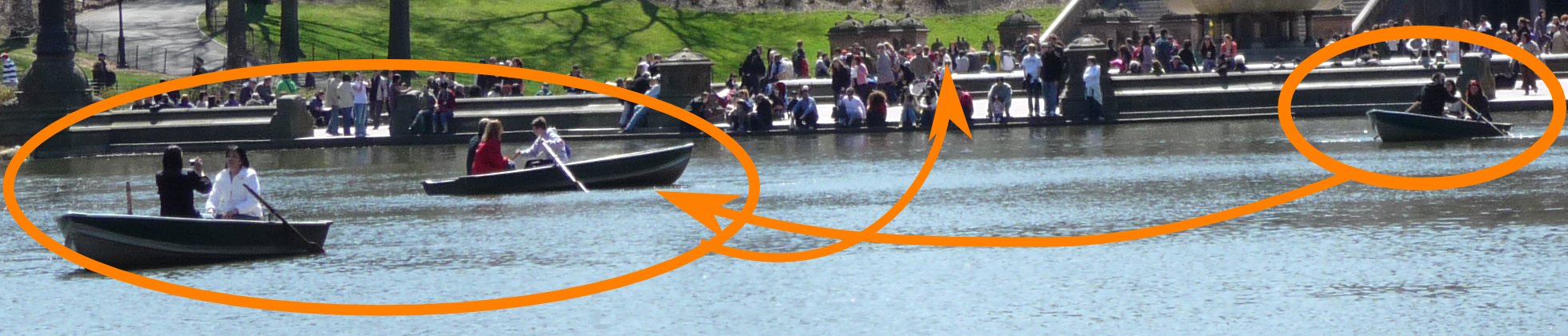}
			\includegraphics[width=0.4\linewidth]{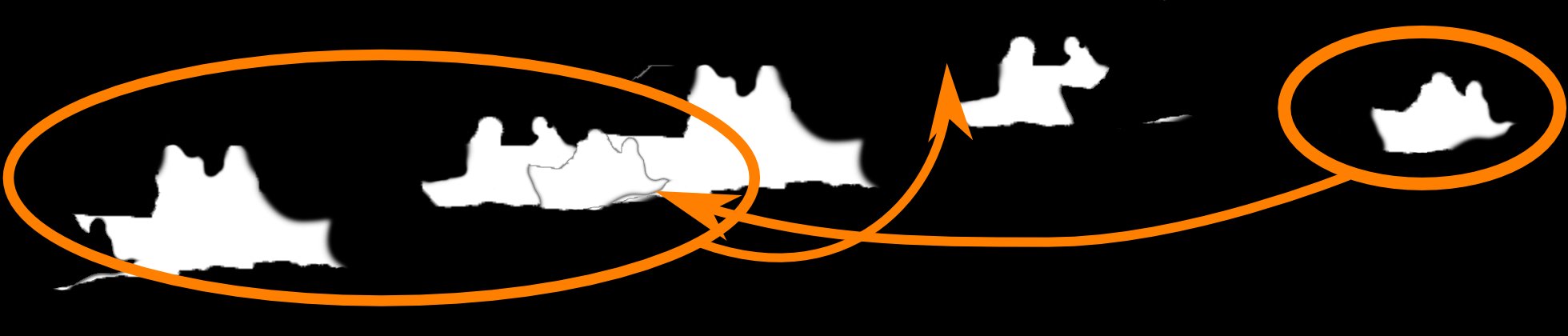}
	\caption{Artificial example of several types of occlusion in the ground truth
	generation. Top left: original
	image. Top right: the two boats from the left and the single boat from
	the right are copied one over another (exaggerated example). Bottom left:
	visualization of where different image parts are copied. Bottom right: When computing the
	ground truth for this copy-move forgery, the occlusions must be computed
	according to the splicing pattern of the image.
	}
	\label{fig:partial_occlusion}
	\end{figure}
\else
	\begin{figure}[!t]
		\centering
			\includegraphics[width=\linewidth]{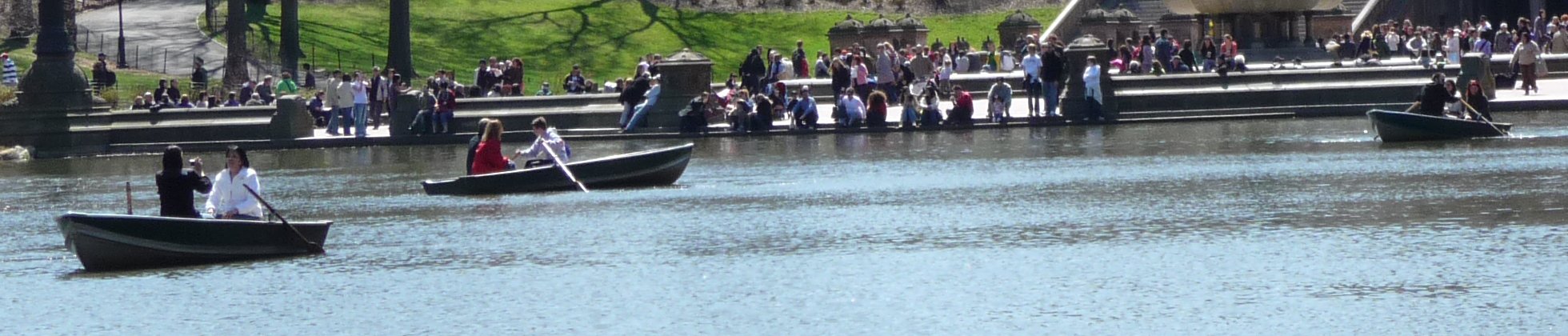}

			\vspace{1mm}
			\includegraphics[width=\linewidth]{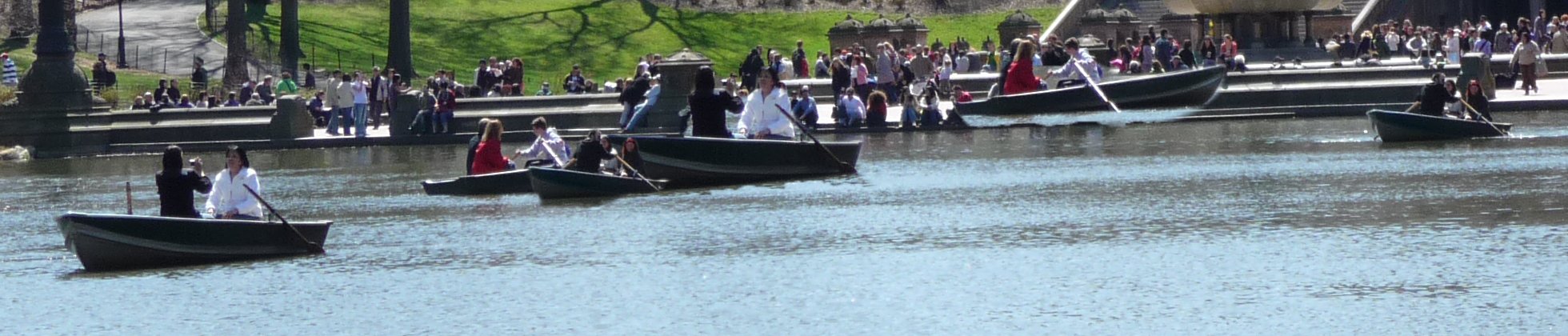}

			\vspace{1mm}
			\includegraphics[width=\linewidth]{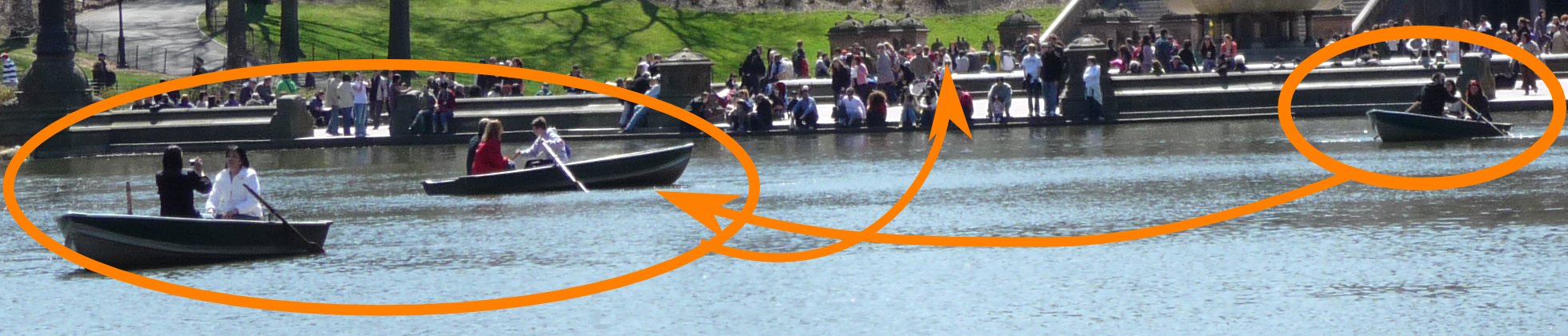}

			\vspace{1mm}
			\includegraphics[width=\linewidth]{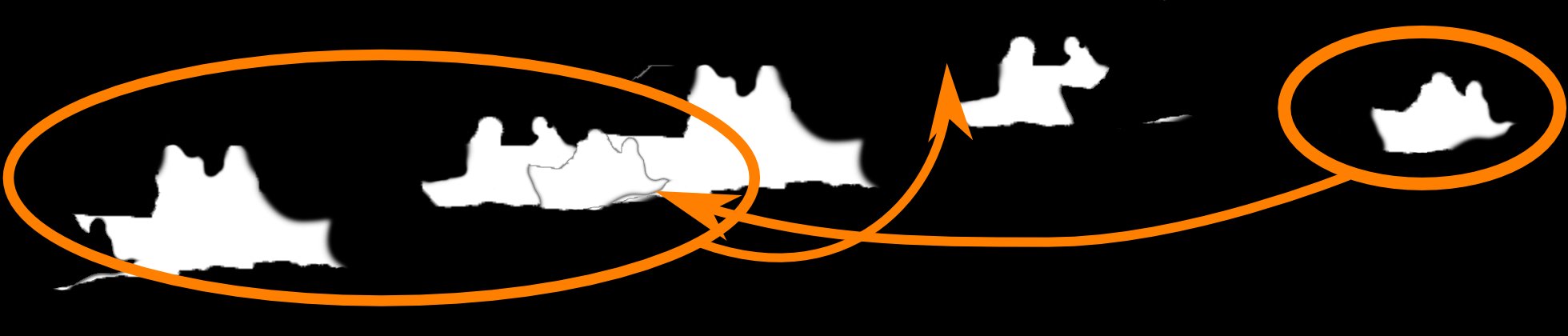}
	\caption{Artificial example of several types of occlusion in the ground truth
	generation. Top: original
	image. Second row: the two boats from the left and the single boat from
	the right are copied one over another (exaggerated example). Third row:
	visualization of where different image parts are copied. Fourth row: When computing the
	ground truth for this copy-move forgery, the occlusions must be computed
	according to the splicing pattern of the image.
	}
	\label{fig:partial_occlusion}
	\end{figure}
\fi
Various kinds of modifications can be applied to the snippet, \eg
the addition of Gaussian noise or the application of an affine transformation
on the snippet.  When affine transformations are used, the ground truth must be
(automatically) adjusted accordingly.  Care has to be taken in the case of
partial occlusion of snippets.
Our software framework allows multiple snippets to be
arbitrarily reinserted in the image. This can result in repeated snippet
overlap. See~\figRef{fig:partial_occlusion} for an exaggerated example.
All occluded pixels have to be removed from both the source and the target
snippet.

\ifCLASSOPTIONdraftcls
	\begin{figure}[!t]
		\centering
			\begin{minipage}[t]{0.42\linewidth}
			\vspace{0pt}
			\includegraphics[width=0.49\linewidth]{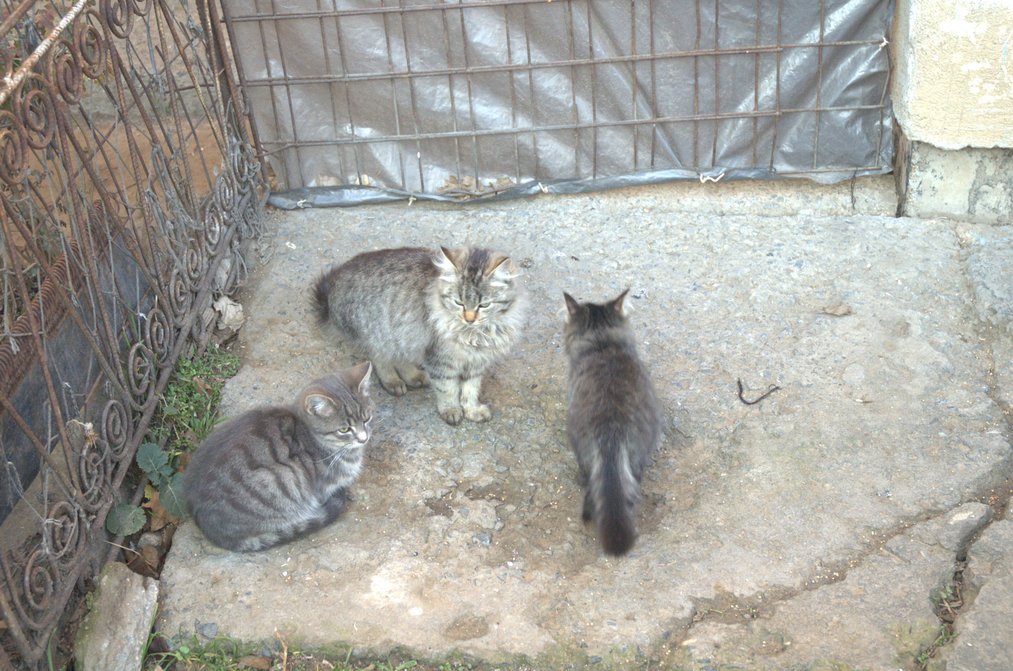}
			\includegraphics[width=0.49\linewidth]{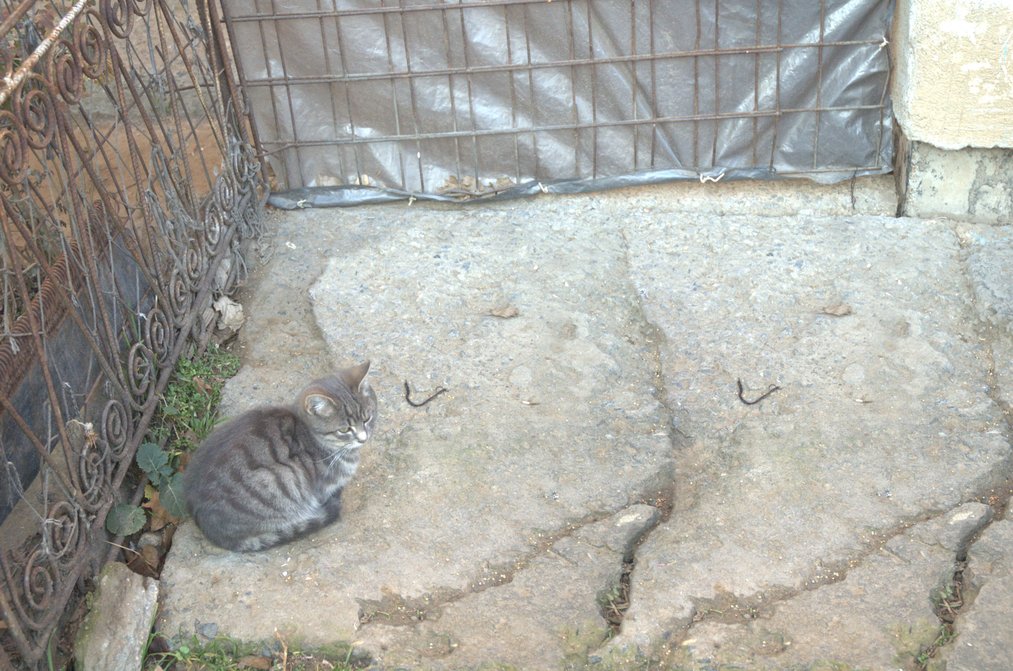}
			\end{minipage}
			\begin{minipage}[t]{0.42\linewidth}
			\vspace{0pt}
			\includegraphics[width=0.49\linewidth]{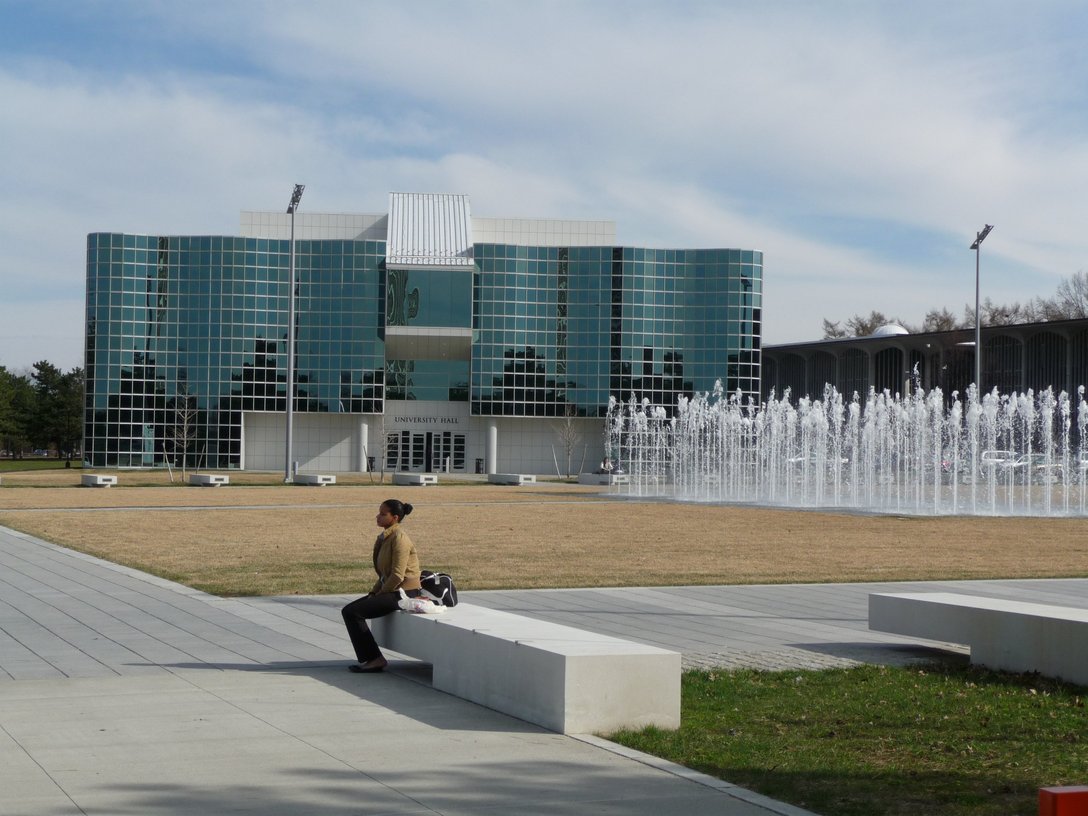}
			\includegraphics[width=0.49\linewidth]{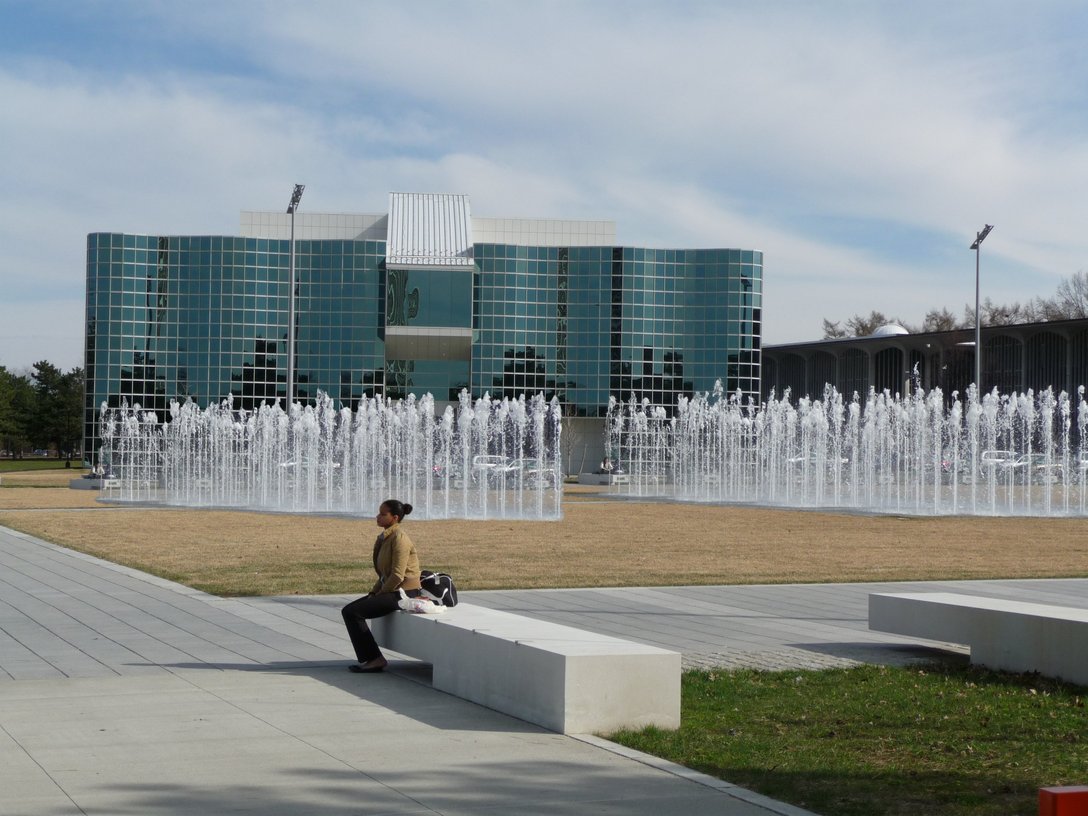}
			\end{minipage}
			\\
			\vspace{1mm}
			\begin{minipage}[t]{0.42\linewidth}
			\vspace{0pt}
			\includegraphics[width=0.49\linewidth]{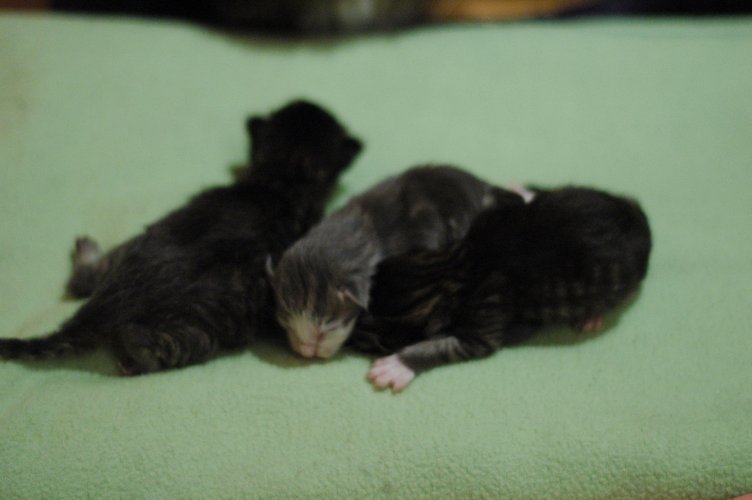}
			\includegraphics[width=0.49\linewidth]{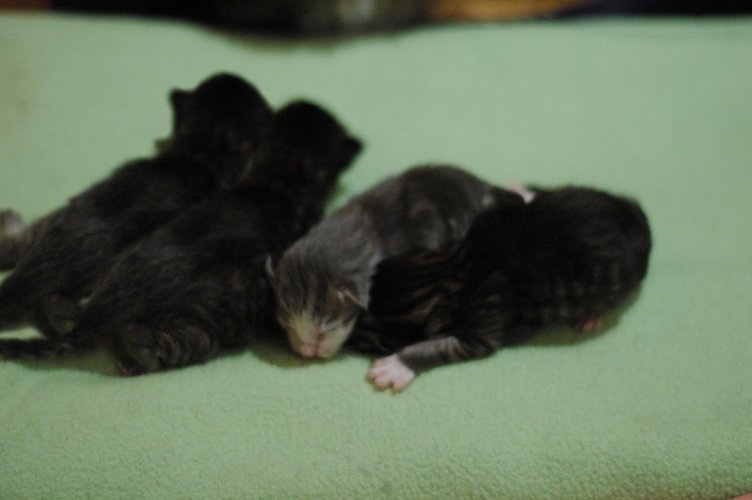}
			\end{minipage}
			\begin{minipage}[t]{0.42\linewidth}
			\vspace{0pt}
			\includegraphics[width=0.49\linewidth]{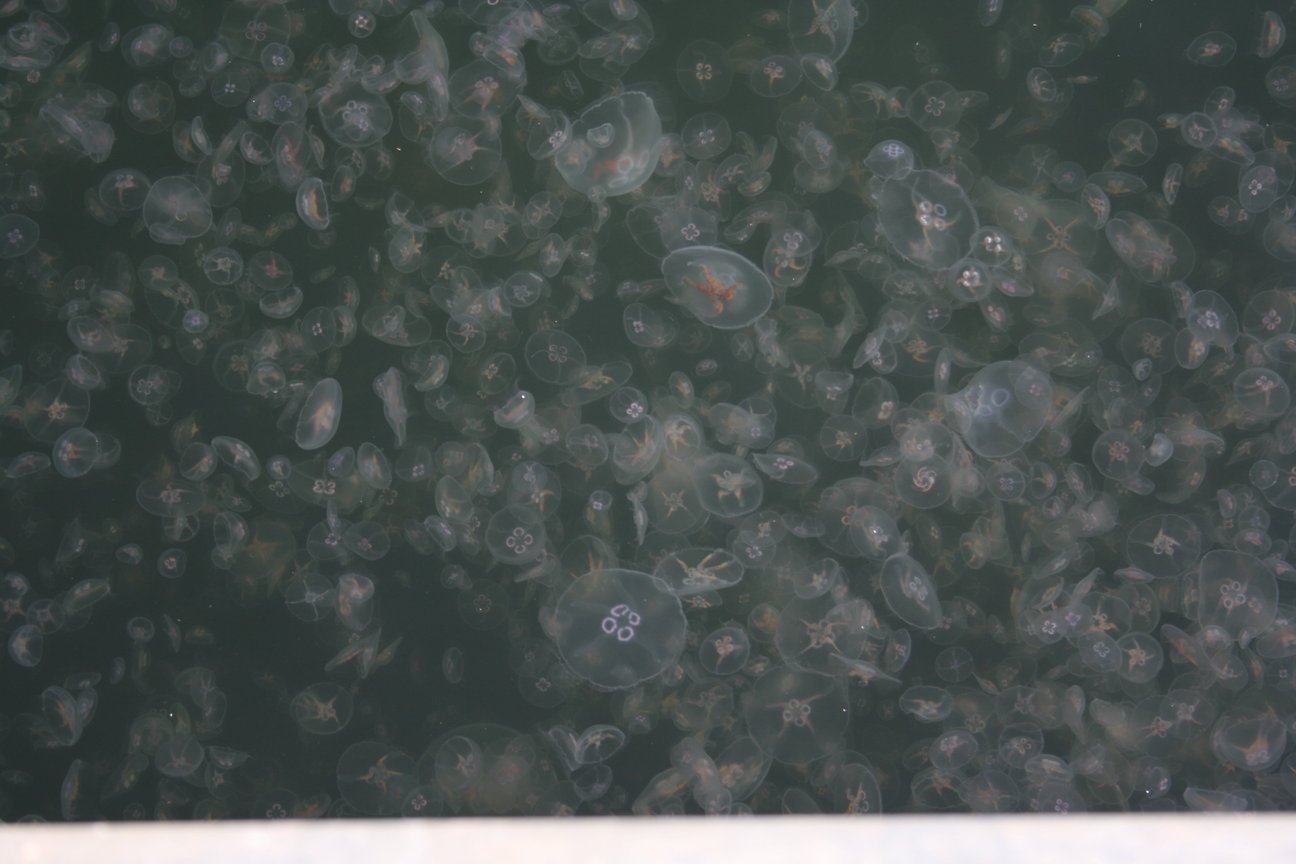}
			\includegraphics[width=0.49\linewidth]{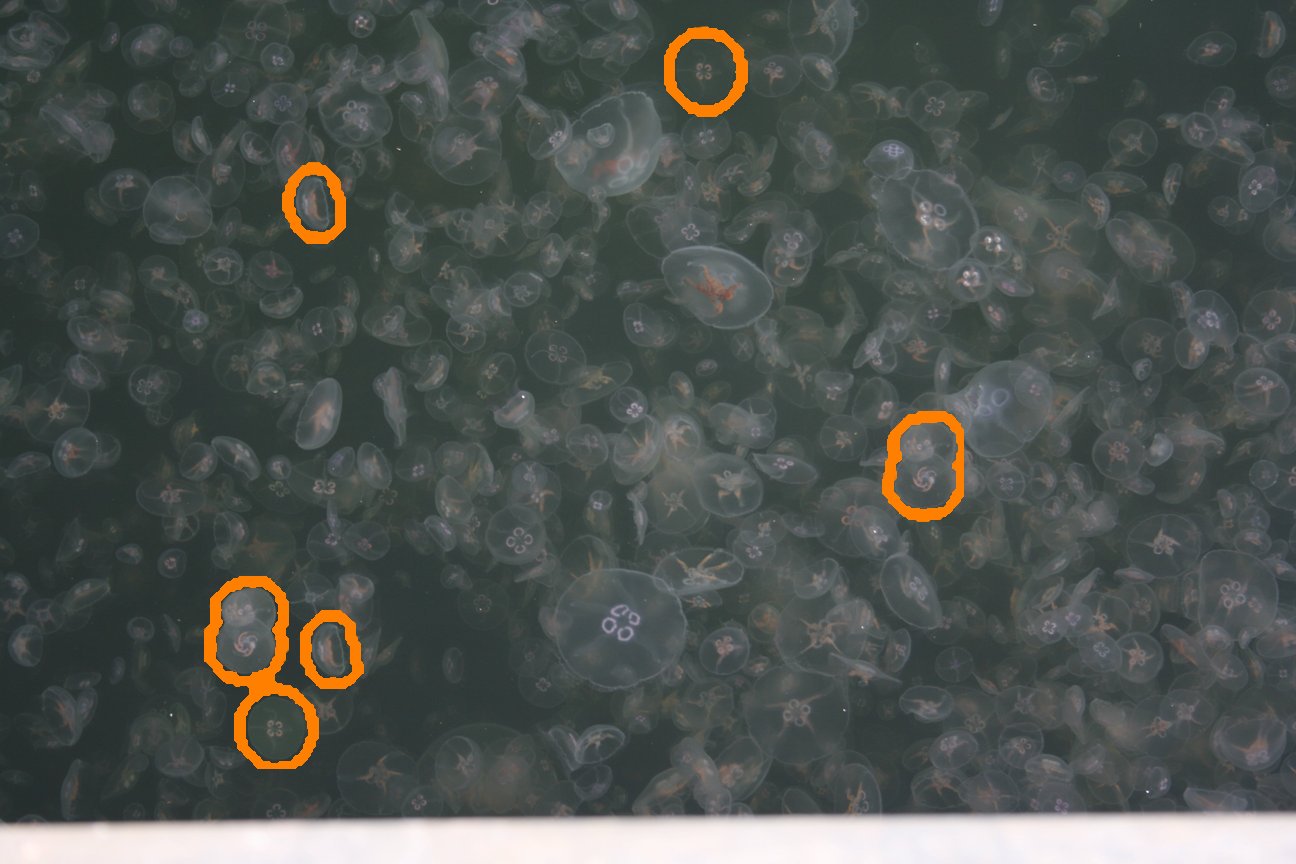}
			\end{minipage}
	\caption{Example test cases. Left: original image,
	right: copy-move forgery. Top left: \emph{lone cat}
	($3039\times{}2014$ pixels) containing a large, moderately textured copy
	region. Top right: \emph{fountain} ($3264\times{}2448$ pixels) with self-similar glass
	front in the background. Bottom left: \emph{four cat babies} ($3008\times{}2000$ pixels).
	The copied cat baby has a very low-contrast body region. Bottom right:
	\emph{jellyfish chaos} ($3888\times{}2592$ pixels) contains high-detail,
	high-contrast features. For better visualization, we highlighted the copied
	elements in \emph{jellyfish chaos}.}
	\label{fig:example_test_case}
	\end{figure}
\else
	\begin{figure}[!t]
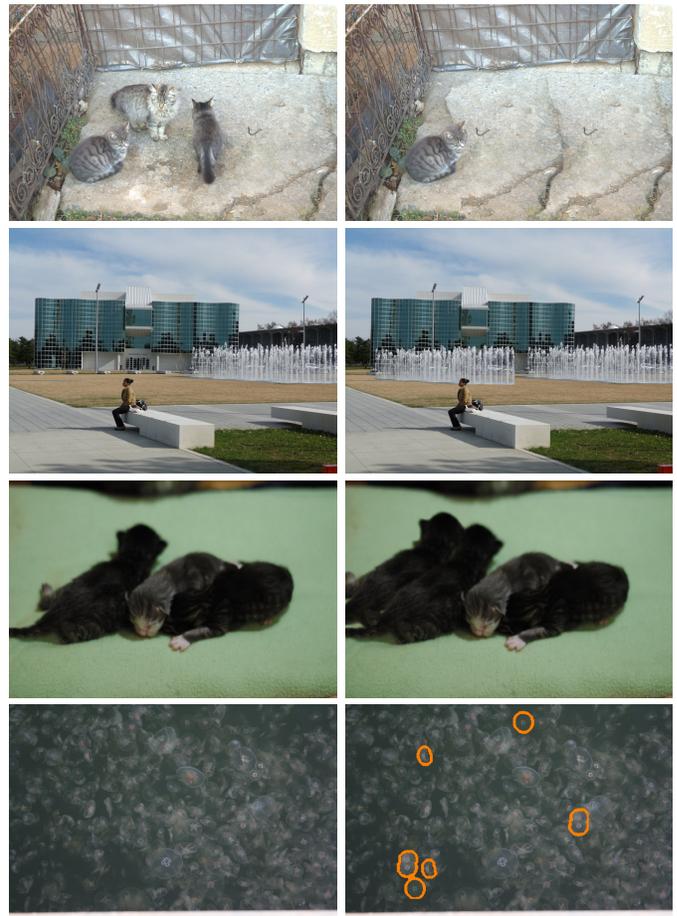

		\centering
			\includegraphics[width=0.49\linewidth]{db_examples/small_lone_cat}
			\includegraphics[width=0.49\linewidth]{db_examples/small_lone_cat_copy}

			\vspace{1mm}
			\includegraphics[width=0.49\linewidth]{db_examples/small_fountain}
			\includegraphics[width=0.49\linewidth]{db_examples/small_fountain_copy}

			\vspace{1mm}
			\includegraphics[width=0.49\linewidth]{db_examples/small_four_babies}
			\includegraphics[width=0.49\linewidth]{db_examples/small_four_babies_copy}

			\vspace{1mm}
			\includegraphics[width=0.49\linewidth]{db_examples/small_jellyfish_chaos}
			\includegraphics[width=0.49\linewidth]{db_examples/small_jellyfish_chaos_copy_annotated}
	\caption{Example test cases (4 out of 48 base scenarios). Left: original image,
	right: copy-move forgery. From top to bottom: \emph{lone cat}
	($3039\times{}2014$ pixels) containing a large, moderately textured copy
	region, \emph{fountain} ($3264\times{}2448$ pixels) with self-similar glass
	front in the background, \emph{four cat babies} ($3008\times{}2000$ pixels)
	where the copied cat baby has a very low-contrast body region, 
	and \emph{jellyfish chaos} ($3888\times{}2592$ pixels)
	containing high-detail, high-contrast features. For better visualization, we
	highlighted the copied elements in \emph{jellyfish chaos}.}
	\label{fig:example_test_case}
	\end{figure}
\fi

\ifCLASSOPTIONdraftcls
\figRef{fig:example_test_case} shows a selection of images from the database.
On the left, the original image is shown, and on the right the
corresponding ``reference manipulation''. Top left, an example of a
relatively straightforward tampering case is presented, \ie a ``large''
copied area with ``good'' contrast. Top right, the building in the
background exhibits a high contrast, regular pattern. Here, our aim was to
create a tampered image that would induce a large number of positive matches.
Bottom left, a very low contrast region, the cat baby, is copied. We
expect keypoint-based methods to have difficulties with such test cases. The
last example contains a large number of jellyfish. Here, though a single
jellyfish contains considerable contrast, the sheer number of jellyfish is
expected to make detection difficult. For better visualization, we highlighted
the copied elements in this image.
\else
\figRef{fig:example_test_case} shows a selection of images from the database.
On the left column, the original image is shown, and on the right column the
corresponding ``reference manipulation''. In the top row, an example of a
relatively straightforward tampering case is presented, \ie a ``large''
copied area with ``good'' contrast. In the second example, the building in the
background exhibits a high contrast, regular pattern. Here, our aim was to
create a tampered image that would induce a large number of positive matches.
In the third row, a very low contrast region, the cat baby, is copied. We
expect keypoint-based methods to have difficulties with such test cases. The
last example contains a large number of jellyfish. Here, though a single
jellyfish contains considerable contrast, the sheer number of jellyfish is
expected to make detection difficult. For better visualization, we highlighted
the copied elements in this image.
\fi

The software together with the images can be downloaded from our web site\footnote{\thisVariantDiff{\texttt{http://www5.cs.fau.de/}}}.

\section{Error Measures}
\label{sec:error}

We focused our evaluation on two performance characteristics. For practical
use, the most important aspect is the ability to distinguish tampered and
original images. However, the power of an algorithm to correctly annotate the
tampered region is also significant, especially when a human expert is visually
inspecting a possible forgery. Thus, when evaluating CMFD algorithms, we
analyze their performance at two levels: at image level, where we focus on
whether the fact that an image has been tampered or not can be detected; at
pixel level, where we evaluate how accurately can tampered regions be
identified.

\subsection{Metrics}

At image level, the important measures are the number of correctly detected
forged images, \TP, the number of images that have been erroneously detected as
forged, \FP, and the falsely missed forged images \FN. From these we computed
the measures \textit{Precision}, $\Precision$, and \textit{Recall}, $\Recall$. They are defined as: 
\begin{equation}
\Precision = \frac{\TPm}{\TPm + \FPm} \mathrm{~,~and~} \Recall = \frac{\TPm}{\TPm + \FNm} \enspace.
\end{equation}
\textit{Precision} denotes the probability that a detected forgery is truly a forgery,
while \textit{Recall} shows the probability that a forged image is
detected. \textit{Recall} is often also called true positive rate. 
In the tables we also give the \FM score as a measure which combines precision
and recall in a single value.
\begin{equation}
\FMm = 2\cdot\frac{\Precision\cdot \Recall}{\Precision + \Recall}\enspace.
\end{equation}

We used these measures at pixel level, too. In that case, \TP are the number of correctly 
detected forged pixels. \FP denotes the number of falsely detected forged pixels and \FN are
the falsely missed pixels. The previous definition of \textit{Precision}, \textit{Recall} and \FM measures
also hold on the pixel level.

\subsection{Protocol}
\begin{figure}[!t]
	\centering
	\ifCLASSOPTIONdraftcls
		\includegraphics[width=0.4\linewidth]{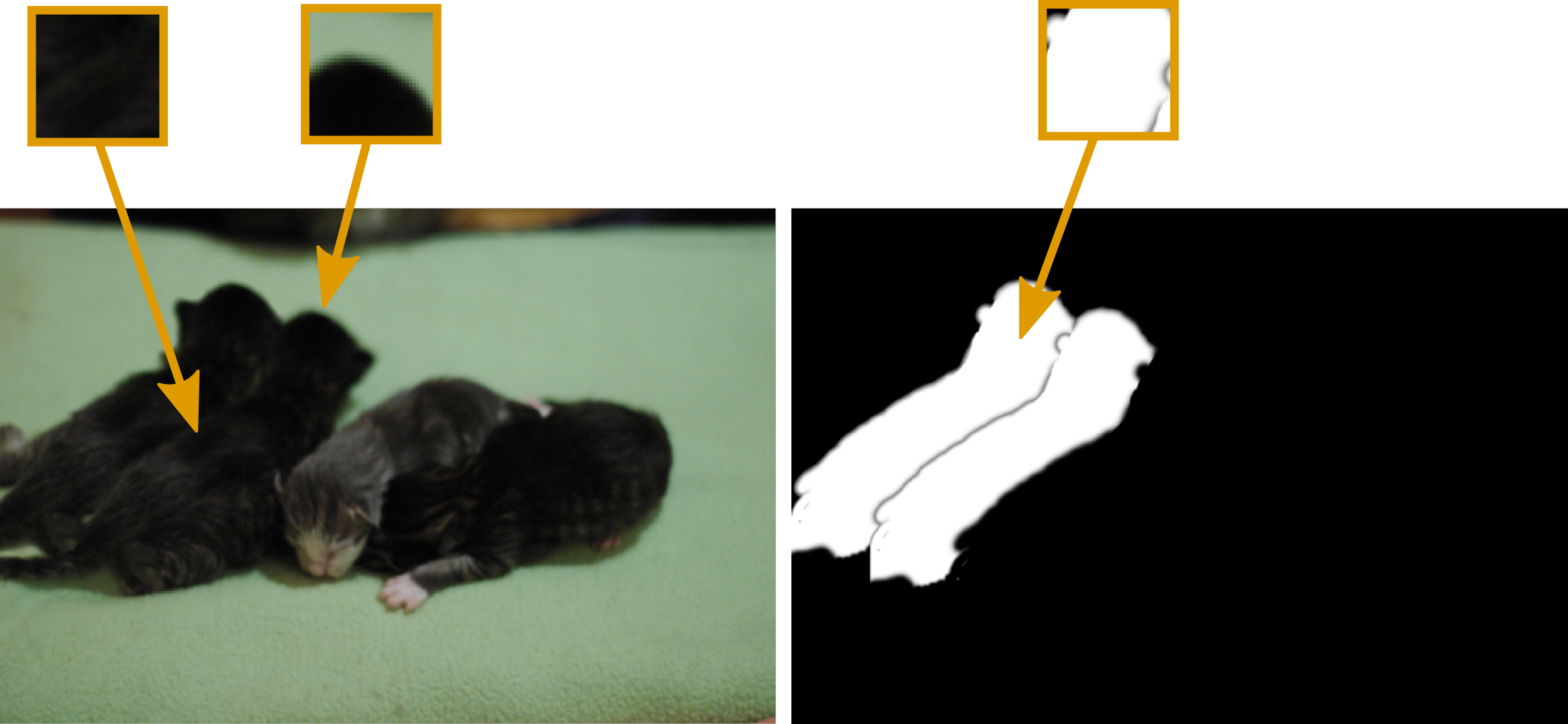}
	\else
		\includegraphics[width=\linewidth]{images/five_babies/illustration_special_cases_bitmap}
	\fi
	\caption{Illustration of three special cases of the ground truth
	generation (see text for details). Left: Pixels at the boundaries of source
	and/or target regions might be intensity-wise indistinguishable.
	Nevertheless, the ground truth differentiates these pixels. Middle: Smeared
	or partially transparent pixels (typically at boundaries) are excluded from
	the evaluation. Right: the ground truth image. Note that by construction
	the ground truth is not affected by the application of JPEG compression or
	additive noise.}
	\label{fig:justification_gt}
\end{figure}
Consider a CMFD algorithm that states per image whether it is tampered or not.
If a copy-operation
has been conducted in the image, it should raise an alarm, and vice
versa. From a practical viewpoint, this can be considered the most important
error measure, as the original goal -- exposing digital forgeries -- is
directly fulfilled. However, when performance is measured on a per-image basis,
the underlying reason that raised the alarm is not really considered.
Image-wise performance charts do not typically distinguish between a method
which correctly marks most forged areas versus a method which almost randomly
marks regions.

To address this issue, we conducted a second series of evaluations. Here, we
examined the performance of detecting copies on a per-pixel basis. In principle,
such an approach allows a much more finegrained statement about the details
that a method is able to detect. Ultimately, we consider these results to offer
more insight on the construction of future CMFD algorithms.
However, such an evaluation requires a careful
definition of ground truth, since it needs to clearly specify which pixels have been copied.
This is often not a simple task. We identified
three cases that need further attention.
\figRef{fig:justification_gt} illustrates these cases. It shows the
tampered \textit{four copied cat babies} and the generated ground truth.  Three particular
regions are shown as closeups.
On the left window, the boundary between source and copy, consisting of black
fur, is shown. The intensities of the source region, as well as the copied
region, are almost indistinguishable. In
such cases we draw the boundary
where the target region has been inserted. This can also be seen from the
grayish border between source and copy in the ground truth image. In the
middle window, a closeup of the head is shown. Next to the head is the seam of the
copied region. However, seams are almost always not 1-to-1 copies, but
partially transparent, smeared and so on. In the ground truth image, such
pixels can be recognized as being neither fully white nor fully black. We
consider such pixels as not well-defined for copy-move forgery detection, as
they exhibit a mixture of the copy and the original background. Thus, for the evaluation
of CMFD, we excluded all such pixels. Finally, on the
right window is a closeup of the ground truth. The interior of the copied area is
fully white, \ie every pixel counts for copy-move forgery detection.
However, further postprocessing, \eg added noise, or JPEG artifacts, can cause
pixels in the copied region to considerably deviate from pixels in the
source region. One could choose to exclude pixels that deviate significantly, but
it is unclear how to accurately define a sufficient amount of deviation. 
Nonetheless, our goal is to examine the robustness of CMFD
approaches under such disturbances. Thus, we chose to leave the ground truth
``clean'', \ie independent of any applied postprocessing.

\section{Experiments}
\label{sec:experiments}

In the first series of experiments, we evaluated the detection rate of tampered
images. In the second series, we evaluated pixelwise the detection of copied
regions, in order to obtain a more detailed assessment of the discriminative properties of the
features.  In total, we conducted experiments with about $4700$ variants of
the forged image (\eg different scales of snippets, different rotation angles
of snippets, different compression rates and their combinations) in order to
better understand the behavior of the different feature sets.  The complete
result tables, as well as the source code to generate these results, are also
available from our web site.

\subsection{Threshold Determination}

Thresholds that are specific to a particular feature set were manually adjusted
to best fit the benchmark dataset.
Most threshold values for the processing pipeline (according to Sec.
\ref{sec:pipeline}) were fixed across the different methods, when possible,
to allow for a fairer comparison of the feature performance.

\noindent
\textbf{Block size $b$:} We chose to use a block size of $16$ pixels.
		We found this to be a good trade-off between detected image details and
		feature robustness. Note that the majority of the original methods also proposed a block size of $16$ pixels.

\noindent
\textbf{Minimum Euclidean distance $\tau_1$:} Spatially close pixels are
	closely correlated. Thus, matches between spatially close blocks should be avoided.  
	In our experiments, we set the minimum Euclidean distance between two matched blocks 
	to $50$ pixels. \thisVariantDiff{Thus, note that we are unable to detect
	copies when the pixels are moved for less than $50$ pixels. However, given
	the high resolution of the benchmark images, this limitation is not
	relevant for this work.}

\begin{figure}[t]
\centering
	\ifCLASSOPTIONdraftcls
		\includegraphics[width=0.5\linewidth]{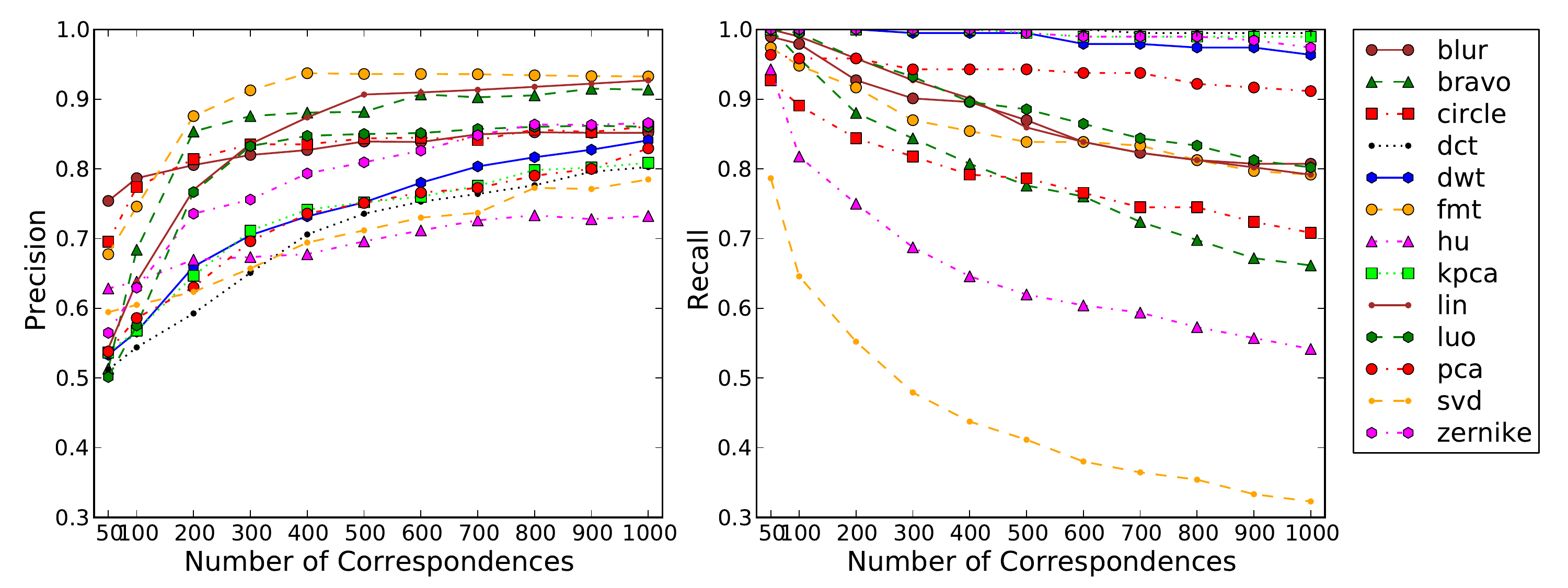}
	\else
		\includegraphics[width=\linewidth]{plots_img_level/1_2/plot_best_jpeg_default_new_i_l_precision_recall.pdf}
	\fi
	\caption{Results at image level for different minimum number of
	correspondences}
	\label{fig:threshold}
\end{figure}

\noindent
\textbf{Minimum number of correspondences $\tau_2$:} This threshold reflects
	the minimum number of pairs which have to fulfill the same
	affine transformation between the copied and the pasted region. Thus, it
	compromises between improved noise suppression, and false rejection of
	small copied regions.
	$\tau_2$ strongly depends on the features, as some features generate denser
	result maps than others. Consequently, $\tau_2$ has to be chosen for each
	feature individually. We empirically
	determined appropriate values $\tau_2$ as follows.
	From our dataset, we created CMFD benchmark images with JPEG quality levels
	between $100$ and $70$ in steps of $10$. Thus, we evaluated on the $48$
	tampered images for $48\times4=192$ images.
	The JPEG artifacts should simulate a training set with
	slight pixel distortions.
	Per block-based feature, we estimated $\tau_2$ by optimizing the \FM-measure at
	image level. The results of the experiment are shown in
	\figRef{fig:threshold}.  Please note that throughout the experiments, we
	were sometimes forced to crop the $y$-axis of the plots, in order to
	increase the visibility of the obtained results.  The feature set-specific
	values for $\tau_2$ are listed in the rightmost column of
	\tabRef{tab:img_nul}. For the sparser keypoint-based methods, we require
	only \thisVariantDiff{$\tau_2=4$} correspondences. 
	
\noindent
\textbf{Area threshold $\tau_3$:} 
	In our experiments, we set $\tau_3=\tau_2$ for the block-based methods and
	$\tau_3=1000$ for the keypoint-based methods \thisVariantDiff{to remove
	spurious matches}\footnote{\thisVariantDiff{Alternatively, it would be
	possible to set the threshold for keypoint matching stricter, and then
	to omit $\tau_3$ completely. However, we preferred this variant (\ie a
	more lenient matching threshold) in order to gain better robustness to
	noise.}}.

\noindent
\textbf{Individual feature parameters:} We omitted the Gaussian pyramid
	decomposition for the Hu-Moments (in contrast to the original
	proposition~\cite{Wang09:FAR}). This variant yields better results gave
	better results on our benchmark data.
	For \cir, we had to use a different
	block size $b=17$, as this feature set requires an odd sized blocks for the radius computation. 
	For \kpca, two parameters had to be determined, namely the number of samples $M$ and the variance
	of the Gaussian kernel $\sigma$. 
	We set up a small
	experiment with two images (with similar proportions as images from 
	our database) in which for both images a block
	of size $128\times128$ was copied and pasted. Then we varied the
	parameters and chose the best result in terms of the \FM-measure.
	We observed that with increasing $\sigma$ and $M$ the results became
	slightly better.  We empirically determined that values of $M=192$ and
	$\sigma=60$ offer an overall good performance. Note that, these values are
	larger than what~\etal{Bashar}~\cite{Bashar10:EDR} used. For the remaining
	features, we used the parameters as suggested in the respective papers.

\subsection{Detection at Image Level}

We split these experiments in a series of separate evaluations. We start
with the baseline results, \ie direct 1-to-1 copies (no postprocessing) of the
pixels. Subsequent
experiments examine the cases of: noise on the copied region, JPEG compression on
the entire image, rotation and scaling of the copied region.

\subsubsection{Plain Copy-Move}

\begin{table}[t]
\centering
	\ifCLASSOPTIONdraftcls
		\renewcommand{\arraystretch}{0.7}
	\fi
	\caption{Results for plain copy-move at image level in percent}
	\label{tab:img_nul}
	\ifCLASSOPTIONdraftcls
		\begin{tabular}{|l|r|r|r||r||l|r|r|r||r|}
			\hline
			\input{tables/nul_default_new_per_img_individual_draft.tex}
	\else
		\begin{tabular}{|l|r|r|r||r|}
			\hline
			\input{tables/nul_default_new_per_img_individual.tex}
	\fi
		\hline
	\end{tabular}
\end{table}

As a baseline, we evaluated how the methods perform under ideal conditions. 
We used the $48$ original images, and spliced $48$
images without any additional modification. We chose per-method optimal thresholds for
classifying these $96$ images. Interestingly, although 
the sizes of the images and the manipulation regions vary greatly on this test set,
$13$ out of the $15$ tested features perfectly solved this CMFD problem with
a recall-rate of $100\%$ (see \tabRef{tab:img_nul}). However, only four
methods have a precision of more than $90\%$. This means that most of the
algorithms, even under these ideal conditions, generate some false alarms.
This comes mainly from the fact that the images in the database impose diverse
challenges, and the large image sizes increase the probability of false positive
matches.

\begin{figure}[t]
\centering
	\ifCLASSOPTIONdraftcls
		\subfigure[Gaussian white noise]{\includegraphics[width=0.45\linewidth]{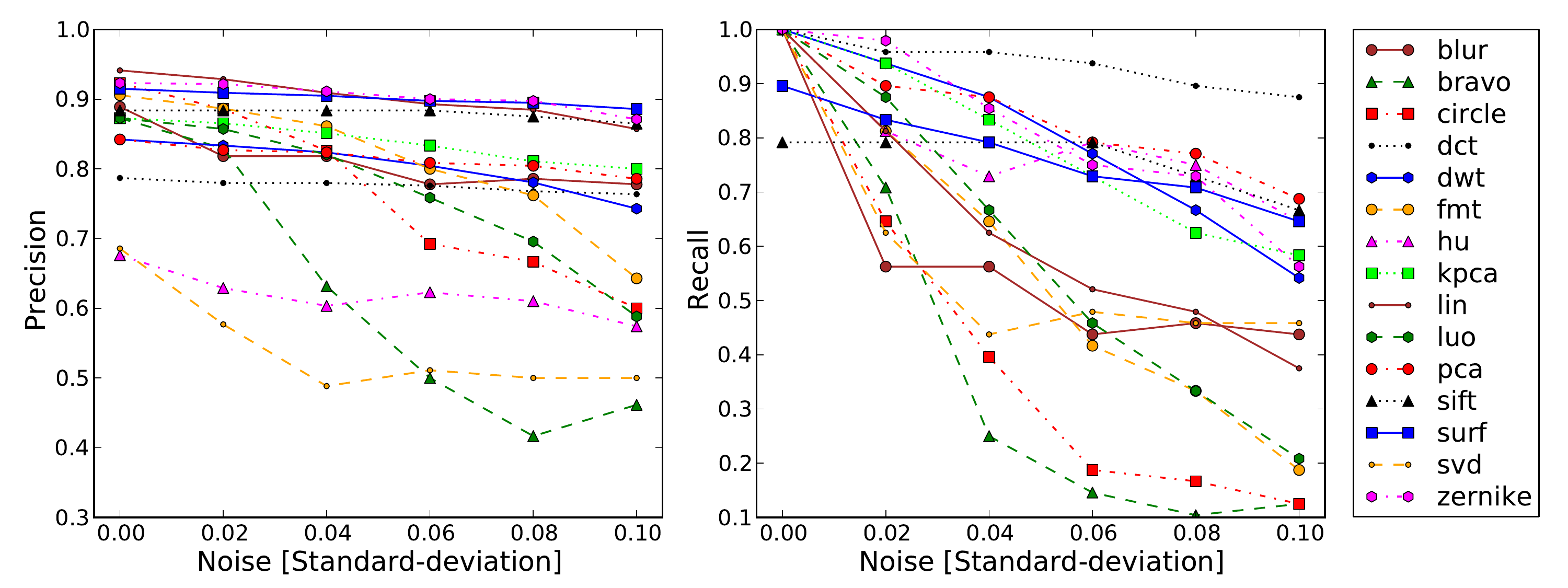}\label{fig:img_noise_results}}
		\subfigure[JPEG compression]{\includegraphics[width=0.45\linewidth]{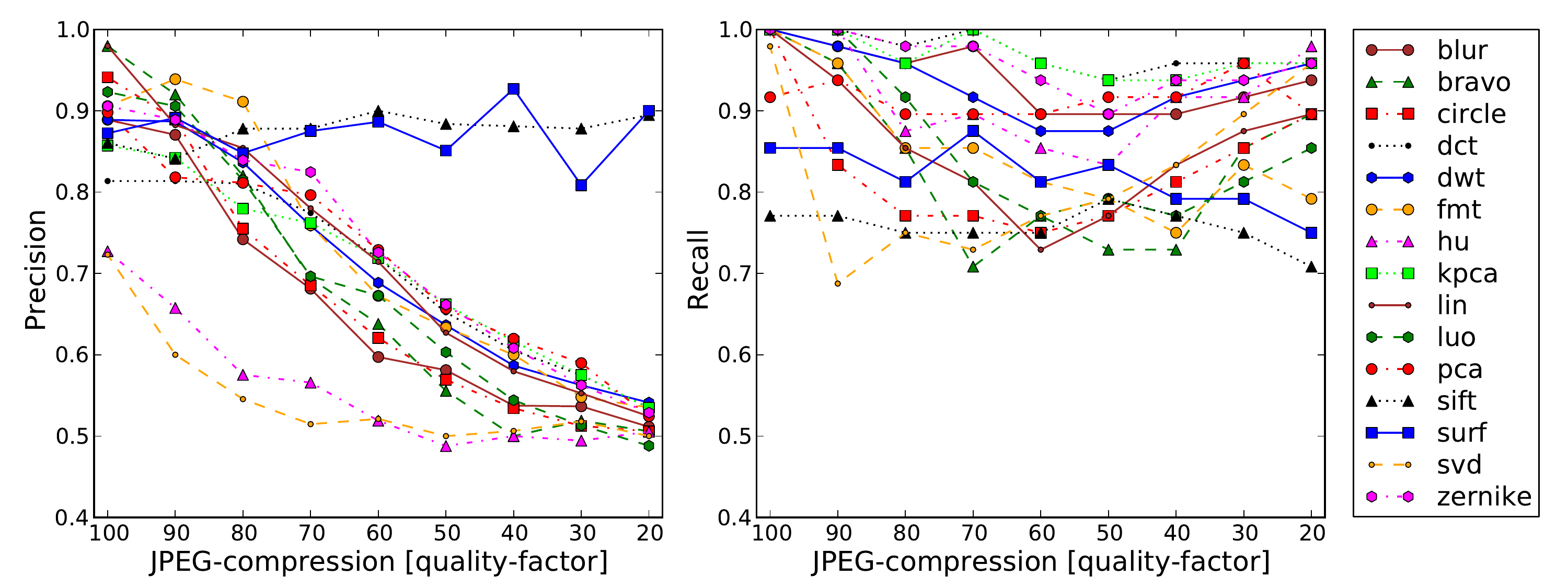}\label{fig:img_jpeg_results}}

		\subfigure[Rescaled copies]{\includegraphics[width=0.45\linewidth]{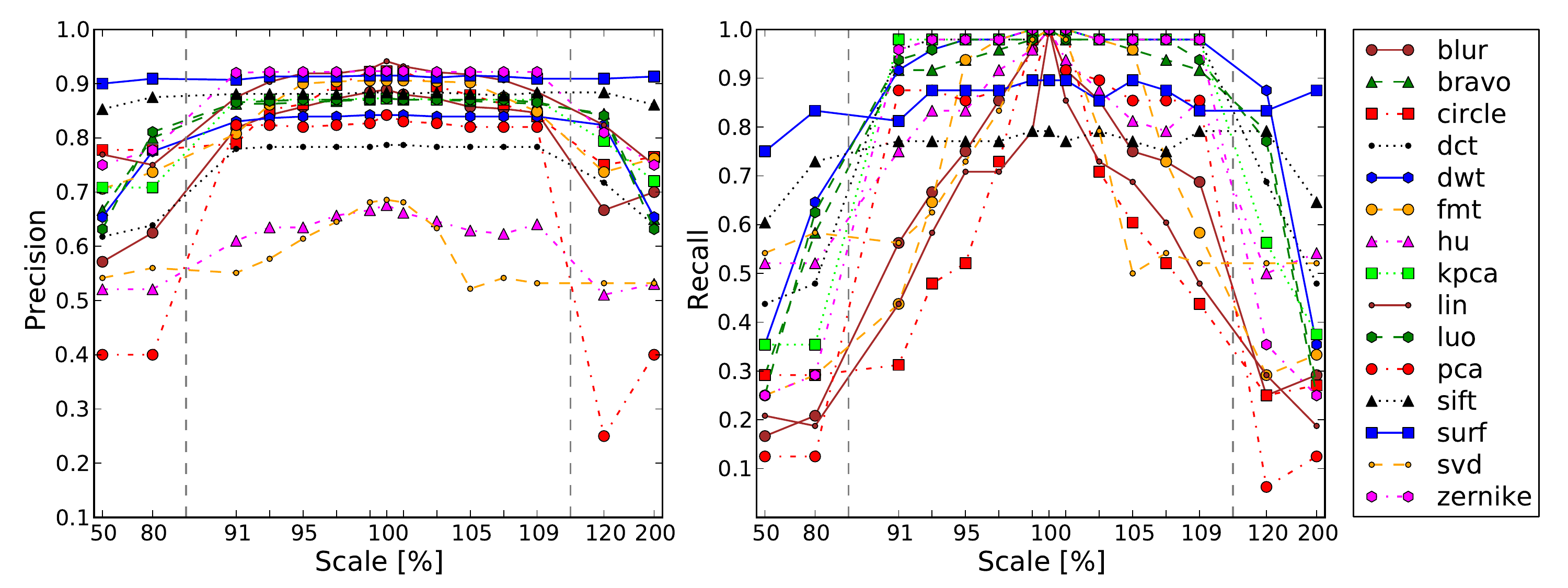}\label{fig:img_scale_results}}
		\subfigure[Rotated copies]{\includegraphics[width=0.45\linewidth]{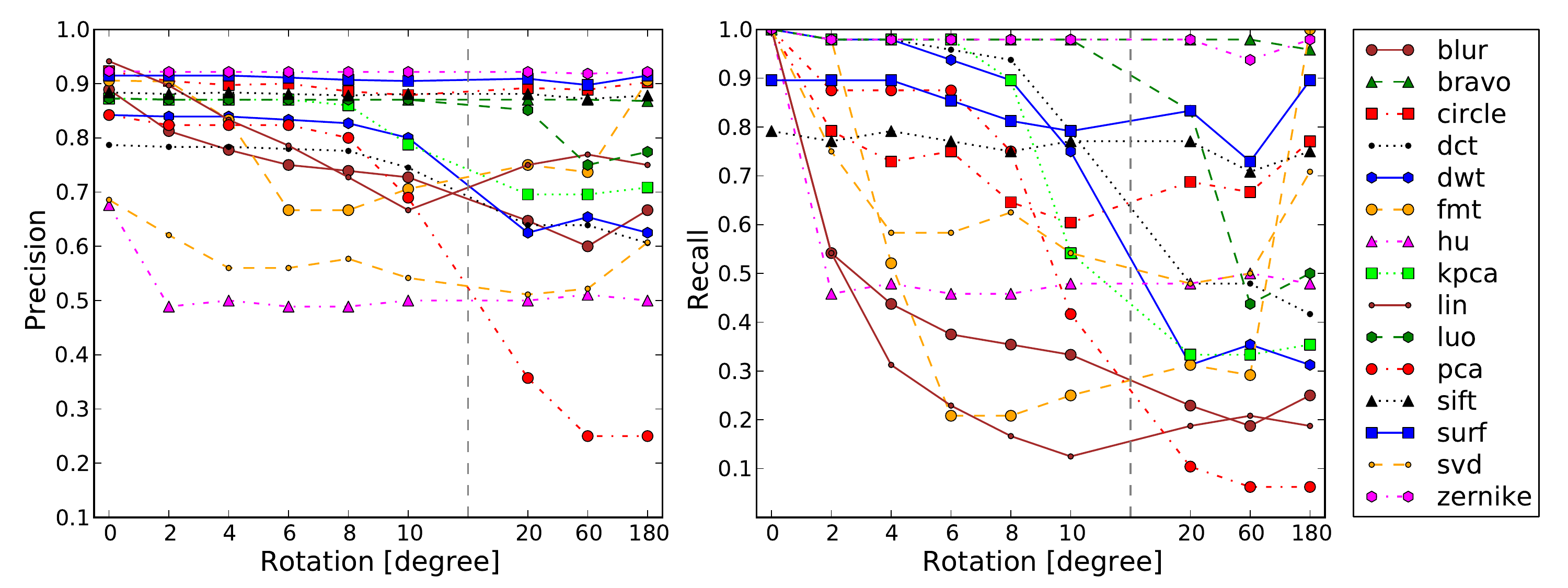}\label{fig:img_rot_results}}
	\else
		\subfigure[Gaussian white noise]{\includegraphics[width=\linewidth]{plots/1_2/plot_lnoise_i_lnoise_i_l_precision_recall_}\label{fig:img_noise_results}}

		\subfigure[JPEG compression]{\includegraphics[width=\linewidth]{plots/1_2/plot_jpeg_i_jpeg_i_l_precision_recall_}\label{fig:img_jpeg_results}}

		\subfigure[Rescaled copies]{\includegraphics[width=\linewidth]{plots/1_2/plot_scale_i_scale_i_l_precision_recall_}\label{fig:img_scale_results}}

		\subfigure[Rotated copies]{\includegraphics[width=\linewidth]{plots/1_2/plot_rot_i_rot_i_l_precision_recall_}\label{fig:img_rot_results}}
	\fi
	\caption{Experimental results at image level (see text for details).}
	\label{fig:img_results}
\end{figure}

\subsubsection{Robustness to Gaussian noise} 
We normalized the image intensities between $0$ and $1$ and added zero-mean Gaussian
noise with standard deviations of $0.02$, $0.04$, $0.06$, $0.08$ and $0.10$ to
the inserted snippets before splicing.
Besides the fact that a standard
deviation of $0.10$ leads to clearly visible artifacts, $7$ out of $15$
features drop to under $50\%$ recall rate, while the precision decreases only
slightly, see \figRef{fig:img_noise_results}. \dct exhibits a remarkably high
recall, even when large amounts of noise are added. \pca, \sift, \surf and \hu also
maintain their good recall, even after the addition of large amounts of noise.
At the same time, several methods exhibit good precision. Among these, \surf
provides a good balance between precision and recall, followed by \pca.

\subsubsection{Robustness to JPEG compression artifacts} 
\label{subsubsec:exp_imagelevel_jpeg}
We introduced a common global disturbance, JPEG
compression artifacts. 
The quality factors varied between $100$ and $20$ in
steps of $10$, as provided by \libJPEG\footnote{\texttt{http://libjpeg.sourceforge.net/}}.
Per evaluated quality level, we applied the same JPEG compression to $48$ forgeries
and $48$ complementary original images.
For very low quality factors, the visual quality of the image is strongly
affected. However, we consider at least quality levels down to $70$ as
reasonable assumptions for real-world forgeries.
\figRef{fig:img_jpeg_results} shows the results for this experiment.
The precision of \surf and \sift remains surprisingly stable, while block-based
methods slowly degenerate to a precision of $0.5$. On the other hand, many
block-based methods exhibit a relatively high recall rate, \ie miss very few
manipulations. Among these, \kpca, \dct, \zernike, \blur and \pca constantly
reach a recall of $90\%$ or higher.

\subsubsection{Scale-invariance} 
One question that recently gained attention was the resilience of CMFD
algorithms to affine transformations, like scaling and rotation. We conducted
an experiment where the inserted snippet was slightly rescaled, as is often
the case in real-world image manipulations. Specifically, we
rescaled the snippet between $91\%$ and $109\%$ of its original size, in steps
of $2\%$. We also evaluated rescaling by $50\%$, $80\%$, $120\%$ and $200\%$ to
test the degradation of algorithms under larger amounts of snippet resizing.
Note that we only scaled the copied region, not the source region.
\figRef{fig:img_scale_results} shows the results for this experiment.
Most features degenerate very fast at low rates of up- or down-sampling.  Some
methods, namely \kpca, \zernike, \luo, \dwt, \dct and \pca are able to
handle a moderate amount of scaling. For more extreme scaling parameters,
keypoint-based methods are the best choice: their performance remains
relatively stable across the whole scaling parameters.

\subsubsection{Rotation-invariance} 
Similar to the previous experiment, we rotated
the snippet between $2^\circ$ to $10^\circ$, in steps of $2^{\circ}$, and also
tested three larger rotation angles of $20^\circ$, $60^\circ$ and $180^\circ$.
In prior
work~\cite{Christlein10:ORI,Christlein10:SFD}, we already
showed that \zernike , \bravo and \cir are particularly well-suited as
rotation-invariant features. Our new results, computed on a much more
extensive data basis, partially confirm this. \figRef{fig:img_rot_results}
shows the results.
\zernike features provide the best precision, followed by \surf, \cir, \luo and
\bravo.
In the recall-rate, \bravo and \zernike yield consistently good results
and thus seem to be very resilient to rotation. For small amounts of rotation,
\kpca and \luo perform also strongly, for higher amounts of rotation, \surf
features perform best. \fmt, \lin, \hu and \blur seem not to be well suited to
handle variations of rotation.

\subsubsection{Robustness to Combined Transformation}
\label{subsubsec:comb_transform_image}

\begin{figure}[t]
\centering
	\ifCLASSOPTIONdraftcls
		\includegraphics[width=0.45\linewidth]{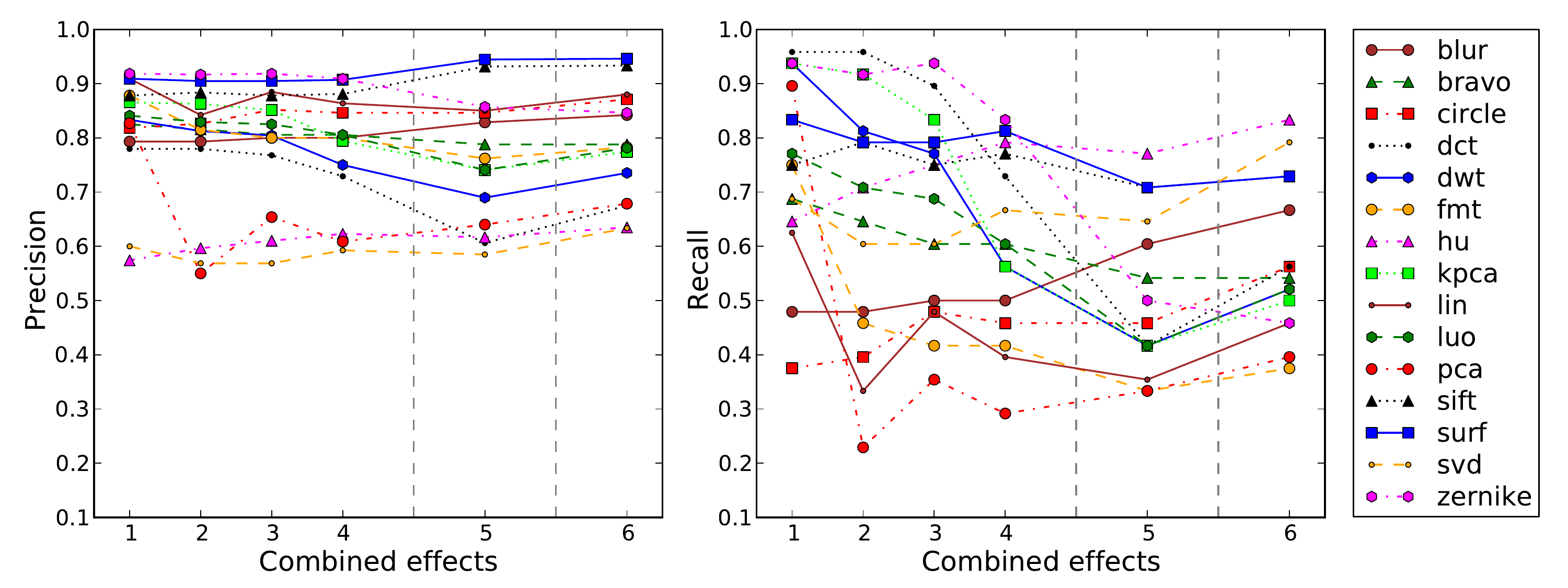}
	\else
		\includegraphics[width=\linewidth]{plots/plot_cmb_easy1_i_cmb_easy1_i_l_precision_recall_}
	\fi
	\caption{Results for increasingly difficult combined transformations at image level.}
	\label{fig:img_cmb}
\end{figure}

In this experiment, we examined the performance under several joint effects.
We rotated the snippet by $2^{\circ}$, scaled it up by $1\%$ and compressed the
image with a JPEG-compression level of $80$. In three subsequent setups, we
increased per step the rotation by $2^\circ$, increased scaling by $2\%$,
and decreased the JPEG quality by $5$ points. \thisVariantDiff{In setup $5$ and $6$,
slightly stronger parameters were chosen: rotation was set to $20^\circ$ and
$60^\circ$, scaling was set to $120\%$ and $140\%$, and JPEG quality was
set to $60$ and $50$, respectively.}
\figRef{fig:img_cmb} shows that high precision can be achieved for several
feature sets. The best recall for small variations is achieved by \dct and
\zernike. For the fourth step, \surf and \sift are almost on a par with
\zernike. Note that also in the fourth step, a number of features exhibit a
recall below $50\%$, and can thus not be recommended for this scenario.
\thisVariantDiff{For large rotations and scaling in the combined effects (see
the scenarios $5$ and $6$), \sift and \surf show best precision and very good
recall.}

\subsection{Detection at Pixel Level}

A second series of experiments considers the accuracy of the features
at pixel level. The goal of this experiment is to evaluate how
precisely a copy-moved region can be marked. However, this testing has a
broader scope, as it is directly related with 
the discriminating abilities of a particular feature set.
Under increasingly challenging evaluation data, experiments on per-match level
allow one
to observe the deterioration of a method in greater detail.  We
repeated the experiments from the previous subsections, with the same test
setups. The only difference is that instead of classifying the image as original
or manipulated, we focused on the number of detected (or missed, respectively)
copied-moved matches.

For each detected match, we check the centers of two matched blocks against the
corresponding (pixelwise) ground truth image.
All boundary pixels are excluded from the evaluation (see also
\figRef{fig:beachwood}).
Please note that all the measures, \eg false positives and false negatives, are
computed using all the pixels in the tampered images only. Note also, that it
is challenging to make the pixelwise comparison of keypoint- and block-based
methods completely fair: as keypoint-based matches are by nature very sparse,
we are not able to directly relate their pixel-wise performance to block-based
methods. Thus, we report the \emph{postprocessed} keypoint matches (as
described in \secRef{sec:pipeline}).

\subsubsection{Plain Copy-Move}
\tabRef{tab:pixel_nul} shows the baseline results for the dataset at
pixel level. Similarly to the experiment at image level, all regions
have been copied and pasted without additional disturbances. Note
that we calibrated the thresholds for all methods in a way that yields very
competitive (comparable) detection performances.

\begin{table}[t]
\centering
	\ifCLASSOPTIONdraftcls
		\renewcommand{\arraystretch}{0.7}
	\fi
	\caption{Results for plain copy-move at pixel level in percent}
	\label{tab:pixel_nul}
	\ifCLASSOPTIONdraftcls
		\begin{tabular}{|l|r|r|r||l|r|r|r|}
			\hline
			\input{tables/nul_default_new_per_pixel_individual_draft.tex}
	\else
		\begin{tabular}{|l|r|r|r|}
			\hline
			\input{tables/nul_default_new_per_pixel_individual.tex}
	\fi
		\hline
	\end{tabular}
\end{table}

\subsubsection{Robustness to Gaussian noise} 
We used the same experimental setup as in the per-image evaluation, \ie
zero-mean Gaussian noise with standard deviations between $0.02$ and $0.1$ has
been added to the copied region. The goal is to simulate further postprocessing
of the copy.
At pixel level, this experiment shows a clearer picture of the performance of
the various algorithms (see \figRef{fig:noise_results}).
\dct, \sift and \surf provide the best recall. \dct also outperforms all other
methods with respect to precision. The performance of the remaining features
splits in two groups: \cir, \blur, \bravo, \svd and \hu are very sensitive to
noise, while \pca, \zernike, \kpca and \dwt deteriorate slightly more gracefully.

\begin{figure}[t]
\centering
	\ifCLASSOPTIONdraftcls
		\subfigure[Gaussian white noise]{\includegraphics[width=0.45\linewidth]{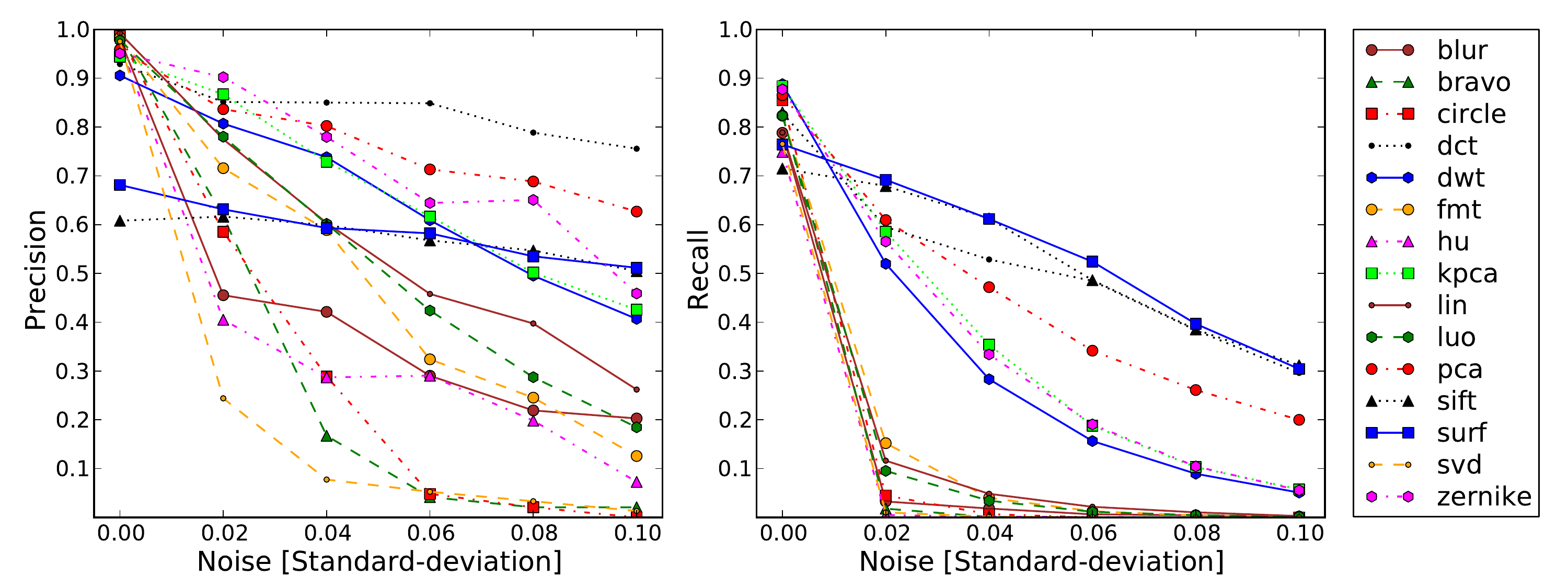}\label{fig:noise_results}}
		\subfigure[JPEG compression]{\includegraphics[width=0.45\linewidth]{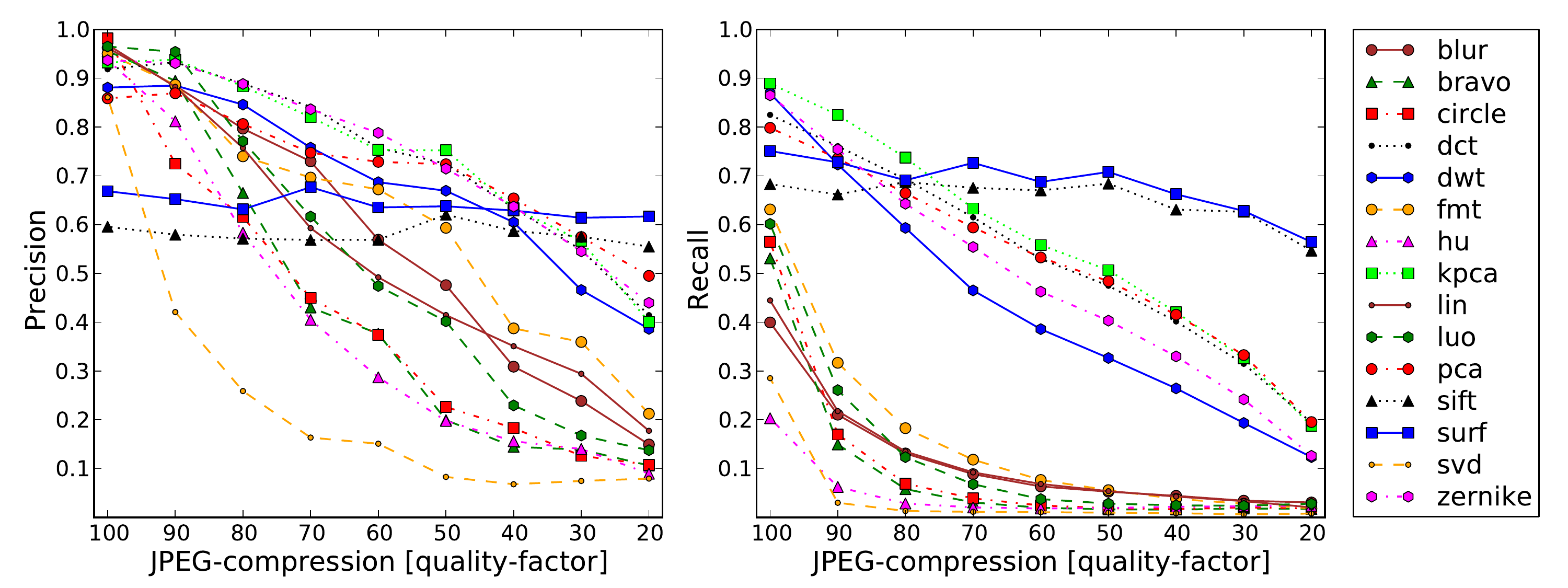}\label{fig:jpeg_results}}

		\subfigure[Rescaled copies]{\includegraphics[width=0.45\linewidth]{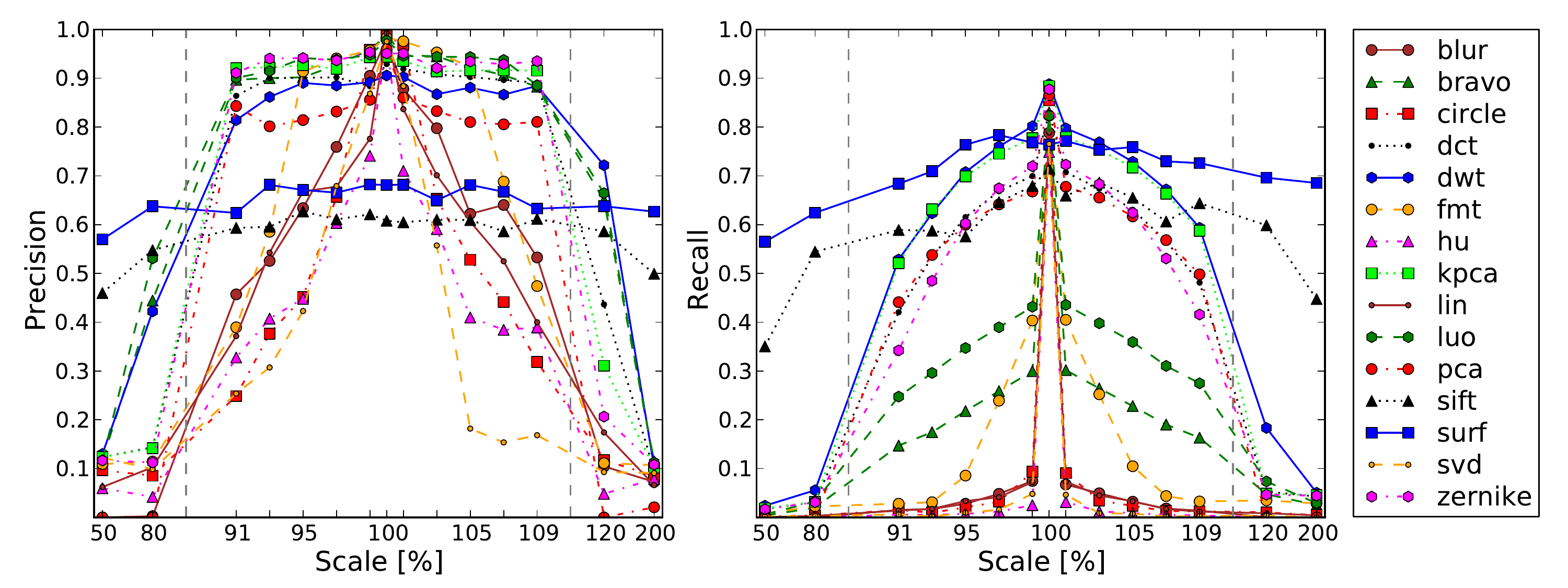}\label{fig:scale_results}}
		\subfigure[Rotated copies]{\includegraphics[width=0.45\linewidth]{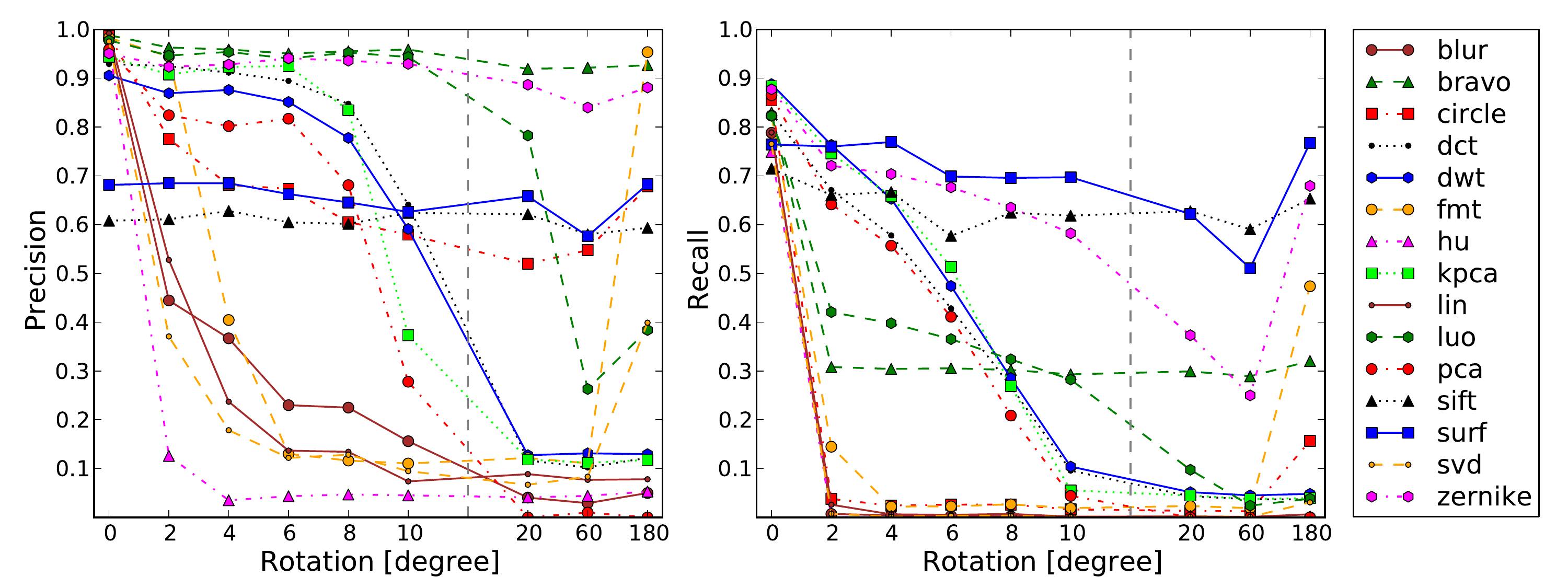}\label{fig:rot_results}}

	\else
		\subfigure[Gaussian white noise]{\includegraphics[width=\linewidth]{plots/1_2/plot_lnoise_lnoise_l_precision_recall_}\label{fig:noise_results}}

		\subfigure[JPEG compression]{\includegraphics[width=\linewidth]{plots/1_2/plot_jpeg_jpeg_l_precision_recall_}\label{fig:jpeg_results}}

		\subfigure[Rescaled copies]{\includegraphics[width=\linewidth]{plots/1_2/plot_scale_scale_l_precision_recall_}\label{fig:scale_results}}

		\subfigure[Rotated copies]{\includegraphics[width=\linewidth]{plots/1_2/plot_rot_rot_l_precision_recall_}\label{fig:rot_results}}
	\fi
	\caption{Experimental results at pixel level (see text for details).}
	\label{fig:pixel_level_results}
\end{figure}

\subsubsection{Robustness to JPEG compression artifacts}
\label{subsubsec:exp_pixellevel_jpeg}
We again used the same experimental setup as in the per-image evaluation, \ie added
JPEG compression between quality levels $100$ and $20$.
\figRef{fig:jpeg_results} shows the resilience at pixel level against these
compression artifacts.
The feature sets forms two clusters: one that is strongly affected by JPEG
compression, and one that is relatively resilient to it.
The resilient methods are \sift, \surf, \kpca, \dct, \pca, \zernike, and
slightly worse, \dwt. The robustness of \dct was foreseeable, as \dct features
are computed from the discrete cosine transform, which is also the basis of the
JPEG algorithm.

\subsubsection{Scale-invariance} 
\label{subsubsec:exp_pixellevel_scale}
The experimental setup is the same as on the per-image level analysis. 
The copy is scaled between $91\%$ and $109\%$ of its original size in
increments of $2\%$. Additionally, we evaluated more extreme scaling
parameters, namely $50\%$, $80\%$, $120\%$ and $200\%$.
As \figRef{fig:scale_results} shows, $7$ feature sets exhibit scaling
invariance for small amounts of scaling. However, the quality strongly varies.
The best performers within these $7$ feature sets are \dwt, \kpca, \zernike,
\pca and \dct. However, for scaling differences of more than $9\%$, the
keypoint-based features \sift and \surf are the only features sets that
preserve acceptable precision and recall.

\subsubsection{Rotation-invariance} 
\label{subsubsec:exp_pixellevel_rot}
We evaluated cases where the copied region has been rotated between $2^\circ$ and
$10^\circ$ (in steps of $2^{\circ}$), as well as for $20^\circ$, $60^\circ$ and
$180^\circ$. We assumed this to be a reasonable range for practical tampering
scenarios. \figRef{fig:rot_results} shows the results. Most feature sets show
only weak invariance to rotation. Similar to the scaling scenario, \sift and
\surf exhibit the most stable recall. From the block-based methods, \zernike,
and also \bravo and \luo are the best features for larger amounts of rotation.
Note that for the special case of $180^\circ$, also \fmt and and \cir perform
very well.

\subsubsection{Robustness to Combined Transformation}\label{subsubsec:comb_transform_pixel}
\begin{figure}[t]
\centering
	\ifCLASSOPTIONdraftcls
		\includegraphics[width=0.45\linewidth]{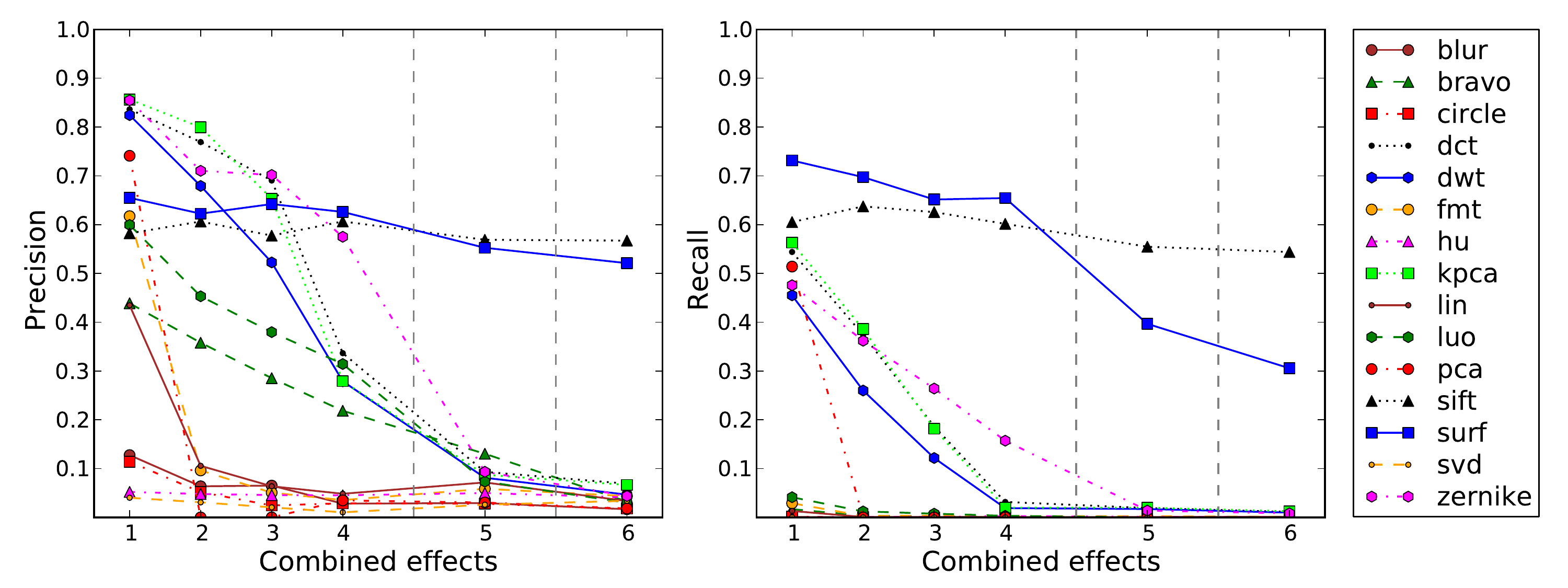}
	\else
		\includegraphics[width=\linewidth]{plots/plot_cmb_easy1_cmb_easy1_l_precision_recall_}
	\fi
	\caption{Results for increasingly difficult combined transformations at pixel level.}
	\label{fig:pixel_cmb}
\end{figure}

Besides the targeted study of single variations in the copied snippet, we
conducted an experiment for evaluating the joint influence of multiple effects.
Thus, we analyzed images where the copied part was increasingly scaled, rotated
and JPEG-compressed. 
The setup was the same as on image level. Thus, scaling varied between $101\%$
and $107\%$ in steps of $2\%$, rotation between $2^\circ$ and $8^\circ$ in
steps of $2^\circ$, and the JPEG quality ranges from $80$ to $65$ in steps of
$5$. Setup $5$ and $6$ have different parameters, namely
a rotation of $20^\circ$ and $60^\circ$, a scaling of $120\%$ and $140\%$,
and the quality of JPEG compression was set to $60$ and $50$, respectively.
The performance results are shown in \figRef{fig:pixel_cmb}. In these 
difficult scenarios, \surf and \sift perform considerably well, followed by
\zernike, \dct, \kpca and \dwt. Note that it is infeasible to cover the whole
joint parameter space experimentally. However, we take this experiment as an
indicator, that the results of the prior experiments approximately transfer to
cases where these transformations jointly occur.

\subsection{Detection of Multiple Copies} 
In recent work, \eg\cite{Amerini11:SFM}, the detection of multiple copies of
the same region has been addressed. While at image level it typically suffices to recognize whether something has been copied, multiple-copies detection requires that all copied regions be identified. 
For such an evaluation, we modified the feature matching as follows.
Instead of considering the nearest neighbor, we implemented the g2NN strategy
by~\cite{Amerini11:SFM}. This method considers not only the single
best-matching feature, but the $n$ best-matching features for detecting
multiple copies.
Although our dataset contains a few cases of single-source multiple-copies, we
created additional synthetic examples.
To achieve this, we
randomly chose for each of the $48$ images a block of $64\times 64$ pixels and
copied it $5$ times.

\begin{table}[t]
\centering
	\ifCLASSOPTIONdraftcls
		\renewcommand{\arraystretch}{0.7}
	\fi
	\caption{Results for multiple copies at pixel level. Left:
	single best match (as conducted in the remainder of the
	paper). Right: g2NN strategy by
	\etal{Amerini}~\cite{Amerini11:SFM}. All results are in percent.}
	\label{tab:multi_paste}
	\begin{tabular}{|l|r|r|r||r|r|r|}
		\hline
	\input{tables/multi_paste_default_vs_multi_corres_per_pixel_individual.tex}
		\hline
	\end{tabular}
\end{table}

\tabRef{tab:multi_paste} shows the results for this scenario at pixel level.
On the left side, we used the same postprocessing method as in the remainder of
the paper, \ie we matched the single nearest neighbor. On the right side, we
present the results using the g2NN strategy. For many feature sets, precision
slightly decreases using g2NN. This is not surprising, as many more
combinations of matched regions are now possible, thus also increasing the
chance for false matches. Still, some methods alike \blur, \bravo, etc. are
relatively unaffected by this change in postprocessing, while
others experience a remarkable performance boost. In particular, \dct, \dwt,
\kpca, \pca, \zernike, \ie the strong feature sets in the prior experiments,
can significantly benefit from the improved matching opportunities of g2NN. As
we discuss later (see \secRef{sec:discussion}), we see this as yet another
indicator that these features exhibit very good discriminating properties.
The performance of \sift and \surf drops considerably, mainly due
to the fact that the random selection of small blocks often yields regions with
very few matched keypoints. Although not explicitly evaluated, we expect that
selecting copy regions with high entropy (instead of a random selection), would
considerably improve the detection rates of \sift and \surf.

\subsection{Downsampling: Computational Complexity versus Detection Performance}

The evaluated methods vary greatly in their demand for resources. One
widely-used workaround is to rescale images
to a size that is computationally efficient. However, this raises the issue of
performance degradation. In order to analyze the effects of downsampling, we
scaled down all $48$ noise-free, one-to-one (\ie without further postprocessing)
forgeries from our database in steps of $10\%$ of
the original image dimensions. 
Note that the detection parameters, as in the whole section,
were globally fixed to avoid overfitting. In this sense, the results in this
section can be seen as a conservative bound on the theoretically best
performance.
We observe that the performance of all features considerably drops.  When
downsampling by a factor of exactly $0.5$, results are
still better than for other scaling amounts
(see \figRef{fig:scale_recomp_results_a} for more
details).
This shows that global resampling
has considerable impact on the results. Some feature sets are almost rendered
unusable.
\kpca, \zernike, \dwt, \pca,
\dct, \luo, \fmt and \bravo perform relatively well. \sift and \surf{}
exhibit slightly worse precision, which might also be due to a
suboptimal choice of $\tau_3$ with respect to the reduced number of keypoints
in the downscaled images.
However, the recall rates are competitive with the
block-based methods. For completeness, we repeated the analysis of subsections
\ref{subsubsec:exp_pixellevel_jpeg},
\ref{subsubsec:exp_pixellevel_scale} and \ref{subsubsec:exp_pixellevel_rot} on
a downsampled ($50\%$) version of the tampered images.
The results are presented in
\figRef{fig:scale_recomp_results_b} to \figRef{fig:scale_recomp_results_d}.
The general shape of the performance curves
is the same as in the previous sections. Note that the performance of recall is more strongly affected by downscaling than precision.

\begin{figure}[t]
\centering
	\ifCLASSOPTIONdraftcls
		\subfigure[Plain copy]{\includegraphics[width=0.45\linewidth]{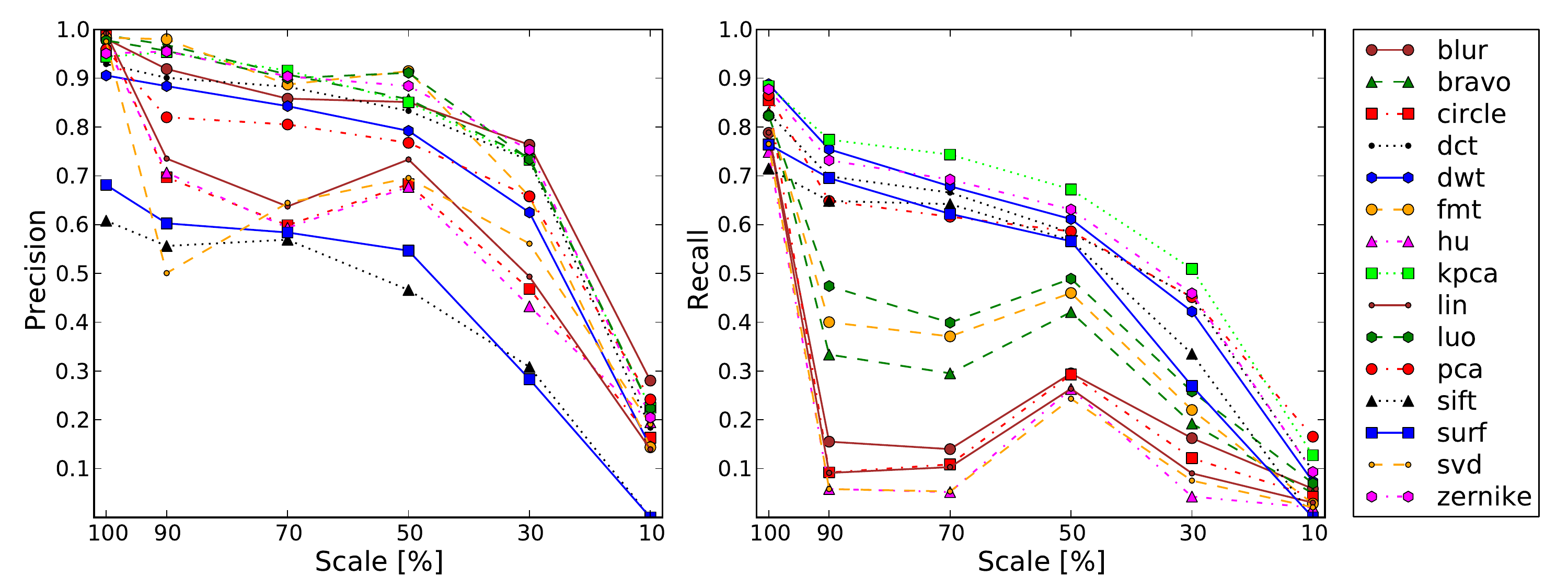}\label{fig:scale_recomp_results_a}}
		\subfigure[JPEG]{\includegraphics[width=0.45\linewidth]{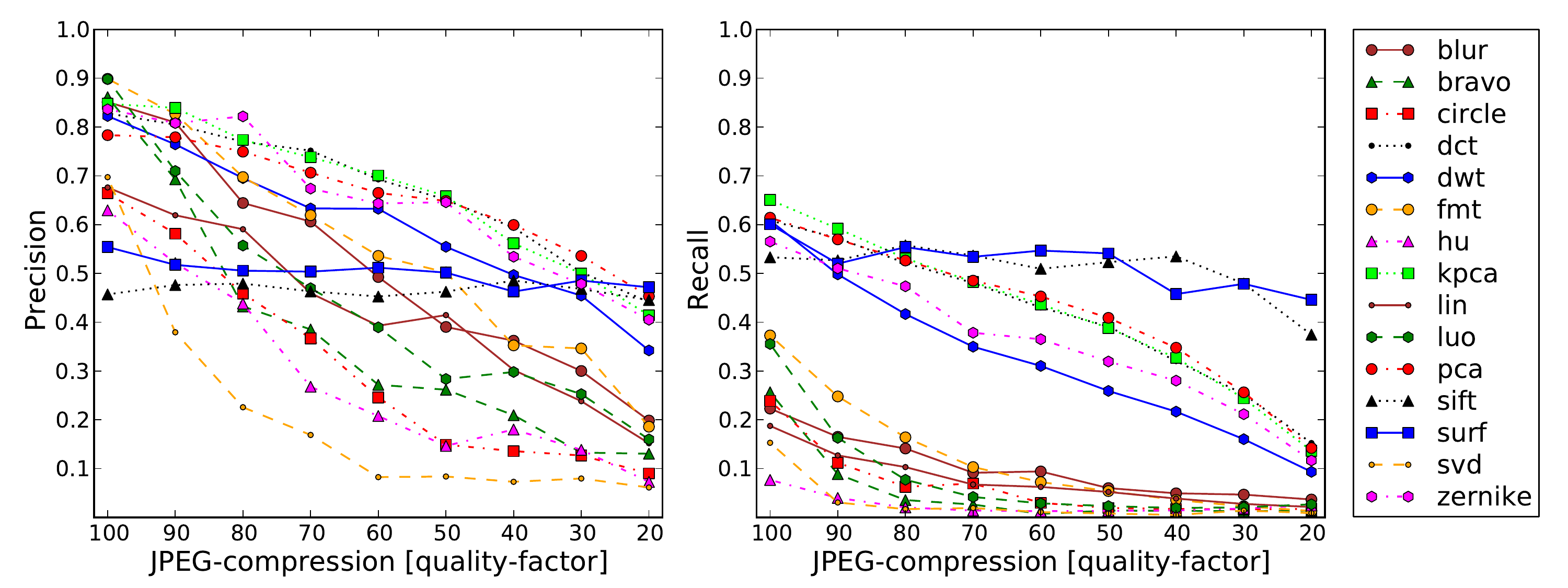}\label{fig:scale_recomp_results_b}}
		\subfigure[Rotation]{\includegraphics[width=0.45\linewidth]{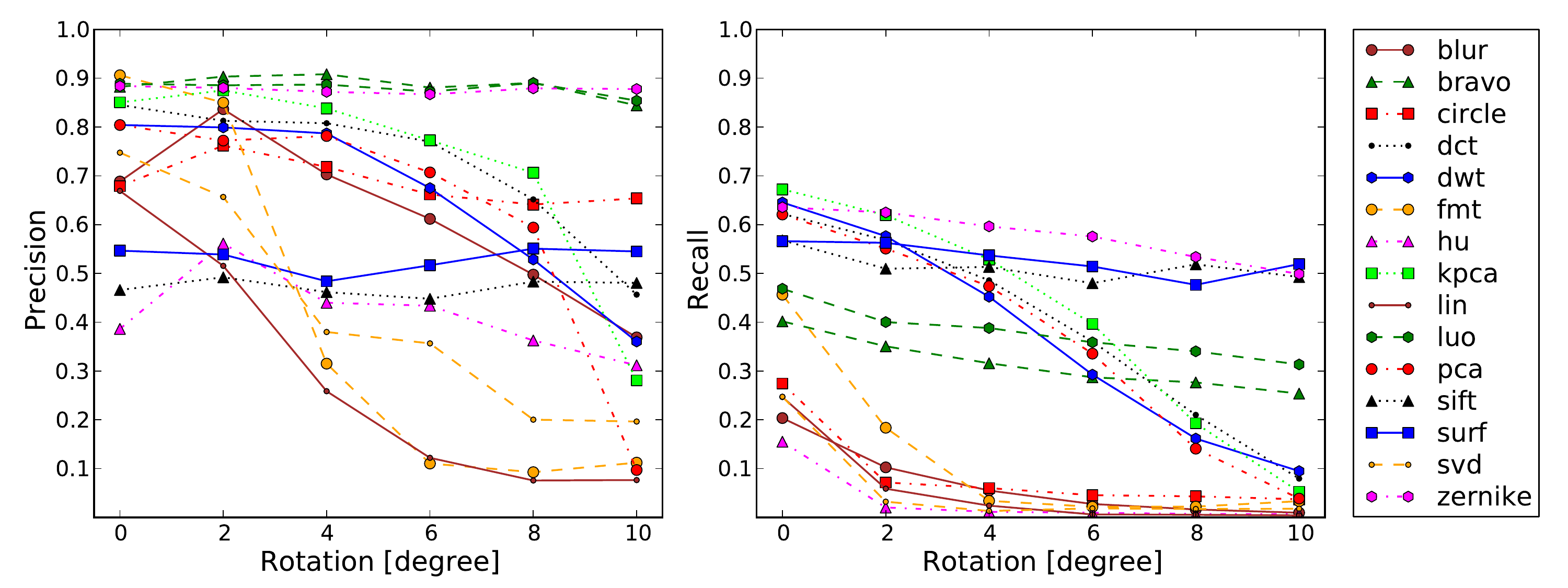}\label{fig:scale_recomp_results_c}}
		\subfigure[Scaling]{\includegraphics[width=0.45\linewidth]{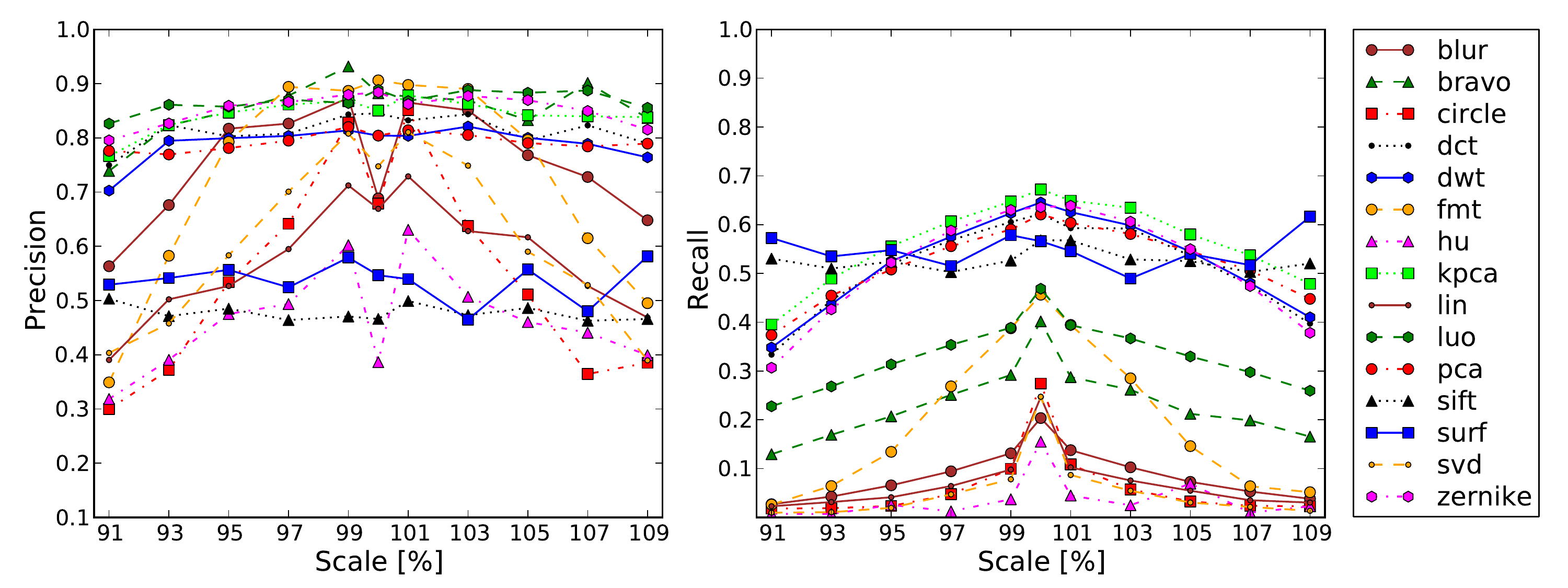}\label{fig:scale_recomp_results_d}}
		\caption{Results for downsampled images: recall is significantly
		worse. From left to right, top to bottom: Plain copy, JPEG, Rotation, Scaling.}
	\else
		\subfigure[Plain copy]{\includegraphics[width=\linewidth]{plots/1_2/plot_scale_down_scale_down_l_precision_recall_}\label{fig:scale_recomp_results_a}}
		\subfigure[JPEG]{\includegraphics[width=\linewidth]{plots/1_2/plot_jpeg_sd_jpeg_sd_l_precision_recall_}\label{fig:scale_recomp_results_b}}
		\subfigure[Rotation]{\includegraphics[width=\linewidth]{plots/1_2/plot_rot_sd_rot_sd_l_precision_recall_}\label{fig:scale_recomp_results_c}}
		\subfigure[Scaling]{\includegraphics[width=\linewidth]{plots/1_2/plot_scale_sd_scale_sd_l_precision_recall_}\label{fig:scale_recomp_results_d}}
		\caption{Results for downsampled images: recall is significantly
		worse. From top to bottom: Plain copy, JPEG, Rotation, Scaling.}
	\fi
	\label{fig:scale_recomp_results}
\end{figure}

\subsection{Resource Requirements}
For block-based methods, the feature-size (see \tabRef{tab:methods}) can lead
to very high memory use. For large images, this can reach several
gigabytes. \tabRef{tab:example_time} (right column) shows the per-method
minimum amount of memory in MB on our largest images, obtained from multiplying
the length of the feature vector with the number of extracted blocks. In our
implementation, the effective memory requirements were more than a factor of
$2$ higher, leading to peak memory usage for \dct and \dwt of more than 20GB.
Note however, that the feature size of \dct and \dwt
depends on the block size. For better comparability, we kept the block size
fixed for all methods. Within a practical setup, the block size of these
feature sets can be reduced. Alternatively, the feature sets can be cropped to
the most significant entries. Some groups explicitly proposed this 
(\eg~\cite{Popescu04:EDFDDIR}, \cite{Bashar10:EDR}). In our experiments, as a
rule of thumb, 8GB of memory sufficed for most feature sets using
\openCV's\footnote{\texttt{http://opencvlibrary.sourceforge.net/}}
implementation for fast approximate nearest neighbor search.

The computation time depends on the complexity of the feature set, and on the
size of the feature vector.
\tabRef{tab:example_time} shows the average running times in seconds over the dataset, split into
feature extraction, matching and postprocessing. Among the block-based methods,
the intensity-based features are very fast to compute.  Conversely, \blur, \dct
and \kpca features are computationally the most costly in our unoptimized
implementation. The generally
good-performing feature
sets \pca, \fmt and \zernike are also relatively computationally demanding.
\begin{table}[t]
\centering
	\ifCLASSOPTIONdraftcls
		\renewcommand{\arraystretch}{0.7}
	\fi
	\caption{Average computation times per image in seconds, and the
	theoretical minimum peak memory requirements in megabytes. Note that this
	is a lower bound, for instance our implementation actually requires more
	than twice of the memory.}
	\label{tab:example_time}
	\input{tables/resource_requirements}
\end{table}

Keypoint-based methods excel in computation time and memory consumption. Their
feature size is relatively large. However, the number of extracted keypoints is
typically an order of magnitude smaller than the number of image blocks. This makes the
whole subsequent processing very lightweight. On average, a result can be
obtained within $10$ minutes, with a remarkably small memory footprint.

\subsection{Qualitative Results}

A more intuitive presentation of the numerical results is provided for
four
selected examples, shown in \figRef{fig:qualitative_results}. On the left,
the extracted contours of the
keypoint-based method \surf are shown. On the right the matches detected by
the block-based \zernike features are depicted. Matched regions are highlighted
as brightly colored areas.
\ifCLASSOPTIONdraftcls
In the top left image, 
the people which formed the top of the ``$5$'' (see \figRef{fig:motivation}) were covered by a region copied
from the right side of the image. Additionally the circle was closed by copying another person.
\surf and
\zernike correctly detected all copied regions. In the top right image,
three passengers were 
copied onto the sea. The image was afterwards compressed with JPEG quality
$70$. \surf yielded one correct match but misses the two other persons.
\zernike marked all passengers correctly. However, it also generated
many false positives in the sky region.
In the bottom left image,
a $20^\circ$ rotation was applied to the copy of the tower. Both
methods accurately detected the copied regions. This observation is easily repeatable,
as long as: a) rotation-invariant descriptors are used, and b) the regions are
sufficiently structured. As with JPEG-compression \zernike produced some false positives above the left tower.
In the bottom right picture, the two stone heads at the edge of the building were copied in the
central part. Each snippet was rotated by $2^\circ$ and scaled by $1\%$. The
entire image was then JPEG compressed at a quality level of $80$. This image is
particularly challenging for
keypoint-based methods, as it contains a number of high-contrast
self-similarities of non-copied regions. \zernike clearly detected the two
copies of the stone heads. \surf also detected these areas, but marked a large
number of the background due to the background symmetries.
\else
In the top image, 
the people which formed the top of the ``$5$'' (see \figRef{fig:motivation}) were covered by a region copied
from the right side of the image. Additionally the circle was closed by copying another person.
\surf and
\zernike correctly detected all copied regions. In the second row,
three passengers were
copied onto the sea. The image was afterwards compressed with JPEG quality
$70$. \surf yielded one correct match but misses the two other persons.
\zernike marked all passengers correctly. However, it also generated
many false positives in the sky region.

In the third image, 
a $20^\circ$ rotation was applied to the copy of the tower. Both
methods accurately detected the copied regions. This observation is easily repeatable
as long as: a) rotation-invariant descriptors are used, and b) the regions are
sufficiently structured. 
As with JPEG-compression \zernike produced some false positives above the left tower.
In the bottom row, the two stone heads at the edge of the building were copied in the
central part. Each snippet was rotated by $2^\circ$ and scaled by $1\%$. The
entire image was then JPEG compressed at a quality level of $80$. This image is
particularly challenging for
keypoint-based methods, as it contains a number of high-contrast
self-similarities of non-copied regions. \zernike clearly detected the two
copies of the stone heads. \surf also detected these areas, but marked a large
number of the background due to the background symmetries.
\fi

\ifCLASSOPTIONdraftcls
	\begin{figure}[!t]
		\centering
			\includegraphics[width=0.24\linewidth]{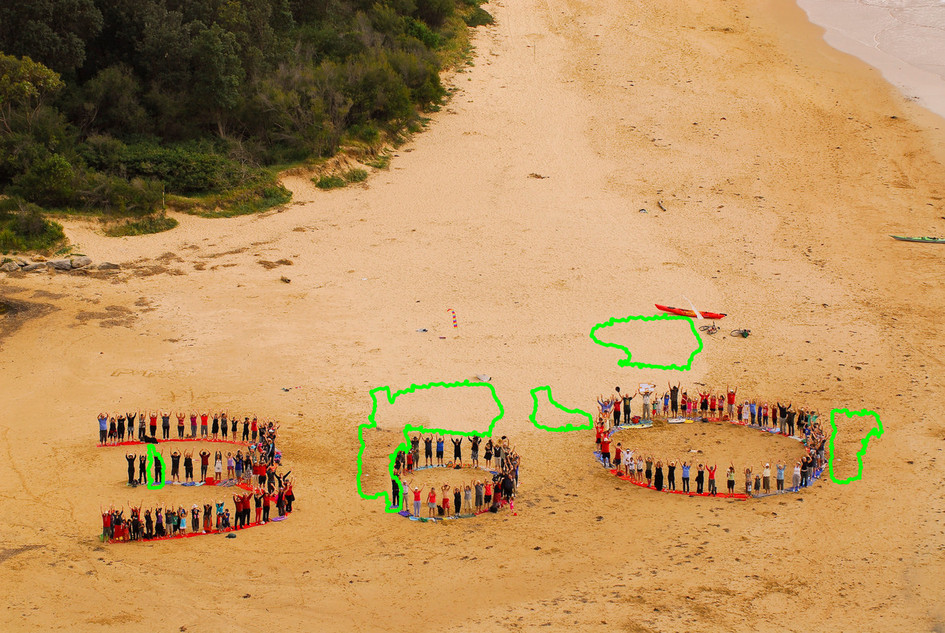}
			\includegraphics[width=0.24\linewidth]{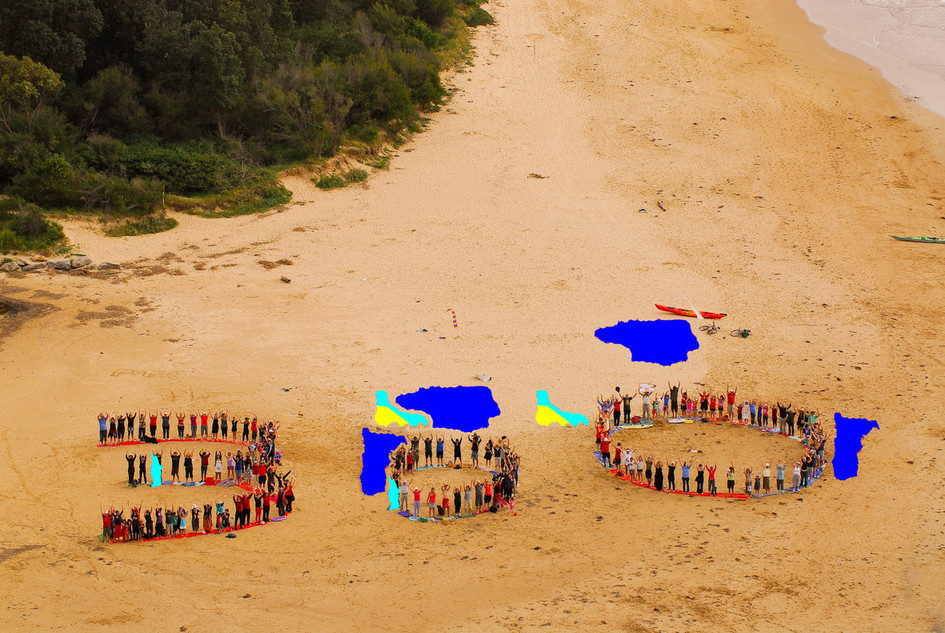}
			\includegraphics[width=0.24\linewidth]{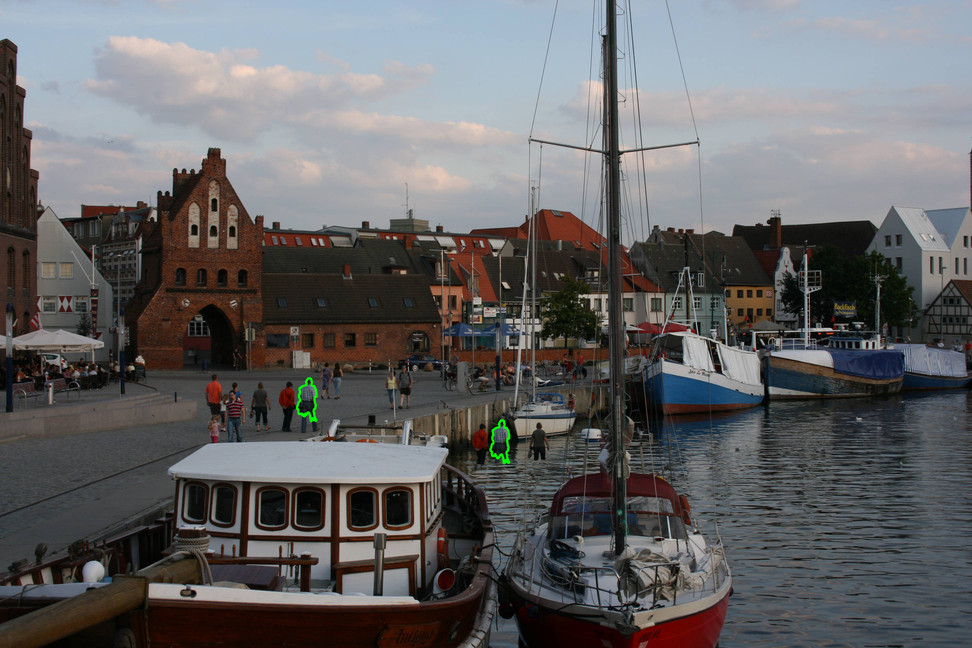}
			\includegraphics[width=0.24\linewidth]{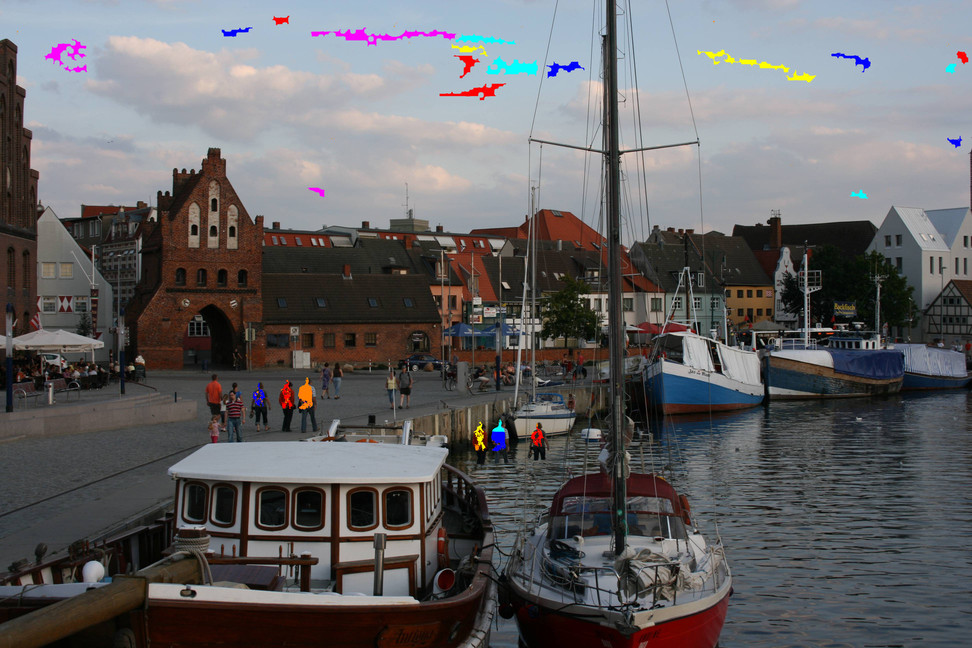}
			
			\vspace{1mm}
			\includegraphics[width=0.24\linewidth]{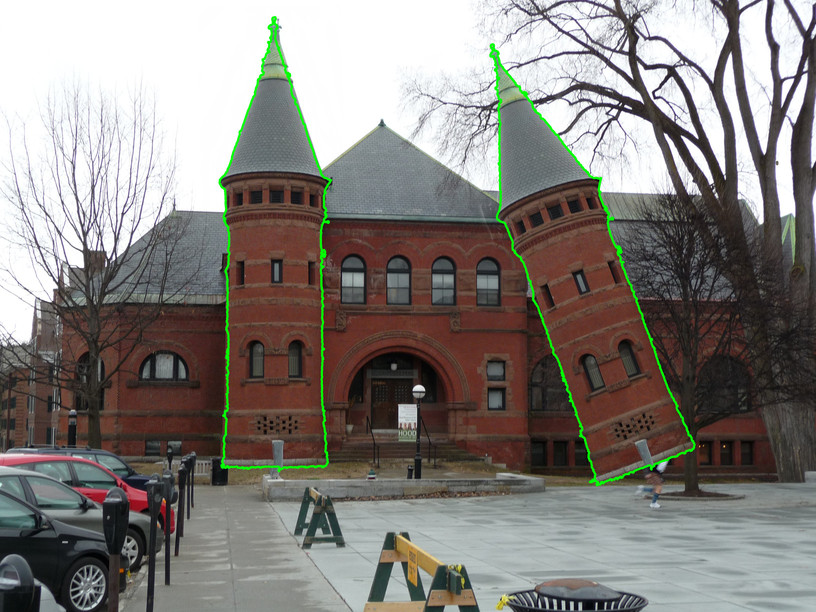}
			\includegraphics[width=0.24\linewidth]{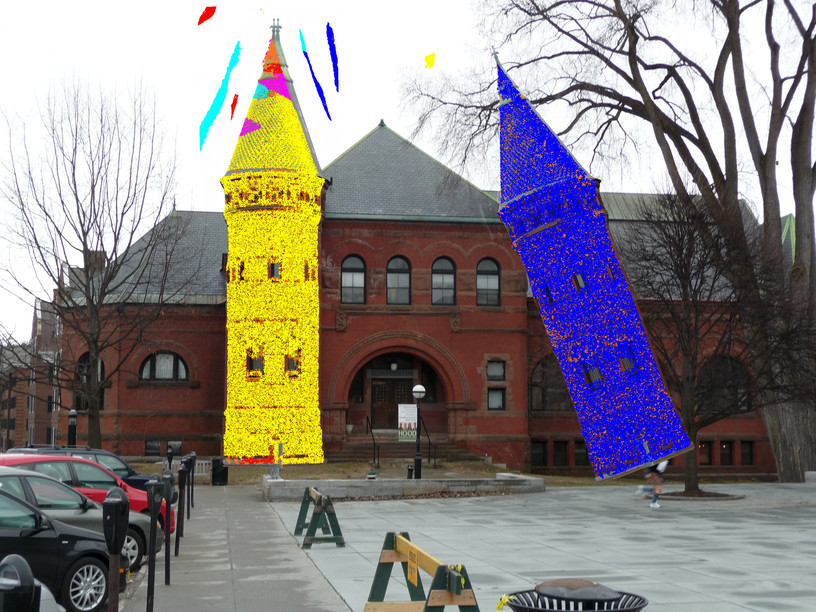}
			\includegraphics[width=0.24\linewidth]{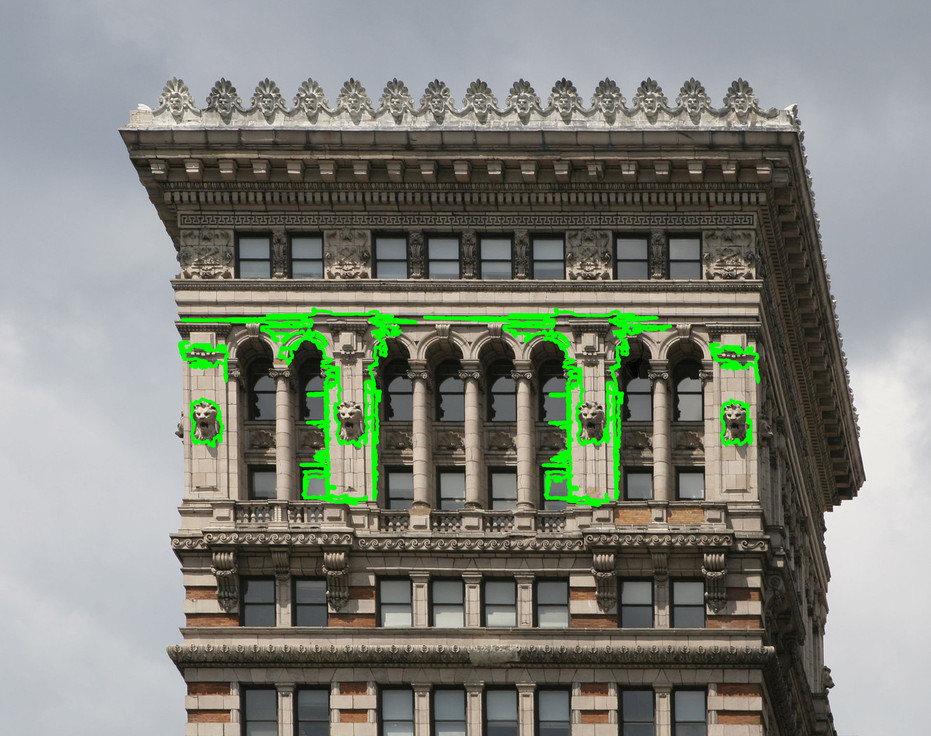}
			\includegraphics[width=0.24\linewidth]{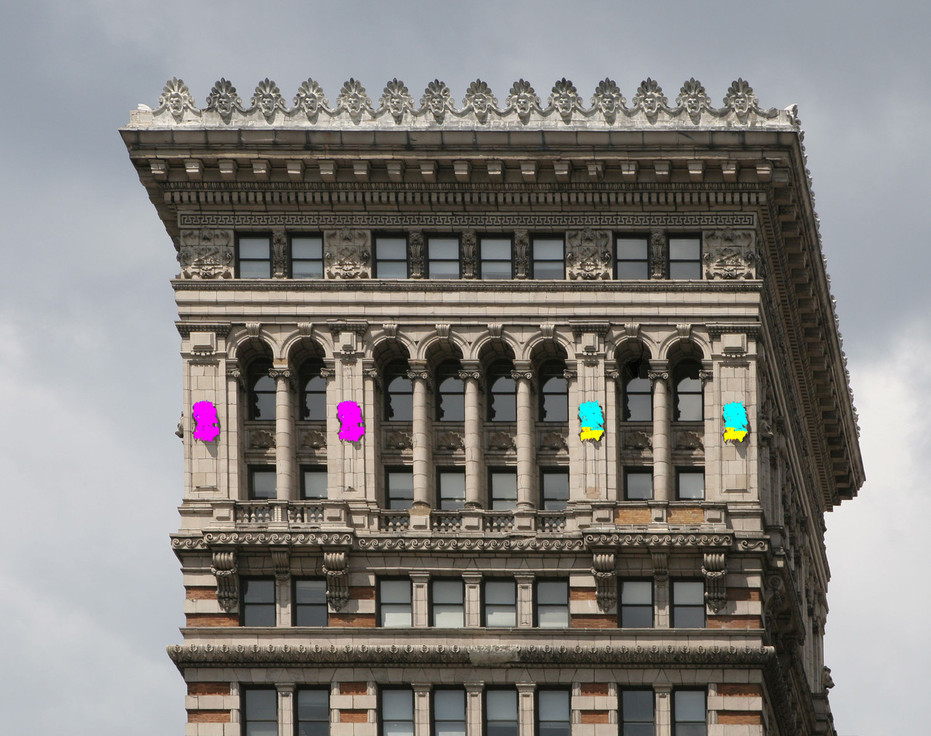}

	\caption{Indicative performance of \surf (left) versus 
	\zernike (right) features. Top left: Plain copy-move example, \surf and
	\zernike detect all copies. Top right: JPEG compression quality of $70$
	hides the copied people from the \surf keypoints. \zernike generates
	many false positives in homogeneous regions. Bottom left: $20^\circ$
	rotation of the snippet is easily handled by both, \surf and \zernike.
	Bottom right: Combined transformations, in addition to highly symmetric
	image content result in \surf producing a large number of false positives.
	The block-based \zernike features correctly detect the copied statues.}

	\label{fig:qualitative_results}
	\end{figure}
\else
	\begin{figure}[!t]
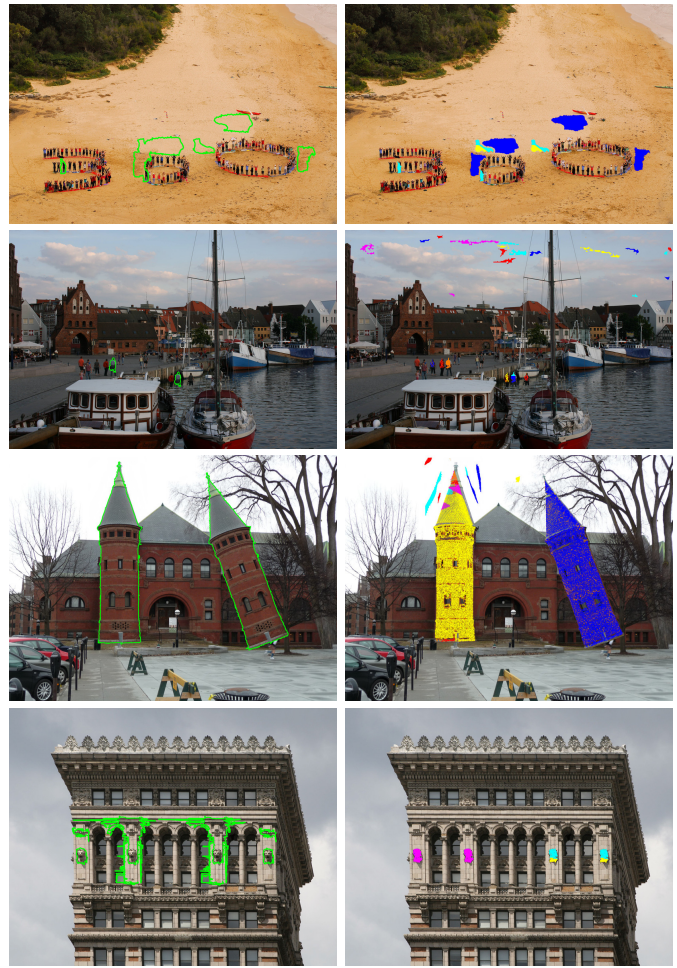

		\centering
			\includegraphics[width=0.49\linewidth]{images/qualitative/nul/surf_threehundred_copy_contour.jpg}
			\includegraphics[width=0.49\linewidth]{images/qualitative/nul/zernike_threehundred_copy_regions_md5ms1000.jpg}

			\vspace{1mm}
			\includegraphics[width=0.49\linewidth]{images/qualitative/jpeg/surf_wading_copy_gj70_contour.jpg}
			\includegraphics[width=0.49\linewidth]{images/qualitative/jpeg/zernike_wading_copy_gj70_regions_ms800.jpg}
			
			\vspace{1mm}
			\includegraphics[width=0.49\linewidth]{images/qualitative/rot/surf_red_tower_copy_r20_contour.jpg}
			\includegraphics[width=0.49\linewidth]{images/qualitative/rot/zernike_red_tower_copy_r20_regions_ms800.jpg}
			
			\vspace{1mm}
			\includegraphics[width=0.49\linewidth]{images/qualitative/cmb_easy1/surf_statue_copy_r2_s1010_gj80_contour.jpg}
			\includegraphics[width=0.49\linewidth]{images/qualitative/cmb_easy1/zernike_statue_copy_r2_s1010_gj80_regions_ms1000.jpg}
	
	\caption{Example qualitative results on the benchmark dataset for \surf
		(left) and \zernike (right). Top row: Plain copy-move, \surf and
			\zernike detect all copies.  Second row: JPEG compression quality
			of $70$ hides the small copied people from the \surf keypoints.
			\zernike tends to strongly overdetecting homogeneous regions. Third
			row: $20^\circ$ rotation poses no problem for both, \surf and
			\zernike. Bottom: combined transformations, in addition to highly
			symmetric image content results in \surf producing a large number
			of false positives. The block-based \zernike features correctly
			detect the copied statues.}

	\label{fig:qualitative_results} \end{figure}
\fi

\subsection{Results by Image Categories}

To investigate performance differences due to different texture of the copied
regions, we computed the performances according to categories.
We subdivided the dataset into the object class categories \emph{living},
\emph{manmade}, \emph{nature} and \emph{mixed}. Although \emph{man-made}
exhibited overall the best performance, the differences between the categories
were relatively small. This finding is in agreement with the intuition that the
descriptors operate on a lower level, such that object types do not lead to
significant performance differences.
In a second series of experiments, we used the artists' categorization of the
snippets into \emph{smooth}, \emph{rough} and \emph{structure} (see
\ref{sec:benchmark}). Overall, these results confirm the intuition that
keypoint-based methods require sufficient entropy in the copied region to
develop their full strength. In the category \emph{rough}, \sift and \surf are
consistently either the best performing features or at least among the best
performers. Conversely, for copied regions from the category \emph{smooth}, the
best block-based methods often outperform \surf and \sift at image or pixel
level.
The category \emph{structure} ranges between these
two extremes.
The
full result tables for both categorization approaches can be found in the supplemental
material\thisVariantSupp{{} in~\cite{SuppMat}}.

\section{Discussion}\label{sec:discussion}

We believe that the obtained insights validate the creation of a new
benchmark dataset. The selection of the evaluation data for the CMFD algorithms
is a
non-trivial task.
To our
knowledge, all existing test sets are somewhat limited in one aspect or
another. For instance, preliminary experiments suggested that image size
strongly influences the detection result of CMFD algorithms. One 
workaround is to scale every input image to a fixed size. However, as we show
in~\figRef{fig:scale_recomp_results}, interpolation itself influences the
detection performance. Furthermore, in the case of downsampling,
the size of the tampered region is also reduced, further inhibiting detection.
Thus, we
conducted all experiments, unless otherwise stated, in the full image
resolution (note, however, that the images themselves had different sizes,
ranging from $800\times 533$ pixels to $3900\times 2613$ pixels). This greatly increased
the risk of undesired matches in feature
space, especially when a feature set exhibits weak discriminative power. Consequently,
differences in the feature performance became more prominent. 

Which CMFD method should be used? During the experiments, we divided the
proposed methods in two groups. \surf and \sift, as keypoint-based methods,
excel in computation time, memory footprint. The advantage in speed is so
significant, that we consider it worth applying these methods always,
independent of the detection goal. \tabRef{tab:img_nul} and subsequent
experiments indicate slightly better result for \surf than for \sift. When a
copied area has undergone large amounts of rotation or scaling, \sift and \surf
are clearly the best choices.
One should be aware that keypoint-based
methods often lack the detail for highly accurate detection
results. When regions with little structure are copied, \eg the cats
image in \figRef{fig:example_test_case}, keypoint-based methods are prone to miss
them. In contrast, highly self-similar image content, as the building in
\figRef{fig:qualitative_results} can provoke false positive matches.

The best-performing block-based features can relatively reliably address these
shortcomings. Experiments on per-image detection indicate that several
block-based features can match the performance of keypoint-based approaches.
We conducted additional experiments to obtain stronger empirical evidence for
the superiority of one block-based method over another. These experiments
measured the pixelwise precision and recall of the block-based approaches.
Experiments on the robustness towards noise and
JPEG artifacts showed similar results. \dct, \pca, \kpca, \zernike and \dwt
outperformed the other methods by a large margin w.r.t. recall. Their precision
also outperformed the other block-based methods for large amounts of noise and
heavy JPEG compression. As shown for example in
\figRef{fig:qualitative_results}, a good precision
leads to a low number of false positive matches. When the copied
region is scaled, the aforementioned five block-based features also perform
well for small amounts of scaling. Note, however, that we were not able to
obtain good results with block-based methods using larger scaling parameters.
For rotated copies, \zernike, \luo and \bravo, \dwt,
\kpca, \dct and \pca can handle small degrees of rotation very well. In
general, for detecting rotated copies, \zernike performed remarkably well. 

In a more practical scenario, running a block-based CMFD algorithm on full-sized images can
easily exceed the available resources. Thus, we examined, how block-based
algorithms perform when the examined image is downsampled by the investigator to
save computation time. Not surprisingly, the overall performance drops.
However, the best performing feature sets remain relatively stable, and confirm
the previous results only at a lower performance level.

In all the previous discussion, we tailored our pipeline for the
detection of a single one-to-one correspondence between source region and
copied region. However, we also evaluated, at a smaller scale, the detection of
multiple copies of the same region. We adapted the matching and filtering steps
to use g2NN, as proposed by \etal{Amerini}~\cite{Amerini11:SFM}, so that 
the $n$ best-matching features were considered.
Interestingly, the already good features \dct, \dwt, \kpca, \pca and \zernike profited
the most from the improved postprocessing. This re-emphasizes the observation
that these feature sets are best at
capturing the relevant information for CMFD. With the improved
postprocessing by \etal{Amerini}, the advantages of these features can be fully
exploited.

In a practical setup, one should consider a two-component CMFD system. One
component involves a keypoint-based system, due to its remarkable computational
efficiency, small memory footprint and very constant performance. This allows
for instance the screening of large image databases.
The second component should be a block-based method, for close and highly
reliable examination of an image. In particular when the added noise or
transformations to the copied area are small, block-based methods are
considerably more accurate. We consider \zernike features as a good choice for
this component. For a block-based method, its memory and runtime requirements
are low, compared to the detection performance.

Note, however, that a final recommendation has of course to be made based upon
the precise detection scenario. We assume that the provided performance
measurements, together with the publicly available dataset, can greatly support
practitioners and researchers to hand-tailor a CMFD pipeline to the task at
hand. For instance, one could imagine a fusion-based system, consisting of the
best features from each category. Alternatively, other forensic tools, like
resampling detection or detection of double JPEG compression can compensate the
current shortcomings of existing CMFD approaches (see, \eg~\cite{Redi11:DIF}).
For instance, under moderate degrees of JPEG compression, rescaled or rotated
copies reliably exhibit traces of resampling. Thus, rotation- and scaling
invariant CMFD can be avoided in such cases. Also, if a region is copied within
a JPEG image, there is a good opportunity that artifacts from double
JPEG-compression can be detected. As an added benefit, these methods are
computationally much more efficient than the average CMFD algorithm. Thus, for
future work, it will be interesting to see how a joint forensic toolbox
performs on manipulated images.

\section{Conclusion}\label{sec:conclusion}

We evaluated the performance of different widely-used features for copy-move forgery detection.
In order to conduct a thorough analysis, we created a challenging evaluation
framework, consisting of: \thisVariantDiff{a) $48$ realistically sized base images, containing b) $87$
copied snippets and c)} a software to replay realistically looking copy-move
forgeries in a controlled environment. We examined various
features within a common processing pipeline. The evaluation is conducted in
two steps. First, on a per-image basis, where only a tampered/original
classification has been done.  In a second step, we performed a more detailed
analysis on a per-pixel basis.

Our results show, that a keypoint-based method, \eg based on SIFT features,
can be very efficiently executed. Its main advantage is the remarkably low
computational load, combined with good performance. Keypoint-based
methods, however, are sensitive to low-contrast
regions and repetitive image content. Here,
block-based methods can clearly improve the detection results.
In a number of experiments, five block-based feature sets stood out, namely
\dct, \dwt, \kpca, \pca and \zernike. Among these, we recommend the use of \zernike,
mainly due to its relatively small memory footprint.

We also quantified the performance loss when
the copy-move forgery detection is not conducted on the original image sizes.
This is an important aspect, as the resource requirements for block-based
CMFD methods on large images \thisVariantArch{(\ie $3000\times 2000$ pixels and more)}{(\eg $3000\times 2000$ pixels)} become
non-trivial.

We believe that the presented results can support the community, particularly
in the development of novel crossover-methods, which combine the advantages of
separate features in a joint super-detector. We also hope that our
insights help forensics practitioners with concrete implementation decisions.

\bibliographystyle{IEEEtran}

\ifCLASSOPTIONcaptionsoff
  \newpage
\fi

\twocolumn[%
\begin{@twocolumnfalse}%
\vskip0.2em{\Huge\centering%
Supplementary Material to ``An Evaluation of Popular Copy-Move Forgery Detection Approaches''\par}\vskip1.0em\par%
{\lineskip.5em\sublargesize\centering Vincent Christlein, Christian Riess, Johannes Jordan, Corinna Riess and Elli Angelopoulou\par
           \par\hfill}\relax%
\end{@twocolumnfalse}%
]%

\section{Contents of the supplemental material}

This document contains additional information about selected topics which were
only briefly addressed in the main paper. We add details on the selection of
$\tau_2$, we state results using the \FM score, and finally add details on the
database, including the reference manipulations.

One such subject is the choice of $\tau_2$, \ie the threshold that filters
out connected matched correspondences below a minimum set size.  As stated in
Sec. V-A of the paper, $\tau_2$ is chosen separately for every feature set. We
explain and justify our strategy for selecting $\tau_2$ in
\secRef{sec:threshold_choice} of this document.  In \secRef{sec:f1_plots}, we
present the \FM score for all evaluations in the main paper. One advantage of
the \FM score is that it subsumes precision and recall in a single number.
In \secRef{sec:postproc_keypoints}, we add details on the postprocessing methods for keypoint-based algorithms.
In \secRef{sec:db_categories}, we describe the results on a per-category
evaluation of the proposed database. Finally, in
\secRef{sec:overview_database}, we add details on the description of the
database, and present all motifs including the associated reference
manipulations.

\section{Threshold Determination for the Minimum Number of Correspondences}
\label{sec:threshold_choice}

Assume that we found a cluster of matches which jointly describes the same
(affine) transformation. If the size of the cluster exceeds the threshold
$\tau_2$, the cluster is reported as a copied area. We
selected $\tau_2$ for every feature set individually. This is necessary,
in order not to punish a slightly weaker performing feature set.
Consider, for example, the case where one feature set roughly matches a copied area, but has
also a considerable number of missed matches within this area. If such a
feature set detects a large number of small, disconnected regions, a high value
of $\tau_2$ would ignore the correct detections. At the same time, if a
feature set typically finds dense, connected sets of matched correspondences,
a low value of $\tau_2$ would increase its sensitivity to noise. This
consideration compelled us to determine this threshold $\tau_2$ for every
feature type individually.

\begin{figure}[!t]
	\includegraphics[width=\linewidth]{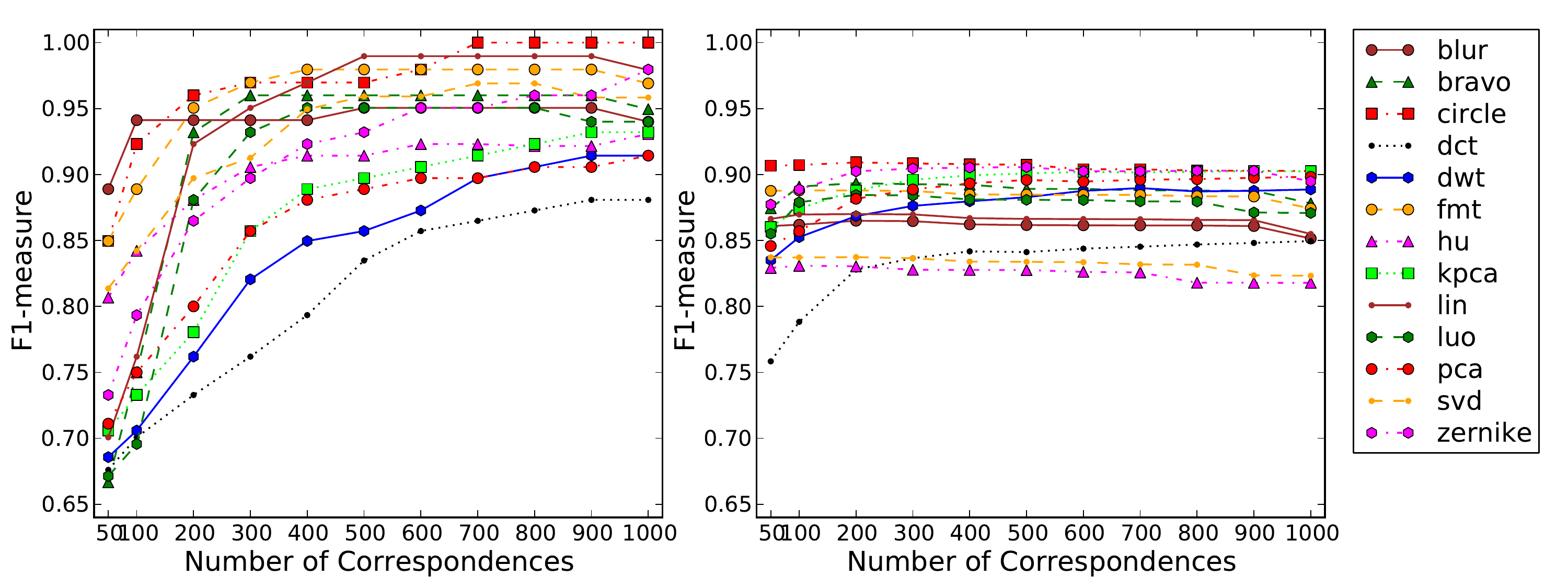}
	\caption{Results for plain copy-move varying the \emph{minimum correspondence threshold} $\tau_2$. Left: per-image evaluation. Right: per-pixel evaluation.}
	\label{fig:nul_thresh}
\end{figure}

We assumed that it is infeasible to conduct a brute-force search over all test
cases for a reasonable value of $\tau_2$. Instead, we searched over
$\tau_2=50,100,\ldots, 1000$ on a very limited postprocessing scenario. As
stated also in the main paper, this scenario included all images from the
dataset. We used original and tampered images without any postprocessing, and
under increasing levels of JPEG compression (conducted with \libJPEG), for
quality levels from $100\%$ down to $70\%$ in steps of $10\%$. The best values of
$\tau_2$ over the entire set of test images were selected for every method. The goal of adding moderate JPEG
compression is to avoid overfitting of $\tau_2$ to ``perfect'' data.
We illustrate this issue with a counter-example (see \figRef{fig:nul_thresh}).
Here, we conducted a
brute-force search for $\tau_2$ only on the clean images, containing only
copied regions without any postprocessing (\ie 1-to-1 copies). If
evaluated at image level (see \figRef{fig:nul_thresh} left), the performance
constantly increases for all methods up to $\tau_2 = 800$, for most of them
until $\tau_2 = 1000$. Evaluating at pixel-level (see
\figRef{fig:nul_thresh} right) yields very similar results. Thus, if we used this
result, we could even set $\tau_2$ globally constant, to a value between $800$
and $1000$. However, such a choice adversely affects most of the feature sets.
To illustrate that, we computed a number of results for $\tau_2 = 1000$.

In \tabRef{tab:corres_thresh}, we report the per-method best thresholds, once
determined from the plain copies (middle column), and once for the mixed clean
and JPEG-compressed images (right column). The average value in the bottom row shows
that the JPEG artifacts decrease the optimum $\tau_2$ value to almost $50\%$.

\begin{table}[!t]
\centering
\caption{Best $\tau_2$, determined using the per-image F1-Measure. In the
middle column, $\tau_2$ is determined on plain 1-to-1 copies, in the right
column on mixed JPEG-compressed images.}
\label{tab:corres_thresh}
	\begin{tabular}{|l|r|r|}
	\hline
	\input{tables/corres_threshold.tex}
	\hline
	\end{tabular}
\end{table}

\begin{table}[!t]
\centering
\caption{Results for plain copy-move, per image. Individual $\tau_2$ (see \tabRef{tab:corres_thresh}, right-column) versus fixed $\tau_2=1000$. }
\label{tab:nul_thresh_img}
	\begin{tabular}{|l|r|r|}
	\hline
	\input{tables/nul_default_per_img_individual_vs_ms1000.tex}
	\hline
	\end{tabular}
\end{table}

\begin{table}[!t]
\centering
\caption{Results for combined transformations ($2^{\circ}$ rotation, upscaling to $101\%$ and $80$ JPEG quality-factor), per image. Individual $\tau_2$ (see \tabRef{tab:corres_thresh}, right-column) versus fixed $\tau_2=1000$. }
\label{tab:cmb_thresh_img}
	\begin{tabular}{|l|r|r|}
	\hline
	\input{tables/cmb_easy1_default_per_img_individual_vs_ms1000.tex}
	\hline
	\end{tabular}
\end{table}

To further demonstrate the sensitivity of $\tau_2$ to JPEG compression, we
performed two further evaluation scenarios (see \tabRef{tab:nul_thresh_img} and
\tabRef{tab:cmb_thresh_img}). \tabRef{tab:nul_thresh_img} reports the \FM
scores on images without postprocessing. \tabRef{tab:cmb_thresh_img} reports
the \FM scores on images with several different types of postprocessing. Both
tables relate
the performance of individually selected thresholds versus fixed thresholds.
In \tabRef{tab:nul_thresh_img}, it can be seen that the individual selection of
$\tau_2$ (middle column) leads to slightly worse performance than a globally
fixed, high $\tau_2=1000$. 
Interestingly, \tabRef{tab:cmb_thresh_img} shows
results for the ``combined transformations'' variant in our evaluation of the
main paper, \ie global JPEG compression at quality level $80$, and slight
modifications on the copied region (rotation of $2^\circ$, upsampling of
$101\%$). Note that for this scenario, we did not conduct dedicated fine-tuning
of $\tau_2$. Nevertheless, the average for method-specific values of $\tau_2$
(\tabRef{tab:cmb_thresh_img} middle) are by a large margin better than the
fixed threshold $\tau_2=1000$ for plain 1-to-1 copies. 
Our further performance analysis was based on the lower thresholds, as one of our goals is
to evaluate the influence of postprocessing artifacts on the features. 

One alternative would be to optimize $\tau_2$ for every experiment (\eg
Gaussian noise, or JPEG compression) separately.  However, we considered such a
complete threshold fine-tuning unrealistic. In practice, it is only expected
that ``some postprocessing'' has been done to better fit a copied region in its
environment. Thus, a successful CMFD algorithm must be able to deal with a
certain amount of uncertainty on the actual type of postprocessing.

\section{Results with \FM score}\label{sec:f1_plots}

All plotted results in the main paper are presented using precision and recall.
However, in some cases (like the determination of $\tau_2$), a single number is
required to more easily compare feature sets. In the tables of the main paper,
we added the \FM score which subsumes precision and recall. For completeness,
we add here the results of the remaining experiments from the main paper with
respect to the \FM score.

\begin{figure}[!t]
\centering
	\subfigure[Gaussian White Noise]{
		\label{fig:lnoise}
		\includegraphics[width=\linewidth]{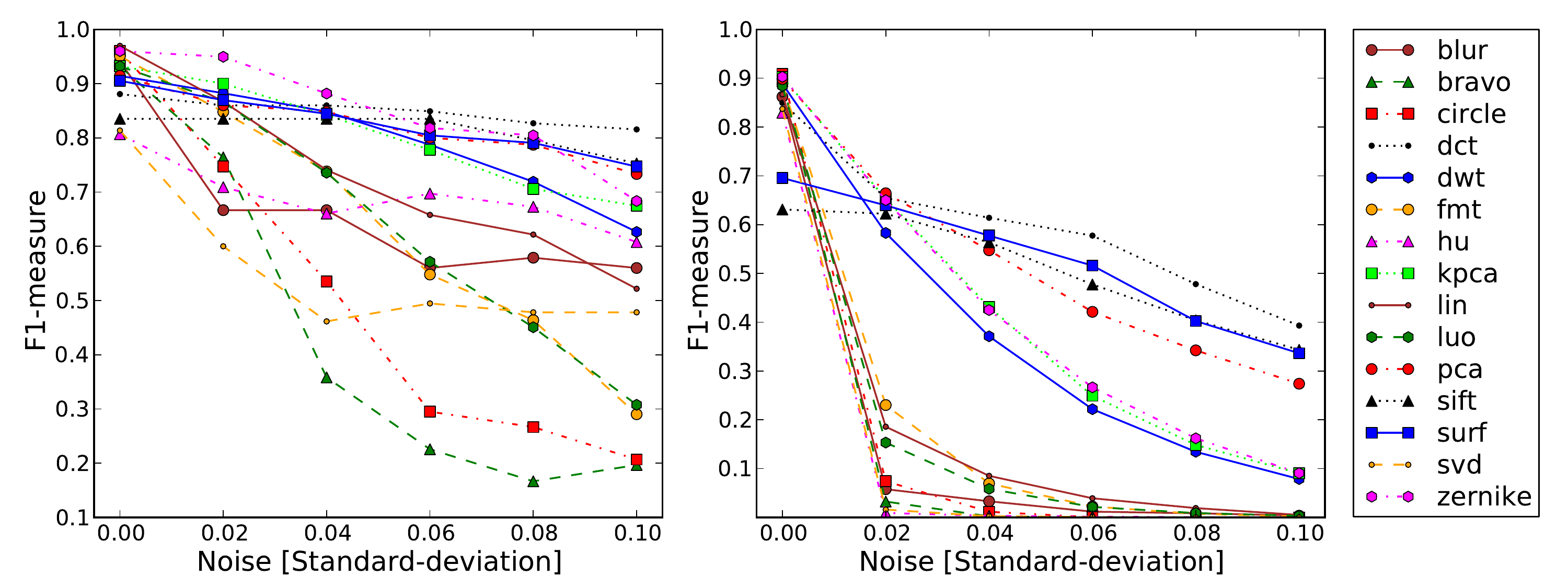}
	}
	\\
	\subfigure[JPEG Compression]{
		\label{fig:jpeg}
		\includegraphics[width=\linewidth]{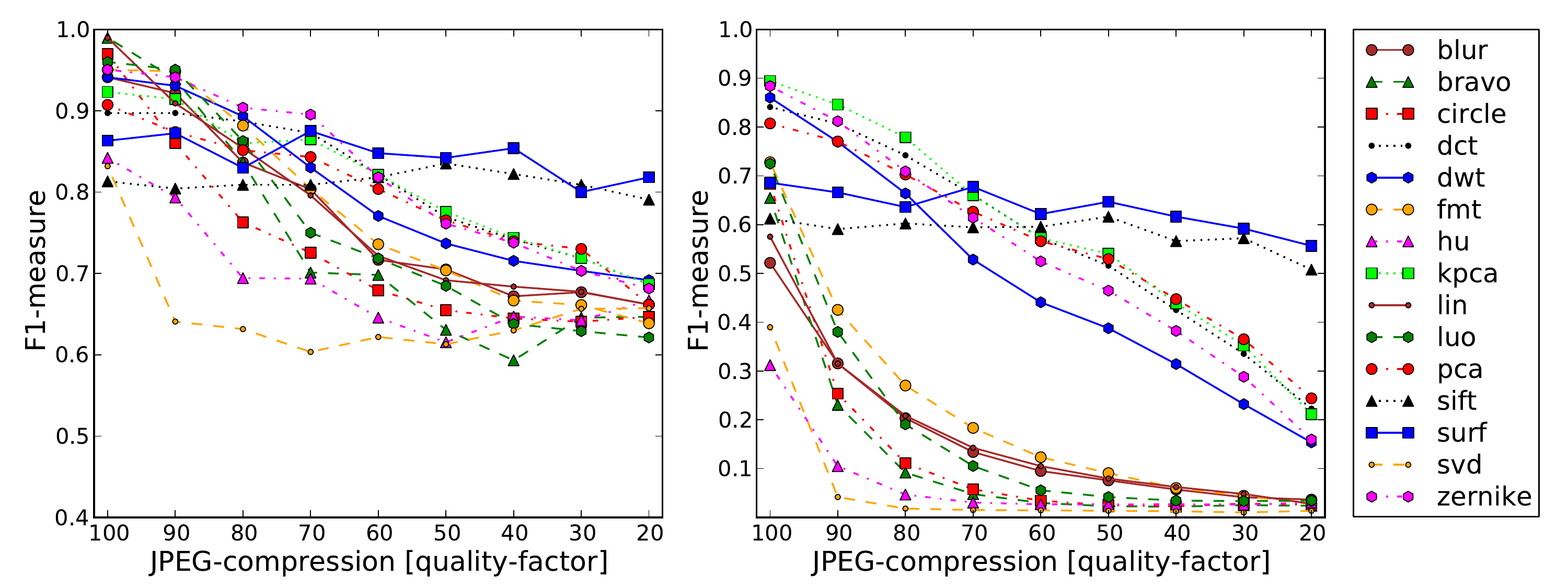}
	}
	\\
	\subfigure[Scaling]{
		\label{fig:scale}
		\includegraphics[width=\linewidth]{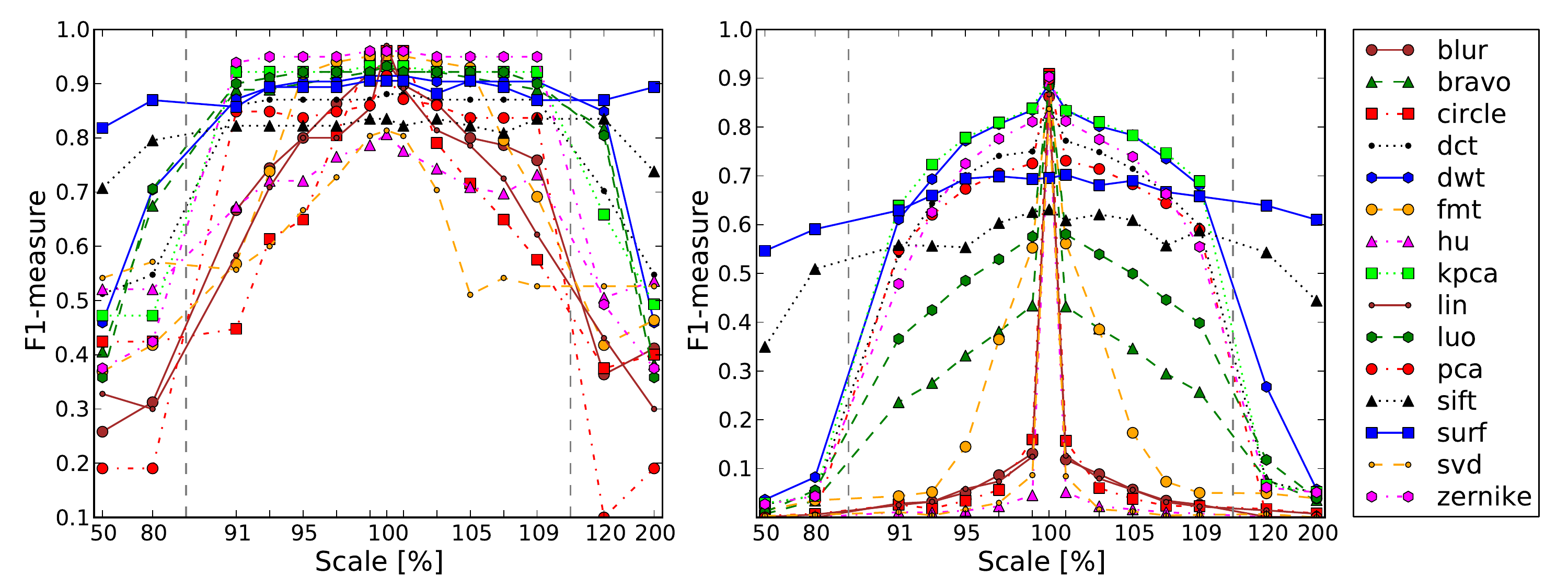}
	}
	\\
	\subfigure[Rotation]{
		\label{fig:rot}
		\includegraphics[width=\linewidth]{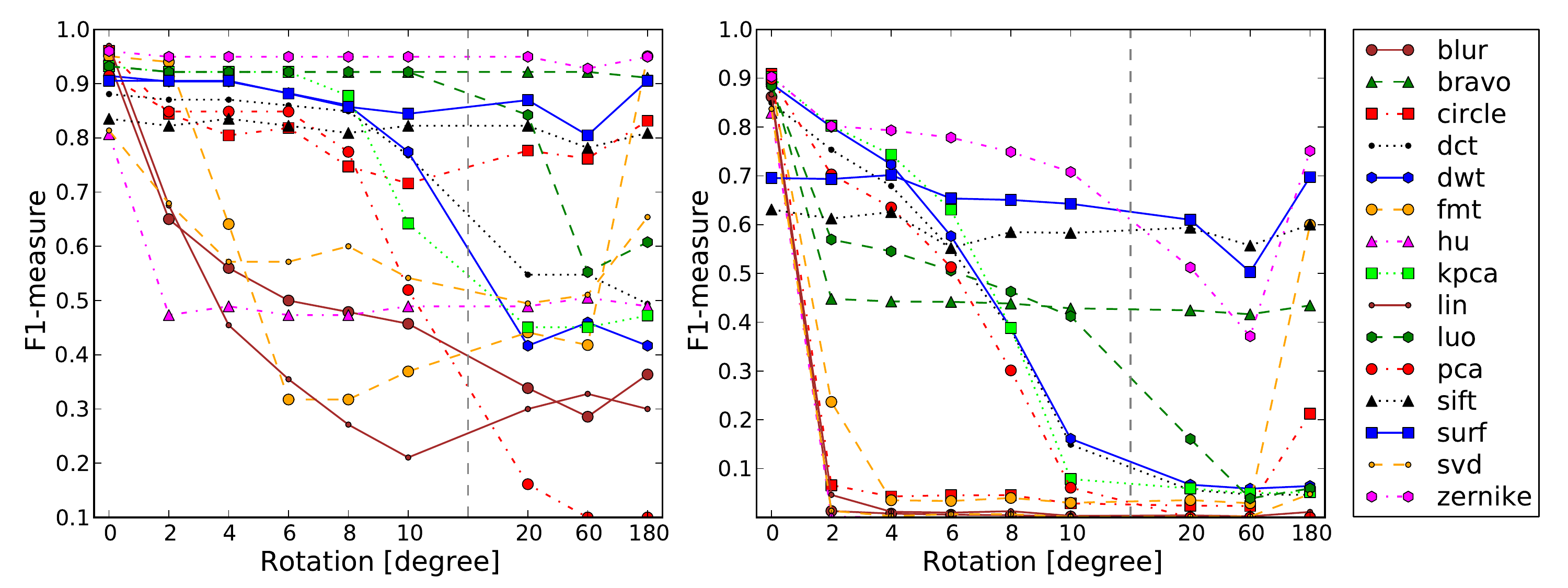}
	}
	\caption{\FM scores for the experiments on the original image sizes
	(corresponds to Fig.~8 and Fig.~9 in the main paper). Left: performance on
	image level, right: performance at pixel level. From top to bottom: Gaussian
	white noise, JPEG compression, rotation, scaling.}
	\label{fig:plots}
\end{figure}

The plots in \figRef{fig:plots} correspond to the results shown in Fig.~8 and
Fig.~9 in the main paper. The left plot shows the results at image level, while
the right side
shows the results for the same experiment at pixel level. The order of the experiments,
and the experimental setup, is the same as in the main paper. Thus, from top to
bottom, results are presented for additive Gaussian noise, varying degrees of
JPEG compression, a limited amount of scaling, and varying degrees of rotation.
We relate the \FM scores to the precision and recall plots from the main paper.
In general, the \FM score captures large dynamics in precision and recall. For
the Gaussian noise, the large spread in the recall mainly shapes both \FM
scores, at image level and pixel level. For the experiment on JPEG compression,
the image level differences in precision and recall are evened out in the \FM
score. At pixel level, the large differences in recall again shape the \FM
score. The recall also mainly influences the \FM scores on scaling and
rotation.

\begin{figure}[!t]
\centering
	\includegraphics[width=\linewidth]{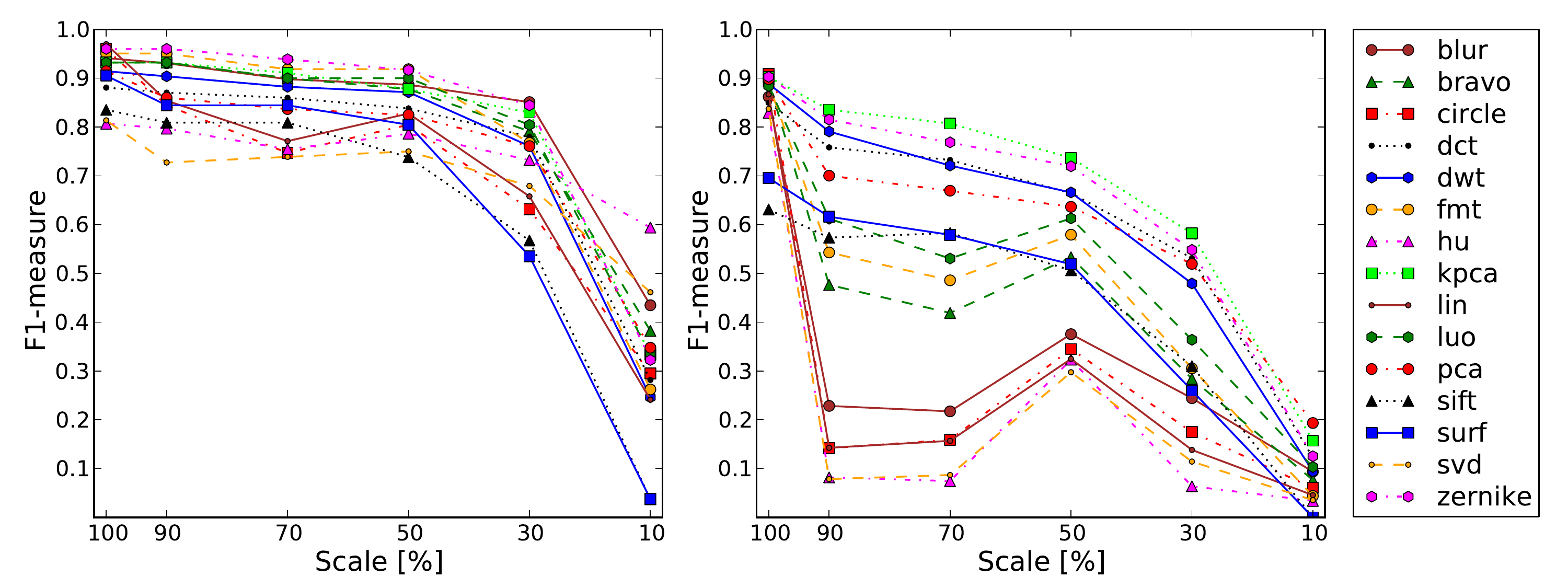}
	\caption{Downsampled versions of the input image up until $50\%$ of its
	original size. Left: performance at image level. Right: performance at
	pixel level.}
	\label{fig:scale_down}
\end{figure}

\figRef{fig:scale_down} shows the \FM scores that correspond to Fig.~10 in the
main paper. It shows the performance deterioriation under increasing
downsampling. On the left side, we evaluated the performance at image level, on
the right side at pixel level. The results of the \FM score are again mostly
influenced by the recall. Note that we omitted a performance evaluation at
image level in the main paper, mainly to improve the flow of the presentation.

\begin{figure}[!t]
\centering
	\subfigure[JPEG Compression]{
		\label{fig:jpeg_sd}
		\includegraphics[width=\linewidth]{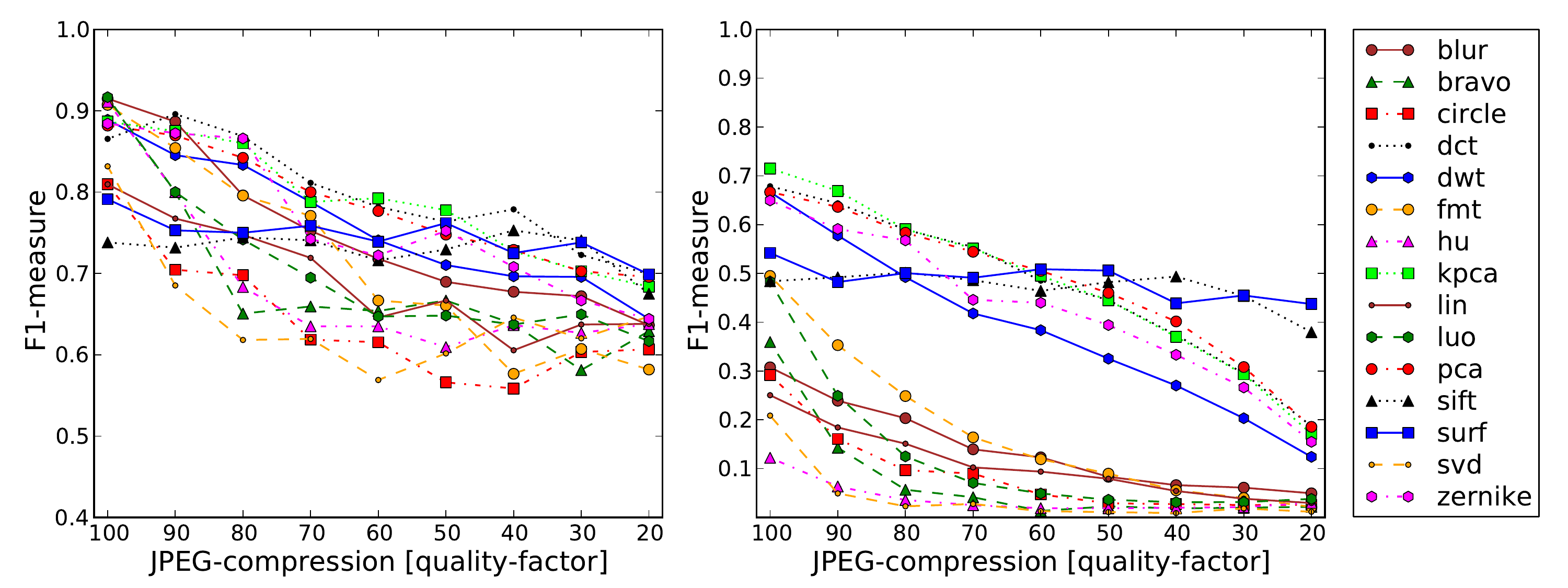}
	}
	\\
	\subfigure[Rotation]{
		\label{fig:rot_sd}
		\includegraphics[width=\linewidth]{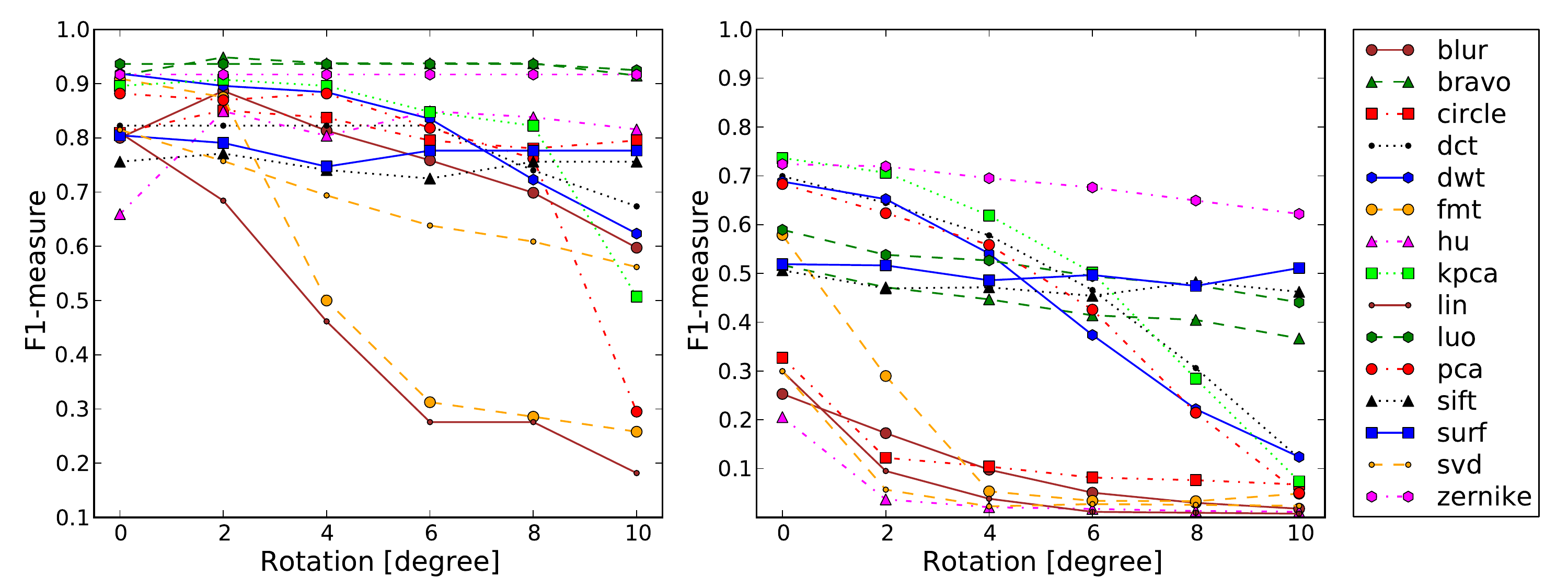}
	}
	\\
	\subfigure[Scale]{
		\label{fig:scale_sd}
		\includegraphics[width=\linewidth]{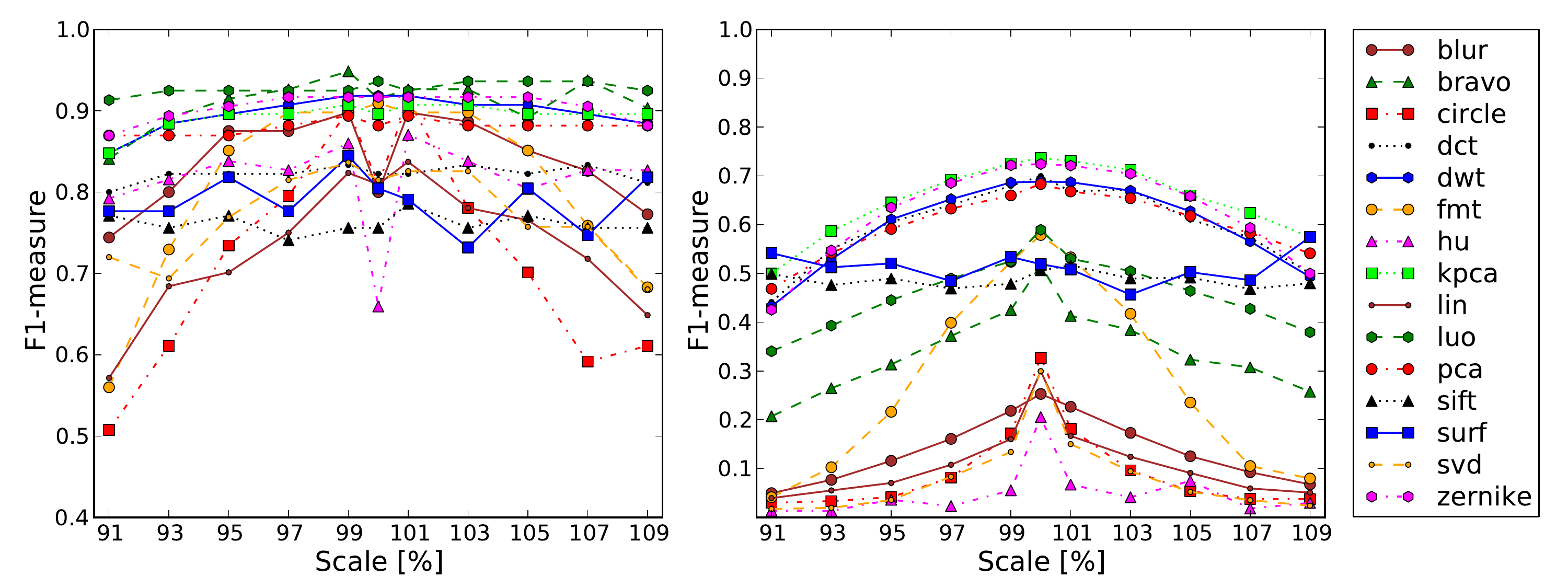}
	}
	\caption{\FM scores for the experiment on downsampled version of the images
	(corresponds to Fig.~11 in the main paper). Left: performance at image
	level. Right: performance at pixel level. From top to bottom:
	JPEG compression, rotation, scaling.}
	\label{fig:sd_plots}
\end{figure}

The \FM scores for varios scenarios applied on the downsampled version of the
images are shown in \figRef{fig:sd_plots} (see Fig.~11 in the main paper).
Again, on the left side, we show the performance at image level, on the right
side at pixel level. From top to bottom, we present results on JPEG
compression, rotation and scaling of the copied snippet. Again, in all three
cases, recall mainly influences the \FM score.

\section{Postprocessing of Keypoint-based Methods}\label{sec:postproc_keypoints}

We apply a hierarchical clustering on matched feature pairs, \ie
assign each point to a cluster and merge them according to a
linkage-method, as described by \etal{Amerini}~\cite{Amerini11:SFM}. For the
linkage method we chose ``single'' linkage as it is very fast to compute and as
the choice of the linkage-method is not
critical~\cite{Amerini11:SFM}. On the other hand, the stop condition for
merging the clusters (``cut-threshold'') plays an important role.
Here, we did not
use the \emph{inconsistency coefficient} as proposed by \etal{Amerini}.
Instead, we rely on the distance between the nearest clusters. 
Two clusters are merged if their distance lies within the cut-threshold.
We chose to use a cut-threshold of $25$ pixels for SIFT and $50$ pixels for
SURF.  As a special case, if we obtained less than $100$ matches, the
cut-threshold is raised to $75$ pixels. Note that the
cut-thresholds are chosen in a defensive way, such that typically multiple
smaller clusters remain, which are merged at a later stage of the algorithm.

If a minimum of $4$ correspondences connects two of these clusters, we estimate
with RANSAC the affine transformation between the points that are connected by
these correspondences, analogously to \etal{Amerini}~\cite{Amerini11:SFM}. In
contrast to prior work, we further compute the optimal transformation matrix
from the inliers according to the gold-standard algorithm for affine homography
matrices~\cite[pp.  130]{Hartley03:MVG}. Transformations with implausibly large
or small values are removed via thresholding.

For large images containing large copied areas the transformation
matrices of the clusters may often be similar. So, we decided to
merge them if the root mean square error (RMSE) between two transformation matrices
is below a threshold of $0.03$ for the scaling and rotation part of the matrix,
and below a threshold of $10$ for the translation part.
For these merged clusters we reestimate the transformation with the
same procedure as above.

For each cluster we warp the image according to the estimated
transformation matrix and compute a correlation map between the image
and its warped version. From here on, we follow the algorithm by Pan and
Lyu~\cite{Pan10:RDD}.
For every pixel and its warped version we
compute the normalized correlation coefficient with a window size of
$5\times 5$ pixels. To remove noise in the correlation map, it is smoothed with a
$7\times7$ pixels Gaussian kernel. Every smoothed correlation map is then binarized
with a threshold of $0.4$, and areas containing less than $1000$ pixels were
removed. Furthermore, areas where no match lies
were removed, too. Then, the outer contours of each area is extracted and the inner
part is flood-filled, which closes holes in the contours.
The output map is the combination of all post-processed correlation
maps. As a final verification step, each area from the combined output
map is tested if it also has a correspondence which lies in another
marked area.

\section{Database Categories}\label{sec:db_categories}

In the main paper, we report results on the full dataset. However, the
performance of the feature descriptors undoubtedly depends on the content of
the image and the copied area. With a subdivision of the dataset in smaller
categories, we make an attempt towards better explaining the performance
differences of the feature sets.

The design of a proper category-driven evaluation in image forensics is still
an open problem.
We made a first attempt towards a proper categorization using two different
approaches. We divided the images in multiple categories, once into object
categories and once in texture categories. The results are reported in
\subsecRef{subsec:object_categories} and \subsecRef{subsec:texture_categories},
respectively.

\subsection{Categorization by Object Classes}\label{subsec:object_categories}

We split the images in categories where the copied regions belong to one of the
object categories \emph{living}, \emph{nature}, \emph{man-made} or
\emph{mixed}. Here, \emph{mixed} denotes copies where arguably multiple object
types occur. Table~\ref{tab:categorization_assignment_semantic} lists which
image belongs to which category. In \secRef{sec:overview_database}, downsampled
versions of the images are presented together with the names of the images.
The number of images varies between the categories.
\begin{table}[tb]
\newlength{\myParWidthA}
\setlength{\myParWidthA}{3.2cm}
\caption{Categorization of the database by object classes.}
\label{tab:categorization_assignment_semantic}
\begin{tabular}{m{1cm}||p{\myParWidthA}|p{\myParWidthA}}
           & \multicolumn{2}{c}{Assigned images} \\
\hline
Category   & Small copied area & Large copied area \\
\hline
\hline
living     & giraffe, jellyfish chaos, cattle, swan  & four babies \\
\hline
nature     & fisherman, no beach, berries, & Scotland, white, hedge, christmas hedge, Malawi, beach wood, tree \\
\hline
man-made    &supermarket, bricks, statue, ship, sailing, dark and bright, sweets, disconnected shift, Egyptian, noise pattern, sails, mask, window, writing history, knight moves & horses, kore, extension, clean walls, tapestry, port, wood carvings, stone ghost, red tower \\
\hline
mixed      & barrier, motorcycle, Mykene, threehundred, Japan tower, wading, central park  & fountain, lone cat \\
\end{tabular}
\end{table}
The smallest category is
\emph{living} containing $5$ test cases, the largest category is
\emph{man-made} ($24$ test cases). Note that the varying category sizes pose no
problem, as we only compute the performance within a category\footnote{For
instance, for classification tasks, a balanced size of each category can
prevent biased results. However, this is not of concern in our copy-move
forgery evaluation.}. We used the same parameters as for the evaluations in the
main paper.  \tabRef{tab:img_nul_categories_object} shows the $F_1$ score for
plain copy-move forgeries at image level (left) and at pixel level (right).
Note that all four categories perform comparably at pixel level. However, at
image level, most feature sets perform best for \emph{living} and
\emph{nature}.

\begin{table}[tb]
\centering
	\caption{Results for plain copy-move per object category at image level (left) and at pixel level (right), in percent.}
	\label{tab:img_nul_categories_object}
	\begin{tabular}{|l|r|r||r|r|}
		\hline
		\input{tables/nul_default_new_living_nature.tex}
		\hline
	\end{tabular}

\mbox{}

\mbox{}

	\begin{tabular}{|l|r|r||r|r|}
		\hline
		\input{tables/nul_default_new_manmade_mixed.tex}

		\hline
	\end{tabular}

\end{table}

\begin{figure}[t!]
\centering
		\includegraphics[width=\linewidth]{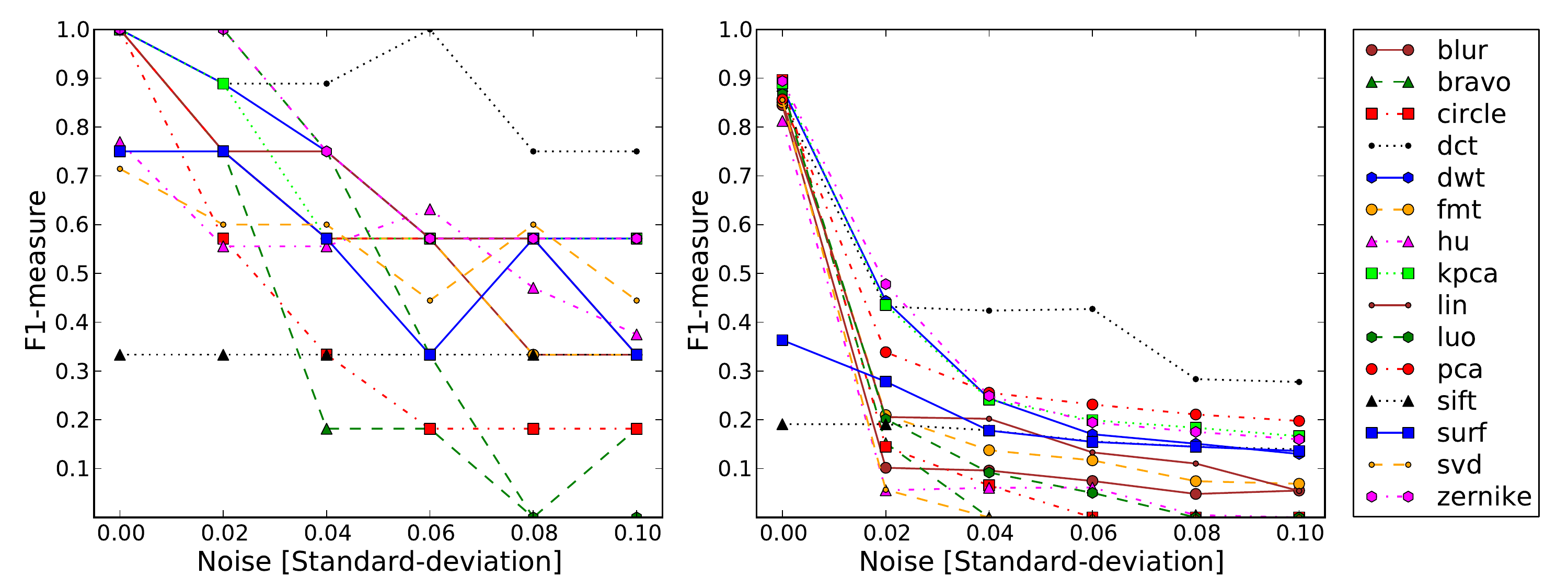}
	\\
		\includegraphics[width=\linewidth]{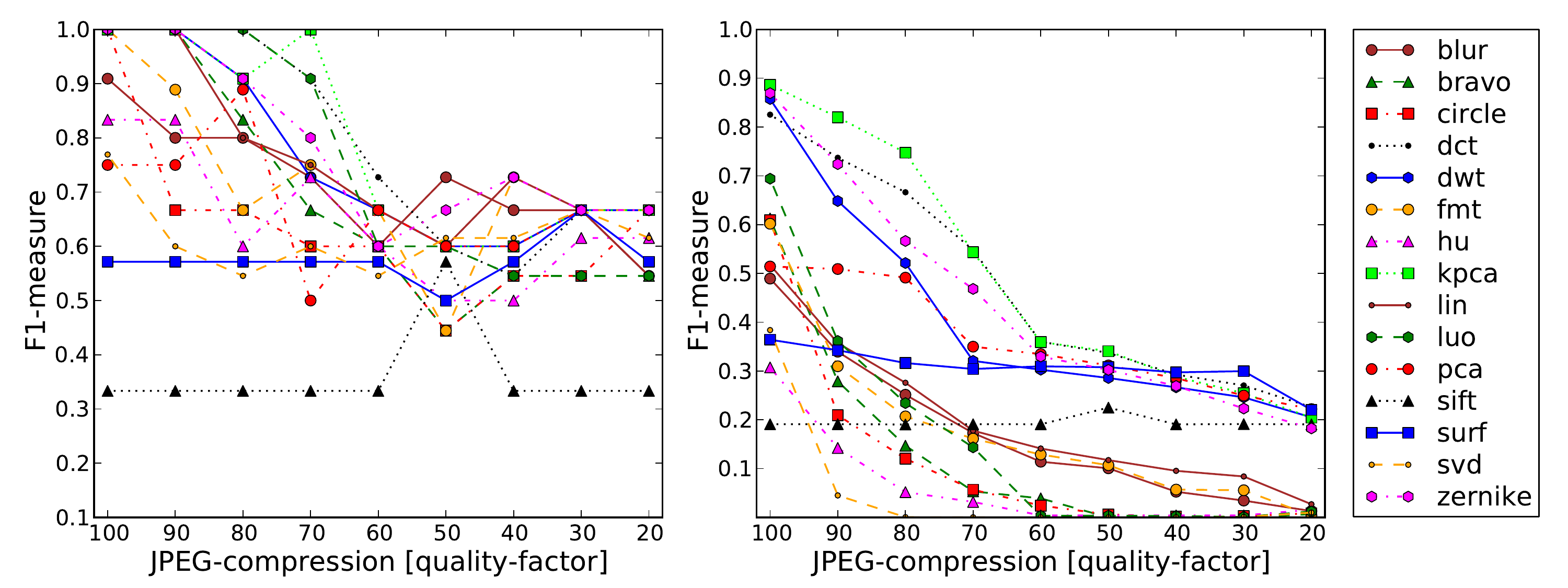}
	\\
		\includegraphics[width=\linewidth]{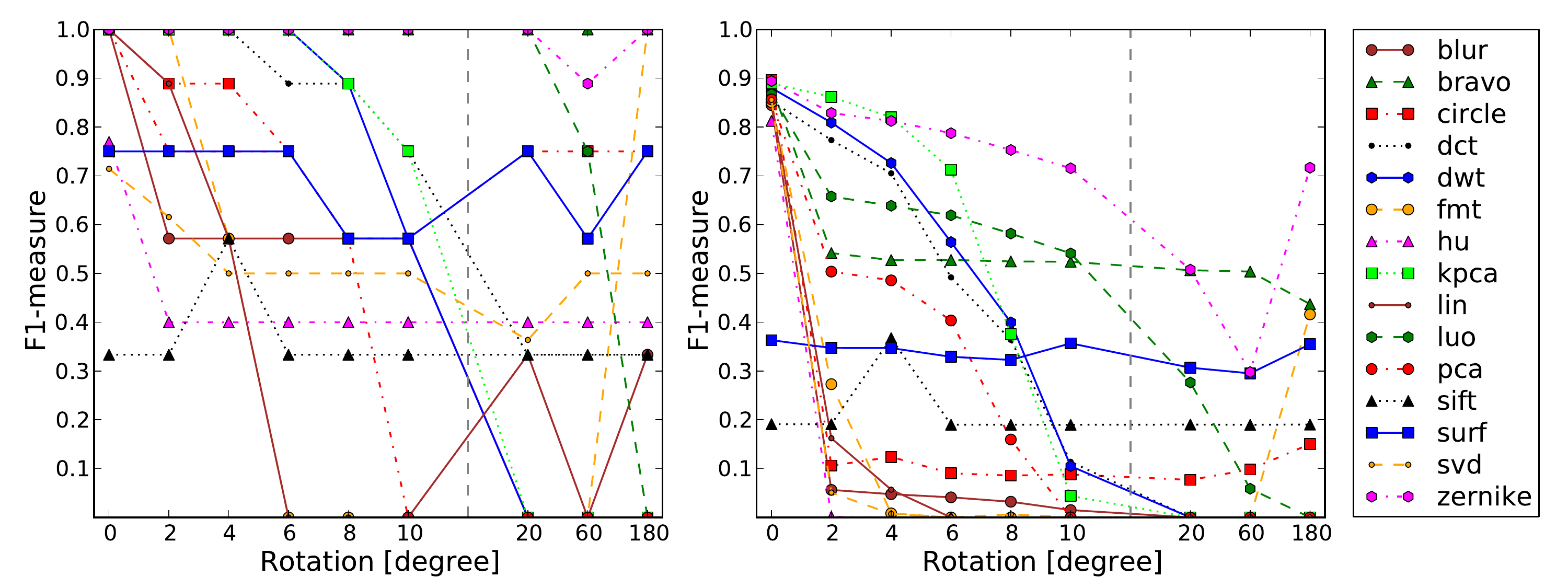}
	\\
		\includegraphics[width=\linewidth]{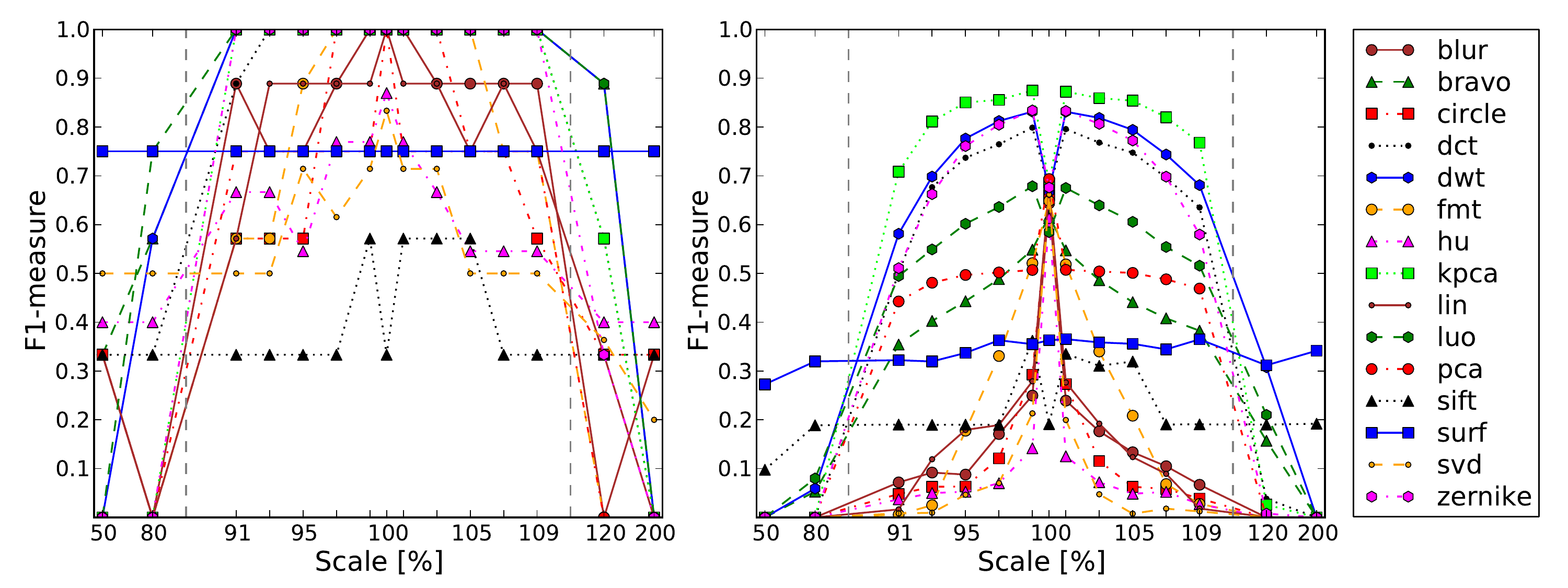}
	\\
		\includegraphics[width=\linewidth]{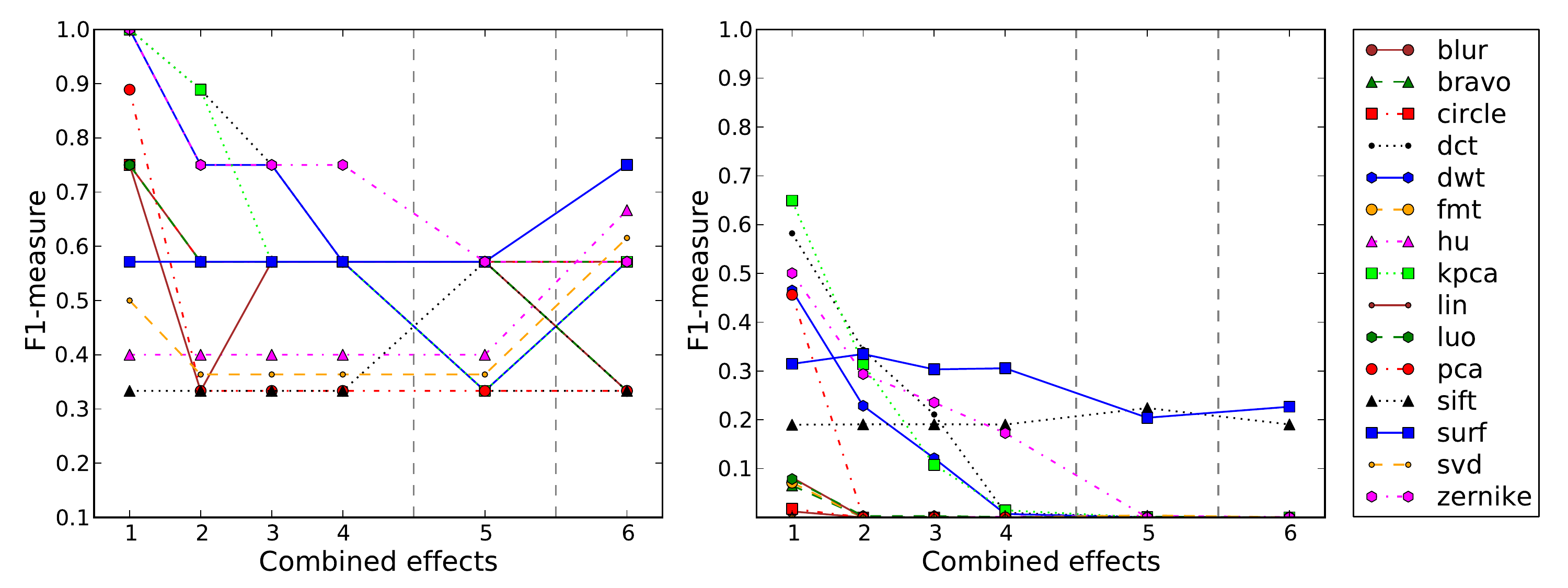}
	\caption{Performance in the category \emph{living} at image level (left) and pixel level (right).}
	\label{fig:cat_living}
\end{figure}

\begin{figure}[t!]
\centering
		\includegraphics[width=\linewidth]{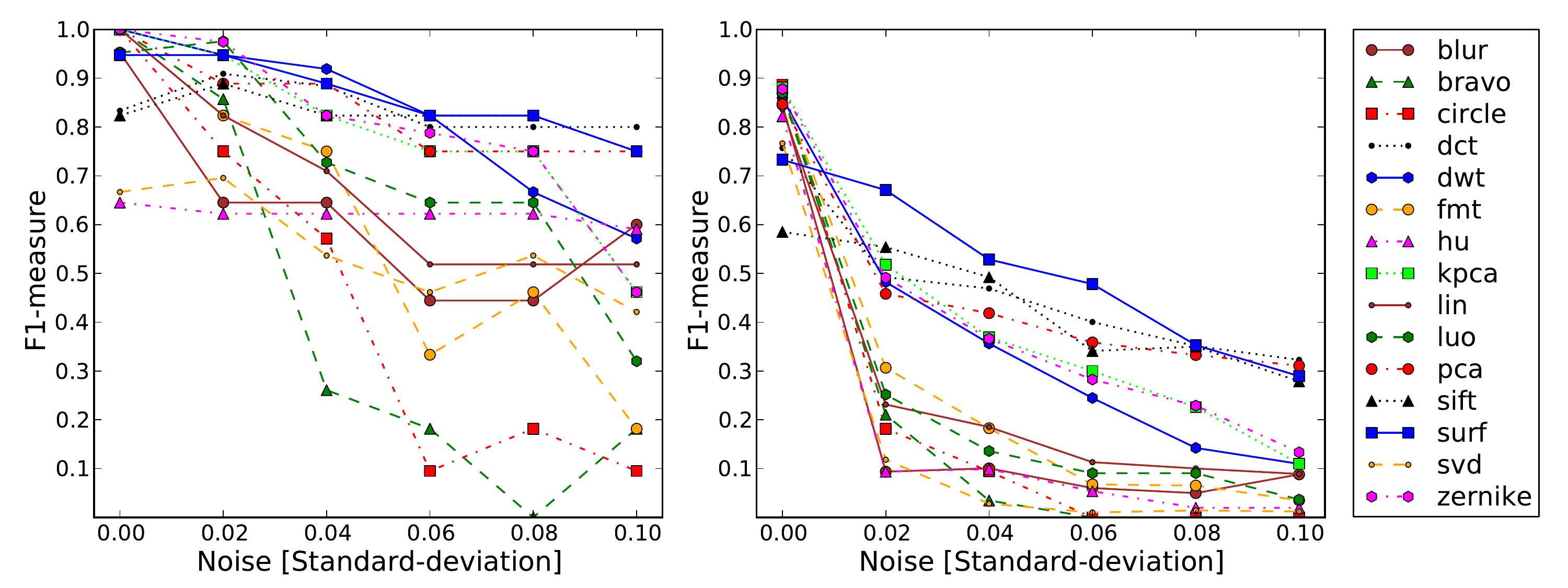}
	\\
		\includegraphics[width=\linewidth]{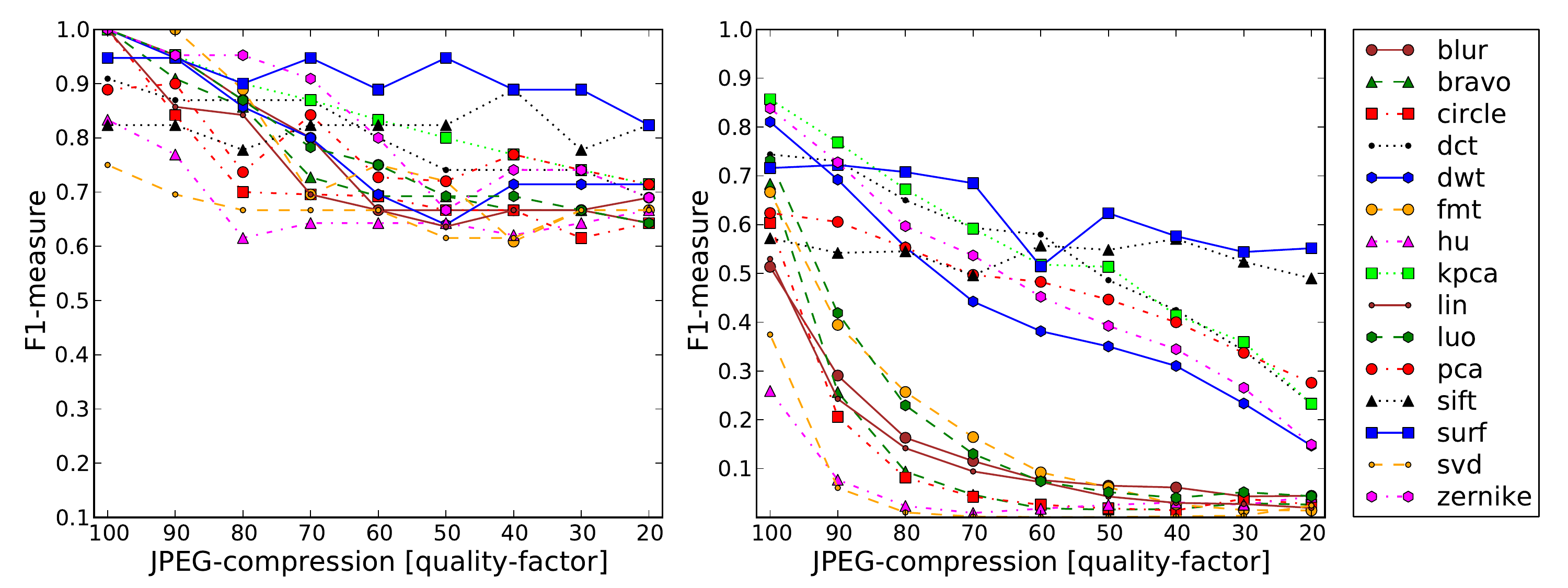}
	\\
		\includegraphics[width=\linewidth]{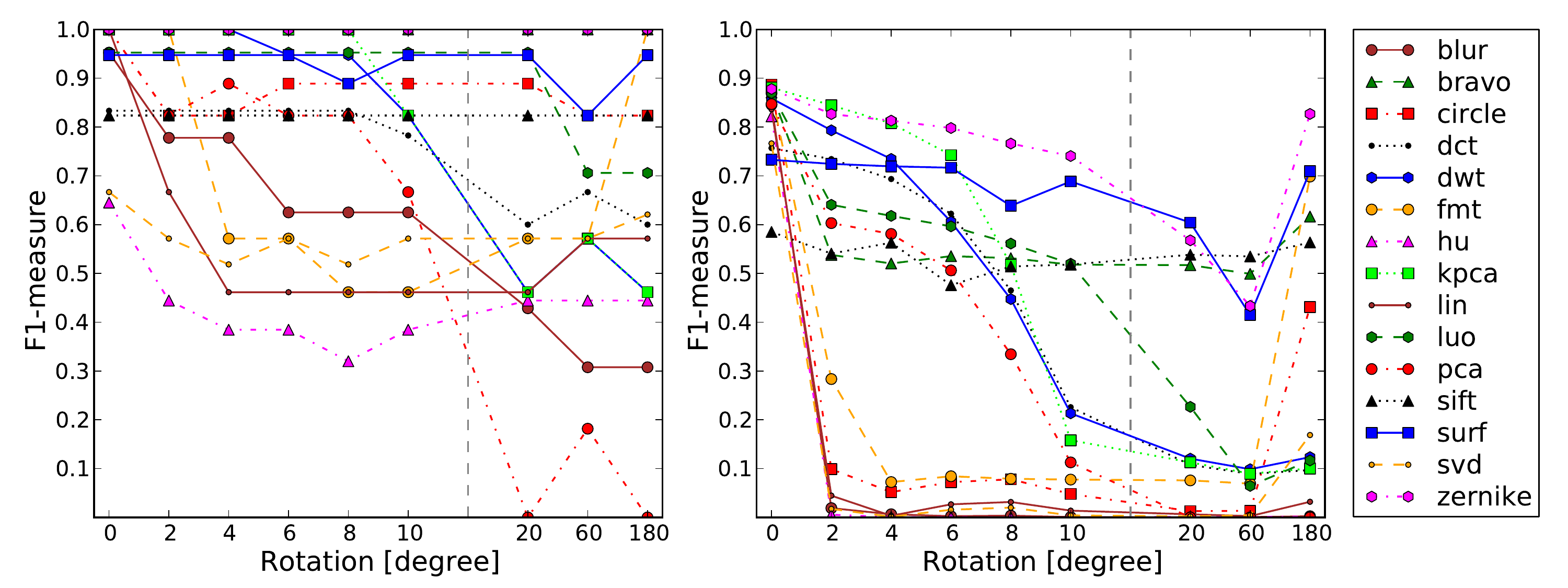}
	\\
		\includegraphics[width=\linewidth]{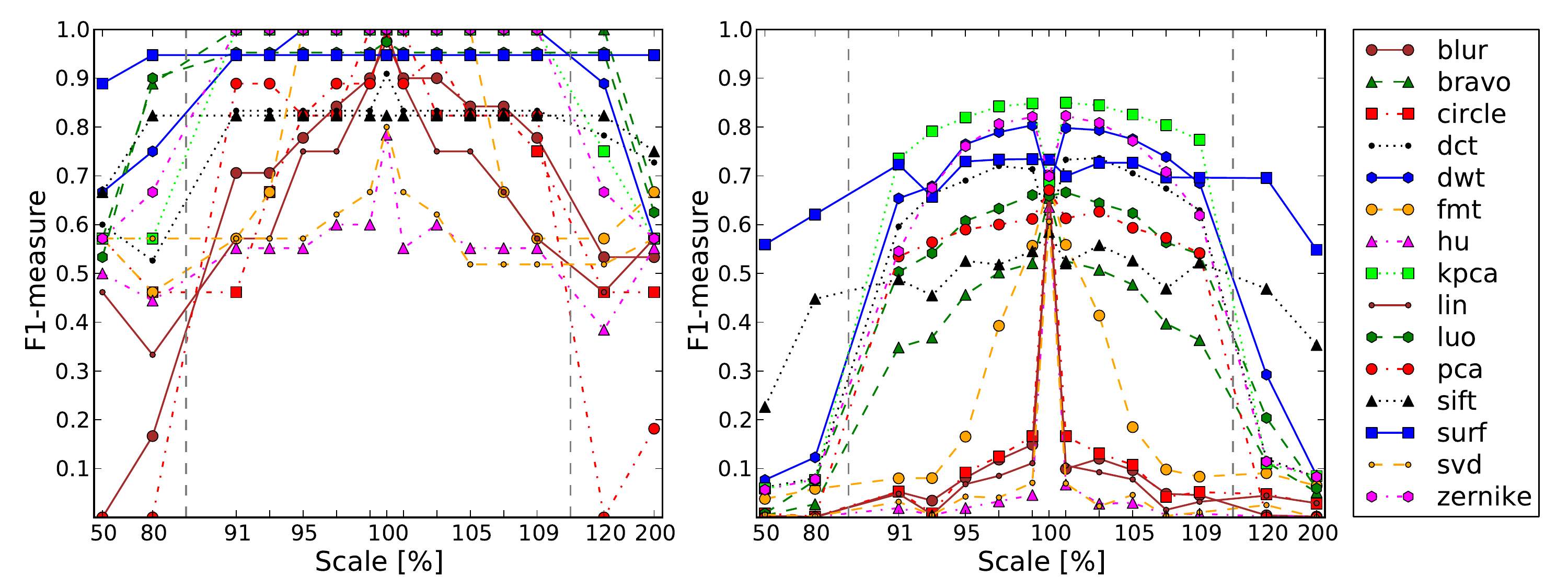}
	\\
		\includegraphics[width=\linewidth]{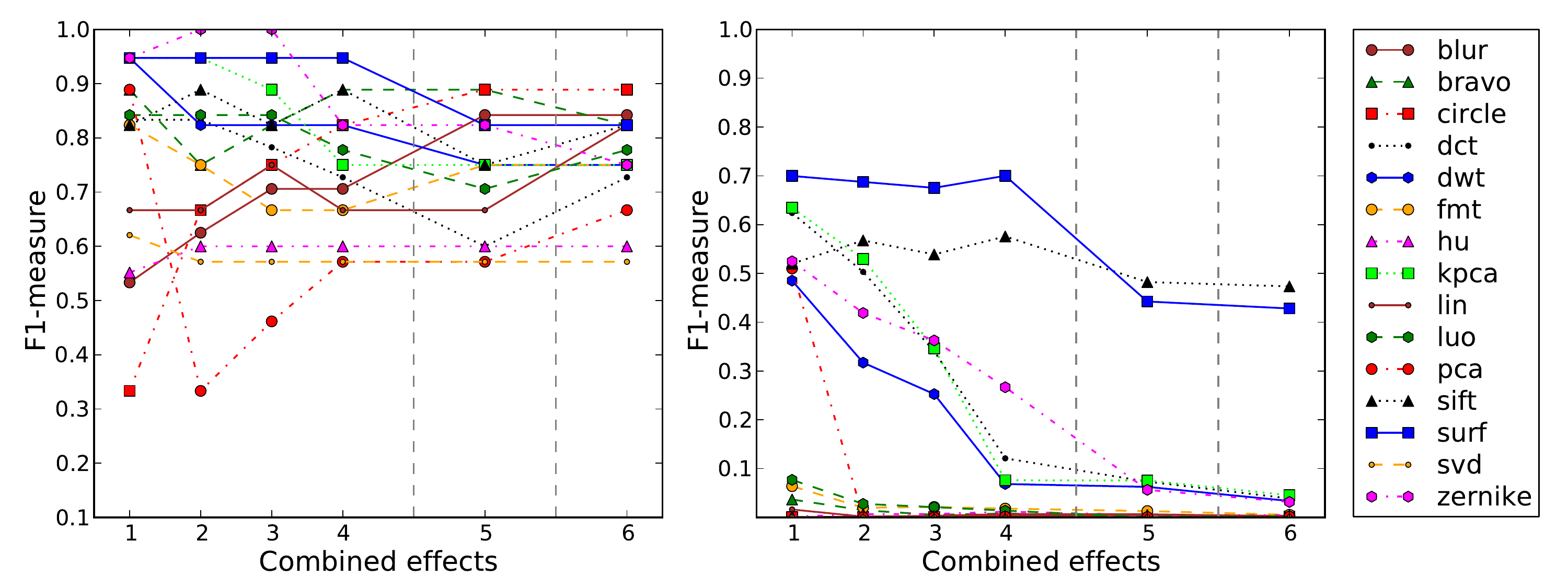}
	\caption{Performance in the category \emph{nature} at image level (left) and pixel level (right).}
	\label{fig:cat_nature}
\end{figure}

\begin{figure}[t!]
\centering
		\includegraphics[width=\linewidth]{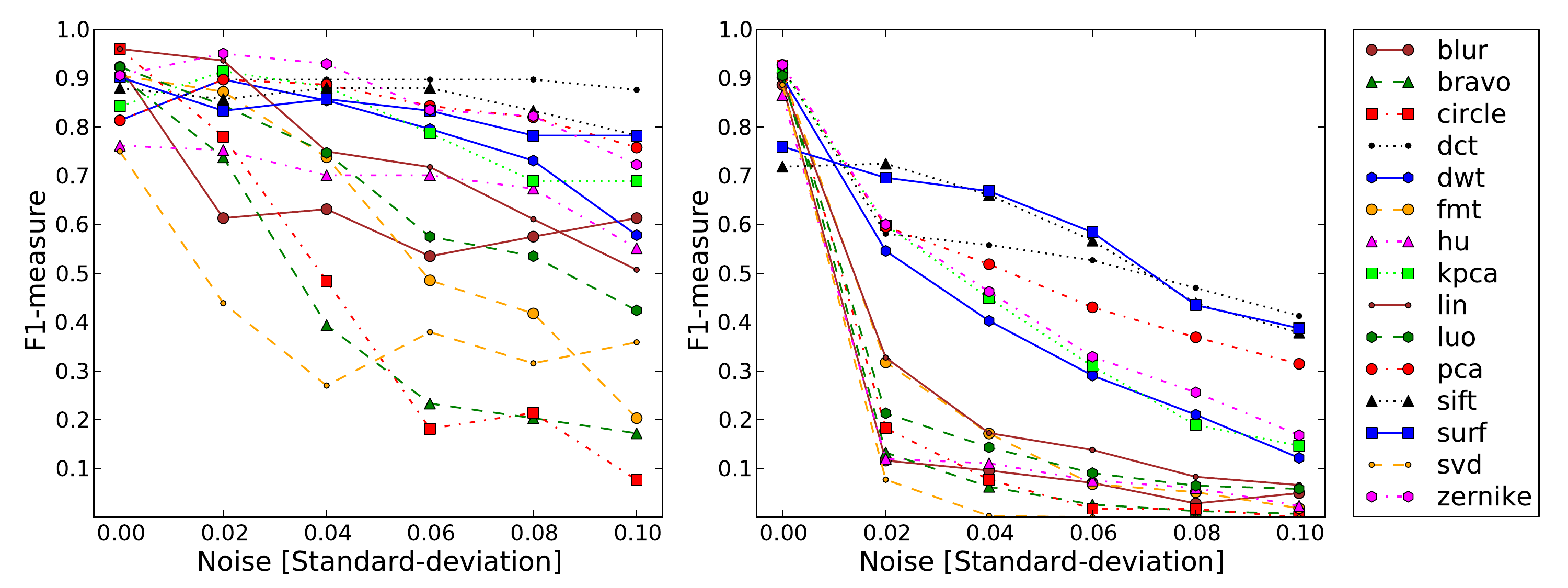}
	\\
		\includegraphics[width=\linewidth]{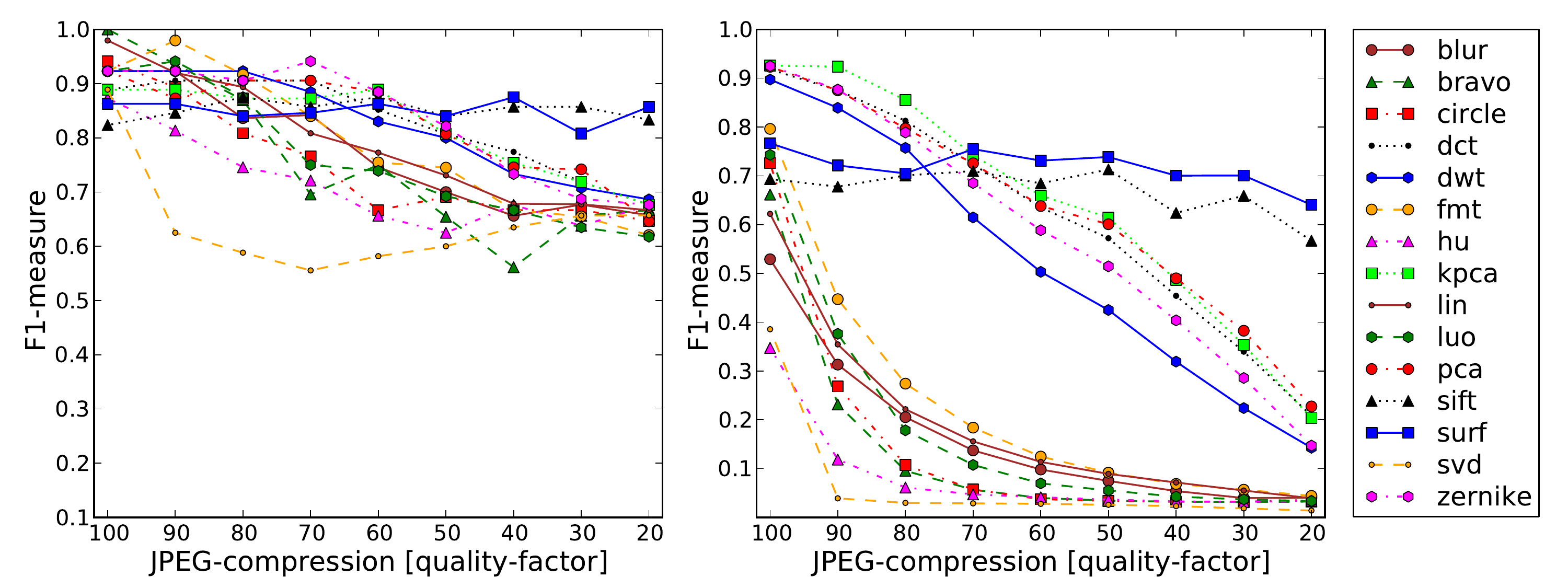}
	\\
		\includegraphics[width=\linewidth]{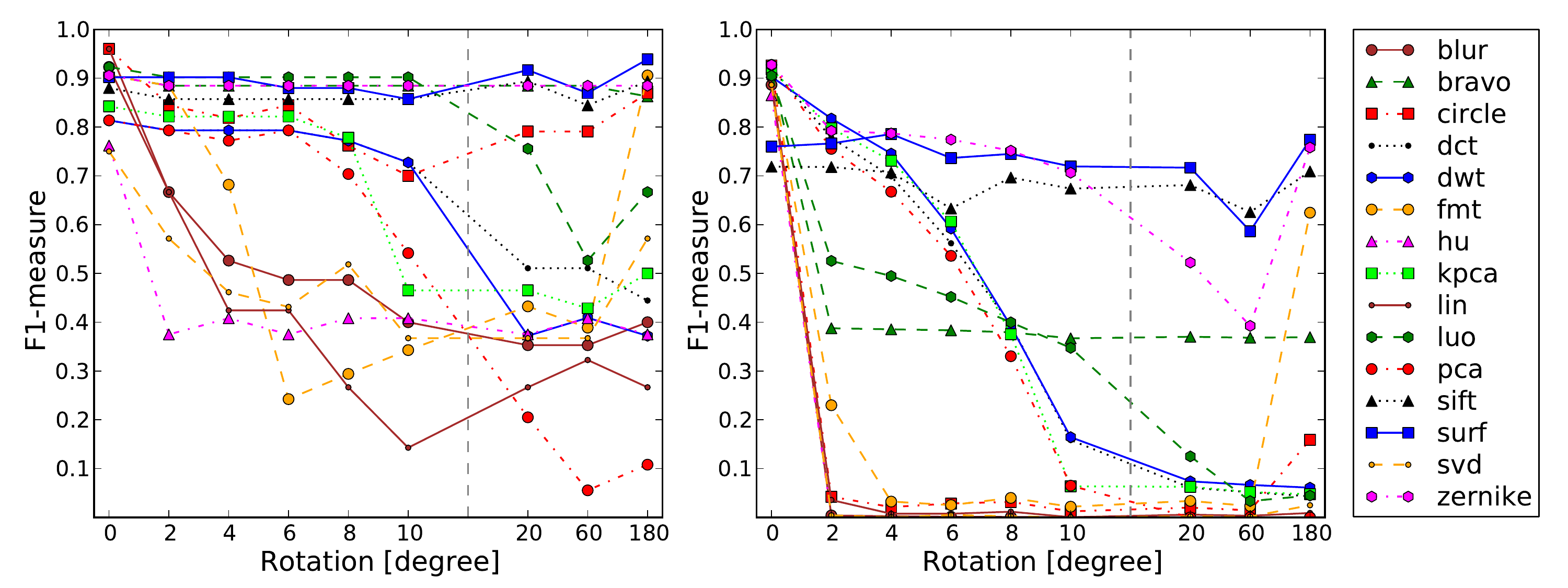}
	\\
		\includegraphics[width=\linewidth]{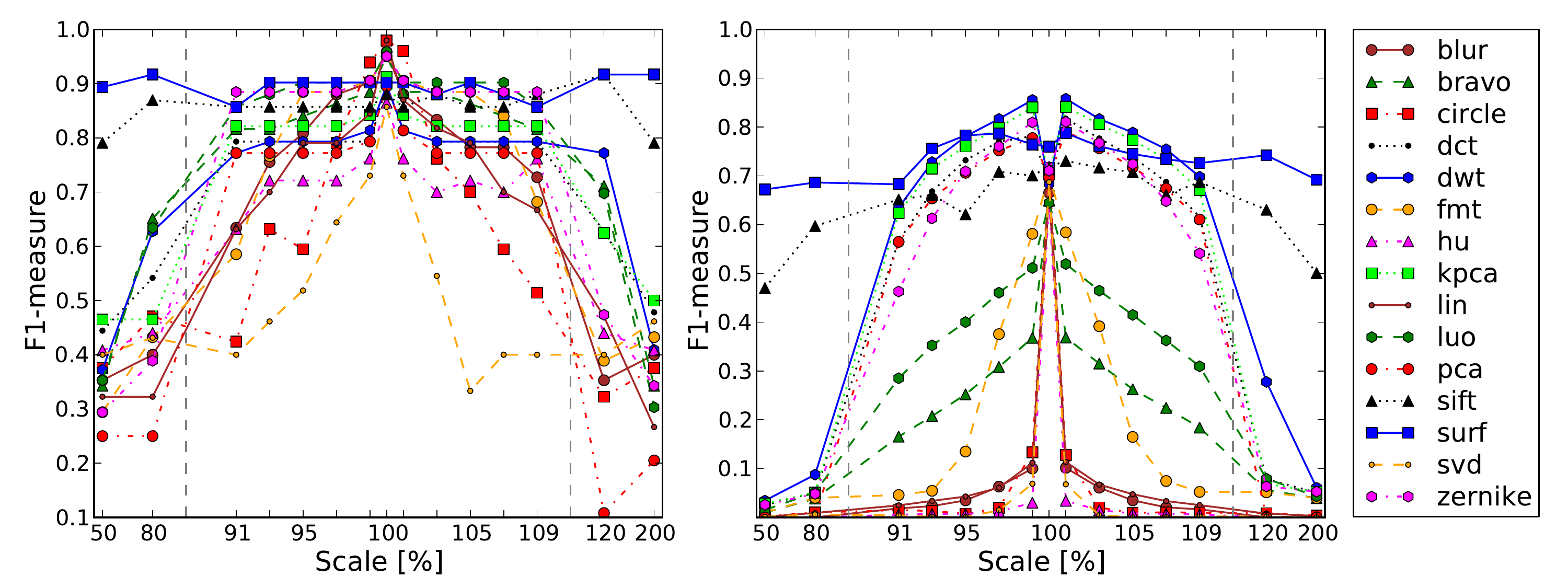}
	\\
		\includegraphics[width=\linewidth]{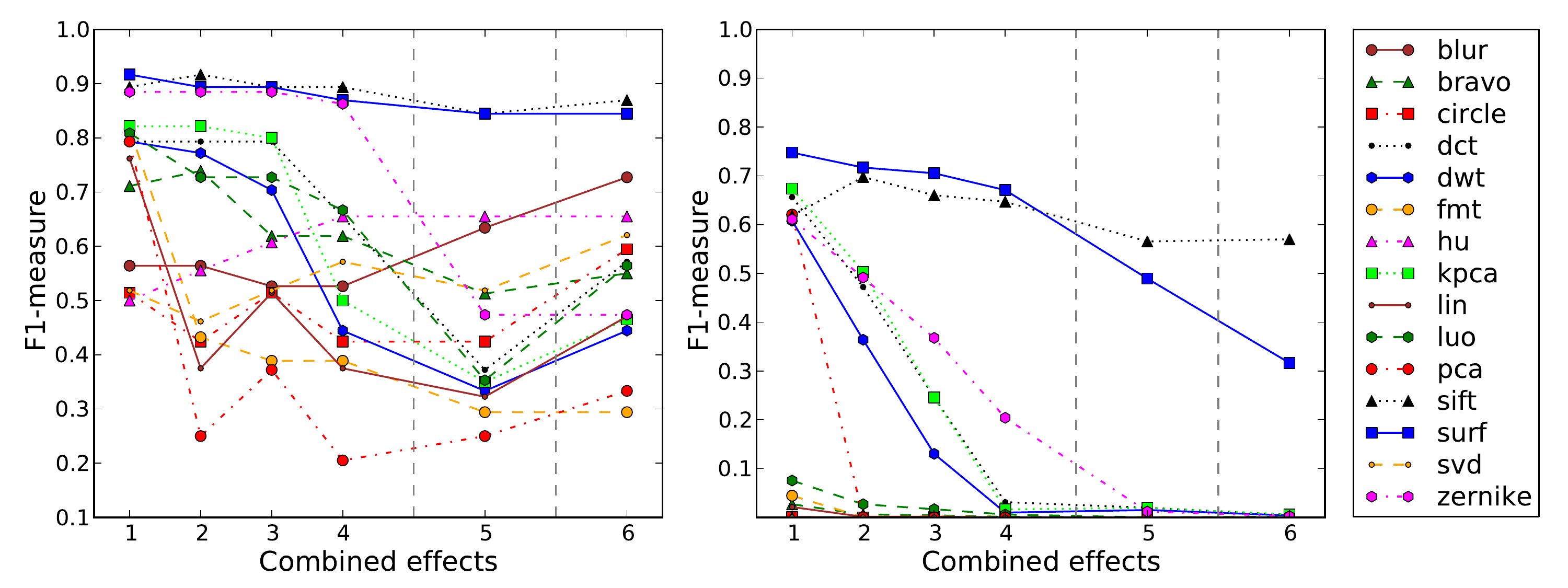}
	\caption{Performance in the category \emph{man-made} at image level (left) and pixel level (right).}
	\label{fig:cat_manmade}
\end{figure}

\begin{figure}[t!]
\centering
		\includegraphics[width=\linewidth]{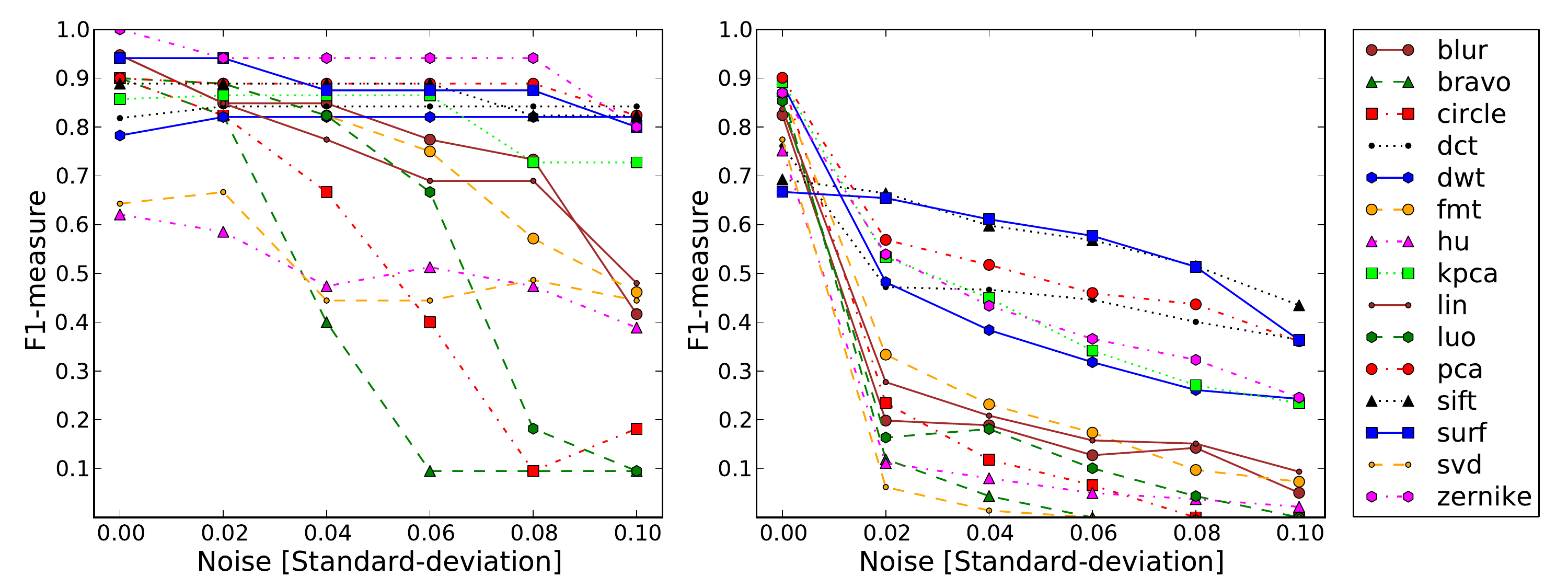}
	\\
		\includegraphics[width=\linewidth]{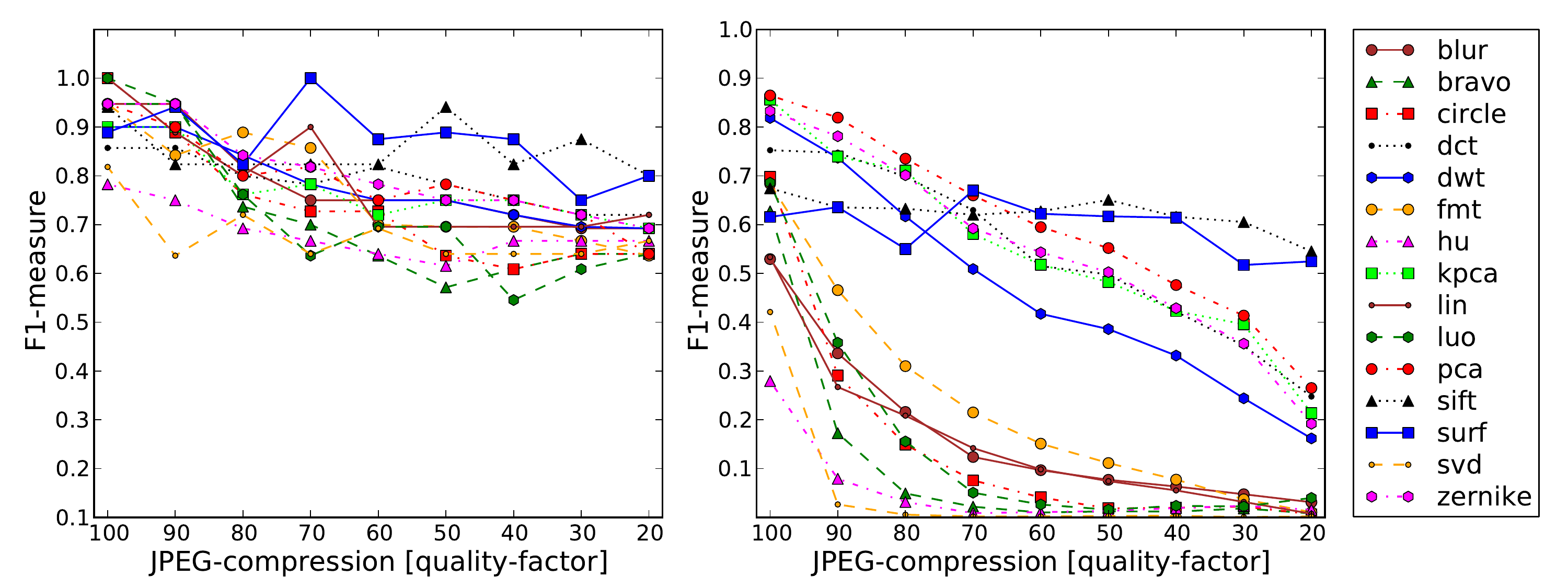}
	\\
		\includegraphics[width=\linewidth]{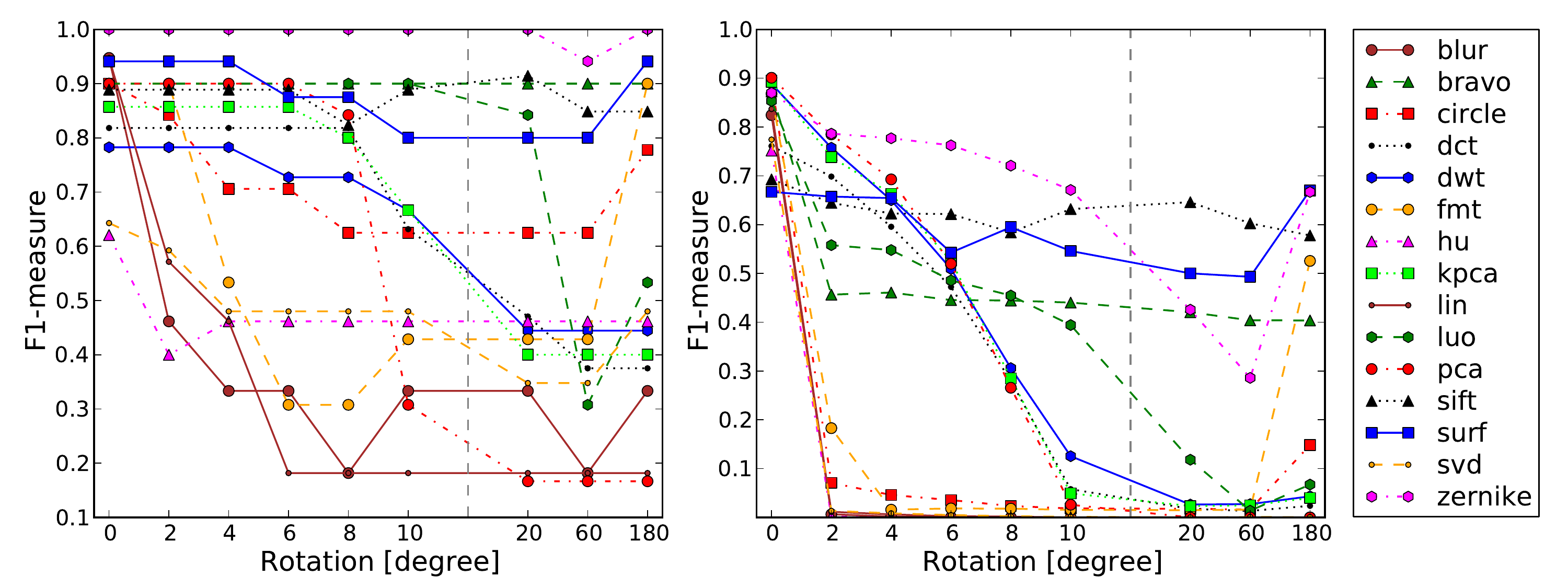}
	\\
		\includegraphics[width=\linewidth]{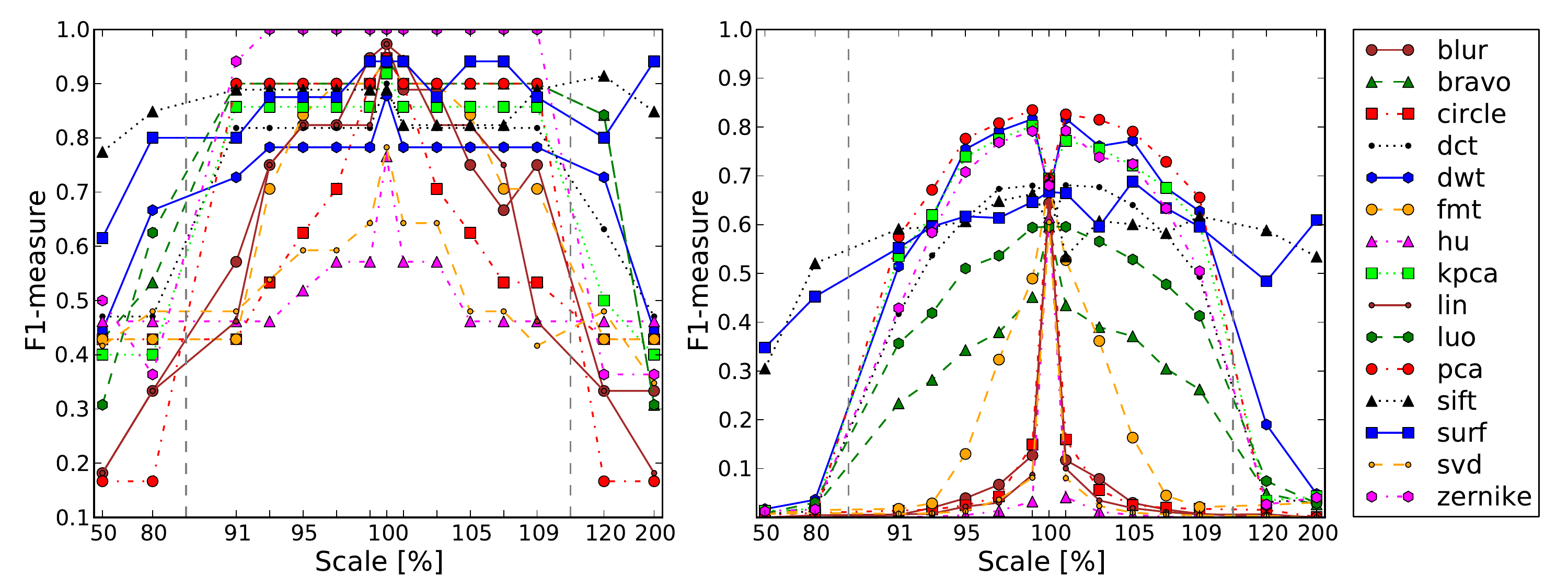}
	\\
		\includegraphics[width=\linewidth]{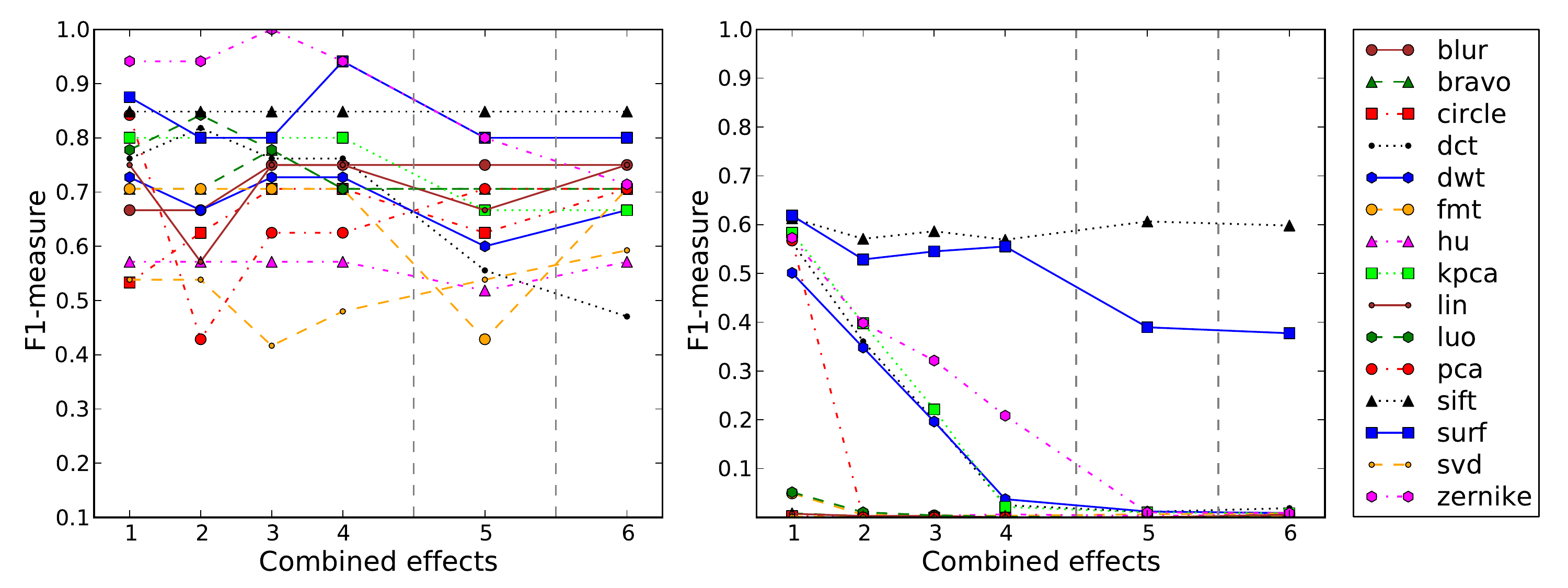}
	\caption{Performance in the category \emph{mixed} at image level (left) and pixel level (right).}
	\label{fig:cat_mixed}
\end{figure}

The Figures~\ref{fig:cat_living}, \ref{fig:cat_nature}, \ref{fig:cat_manmade}
and \ref{fig:cat_mixed} show the results for \emph{living}, \emph{nature},
\emph{man-made} and \emph{mixed} under Gaussian noise, JPEG compression,
rotation, scaling and joint effects. Again, the evaluation parameters were the
same as in the main paper. At pixel level, several feature sets perform better
for \emph{nature} than for \emph{living}. However, in several scenarios, the
best results are obtained in the categories \emph{man-made} and \emph{mixed}.
We assume that this comes from the fact that man-made objects often exhibit a
clearer structure, and as such encompass the task of copy-move forgery
detection.

\subsection{Categorization by Texture}\label{subsec:texture_categories}

We investigated a second categorization of the dataset, this time by texture.
The dataset is divided in copied areas providing \emph{smooth}, \emph{rough}
and \emph{structured} content. Here, \emph{smooth} and \emph{rough} serves as a
distinction of texture properties, which can be approximately seen as low or
high entropy in the snippet. The third category, \emph{structured}, refers in
most cases to man-made structures, like buildings: regular, clearly pronounced
edges and corners.

This categorization was already prepared during the creation of the dataset. We
aimed to use a diverse set of scenes, providing various challenges to the
detectors.
One challenge of real-world forgeries is the fact that we have little control
over the creation of the manipulation. As a consequence, the texture categories
are not based on quantitative measures. Instead, we used the artists' result on
a fuzzy task description, like to ``create a copy with little
texture''.  Given the fact that real-world copy-move forgeries are done from
artists as well, we found it reasonable to adopt the artists' viewpoint on
CMFD.

\tabRef{tab:categorization_assignment} shows the assignment of images to
categories. For our evaluation, we did not distinguish the size of the copied
areas. Thus, we compare the three major categories \emph{smooth}, \emph{rough}
and \emph{structured}. The number of motifs per category is $17$, $16$ and
$15$, respectively.

\begin{table}[tb]
\newlength{\myParWidth}
\setlength{\myParWidth}{3.2cm}
\caption{Categorization of the database by texture properties.}
\label{tab:categorization_assignment}
\begin{tabular}{m{1cm}||p{\myParWidth}|p{\myParWidth}}
           & \multicolumn{2}{c}{Assigned images} \\
\hline
Category   & Small copied area & Large copied area \\
\hline
\hline
Smooth     & ship, motorcycle, sailing, disconnected shift, noise pattern, berries, sails, mask, cattle, swan, Japan tower, wading & four babies, Scotland, hedge, tapestry, Malawi \\
\hline
Rough      & supermarket, no beach, fisherman, barrier, threehundred, writing history, central park & lone cat, kore, white, clean walls, tree, christmas hedge, stone ghost, beach wood, red tower \\
\hline
Structured & bricks, statue, giraffe, dark and bright, sweets, Mykene, jellyfish chaos, Egyptian, window, knight moves & fountain, horses, port, wood carvings, extension  \\
\end{tabular}
\end{table}

\begin{table}[tb]
\centering
	\caption{Results for plain copy-move at image level (left) and at pixel level (right), in percent.}
	\label{tab:img_nul_categories}
	\begin{tabular}{|l|r|r|r||r|r|r|}
		\hline
		\input{tables/nul_default_new_smooth_rough_structure.tex}
		\hline
	\end{tabular}
\end{table}

\begin{figure}[t!]
\centering
		\includegraphics[width=\linewidth]{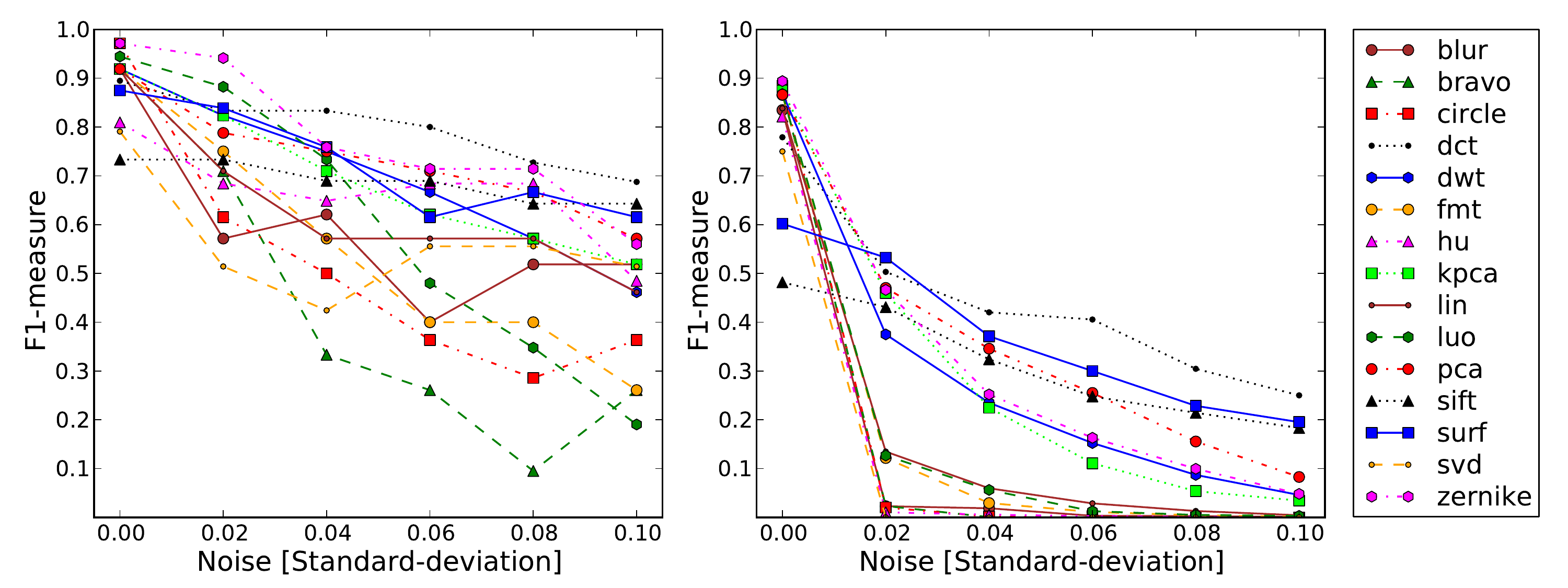}
	\\
		\includegraphics[width=\linewidth]{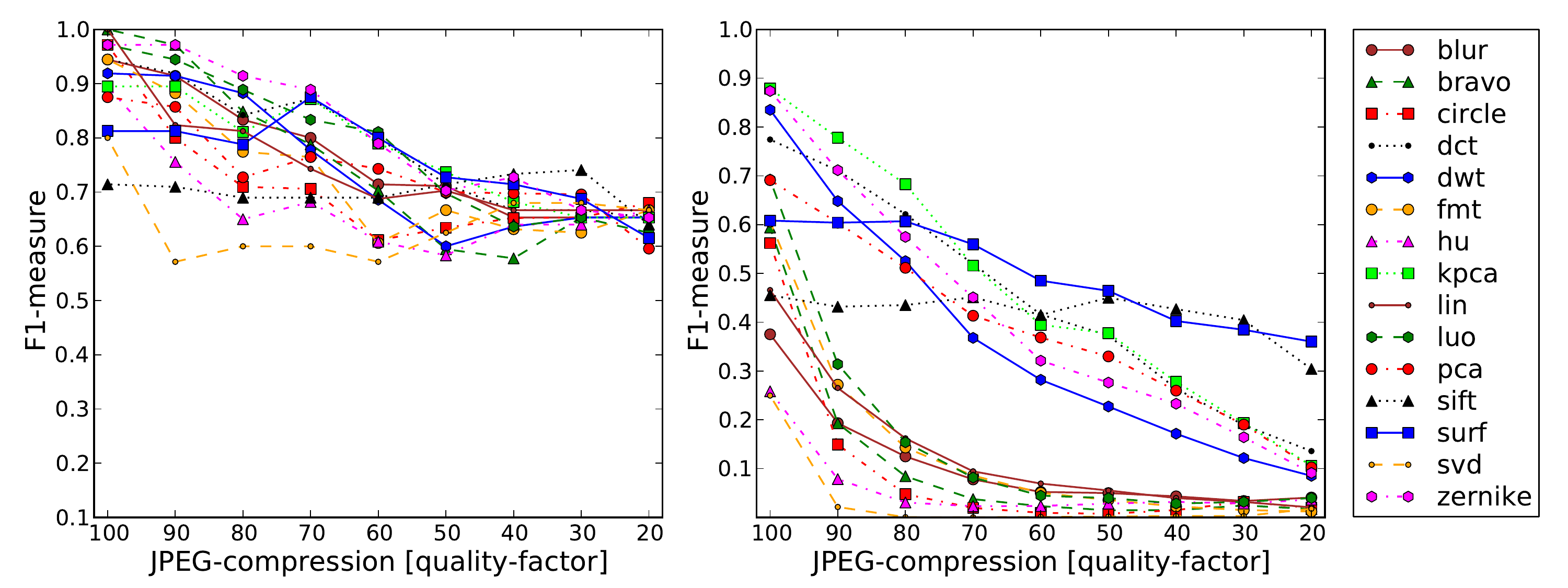}
	\\
		\includegraphics[width=\linewidth]{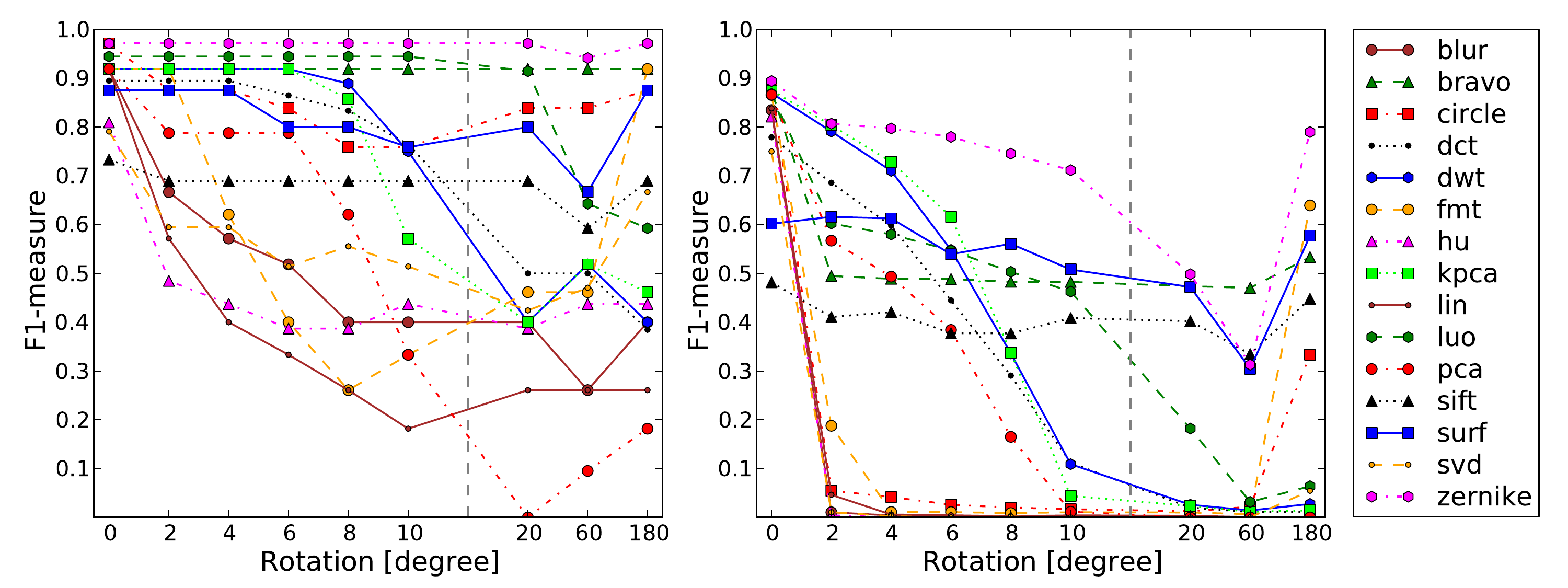}
	\\
		\includegraphics[width=\linewidth]{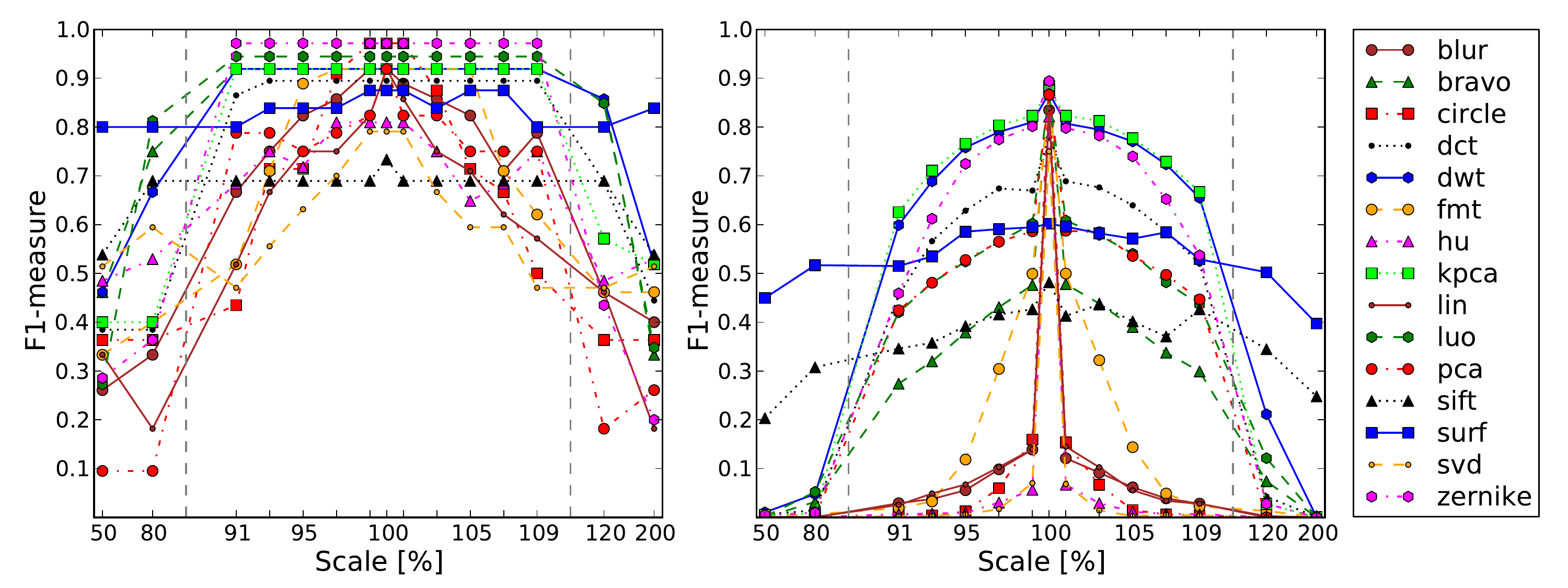}
	\\
		\includegraphics[width=\linewidth]{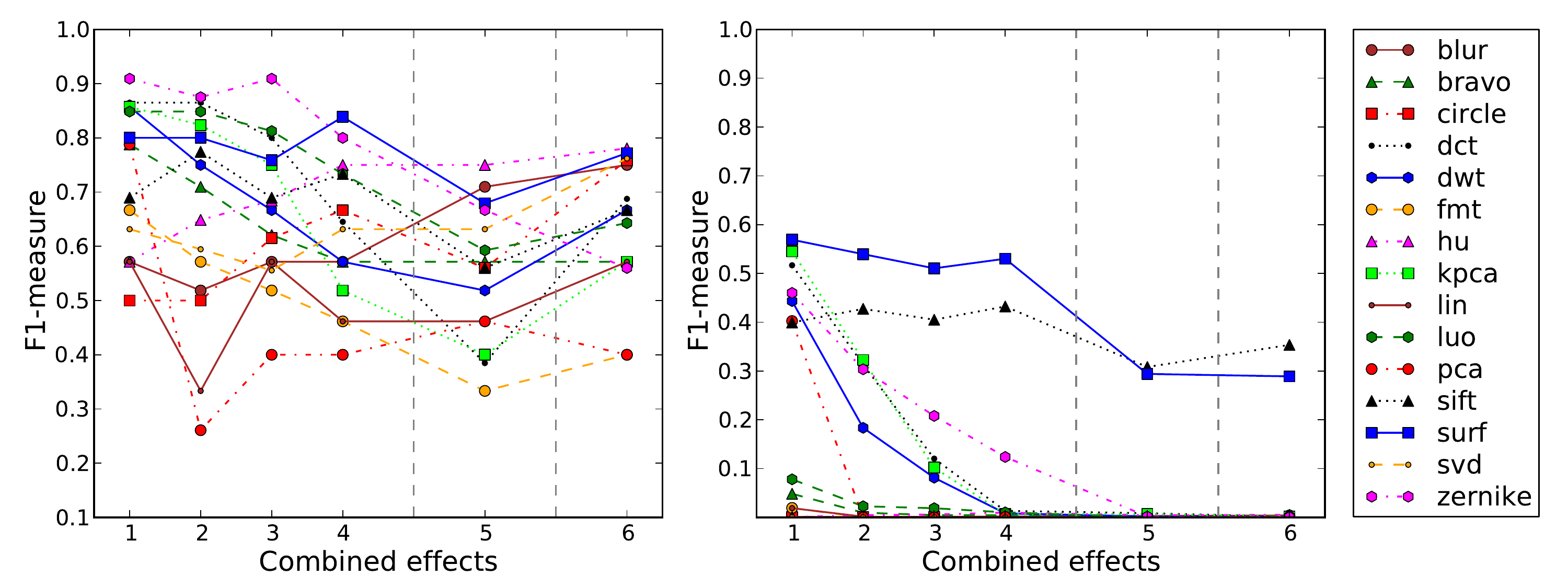}
	\caption{Performance in the category \emph{smooth} at image level (left) and pixel level (right).}
	\label{fig:cat_smooth}
\end{figure}

\begin{figure}[t!]
\centering
		\includegraphics[width=\linewidth]{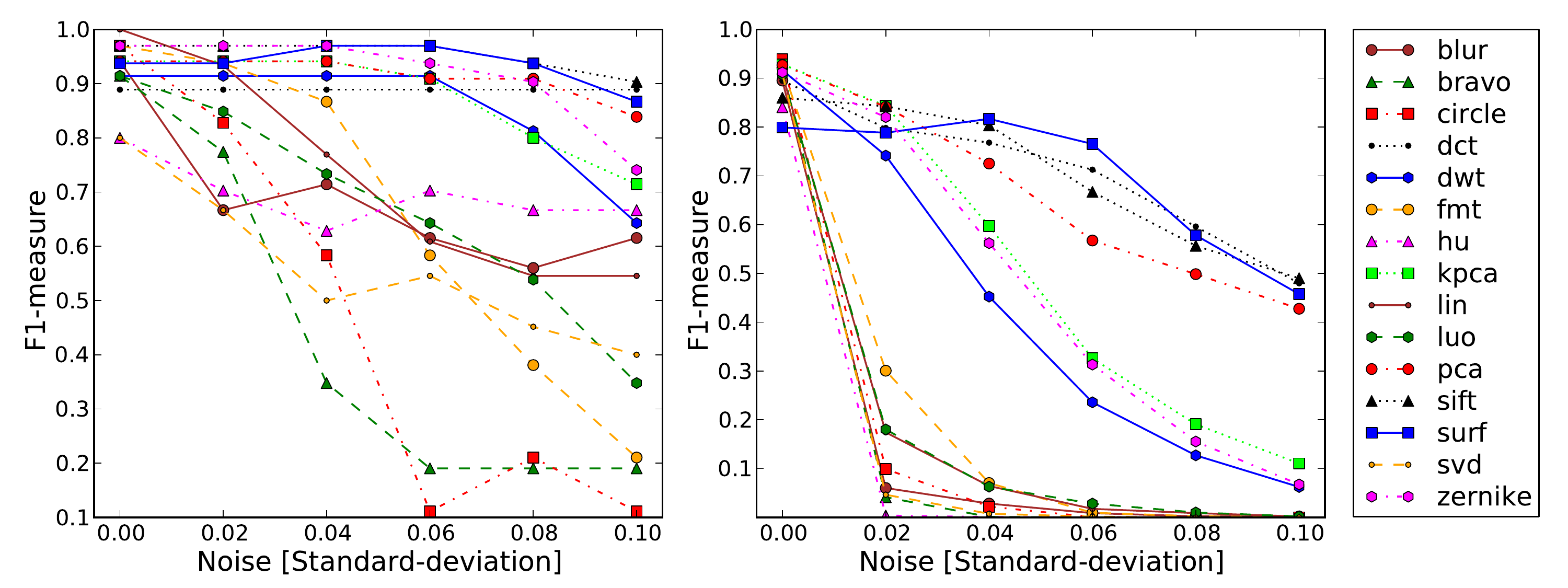}
	\\
		\includegraphics[width=\linewidth]{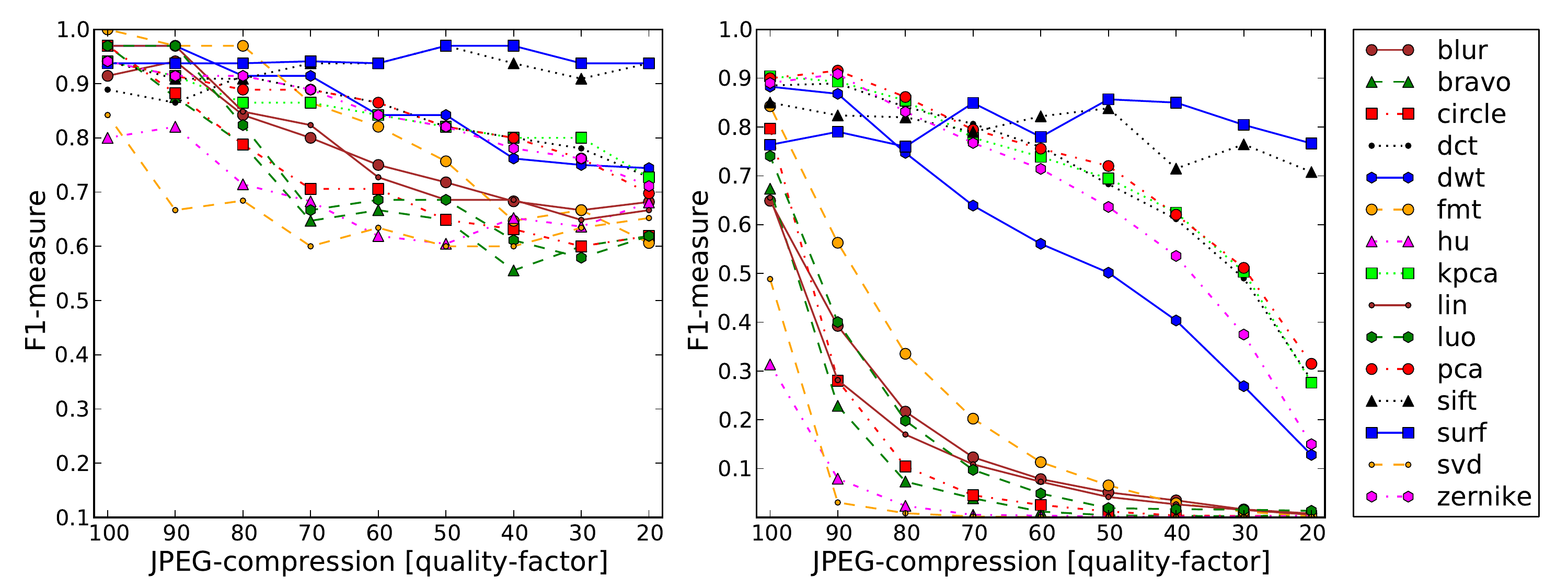}
	\\
		\includegraphics[width=\linewidth]{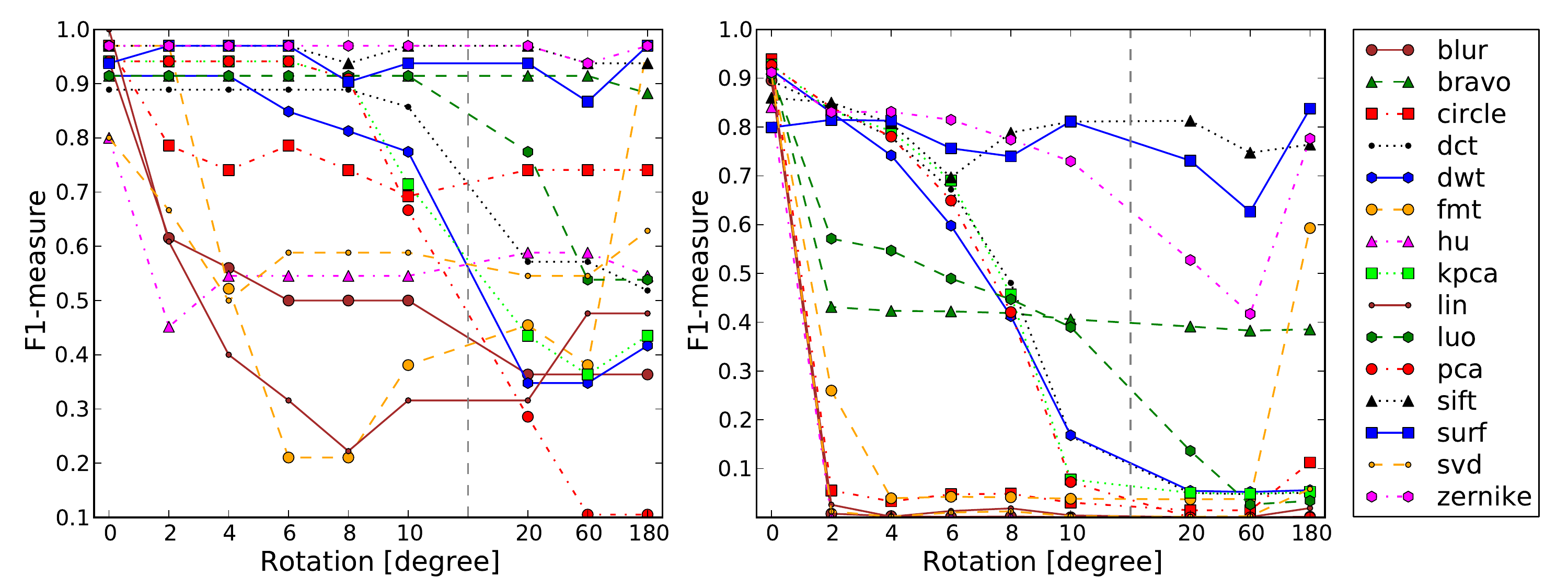}
	\\
		\includegraphics[width=\linewidth]{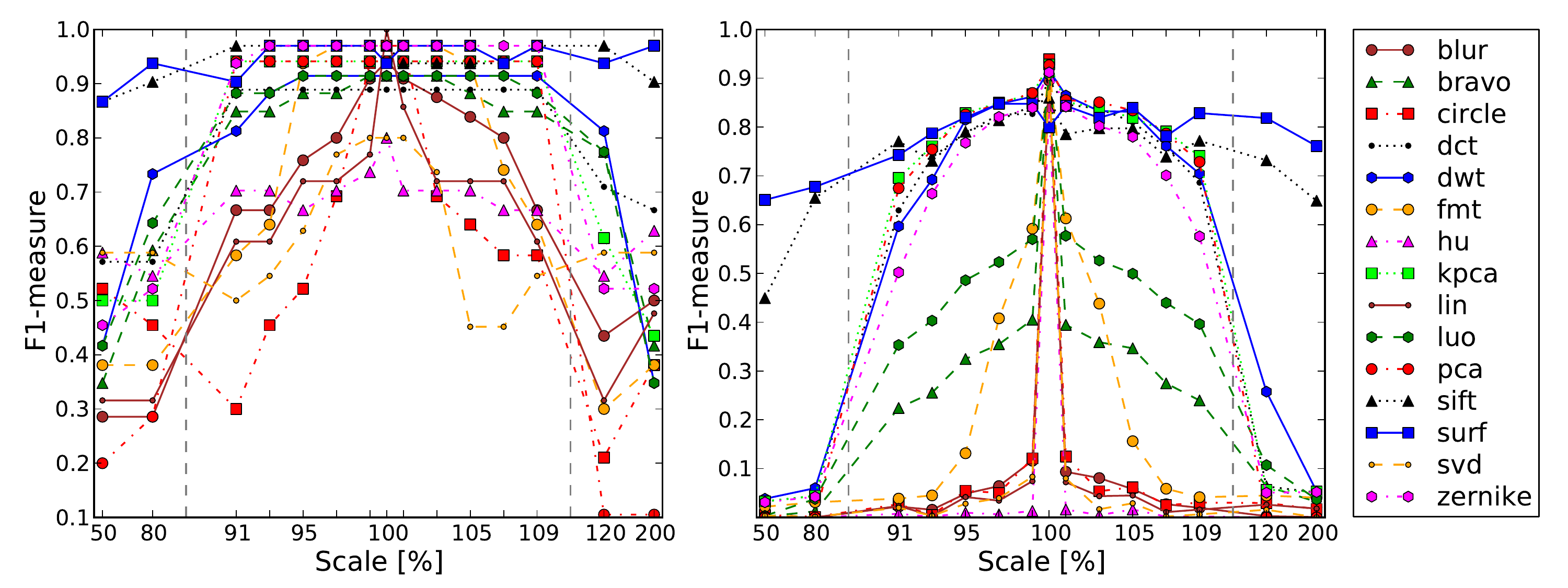}
	\\
		\includegraphics[width=\linewidth]{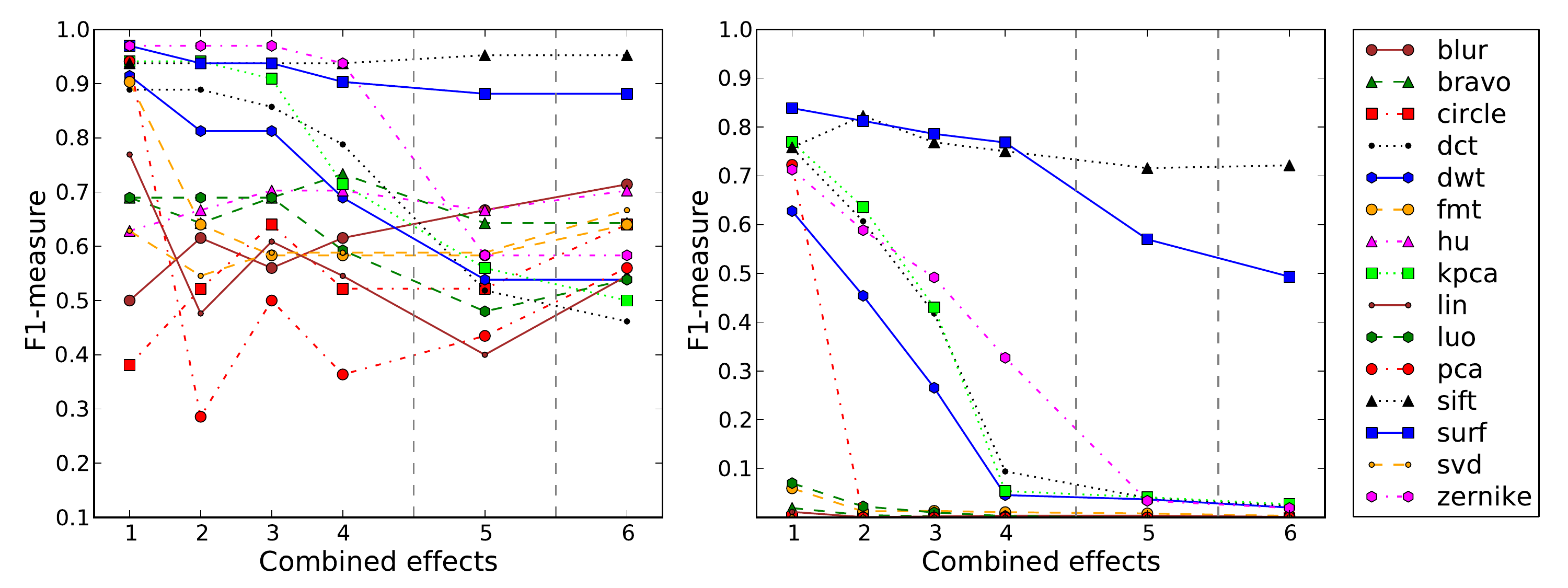}
	\caption{Performance in the category \emph{rough} at image level (left) and pixel level (right).}
	\label{fig:cat_rough}
\end{figure}

\begin{figure}[t!]
\centering
		\includegraphics[width=\linewidth]{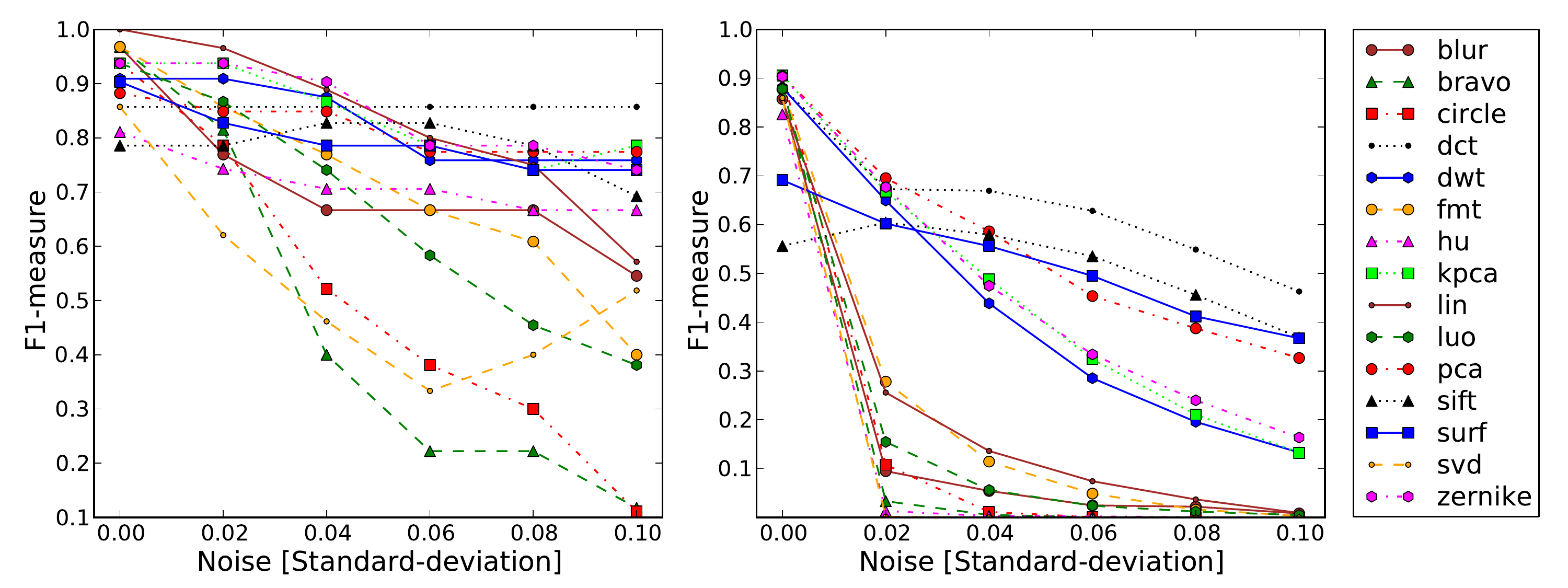}
	\\
		\includegraphics[width=\linewidth]{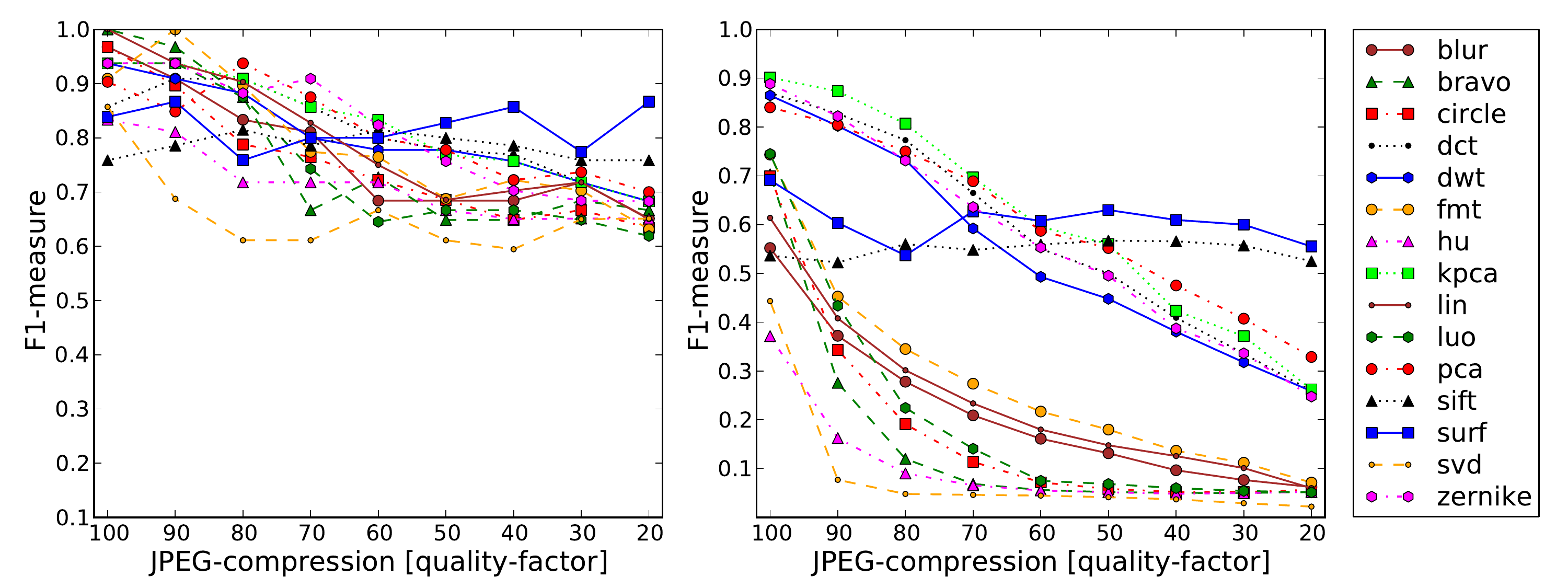}
	\\
		\includegraphics[width=\linewidth]{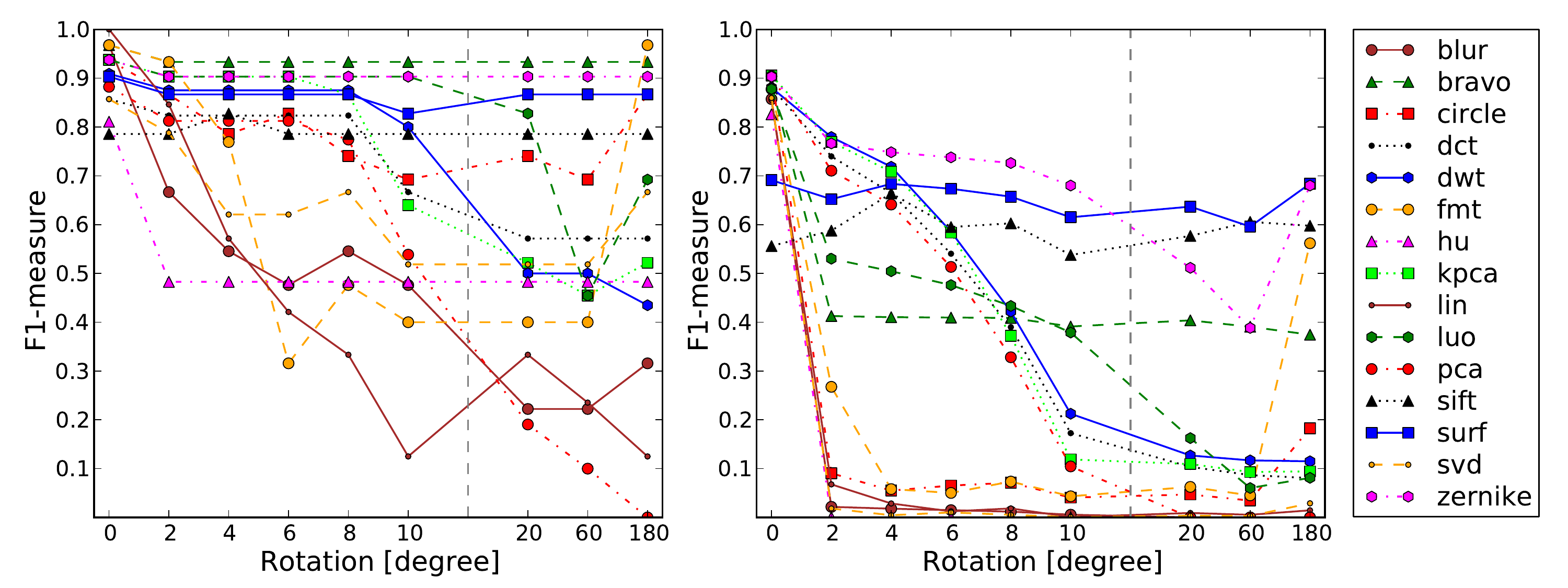}
	\\
		\includegraphics[width=\linewidth]{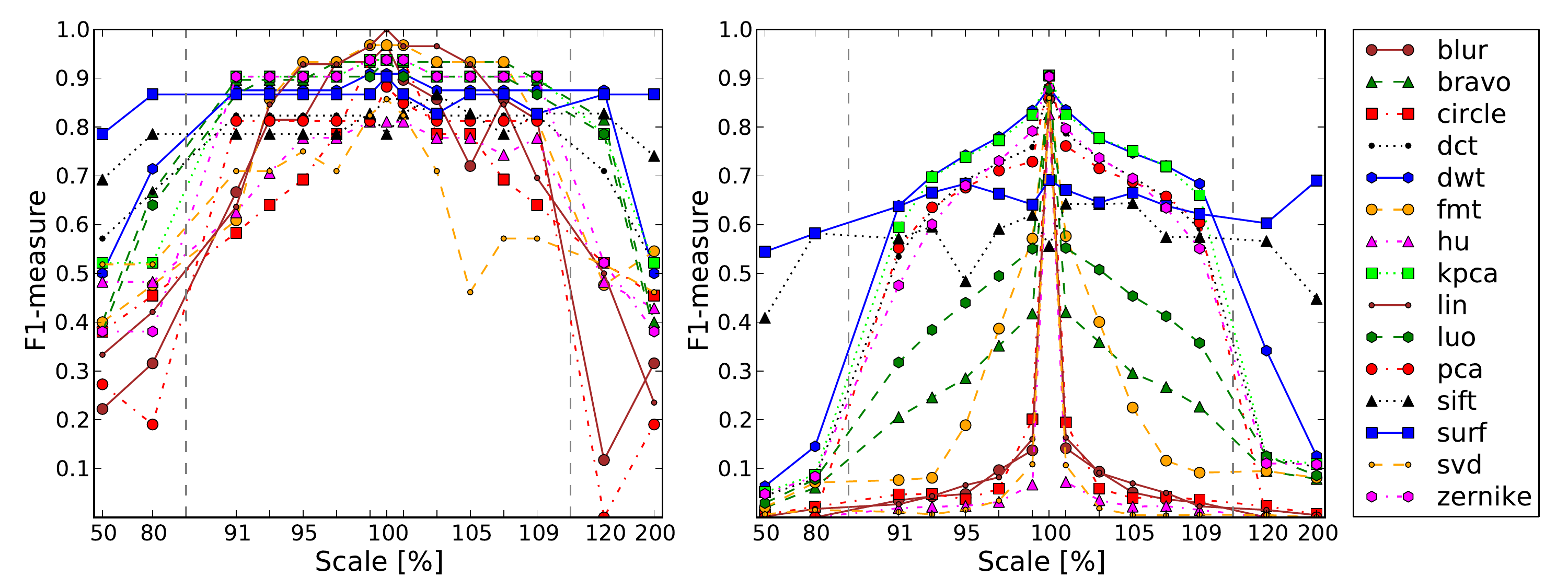}
	\\
		\includegraphics[width=\linewidth]{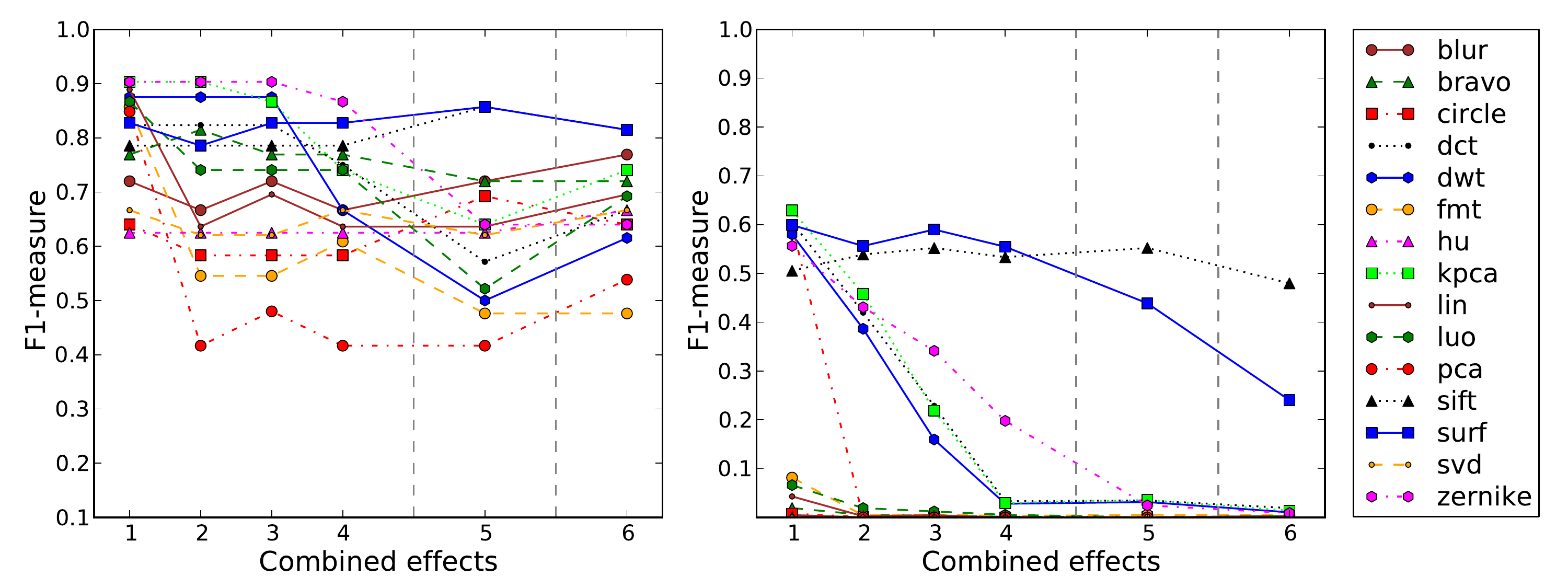}
	\caption{Performance in the category \emph{structure} at image level (left) and pixel level (right).}
	\label{fig:cat_structure}
\end{figure}

\tabRef{tab:img_nul_categories}, \figRef{fig:cat_smooth}, \figRef{fig:cat_rough} and \figRef{fig:cat_structure}
show the results per category. On the left side, the results are shown at image
level, on the right side at pixel level. The most notable performance shift
across categories is the relation between keypoint- and block-based methods.
\surf and \sift perform best in \emph{rough} (see \figRef{fig:cat_rough}),
while block-based methods often have an advantage in \emph{smooth} (see
\figRef{fig:cat_smooth}). In the category \emph{structure} (see
\figRef{fig:cat_structure}), block-based and keypoint-based methods perform
similarly well.

\section{Overview on the Forgery Database}\label{sec:overview_database}

We comment on the implementation of the benchmarking framework, with emphasis
on its extensibility. Then, we briefly discuss the choice of the manipulated
regions. Finally, we show all reference manipulations of the database, and
their associated ground truth.

Recall that the database consists of $48$ base images and $87$ prepared image
regions from
these images, called \emph{snippets}. Base images and snippets are spliced, to
simulate a close-to-real-world copy-move forgery. During splicing,
postprocessing artifacts can be added to the snippets and the final output
images. The software to create tampered images and the associated ground truth
is written in C++ and is best used with scripts written in Perl. Within the
Perl-scripts, a series of output images can be created by iterating over a
parameter space. For instance, all spliced images with JPEG compression are
obtained by iterating over the JPEG quality parameter space. We call one such
parameterization a \emph{configuration}. 
Upon acceptance of the paper, all images, snippets, code, scripts and
configuration files are made publicly available from our web page.  Note that
with the separate building blocks, it is straightforward to add a copy-move
tampering scenario that has not been addressed so far. For instance, assume
(hypothetically) that one aims to evaluate instead of Gaussian noise Laplacian
noise on the inserted regions.  Then, all that is required from the author is to
add a Laplacian noise function to the C++ code, and to add a matching
configuration to the perl scripts. 

\begin{figure*}[t!]
	\centering
\subfigure[ship]{\includegraphics[height=2.3cm]{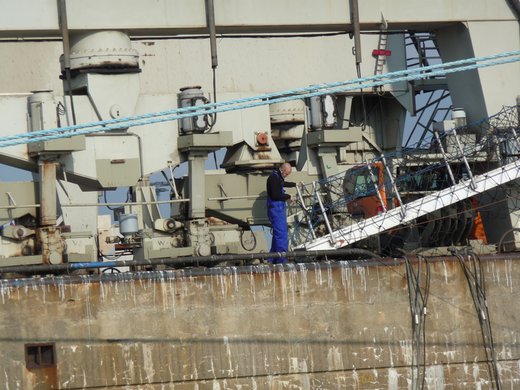}}       {} 
\subfigure[tapestry]{\includegraphics[height=2.3cm]{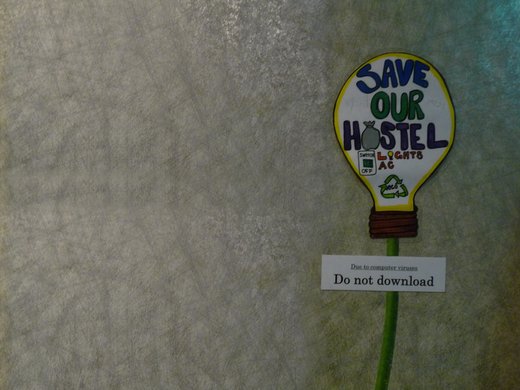}}   {} 
\subfigure[noise pattern]{\includegraphics[height=2.3cm]{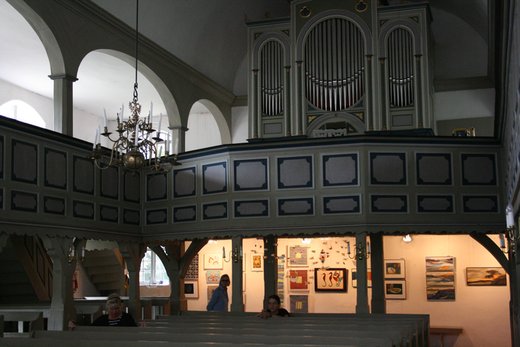}} {} 
\subfigure[four babies]{\includegraphics[height=2.3cm]{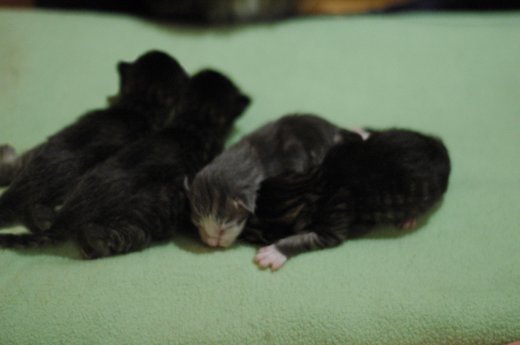}} {} 
\subfigure[cattle]{\includegraphics[height=2.3cm]{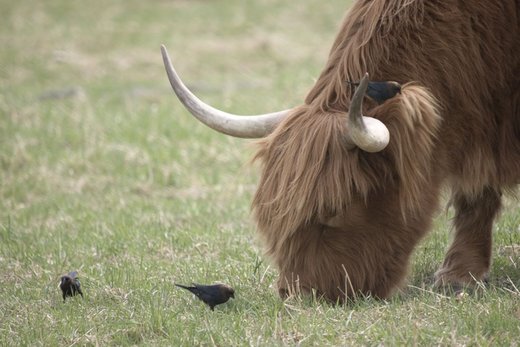}} 

	\vspace{1mm}
\subfigure[gt ship]{\includegraphics[height=2.3cm]{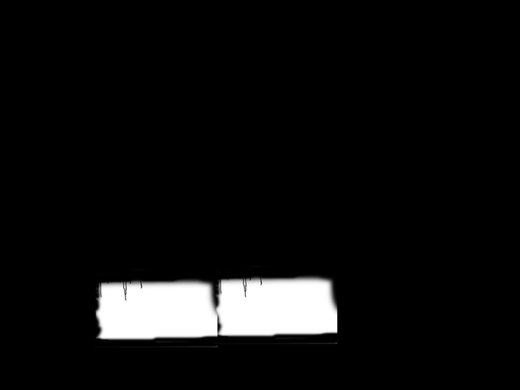}}       {} 
\subfigure[gt tapestry]{\includegraphics[height=2.3cm]{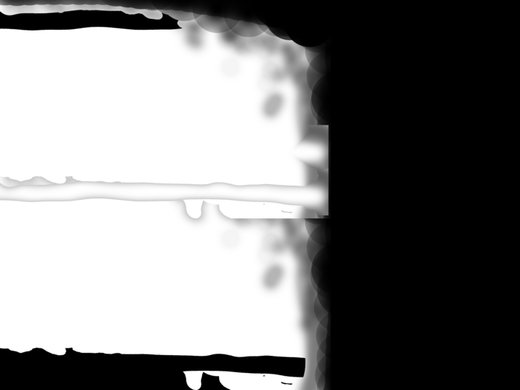}}   {} 
\subfigure[gt noise pattern]{\includegraphics[height=2.3cm]{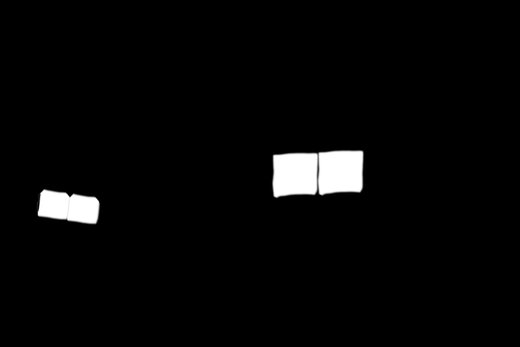}} {} 
\subfigure[gt four babies]{\includegraphics[height=2.3cm]{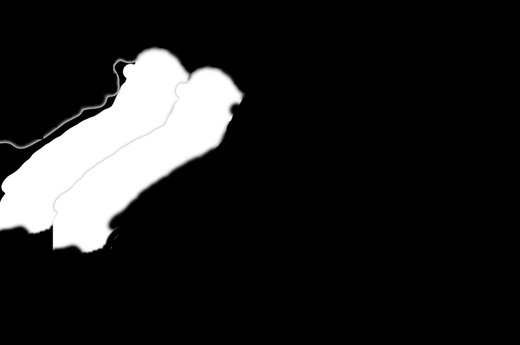}} {} 
\subfigure[gt cattle]{\includegraphics[height=2.3cm]{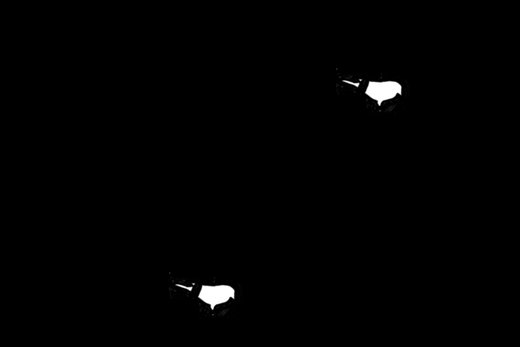}} 

	\vspace{1mm}
\subfigure[Scotland]{\includegraphics[height=2.3cm]{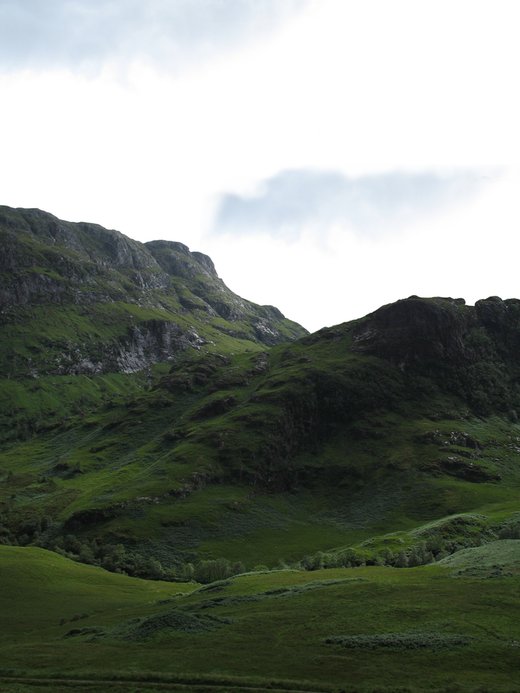}} {} 
\subfigure[hedge]{\includegraphics[height=2.3cm]{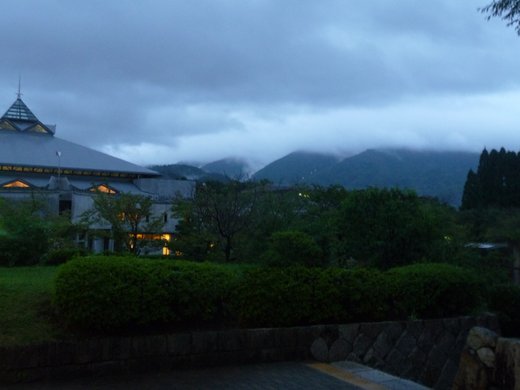}}   {} 
\subfigure[Japan tower]{\includegraphics[height=2.3cm]{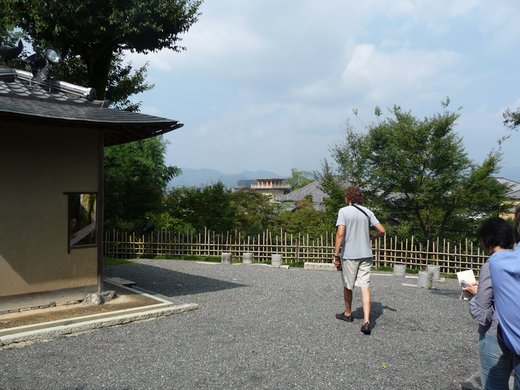}} {} 
\subfigure[motorcycle]{\includegraphics[height=2.3cm]{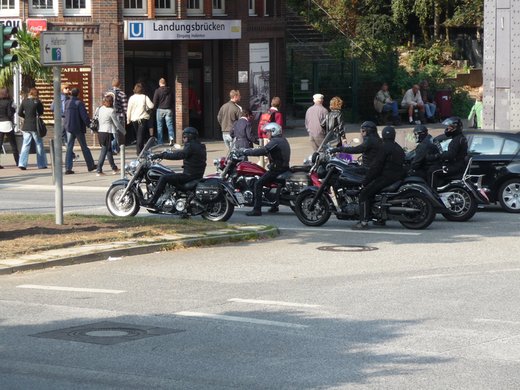}} {} 
\subfigure[mask]{\includegraphics[height=2.3cm]{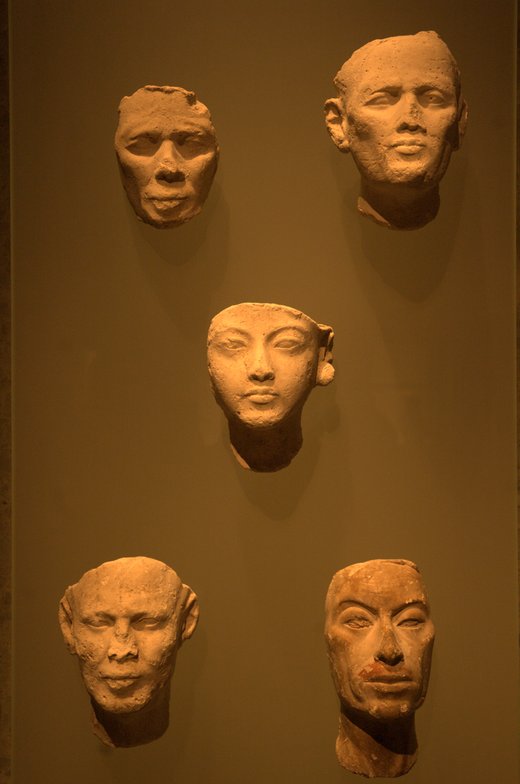}} {} 
\subfigure[berries]{\includegraphics[height=2.3cm]{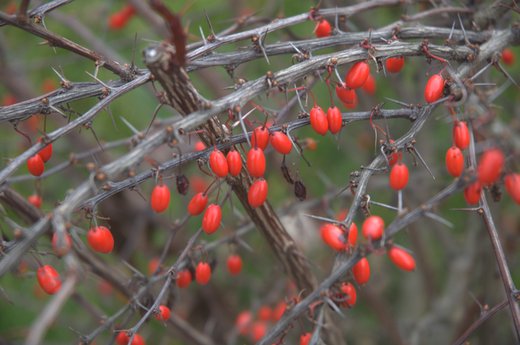}} 

	\vspace{1mm}
\subfigure[gt Scotland]{\includegraphics[height=2.3cm]{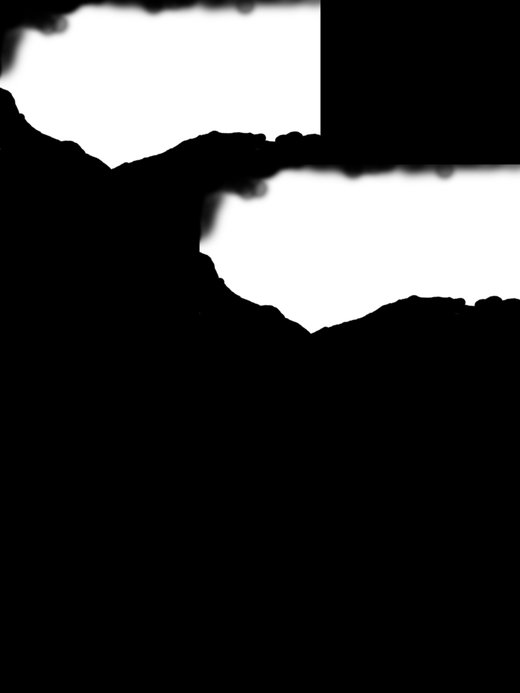}} {} 
\subfigure[gt hedge]{\includegraphics[height=2.3cm]{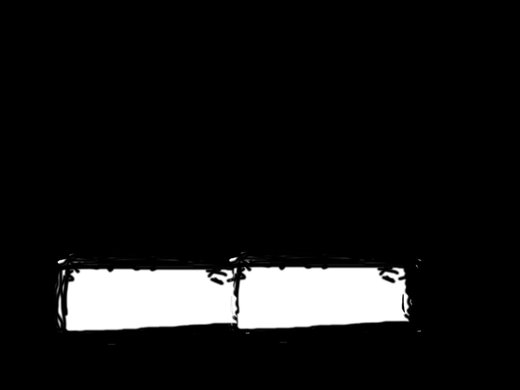}} {} 
\subfigure[gt Japan tower]{\includegraphics[height=2.3cm]{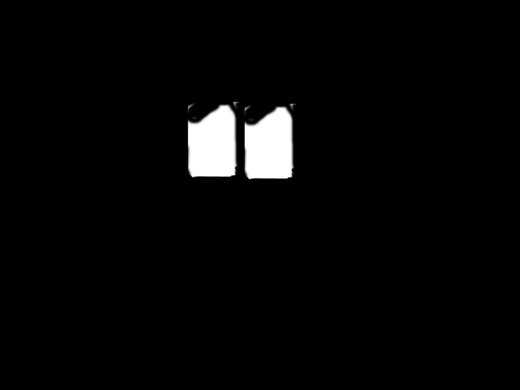}} {} 
\subfigure[gt motorcycle]{\includegraphics[height=2.3cm]{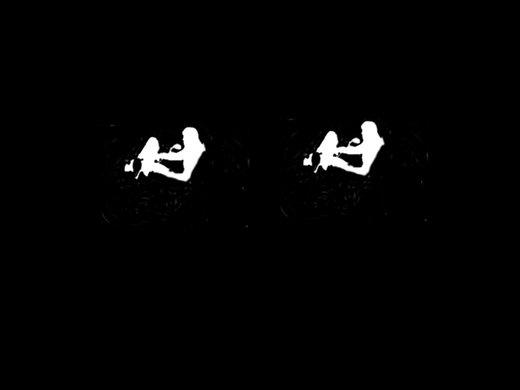}} {} 
\subfigure[gt mask]{\includegraphics[height=2.3cm]{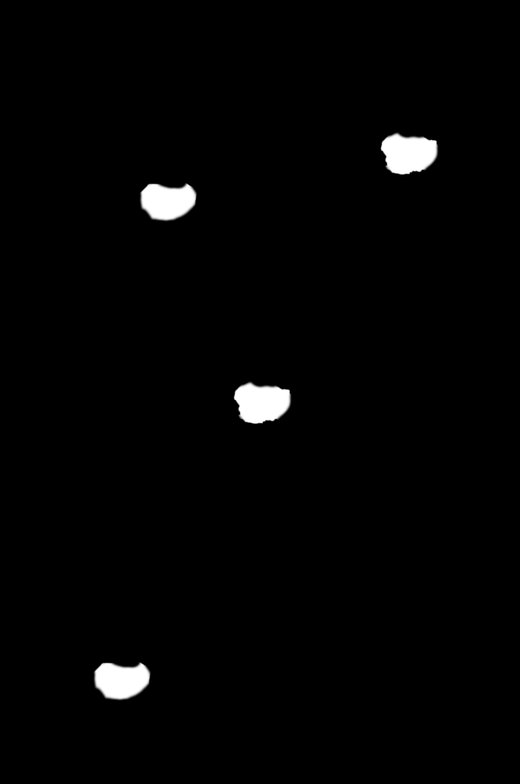}} {} 
\subfigure[gt berries]{\includegraphics[height=2.3cm]{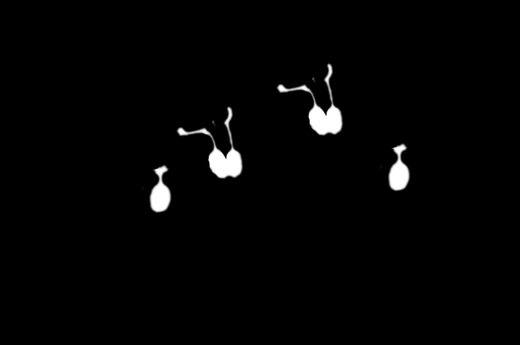}} 

	\vspace{1mm}
\subfigure[Malawi]{\includegraphics[height=2.3cm]{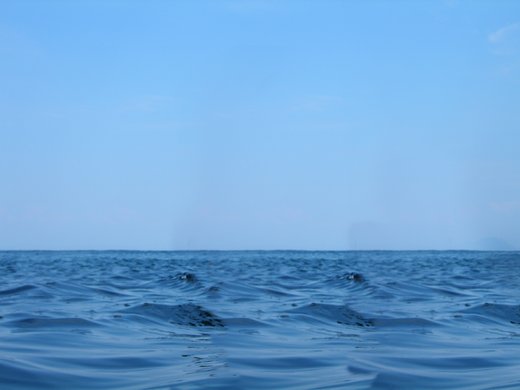}} {} 
\subfigure[sailing]{\includegraphics[height=2.3cm]{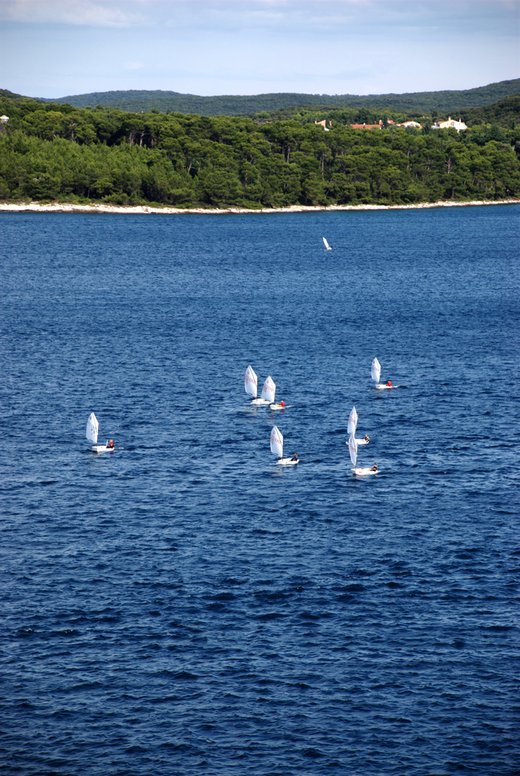}} {} 
\subfigure[disconnected shift]{\includegraphics[height=2.3cm]{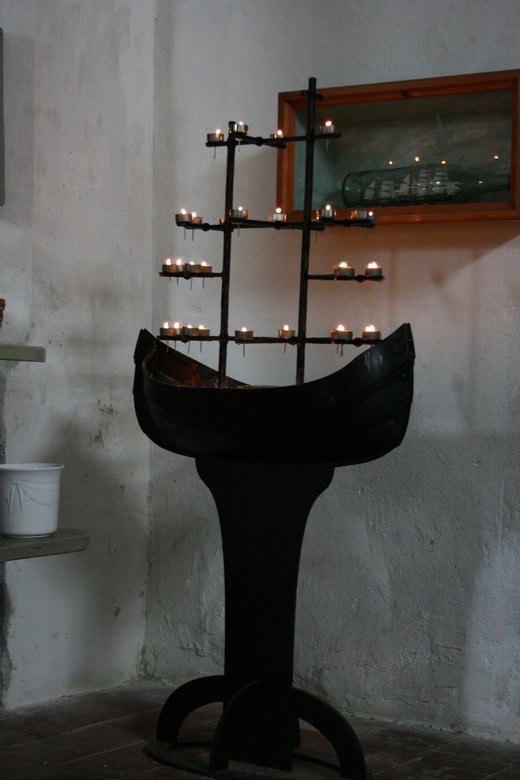}} {} 
\subfigure[swan]{\includegraphics[height=2.3cm]{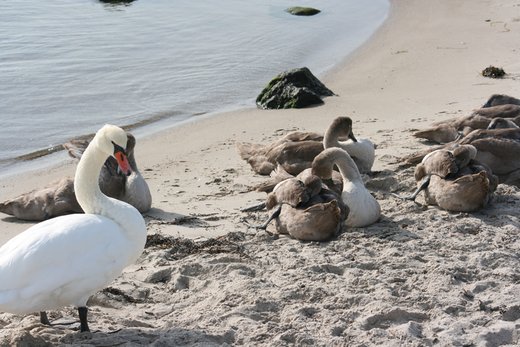}} {} 
\subfigure[wading]{\includegraphics[height=2.3cm]{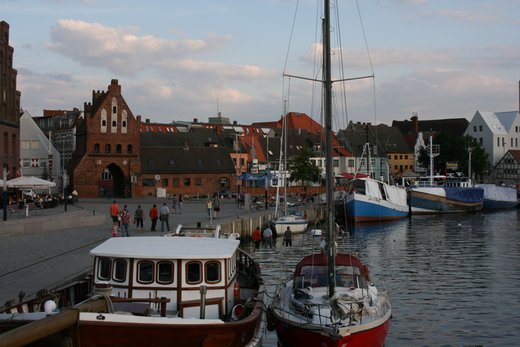}}  {} 
\subfigure[sails]{\includegraphics[height=2.3cm]{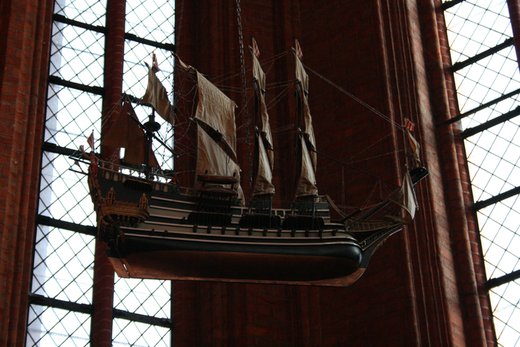}} 

	\vspace{1mm}
\subfigure[gt Malawi]{\includegraphics[height=2.3cm]{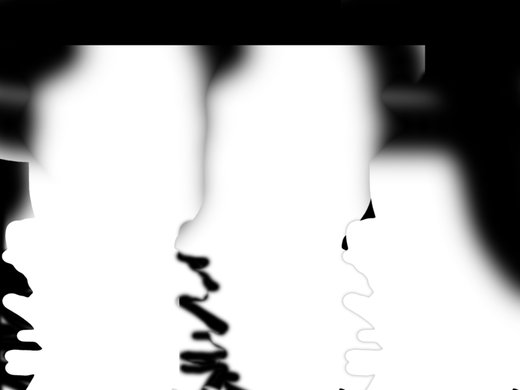}} {} 
\subfigure[gt sailing]{\includegraphics[height=2.3cm]{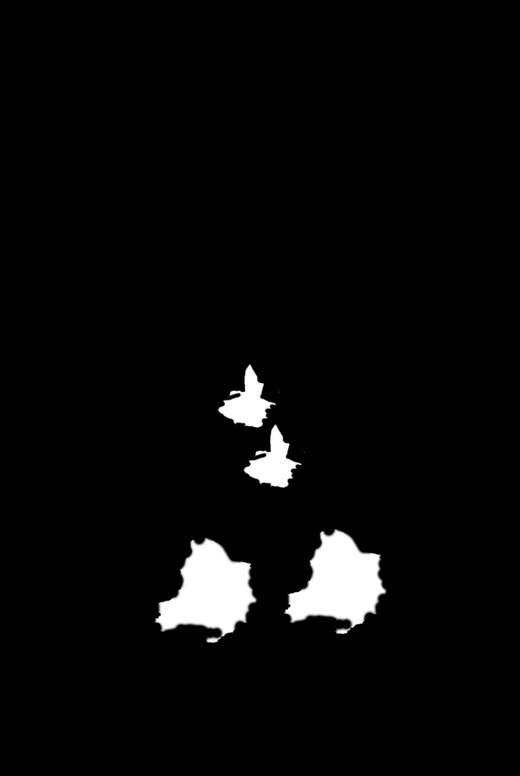}} {} 
\subfigure[gt disconnected shift]{\includegraphics[height=2.3cm]{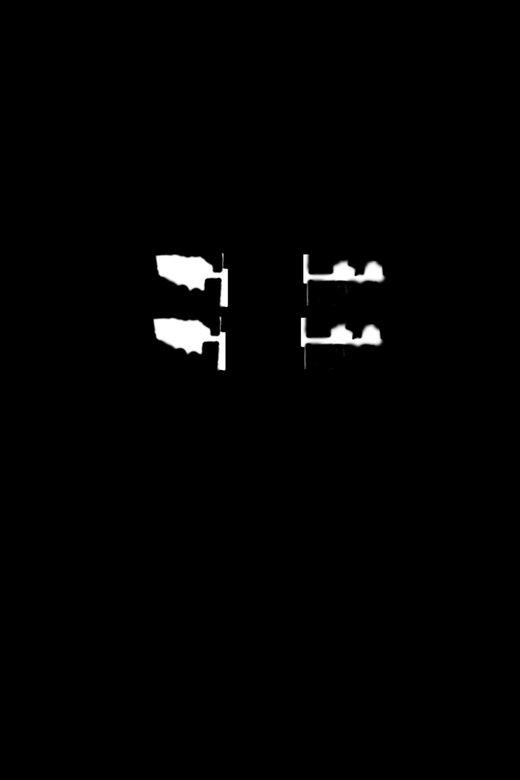}} {} 
\subfigure[gt swan]{\includegraphics[height=2.3cm]{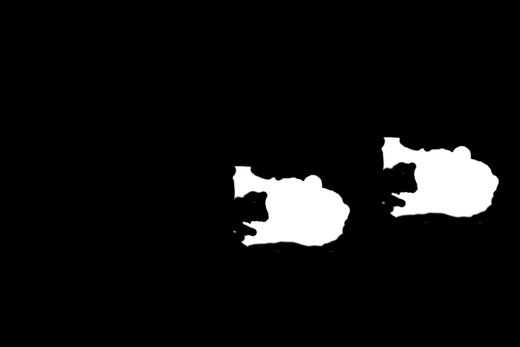}} {} 
\subfigure[gt wading]{\includegraphics[height=2.3cm]{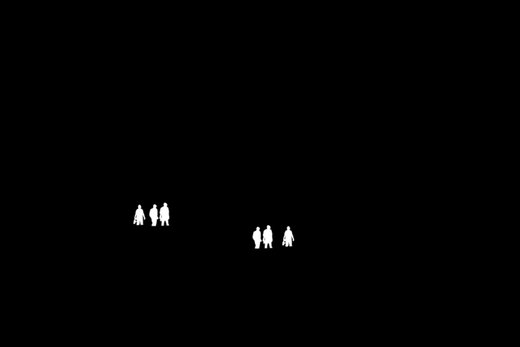}} {} 
\subfigure[gt sails]{\includegraphics[height=2.3cm]{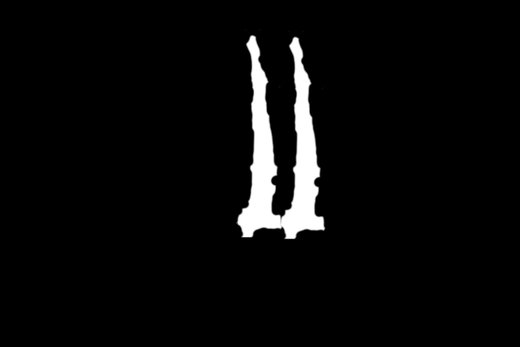}} 
\caption{Database images from the category \emph{smooth}, with annotated ground truth for the
``reference forgery'', \ie without rotation or scaling of the copied region.}
\label{fig:db_images1}
\end{figure*}

\begin{figure*}[t!]
	\centering
\subfigure[no beach]{\includegraphics[height=2.3cm]{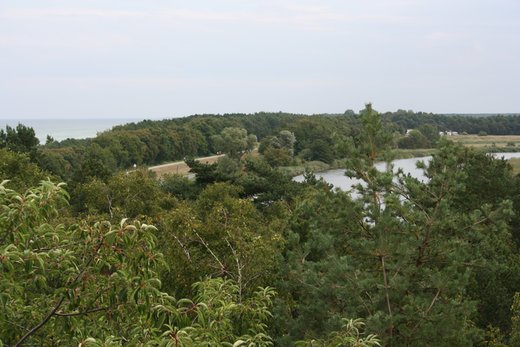}} {} 
\subfigure[central park]{\includegraphics[height=2.3cm]{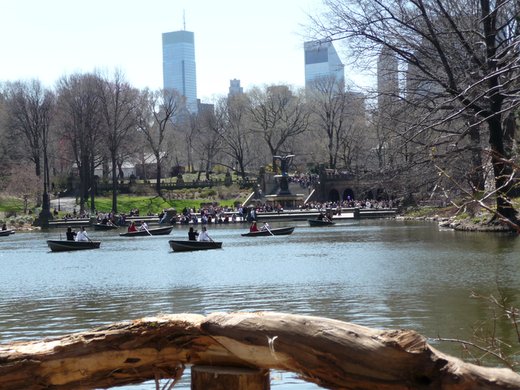}} {} 
\subfigure[supermarket]{\includegraphics[height=2.3cm]{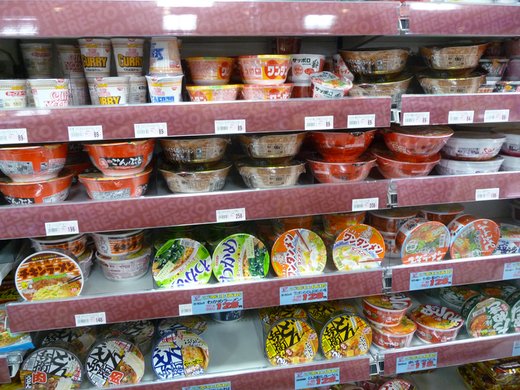}} {} 
\subfigure[fisherman]{\includegraphics[height=2.3cm]{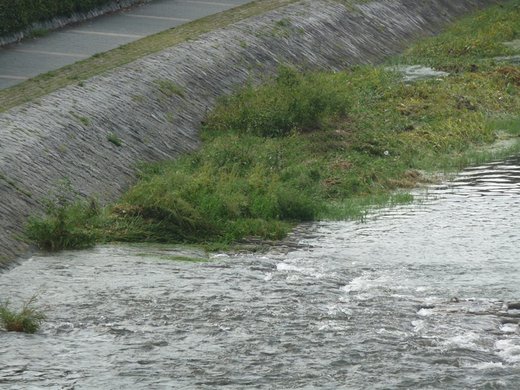}} {}  
\subfigure[clean walls]{\includegraphics[height=2.3cm]{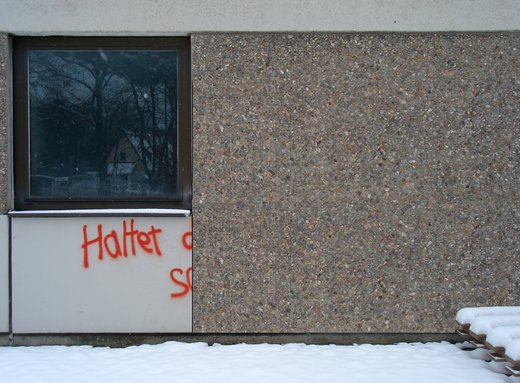}} 

	\vspace{1mm}
\subfigure[gt no beach]{\includegraphics[height=2.3cm]{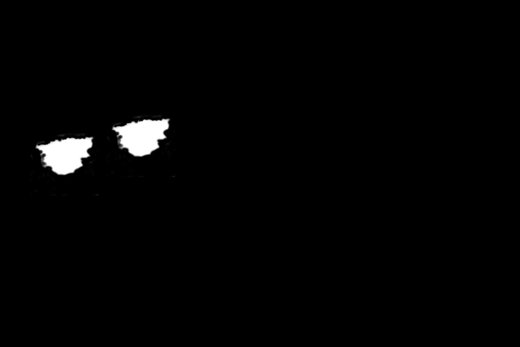}} {} 
\subfigure[gt central park]{\includegraphics[height=2.3cm]{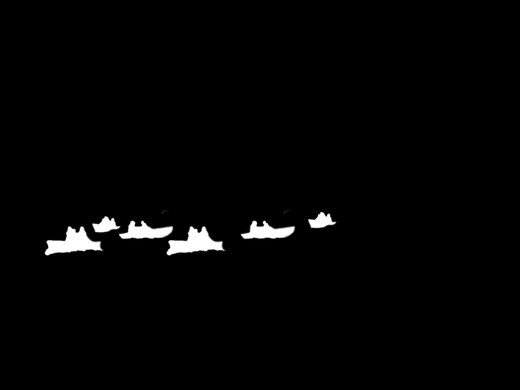}} {} 
\subfigure[gt supermarket]{\includegraphics[height=2.3cm]{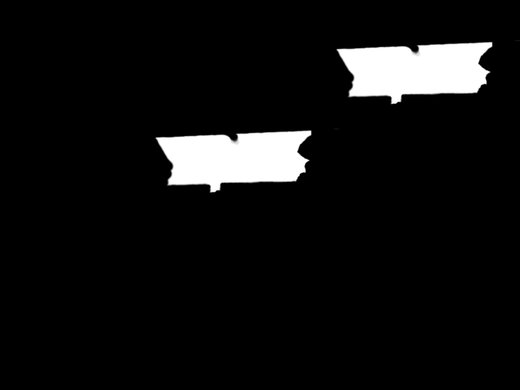}} {}  
\subfigure[gt fisherman]{\includegraphics[height=2.3cm]{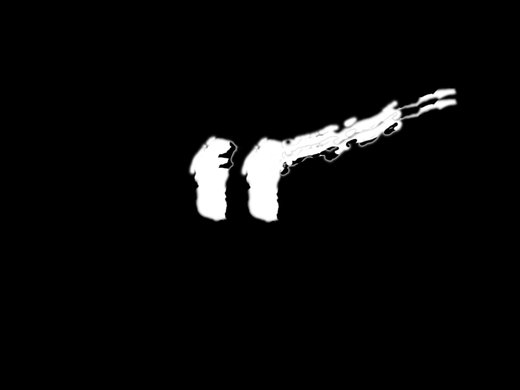}} {}  
\subfigure[gt clean walls]{\includegraphics[height=2.3cm]{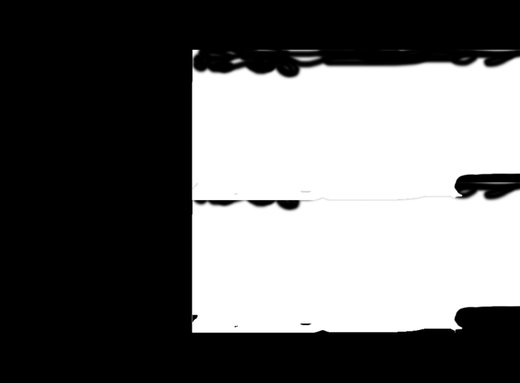}} 

	\vspace{1mm}
\subfigure[stone ghost]{\includegraphics[height=2.3cm]{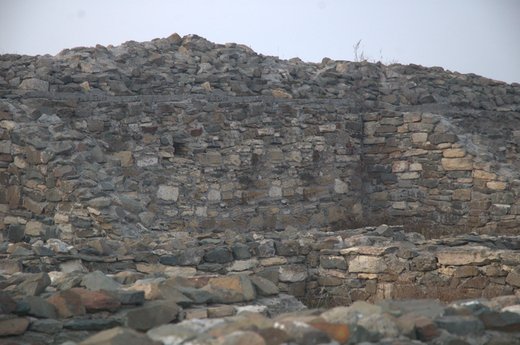}} {} 
\subfigure[white]{\includegraphics[height=2.3cm]{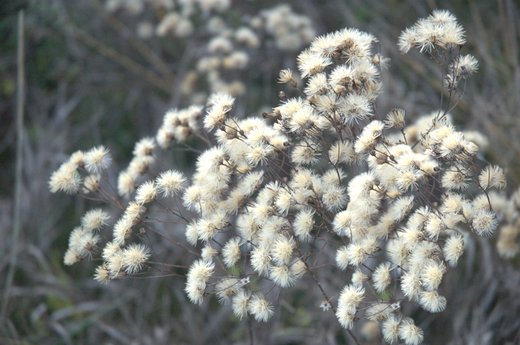}} {} 
\subfigure[writing history]{\includegraphics[height=2.3cm]{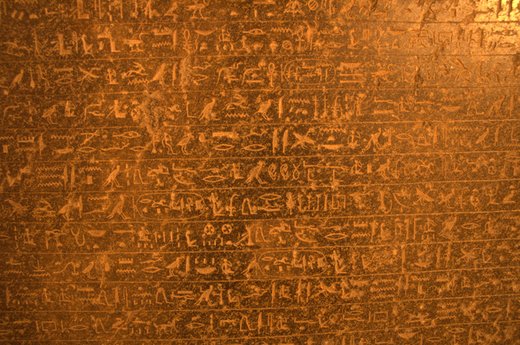}} {} 
\subfigure[barrier]{\includegraphics[height=2.3cm]{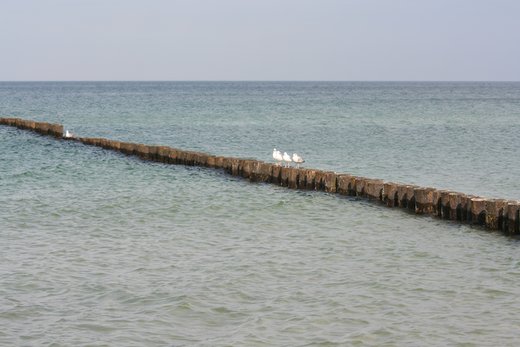}} {} 
\subfigure[red tower]{\includegraphics[height=2.3cm]{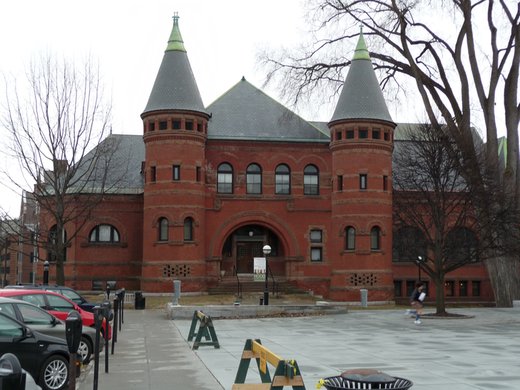}} 

	\vspace{1mm}
\subfigure[gt stone ghost]{\includegraphics[height=2.3cm]{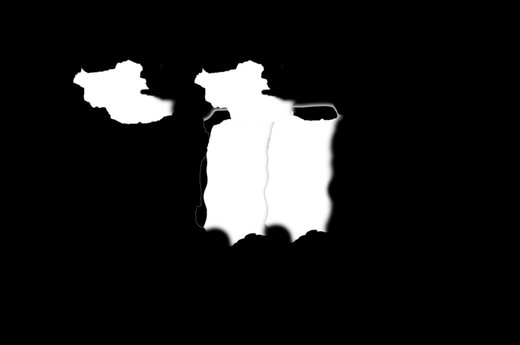}} {} 
\subfigure[gt white]{\includegraphics[height=2.3cm]{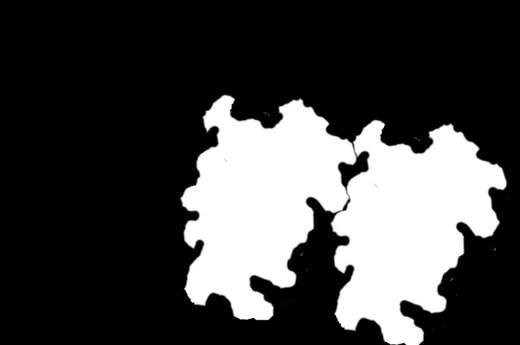}} {} 
\subfigure[gt writing history]{\includegraphics[height=2.3cm]{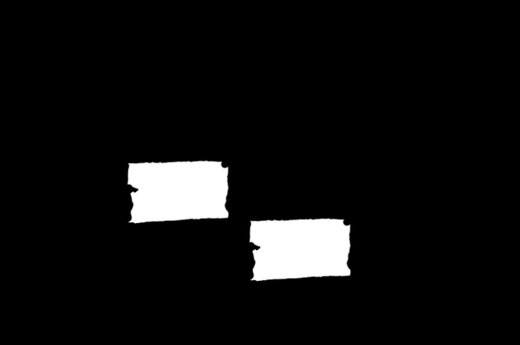}} {} 
\subfigure[gt barrier]{\includegraphics[height=2.3cm]{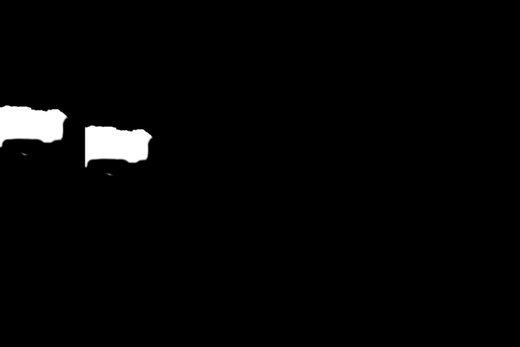}} {} 
\subfigure[gt red tower]{\includegraphics[height=2.3cm]{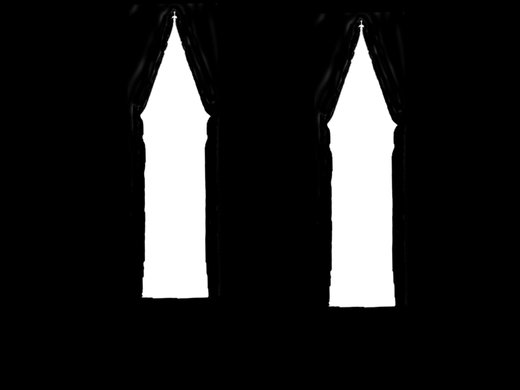}} 

	\vspace{1mm}
\subfigure[christmas hedge]{\includegraphics[height=2.3cm]{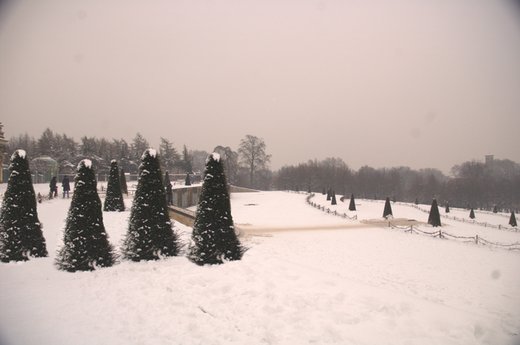}} {} 
\subfigure[lone cat]{\includegraphics[height=2.3cm]{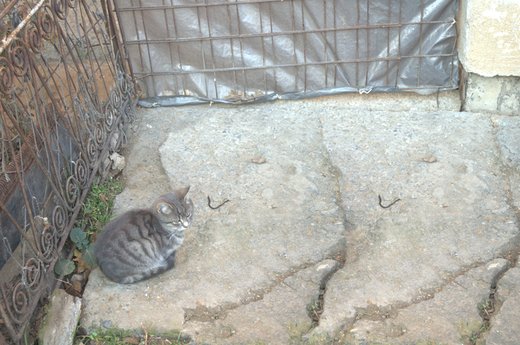}} {} 
\subfigure[beach wood]{\includegraphics[height=2.3cm]{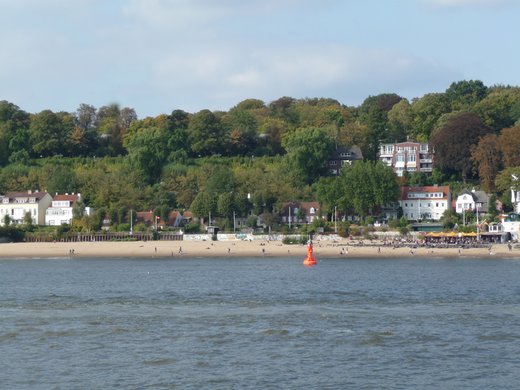}} {} 
\subfigure[kore]{\includegraphics[height=2.3cm]{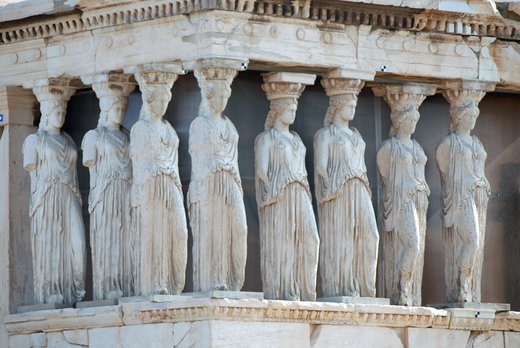}} {} 
\subfigure[tree]{\includegraphics[height=2.3cm]{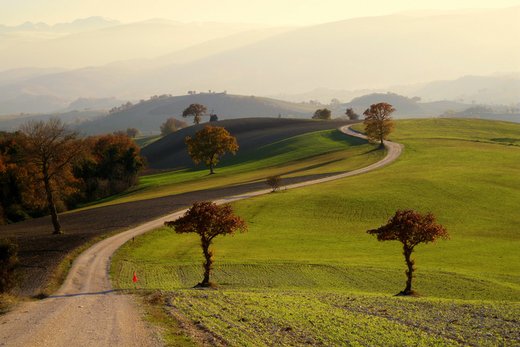}} 

	\vspace{1mm}
\subfigure[gt christmas hedge]{\includegraphics[height=2.3cm]{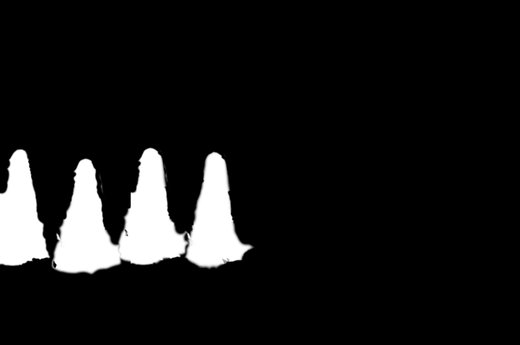}} {} 
\subfigure[gt lone cat]{\includegraphics[height=2.3cm]{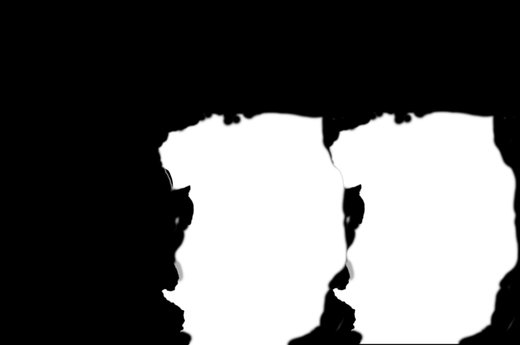}} {} 
\subfigure[gt beach wood]{\includegraphics[height=2.3cm]{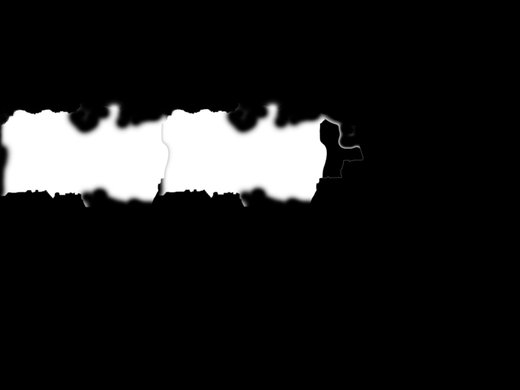}} {} 
\subfigure[gt kore]{\includegraphics[height=2.3cm]{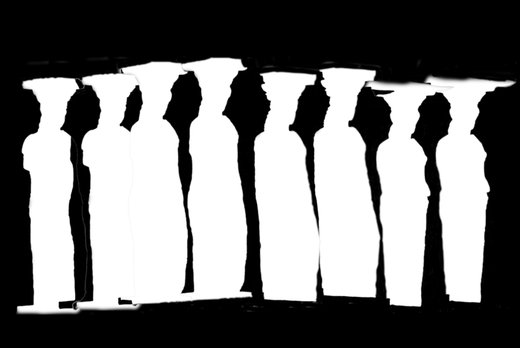}} {} 
\subfigure[gt tree]{\includegraphics[height=2.3cm]{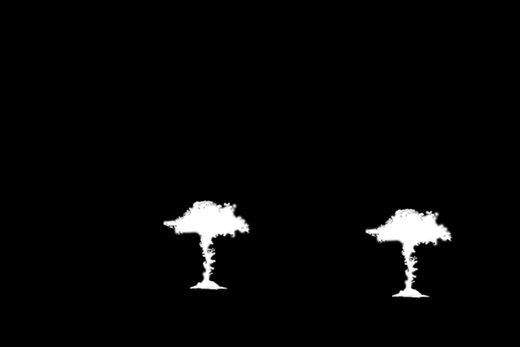}} 

	\vspace{1mm}
\subfigure[threehundred]{\includegraphics[height=2.3cm]{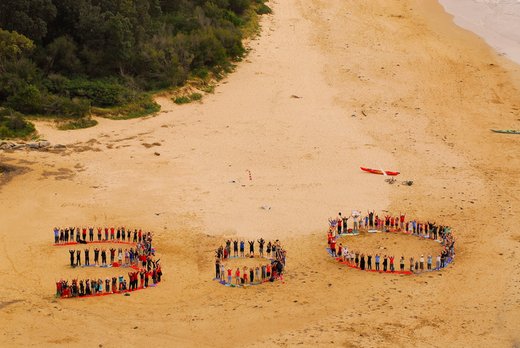}} {} 
\subfigure[gt threehundred]{\includegraphics[height=2.3cm]{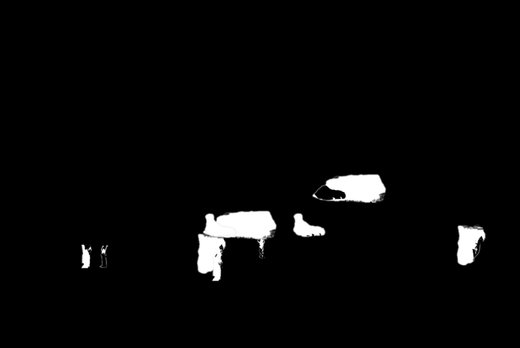}} 
\caption{Database images from the category \emph{rough}, with annotated ground truth for the
``reference forgery'', \ie without rotation or scaling of the copied region.}
\label{fig:db_images2}
\end{figure*}

\begin{figure*}[t!]
	\centering
\subfigure[horses]{\includegraphics[height=2.25cm]{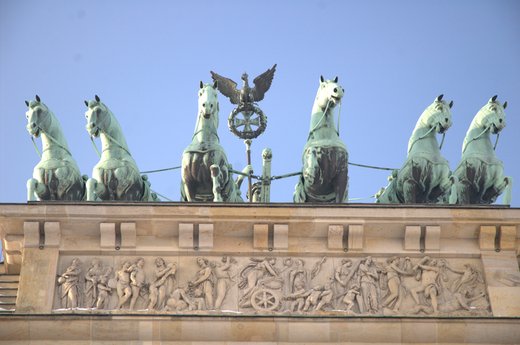}} {} 
\subfigure[giraffe]{\includegraphics[height=2.25cm]{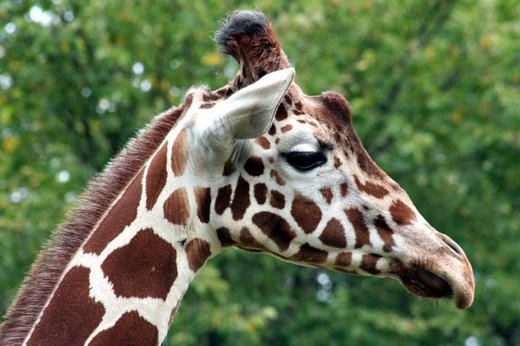}} {} 
\subfigure[extension]{\includegraphics[height=2.25cm]{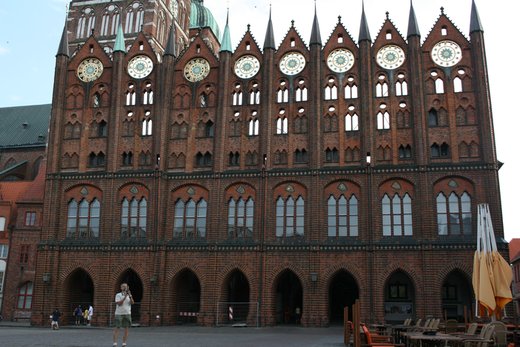}} {} 
\subfigure[jellyfish]{\includegraphics[height=2.25cm]{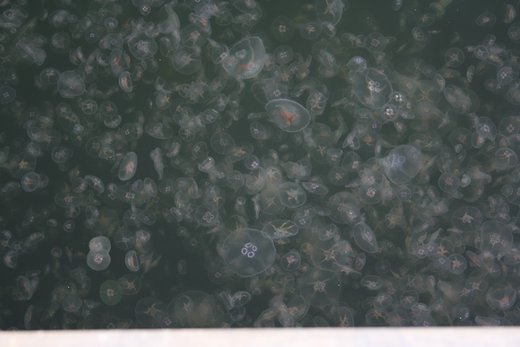}} {} 
\subfigure[sweets]{\includegraphics[height=2.25cm]{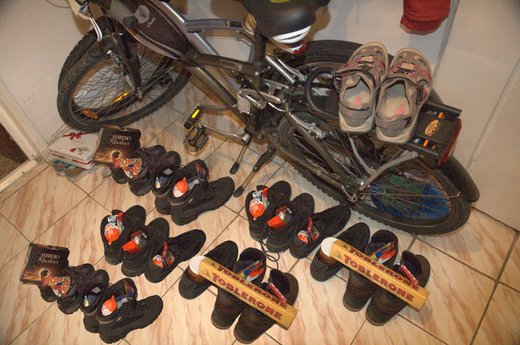}} 

	\vspace{1mm}
\subfigure[gt horses]{\includegraphics[height=2.25cm]{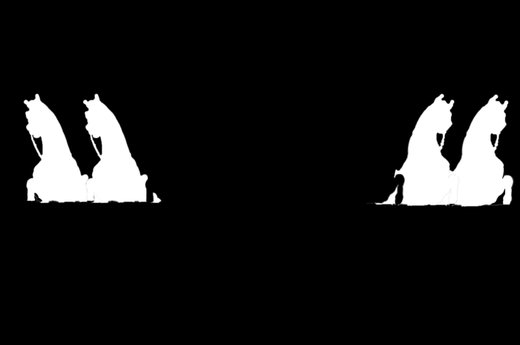}} {} 
\subfigure[gt giraffe]{\includegraphics[height=2.25cm]{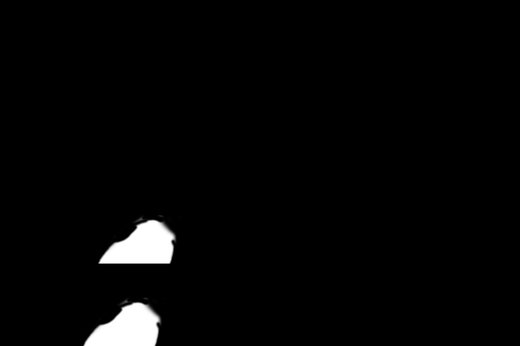}} {} 
\subfigure[gt extension]{\includegraphics[height=2.25cm]{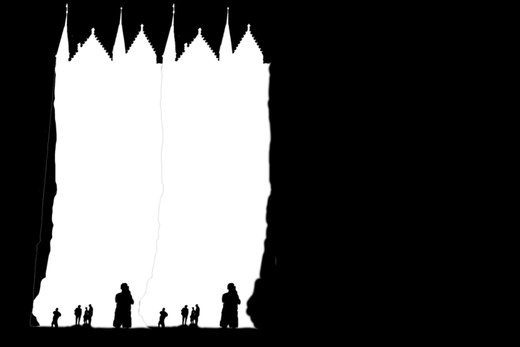}} {} 
\subfigure[gt jellyfish chaos]{\includegraphics[height=2.25cm]{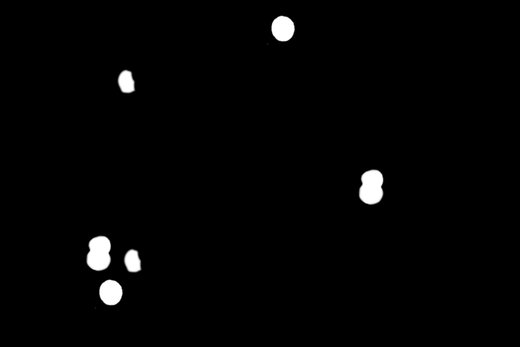}} {} 
\subfigure[gt sweets]{\includegraphics[height=2.25cm]{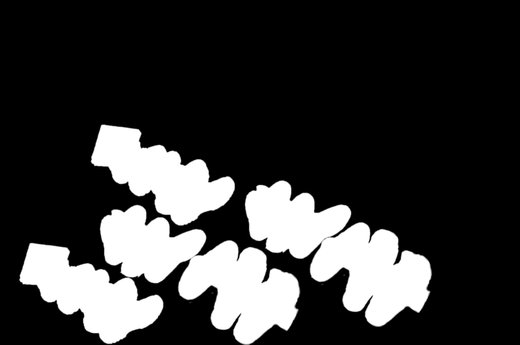}} 

	\vspace{1mm}
\subfigure[window]{\includegraphics[height=2.3cm]{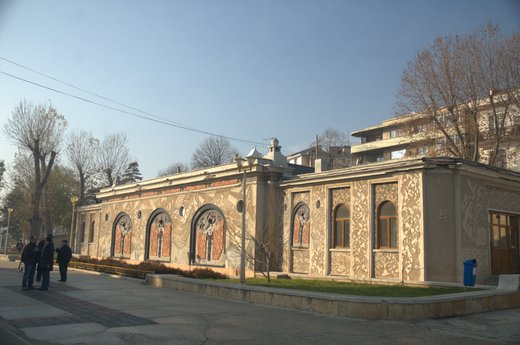}} {} 
\subfigure[Egyptian]{\includegraphics[height=2.3cm]{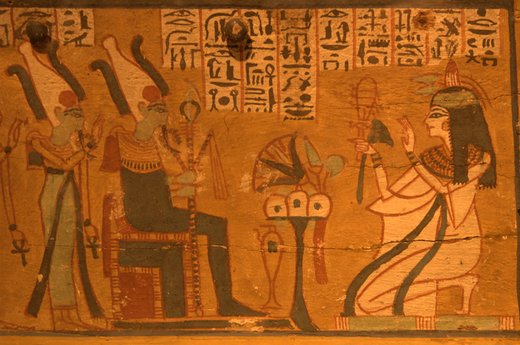}} {} 
\subfigure[statue]{\includegraphics[height=2.3cm]{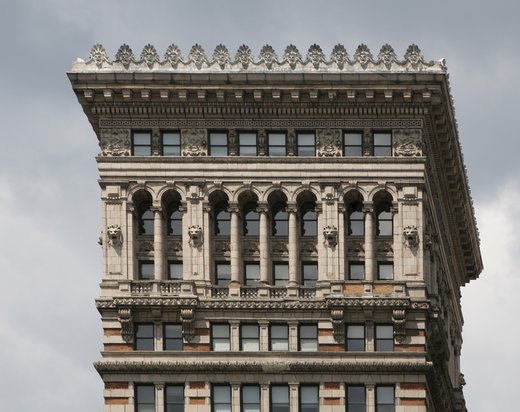}} {} 
\subfigure[Mykene]{\includegraphics[height=2.3cm]{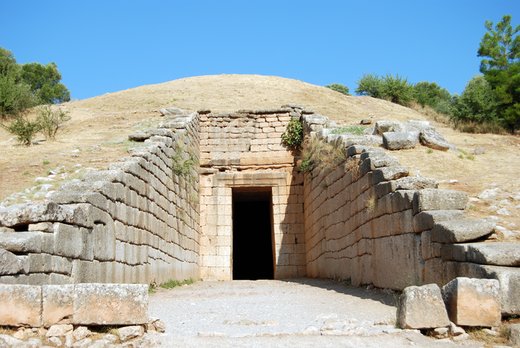}} {} 
\subfigure[knight moves]{\includegraphics[height=2.3cm]{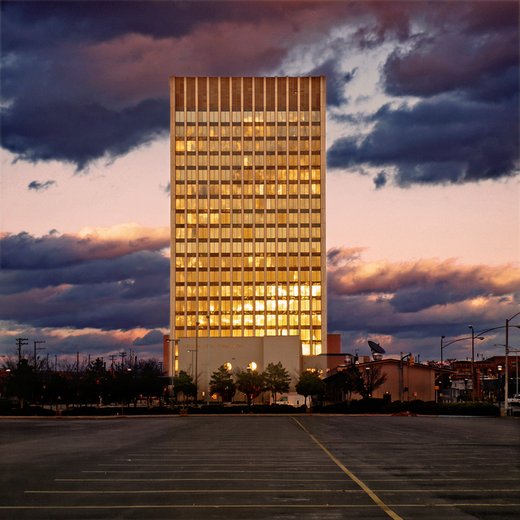}} 

	\vspace{1mm}
\subfigure[gt window]{\includegraphics[height=2.3cm]{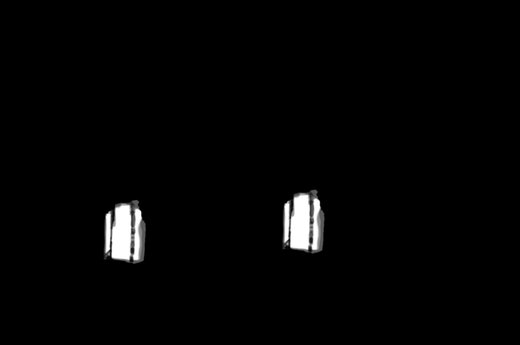}} {} 
\subfigure[gt Egyptian]{\includegraphics[height=2.3cm]{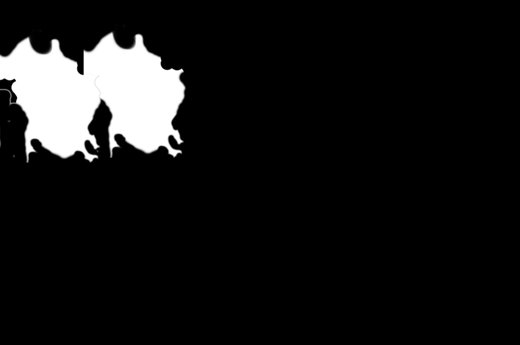}} {} 
\subfigure[gt statue]{\includegraphics[height=2.3cm]{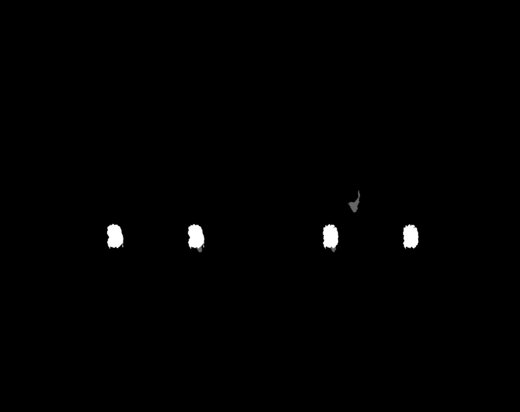}} {} 
\subfigure[gt Mykene]{\includegraphics[height=2.3cm]{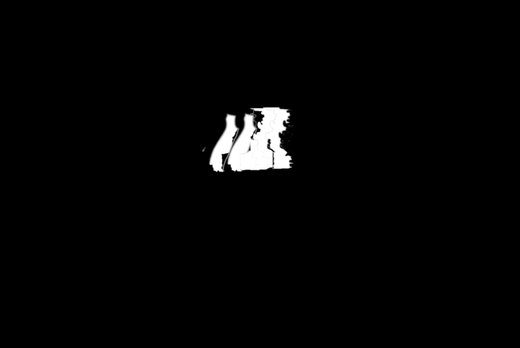}} {} 
\subfigure[gt knight moves]{\includegraphics[height=2.3cm]{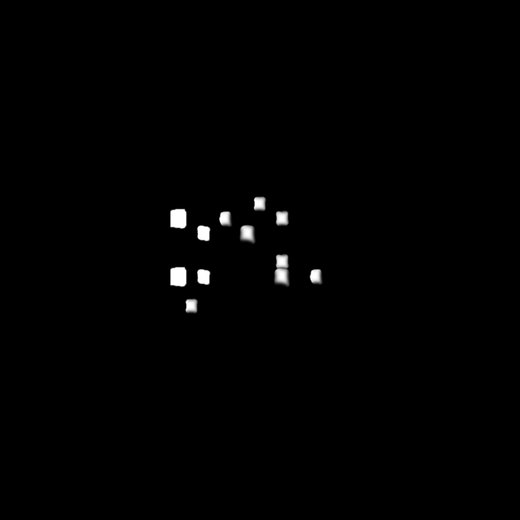}} 

	\vspace{1mm}
\subfigure[port]{\includegraphics[height=2.3cm]{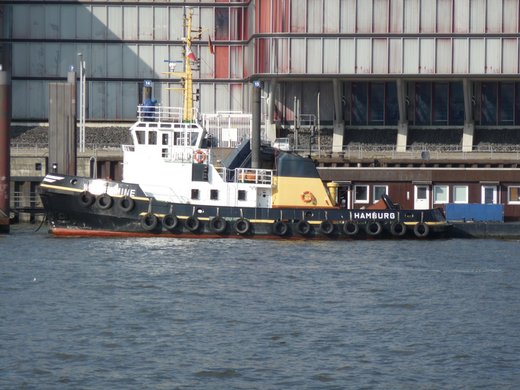}} {}    
\subfigure[fountain]{\includegraphics[height=2.3cm]{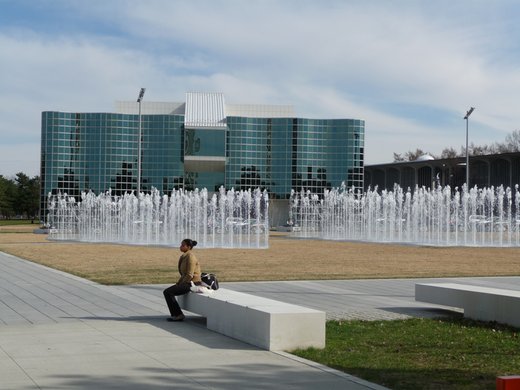}} {} 
\subfigure[dark and bright]{\includegraphics[height=2.3cm]{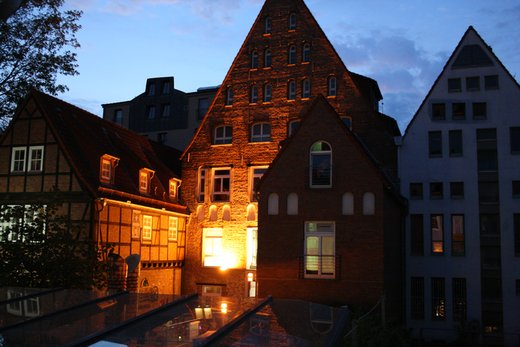}} {} 
\subfigure[bricks]{\includegraphics[height=2.3cm]{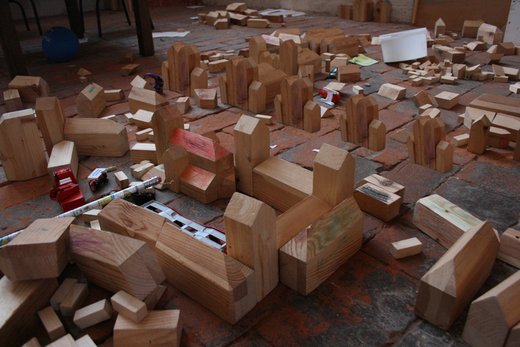}} {}  
\subfigure[wood carvings]{\includegraphics[height=2.3cm]{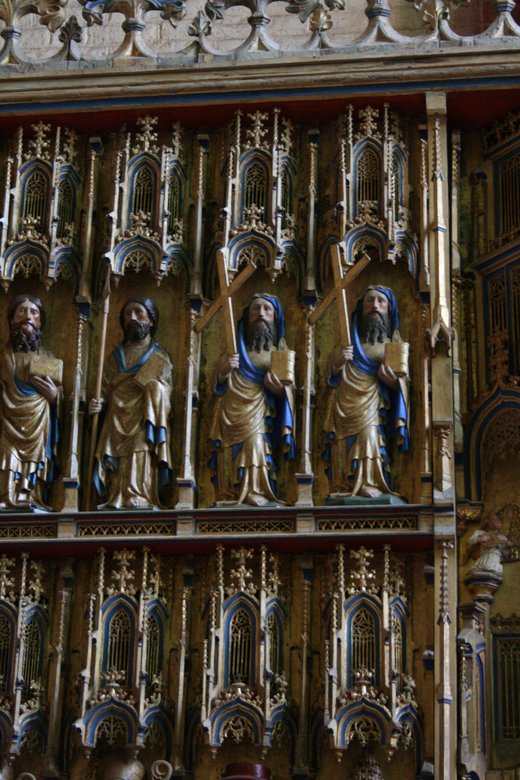}} 

	\vspace{1mm}
\subfigure[gt port]{\includegraphics[height=2.3cm]{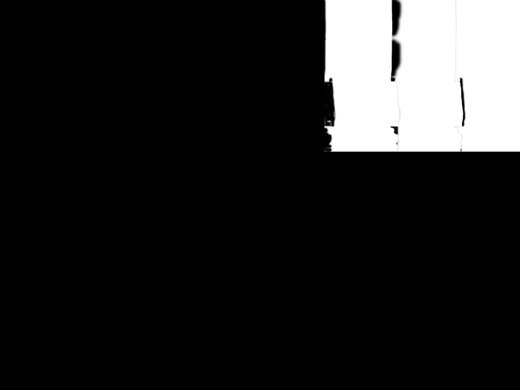}} {} 
\subfigure[gt fountain]{\includegraphics[height=2.3cm]{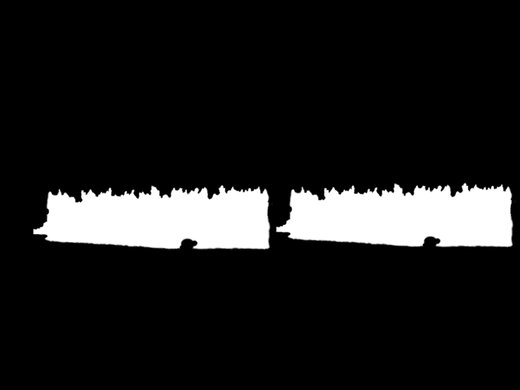}} {} 
\subfigure[gt dark and bright]{\includegraphics[height=2.3cm]{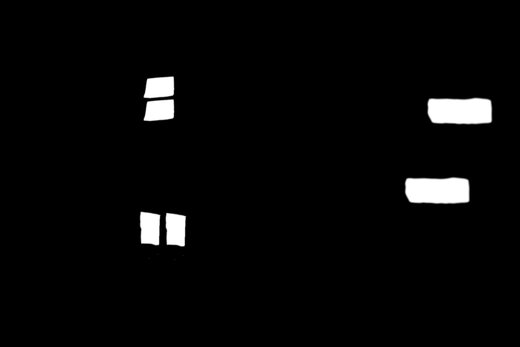}} {} 
\subfigure[gt bricks]{\includegraphics[height=2.3cm]{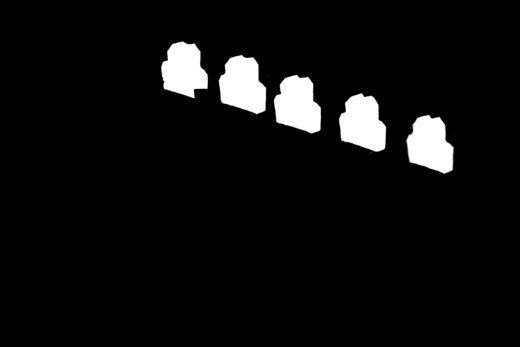}} {} 
\subfigure[gt wood carvings]{\includegraphics[height=2.3cm]{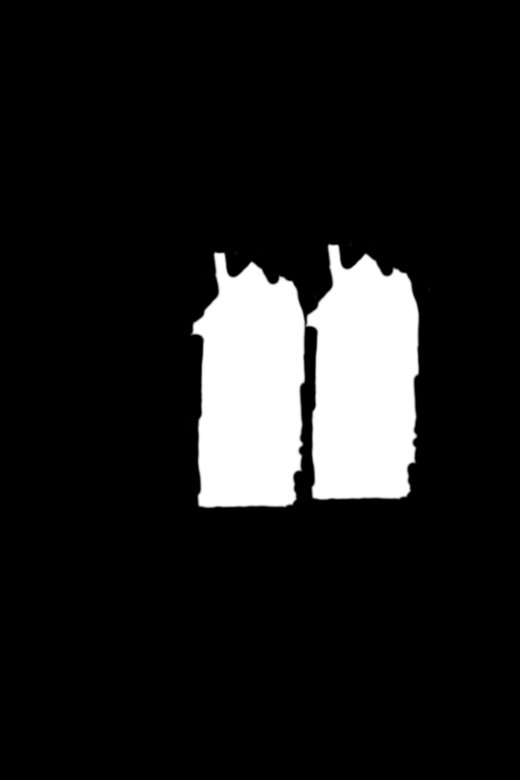}} 

\caption{Database images from the category \emph{structure}, with annotated ground truth for the
``reference forgery'', \ie without rotation or scaling of the copied region.}
\label{fig:db_images3}
\end{figure*}

\figRef{fig:db_images1}, \figRef{fig:db_images2} and \figRef{fig:db_images3}
show a preview of the images and the regions of plain copy-move forgeries,
sorted by the categories \emph{smooth}, \emph{rough} and \emph{structure}.
For every tampering example we show at the top the image containing the
``reference'' tampered regions. At the bottom we
show the associated ground truth (with white being the copy-source or
copy-target regions). Note that several aspects vary over the images, \eg
the size of the copied regions, or the level of detail in the copied
region. Note also, that the assignment of categories is debatable. For
instance, in \figRef{fig:db_images3} the jellyfish image (top row, third
image). Upon close examination, the jellyfish exhibit pronounced edges, although it is not a man-made structure.
As presented, the copied regions are meaningful, \ie
either they hide image content, or they emphasize an element of the picture.
Note, however, that the software allows the snippets to be inserted at
arbitrary positions. Thus, one could equally well create semantically
meaningless forgeries. This is often the case when the copied region is
rotated and resampled. In such cases, the image content becomes (naturally)
implausible.
\clearpage

\renewcommand{\thesubfigure}{\thefigure.\arabic{subfigure}}

\clearpage

\begin{biography}[{\includegraphics[width=1in,height=1.25in,clip,keepaspectratio]{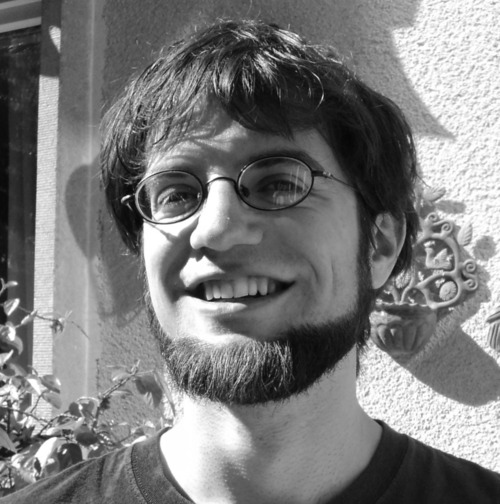}}]{Vincent Christlein}
received his Diploma degree in computer science in 2012 from
the University of Erlangen-Nuremberg, Germany.  During winter 2010, he was a
visiting research student at the Computational Biomedicine Lab, University of
Houston, USA.  Currently, he is a doctoral student at the Pattern Recognition
Lab, University of Erlangen-Nuremberg.  His research interests lie in the field
of computer vision and computer graphics, particularly in image forensics,
reflectance modeling, and historical document analysis.
 \end{biography}
     
\begin{biography}[{\includegraphics[width=1in,height=1.25in,clip,keepaspectratio]{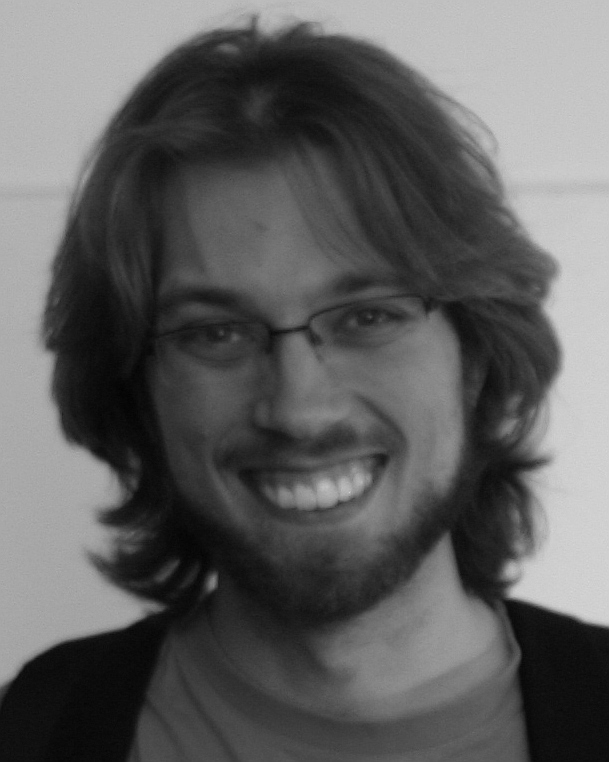}}]{Christian Riess}
received his Diploma degree in computer science in
2007 from the University of Erlangen-Nuremberg, Germany. He was working
on an industry project with Giesecke+Devrient on optical inspection from
2007 to 2010. He is currently pursuing a Ph.D. degree at the Pattern
Recognition Lab, at the University of Erlangen-Nuremberg, Germany. His
research interests
include computer vision and image processing and in particular illumination
and reflectance analysis and image forensics.
     \end{biography}

     \begin{IEEEbiography}[{\includegraphics[width=1in,height=1.25in,clip,keepaspectratio]{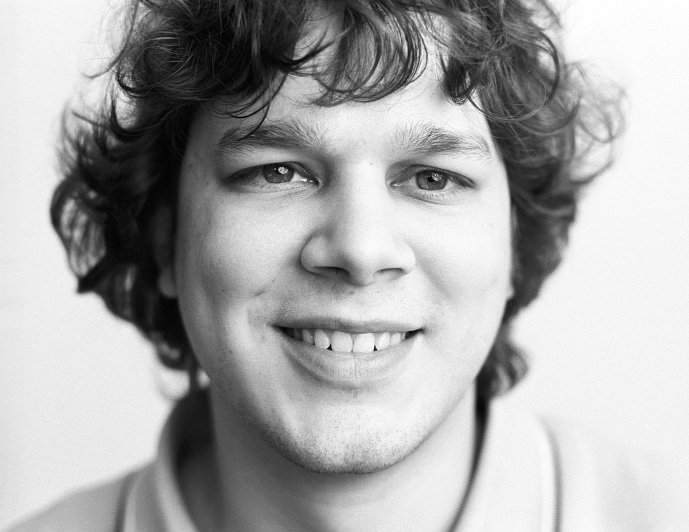}}]{Johannes Jordan}
received his Diploma degree in computer science in 2009 from the University of Erlangen-Nuremberg, Germany. He was a visiting scholar at Stony Brook University, Stony Brook, NY, in 2008. Currently, he is pursuing a Ph.D. degree at the Pattern Recognition Lab, University of Erlangen-Nuremberg, Germany. His research interests focus on computer vision and image processing, particularly in reflectance analysis, multispectral image analysis and image forensics.

     \end{IEEEbiography}

     \begin{biography}[{\includegraphics[width=1in,height=1.25in,clip,keepaspectratio]{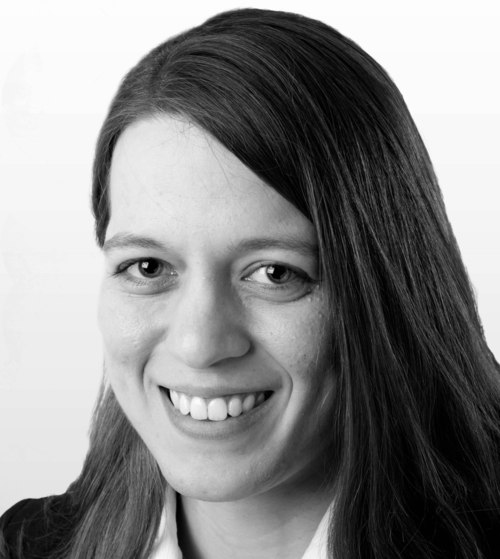}}]{Corinna Riess}
received her Diploma degree in social sciences in 2011 from the
University of Erlangen-Nuremberg, Germany. She is currently working as online
marketing manager and web developer for xeomed, Nuremberg, Germany. Her
further interests include print media layout and image processing and
editing.
     \end{biography}

\begin{biography}[{\includegraphics[width=1in,height=1.25in,clip,keepaspectratio]{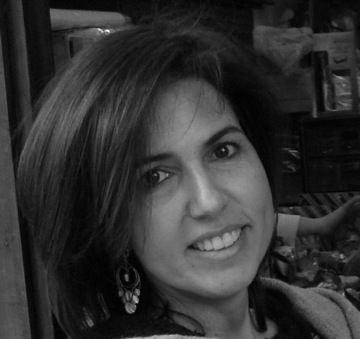}}]{Elli Angelopoulou}
received her Ph.D. in Computer Science from the Johns
Hopkins University in 1997. She did her postdoc at the General
Robotics, Automation, Sensing and Perception (GRASP) Laboratory at the
University of Pennsylvania. She then became an assistant professor at Stevens
Institute of Technology. She is currently an associate research professor at
the University of Erlangen-Nuremberg. Her research focuses on multispectral
imaging, skin reflectance, reflectance analysis in support of shape recovery,
image forensics, image retrieval and reflectance analysis in medical imaging
(e.g. capsule endoscopy). She has over 50 publications, multiple patents and
has received numerous grants, including an NSF CAREER award. Dr. Angelopoulou has
served on the program committees of ICCV, CVPR and ECCV and is an associate
editor of Machine Vision and Applications (MVA) and the Journal of Intelligent
Service Robotics (JISR).
\end{biography}
     
\vfill

\end{document}

%% file: tables/nul_default_new_per_img_individual_draft.tex
Method & Precision & Recall & \FM & $\tau_2$ & Method & Precision & Recall & \FM & $\tau_2$ \\
\hline
\blur	 & 88.89 & 100.00 & 94.12 & 100  &\lin	 & 94.12 & 100.00 & 96.97 & 400 \\
\bravo	 & 87.27 & 100.00 & 93.20 & 200  &\luo	 & 87.27 & 100.00 & 93.20 & 300 \\
\cir	 & 92.31 & 100.00 & 96.00 & 200  &\pca	 & 84.21 & 100.00 & 91.43 & 1000\\
\dct	 & 78.69 & 100.00 & 88.07 & 1000 &\sift	 & 88.37 & 79.17 & 83.52 & 4    \\
\dwt	 & 84.21 & 100.00 & 91.43 & 1000 &\surf	 & 91.49 & 89.58 & 90.53 & 4    \\
\fmt	 & 90.57 & 100.00 & 95.05 & 200  &\svd	 & 68.57 & 100.00 & 81.36 & 50  \\
\hu	     & 67.61 & 100.00 & 80.67 & 50   &\zernike & 92.31 & 100.00 & 96.00 & 800 \\
\kpca    & 87.27 & 100.00 & 93.20 & 1000 &       &       &        &       &      \\
\hline
\hline
Average  &       &        &       &      &       & 85.54 & 97.92 & 90.98 & --\\

%% file: tables/nul_default_new_per_img_individual.tex
Method & Precision & Recall & \FM & $\tau_2$\\
\hline
\blur	 & 88.89 & 100.00 & 94.12 & 100 \\
\bravo	 & 87.27 & 100.00 & 93.20 & 200 \\
\cir	 & 92.31 & 100.00 & 96.00 & 200 \\
\dct	 & 78.69 & 100.00 & 88.07 & 1000\\
\dwt	 & 84.21 & 100.00 & 91.43 & 1000 \\
\fmt	 & 90.57 & 100.00 & 95.05 & 200\\
\hu	 & 67.61 & 100.00 & 80.67 & 50\\
\kpca & 87.27 & 100.00 & 93.20 & 1000\\
\lin	 & 94.12 & 100.00 & 96.97 & 400\\
\luo	 & 87.27 & 100.00 & 93.20 & 300 \\
\pca	 & 84.21 & 100.00 & 91.43 & 1000 \\
\sift	 & 88.37 & 79.17 & 83.52 & 4\\
\surf	 & 91.49 & 89.58 & 90.53 & 4\\
\svd	 & 68.57 & 100.00 & 81.36 & 50 \\
\zernike	 & 92.31 & 100.00 & 96.00 & 800 \\
\hline
Average & 85.54 & 97.92 & 90.98 & --\\

%% file: tables/nul_default_new_per_pixel_individual_draft.tex
Method   & Precision & Recall & \FM & Method & Precision & Recall & \FM  \\
\hline
\blur	 & 98.07 & 78.81 & 86.19 &\lin	 & 99.21 & 78.87 & 86.69 \\
\bravo	 & 98.81 & 82.98 & 89.34 &\luo	 & 97.75 & 82.31 & 88.41 \\
\cir	 & 98.69 & 85.44 & 90.92 &\pca	 & 95.88 & 86.51 & 89.82 \\
\dct	 & 92.90 & 82.85 & 84.95 &\sift	 & 60.80 & 71.48 & 63.10 \\
\dwt	 & 90.55 & 88.78 & 88.86 &\surf	 & 68.13 & 76.43 & 69.54 \\
\fmt	 & 98.29 & 82.33 & 88.79 &\svd	 & 97.53 & 76.53 & 83.71 \\
\hu	     & 97.08 & 74.89 & 82.92 &\zernike & 95.07 & 87.72 & 90.29 \\
\kpca	 & 94.38 & 88.36 & 90.24 &       &       &       &  \\
\hline
\hline
Average &        &       &       &       & 92.21 & 81.62 & 84.92 \\

%% file: tables/nul_default_new_per_pixel_individual.tex
Method & Precision & Recall & \FM \\
\hline
\blur	 & 98.07 & 78.81 & 86.19\\
\bravo	 & 98.81 & 82.98 & 89.34\\
\cir	 & 98.69 & 85.44 & 90.92\\
\dct	 & 92.90 & 82.85 & 84.95\\
\dwt	 & 90.55 & 88.78 & 88.86\\
\fmt	 & 98.29 & 82.33 & 88.79\\
\hu	 & 97.08 & 74.89 & 82.92\\
\kpca	 & 94.38 & 88.36 & 90.24\\
\lin	 & 99.21 & 78.87 & 86.69\\
\luo	 & 97.75 & 82.31 & 88.41\\
\pca	 & 95.88 & 86.51 & 89.82\\
\sift	 & 60.80 & 71.48 & 63.10\\
\surf	 & 68.13 & 76.43 & 69.54\\
\svd	 & 97.53 & 76.53 & 83.71\\
\zernike	 & 95.07 & 87.72 & 90.29\\
\hline
Average & 92.21 & 81.62 & 84.92\\

%% file: tables/multi_paste_default_vs_multi_corres_per_pixel_individual.tex
Method	&Precision	&Recall	&\FM	&Precision	&Recall	&\FM\\
\hline
\blur	&95.24	&52.50	&67.31	&89.91	&54.11	&65.20\\
\bravo	&97.54	&52.58	&68.16	&88.75	&58.27	&67.58\\
\cir    &95.12	&60.90	&73.75	&89.60	&62.48	&71.43\\
\dct	&19.15	&5.37	&8.02	&66.11	&55.76	&55.06\\
\dwt	&52.15	&14.55	&21.21	&81.88	&69.15	&71.84\\
\fmt	&94.42	&54.07	&68.14	&88.85	&60.50	&69.91\\
\hu	    &94.98	&54.08	&68.64	&89.98	&54.61	&65.99\\
\kpca	&37.01	&7.50	&12.05	&87.79	&62.27	&70.06\\
\lin	&96.84	&51.04	&66.61	&90.86	&59.96	&70.63\\
\luo	&95.53	&51.70	&66.72	&89.32	&58.95	&68.47\\
\pca	&37.79	&9.05	&13.95	&88.20	&61.95	&71.77\\
\sift	& 11.37 & 4.95 & 6.74	& 17.00 & 7.34 & 10.07\\
\surf	& 37.49 & 21.86 & 26.15 & 38.31 & 22.93 & 26.79\\
\svd	&91.91	&59.06	&71.51	&71.98	&58.91	&59.33\\
\zernike&83.15	&22.00	&33.52	&87.55	&61.87	&69.64\\
\hline
Average	& 69.31	&34.75	&44.83	&77.31	&53.71	&60.65\\

%% file: tables/resource_requirements.tex
\begin{tabular}{|l|r|r|r|r|r|}
\hline
Method		&Feature 	&Matching 	&Postpr. 	&Total 		& Mem.\\
\hline
\blur		&12059.66	&4635.98	&12.81		&16712.19	&924.06	\\
\bravo		&488.23		&5531.52	&156.27		&6180.42	&154.01	\\
\cir		&92.29		&4987.96	&19.45		&5103.43	&308.02	\\
\dct		&28007.86	&7365.53	&213.06		&35590.03	&9856.67	\\
\dwt		&764.49		&7718.25	&119.66		&8606.50	&9856.67	\\
\fmt		&766.60		&6168.73	&8.07		&6948.03	&1732.62	\\
\hu			&7.04		&4436.63	&5.36		&4452.77	&192.51	\\
\kpca		&6451.34	&7048.83	&88.58		&13592.08	&7392.50	\\
\lin		&12.41		&4732.88	&36.73		&4785.71	&346.52	\\
\luo		&42.90		&4772.67	&119.04		&4937.81	&269.52	\\
\pca		&1526.92	&4322.84	&7.42		&5861.01	&1232.08	\\
\sift		&15.61		&126.15		&469.14		&610.96		&17.18	\\
\surf		&31.07		&725.68		&295.34		&1052.12	&19.92	\\
\svd		&843.52		&4961.11	&7.65		&5816.15	&1232.08	\\
\zernike	&2131.27	&4903.59	&27.23		&7065.18	&462.03	\\
\hline		
Average		&3549.41	& 4829.22	&105.72	&8487.63	&2266.43	\\
\hline
\end{tabular}

%% file: tables/corres_threshold.tex
Method	&Plain Copy	& Copy+JPEG \\
\hline
\blur	&900	&100\\
\bravo	&900	&200\\
\cir	&900	&200\\
\dct	&900	&1000\\
\dwt	&900	&1000\\
\fmt	&900	&200\\
\hu	&1000	&50\\
\kpca	&900	&1000\\
\lin	&900	&400\\
\luo	&800	&300\\
\pca	&1000	&1000\\
\sift	&4	&4\\
\surf	&4	&4\\
\svd	&800	&50\\
\zernike	&1000	&800\\
\hline
Average	&787	&421\\

%% file: tables/nul_default_per_img_individual_vs_ms1000.tex
Method	& \FM (individual)	& \FM ($\tau_2=1000$)\\
\hline
\blur	  &  94.12	&  94.00	\\
\bravo	  &  93.20	&  94.95	\\
\cir	  &  96.00	& 100.00	\\
\dct	  &  88.07	&  88.07	\\
\dwt	  &  91.43	&  91.43	\\
\fmt	  &  95.05	&  96.91	\\
\hu	      &  80.67	&  93.07	\\
\kpca	  &  93.20	&  93.20	\\
\lin	  &  96.97	&  97.92	\\
\luo	  &  93.20	&  94.00	\\
\pca	  &  91.43	&  91.43	\\
\svd	  &  81.36	&  95.83	\\
\zernike  &  96.00	&  96.00\\
\hline
Average	&91.59	&94.37\\

%% file: tables/cmb_easy1_default_per_img_individual_vs_ms1000.tex
Method	&\FM (individual)	&\FM ($\tau_2=1000$)\\
\hline
\blur		& 59.74	& 36.92 \\
\bravo		& 75.00	& 49.28 \\
\cir   		& 51.43	& 42.62 \\
\dct		& 85.98 & 85.98 \\
\dwt		& 88.24	& 88.24 \\
\fmt		& 80.90	& 55.07 \\
\hu			& 60.78 & 41.18 \\
\kpca		& 90.00	& 90.00 \\
\lin		& 74.07 & 36.67 \\
\luo		& 80.43	& 65.82 \\
\pca		& 86.00	& 86.00 \\
\svd		& 64.08	& 33.33 \\
\zernike	& 92.78	& 92.78 \\
\hline
Average	&76.11	&61.84 \\

%% file: tables/nul_default_new_living_nature.tex
         & \multicolumn{2}{c||}{Image level} & \multicolumn{2}{|c|}{Pixel level} \\
\hline
Method	 & Living & Nature & Living & Nature \\
\hline
\blur	 & 100.00 &  97.56 &  64.37 & 65.83 \\
\bravo	 & 100.00 & 100.00 &  61.82 & 66.63 \\
\cir	 & 100.00 & 100.00 &  66.09 & 69.97 \\
\dct	 & 100.00 &  90.91 &  67.27 & 59.47 \\
\dwt	 & 100.00 & 100.00 &  66.77 & 67.51 \\
\fmt	 & 100.00 & 100.00 &  64.85 & 68.35 \\
\hu	     &  86.96 &  78.43 &  61.36 & 63.62 \\
\kpca	 & 100.00 & 100.00 &  68.46 & 68.77 \\
\lin	 & 100.00 & 100.00 &  67.00 & 67.35 \\
\luo	 & 100.00 &  97.56 &  58.46 & 65.98 \\
\pca	 & 100.00 & 100.00 &  71.01 & 67.08 \\
\sift	 &  33.33 &  82.35 &  19.12 & 58.51 \\
\surf	 &  75.00 &  94.74 &  36.32 & 73.28 \\
\svd	 &  83.33 &  80.00 &  66.14 & 60.62 \\
\zernike & 100.00 & 100.00 &  67.60 & 68.59 \\
\hline                                      
Average  &  91.91 &  94.77 &  60.44 & 66.10 \\

%% file: tables/nul_default_new_manmade_mixed.tex
         & \multicolumn{2}{c||}{Image level} & \multicolumn{2}{|c|}{Pixel level} \\
\hline
Method   & man-made & mixed   & man-made & mixed  \\
\hline                                           
\blur	 & 96.00    &  97.30  & 65.64    & 64.40  \\
\bravo	 & 95.05    &  94.74  & 65.00    & 59.59  \\
\cir	 & 97.96    &  94.74  & 71.62    & 68.91  \\
\dct	 & 89.72    &  90.00  & 68.82    & 58.86  \\
\dwt	 & 89.72    &  87.80  & 68.42    & 66.64  \\
\fmt	 & 95.05    &  94.74  & 69.95    & 67.98  \\
\hu	     & 86.49    &  76.60  & 64.89    & 61.50  \\
\kpca	 & 91.43    &  92.31  & 70.56    & 70.37  \\
\lin	 & 97.96    &  97.30  & 69.57    & 66.11  \\
\luo	 & 96.00    &  94.74  & 65.03    & 59.59  \\
\pca	 & 89.72    &  94.74  & 69.88    & 69.35  \\
\sift	 & 88.00    &  88.89  & 71.87    & 69.26  \\
\surf	 & 90.20    &  94.12  & 75.97    & 66.71  \\
\svd	 & 85.71    &  78.26  & 68.71    & 60.49  \\
\zernike & 95.05    & 100.00  & 71.04    & 68.00  \\
\hline                                           
Average  & 92.27    &  91.75  & 69.13    & 65.18  \\

%% file: tables/nul_default_new_smooth_rough_structure.tex
Method	&Smooth	&Rough	&Struct.	&Smooth	&Rough	&Struct.\\
\hline			
\blur	&91.89	&94.12	&96.77	&83.47	&89.52	&85.73\\
\bravo	&91.89	&91.43	&96.77	&87.51	&91.92	&88.67\\
\cir	&97.14	&96.97	&93.75	&88.48	&93.86	&90.54\\
\dct	&89.47	&88.89	&85.71	&77.90	&89.47	&88.12\\
\dwt	&91.89	&91.43	&90.91	&86.94	&91.65	&88.06\\
\fmt	&91.89	&96.97	&96.77	&86.56	&92.13	&87.76\\
\hu		&80.95	&80.00	&81.08	&82.13	&84.07	&82.59\\
\kpca	&91.89	&94.12	&93.75	&87.48	&92.83	&90.59\\
\lin	&91.89	&100.00	&100.00	&83.82	&90.81	&85.56\\
\luo	&94.44	&91.43	&93.75	&86.77	&90.72	&87.79\\
\pca	&91.89	&94.12	&88.24	&86.62	&92.68	&90.40\\
\sift	&73.33	&96.97	&78.57	&48.18	&85.99	&55.61\\
\surf	&87.50	&93.75	&90.32	&60.18	&79.90	&69.11\\
\svd	&79.07	&80.00	&85.71	&75.00	&90.87	&85.96\\
\zernike &97.14	&96.97	&93.75	&89.41	&91.19	&90.32\\
\hline
Average	&89.48	&92.48	&91.06	&80.70	&89.84	&84.45\\

%% file: archive_cmfd.bbl
\begin{thebibliography}{10}
\providecommand{\url}[1]{#1}
\csname url@samestyle\endcsname
\providecommand{\newblock}{\relax}
\providecommand{\bibinfo}[2]{#2}
\providecommand{\BIBentrySTDinterwordspacing}{\spaceskip=0pt\relax}
\providecommand{\BIBentryALTinterwordstretchfactor}{4}
\providecommand{\BIBentryALTinterwordspacing}{\spaceskip=\fontdimen2\font plus
\BIBentryALTinterwordstretchfactor\fontdimen3\font minus
  \fontdimen4\font\relax}
\providecommand{\BIBforeignlanguage}[2]{{%
\expandafter\ifx\csname l@#1\endcsname\relax
\typeout{** WARNING: IEEEtran.bst: No hyphenation pattern has been}%
\typeout{** loaded for the language `#1'. Using the pattern for}%
\typeout{** the default language instead.}%
\else
\language=\csname l@#1\endcsname
\fi
#2}}
\providecommand{\BIBdecl}{\relax}
\BIBdecl

\bibitem{Redi11:DIF}
J.~Redi, W.~Taktak, and J.-L. Dugelay, ``{Digital Image Forensics: A Booklet
  for Beginners},'' \emph{Multimedia Tools and Applications}, vol.~51, no.~1,
  pp. 133--162, Jan. 2011.

\bibitem{Farid09:ASO}
H.~Farid, ``{A Survey of Image Forgery Detection},'' \emph{Signal Processing
  Magazine}, vol.~26, no.~2, pp. 16--25, Mar. 2009.

\bibitem{Amerini11:SFM}
I.~Amerini, L.~Ballan, R.~Caldelli, A.~D. Bimbo, and G.~Serra, ``{A SIFT-based
  Forensic Method for Copy-Move Attack Detection and Transformation
  Recovery},'' \emph{IEEE Transactions on Information Forensics and Security},
  vol.~6, no.~3, pp. 1099--1110, Sep. 2011.

\bibitem{Bashar10:EDR}
M.~Bashar, K.~Noda, N.~Ohnishi, and K.~Mori, ``{Exploring Duplicated Regions in
  Natural Images},'' \emph{IEEE Transactions on Image Processing}, Mar. 2010,
  accepted for publication.

\bibitem{Bayram05:IMD}
S.~Bayram, {\.{I}}.~Avc{\i}ba\c{s}, B.~Sankur, and N.~Memon, ``{Image
  Manipulation Detection with Binary Similarity Measures},'' in \emph{European
  Signal Processing Conference}, Sep. 2005.

\bibitem{Bayram09:AEA}
S.~Bayram, H.~Sencar, and N.~Memon, ``An efficient and robust method for
  detecting copy-move forgery,'' in \emph{IEEE International Conference on
  Acoustics, Speech, and Signal Processing}, Apr. 2009, pp. 1053--1056.

\bibitem{Bravo11:EDR}
S.~Bravo-Solorio and A.~K. Nandi, ``{Exposing Duplicated Regions Affected by
  Reflection, Rotation and Scaling},'' in \emph{International Conference on
  Acoustics, Speech and Signal Processing}, May 2011, pp. 1880--1883.

\bibitem{Christlein10:ORI}
V.~Christlein, C.~Riess, and E.~Angelopoulou, ``{On Rotation Invariance in
  Copy-Move Forgery Detection},'' in \emph{IEEE Workshop on Information
  Forensics and Security}, Dec. 2010.

\bibitem{Dybala07:DFC}
B.~Dybala, B.~Jennings, and D.~Letscher, ``{Detecting Filtered Cloning in
  Digital Images},'' in \emph{Workshop on Multimedia \& Security}, Sep. 2007,
  pp. 43--50.

\bibitem{Fridrich03:DOC}
J.~Fridrich, D.~Soukal, and J.~Luk{\'a}{\v{s}}, ``Detection of copy-move
  forgery in digital images,'' in \emph{Proceedings of Digital Forensic
  Research Workshop}, Aug. 2003.

\bibitem{Huang08:DOC}
H.~Huang, W.~Guo, and Y.~Zhang, ``{Detection of Copy-Move Forgery in Digital
  Images Using SIFT Algorithm},'' in \emph{Pacific-Asia Workshop on
  Computational Intelligence and Industrial Application}, vol.~2, Dec. 2008,
  pp. 272--276.

\bibitem{Ju07:AAM}
S.~Ju, J.~Zhou, and K.~He, ``{An Authentication Method for Copy Areas of
  Images},'' in \emph{International Conference on Image and Graphics}, Aug.
  2007, pp. 303--306.

\bibitem{Kang08:ITR}
X.~Kang and S.~Wei, ``{Identifying Tampered Regions Using Singular Value
  Decomposition in Digital Image Forensics},'' in \emph{International
  Conference on Computer Science and Software Engineering}, vol.~3, 2008, pp.
  926--930.

\bibitem{Ke04:AEP}
Y.~Ke, R.~Sukthankar, and L.~Huston, ``{An Efficient Parts-Based Near-Duplicate
  and Sub-Image Retrieval System},'' in \emph{ACM International Conference on
  Multimedia}, Oct. 2004, pp. 869--876.

\bibitem{Li07:ASN}
G.~Li, Q.~Wu, D.~Tu, and S.~Sun, ``{A Sorted Neighborhood Approach for
  Detecting Duplicated Regions in Image Forgeries Based on DWT and SVD},'' in
  \emph{IEEE International Conference on Multimedia and Expo}, Jul. 2007, pp.
  1750--1753.

\bibitem{Langille06:AEM}
A.~Langille and M.~Gong, ``{An Efficient Match-based Duplication Detection
  Algorithm},'' in \emph{Canadian Conference on Computer and Robot Vision},
  Jun. 2006, pp. 64--71.

\bibitem{Lin01:RSA}
C.-Y. Lin, M.~Wu, J.~Bloom, I.~Cox, M.~Miller, and Y.~Lui, ``{Rotation, Scale,
  and Translation Resilient Watermarking for Images},'' \emph{IEEE Transactions
  on Image Processing}, vol.~10, no.~5, pp. 767--782, May 2001.

\bibitem{Lin09:FCM}
H.~Lin, C.~Wang, and Y.~Kao, ``{Fast Copy-Move Forgery Detection},''
  \emph{WSEAS Transactions on Signal Processing}, vol.~5, no.~5, pp. 188--197,
  2009.

\bibitem{Luo06:RDO}
W.~Luo, J.~Huang, and G.~Qiu, ``{Robust Detection of Region-Duplication Forgery
  in Digital Images},'' in \emph{International Conference on Pattern
  Recognition}, vol.~4, Aug. 2006, pp. 746--749.

\bibitem{Mahdian07:DOC}
B.~Mahdian and S.~Saic, ``{Detection of Copy-Move Forgery using a Method Based
  on Blur Moment Invariants},'' \emph{Forensic Science International}, vol.
  171, no.~2, pp. 180--189, Dec. 2007.

\bibitem{Myrna07:DOR}
A.~N. Myrna, M.~G. Venkateshmurthy, and C.~G. Patil, ``{Detection of Region
  Duplication Forgery in Digital Images Using Wavelets and Log-Polar
  Mapping},'' in \emph{IEEE International Conference on Computational
  Intelligence and Multimedia Applications}, Dec. 2007, pp. 371--377.

\bibitem{Pan10:RDD}
X.~Pan and S.~Lyu, ``{Region Duplication Detection Using Image Feature
  Matching},'' \emph{IEEE Transactions on Information Forensics and Security},
  vol.~5, no.~4, pp. 857--867, Dec. 2010.

\bibitem{Popescu04:EDFDDIR}
\BIBentryALTinterwordspacing
A.~Popescu and H.~Farid, ``Exposing digital forgeries by detecting duplicated
  image regions,'' Department of Computer Science, Dartmouth College, Tech.
  Rep. TR2004-515, 2004. [Online]. Available:
  \url{www.cs.dartmouth.edu/farid/publications/tr04.html}
\BIBentrySTDinterwordspacing

\bibitem{Ryu10:DCR}
S.~Ryu, M.~Lee, and H.~Lee, ``{Detection of Copy-Rotate-Move Forgery using
  Zernike Moments},'' in \emph{Information Hiding Conference}, Jun. 2010, pp.
  51--65.

\bibitem{Shieh06:ASB}
J.-M. Shieh, D.-C. Lou, and M.-C. Chang, ``{A Semi-Blind Digital Watermarking
  Scheme based on Singular Value Decomposition},'' \emph{Computer Standards \&
  Interfaces}, vol.~28, no.~4, pp. 428--440, 2006.

\bibitem{Wang09:FAR}
J.~Wang, G.~Liu, Z.~Zhang, Y.~Dai, and Z.~Wang, ``{Fast and Robust Forensics
  for Image Region-Duplication Forgery},'' \emph{Acta Automatica Sinica},
  vol.~35, no.~12, pp. 1488–--1495, Dec. 2009.

\bibitem{Wang09:DOI}
J.~Wang, G.~Liu, H.~Li, Y.~Dai, and Z.~Wang, ``{Detection of Image Region
  Duplication Forgery Using Model with Circle Block},'' in \emph{International
  Conference on Multimedia Information Networking and Security}, Jun. 2009, pp.
  25--29.

\bibitem{Zhang08:ANA}
J.~Zhang, Z.~Feng, and Y.~Su, ``{A New Approach for Detecting Copy-Move Forgery
  in Digital Images},'' in \emph{International Conference on Communication
  Systems}, Nov. 2008, pp. 362--366.

\bibitem{Bayram09:ASO}
S.~Bayram, H.~T. Sencar, and N.~Memon, ``A survey of copy-move forgery
  detection techniques,'' in \emph{IEEE Western New York Image Processing
  Workshop}, 2009.

\bibitem{Beis97:SIU}
J.~S. Beis and D.~G. Lowe, ``{Shape Indexing Using Approximate
  Nearest-Neighbour Search in High-Dimensional Spaces},'' in \emph{IEEE
  Conference on Computer Vision and Pattern Recognition}, Jun. 1997, pp.
  1000--1006.

\bibitem{Christlein10:SFD}
V.~Christlein, C.~Riess, and E.~Angelopoulou, ``{A Study on Features for the
  Detection of Copy-Move Forgeries},'' in \emph{{GI SICHERHEIT}}, Oct. 2010.

\bibitem{Bashar07:WBM}
M.~K. Bashar, K.~Noda, N.~Ohnishi, H.~Kudo, T.~Matsumoto, and Y.~Takeuchi,
  ``Wavelet-based multiresolution features for detecting duplications in
  images,'' in \emph{Conference on Machine Vision Application}, May 2007, pp.
  264--267.

\bibitem{Muja09:FAN}
M.~Muja and D.~G. Lowe, ``{Fast Approximate Nearest Neighbors with Automatic
  Algorithm Configuration},'' in \emph{International Conference on Computer
  Vision Theory and Applications}, Feb. 2009, pp. 331--340.

\bibitem{Hartley03:MVG}
R.~Hartley and A.~Zisserman, \emph{{Multiple View Geometry}}.\hskip 1em plus
  0.5em minus 0.4em\relax Cambridge University Press, 2003.

\bibitem{SuppMat}
V.~Christlein, C.~Riess, J.~Jordan, C.~Riess, and E.~Angelopoulou,
  ``{Supplemental Material to ``An Evaluation of Popular Copy-Move Forgery
  Detection Approaches},'' \url{http://www.arxiv.org/abs/1208.3665}, Aug. 2012.

\bibitem{Shivakumar11:DOR}
B.~L. Shivakumar and S.~Baboo, ``{Detection of Region Duplication Forgery in
  Digital Images Using SURF},'' \emph{International Journal of Computer Science
  Issues}, vol.~8, no.~4, pp. 199--205, 2011.

\bibitem{Bo10:ICP}
X.~Bo, W.~Junwen, L.~Guangjie, and D.~Yuewei, ``{Image Copy-Move Forgery
  Detection Based on SURF},'' in \emph{Multimedia Information Networking and
  Security}, Nov. 2010, pp. 889--892.

\bibitem{Ng04:ADS}
T.~Ng and S.~Chang, ``{A Data Set of Authentic and Spliced Image Blocks},''
  Columbia University, Tech. Rep. 20320043, Jun. 2004.

\bibitem{CASIA}
\emph{{CASIA Image Tampering Detection Evaluation Database}}, National
  Laboratory of Pattern Recognition, Institute of Automation, Chinese Academy
  of Science, \url{http://forensics.idealtest.org}.

\bibitem{Battiato10:DFE}
S.~Battiato and G.~Messina, ``{Digital Forgery Estimation into DCT Domain - A
  Critical Analysis},'' in \emph{ACM Multimedia Workshop Multimedia in
  Forensics}, Oct. 2009, pp. 37--42.

\bibitem{Gloe10:DID}
T.~Gloe and R.~B{\"o}hme, ``{The `Dresden Image Database' for benchmarking
  digital image forensics},'' in \emph{25th Symposium On Applied Computing},
  vol.~2, Mar. 2010, pp. 1585--1591.

\bibitem{Goljan09:LST}
M.~Goljan, J.~Fridrich, and T.~Filler, ``{Large Scale Test of Sensor
  Fingerprint Camera Identification},'' in \emph{SPIE Media Forensics and
  Security}, vol. 7254, Jan. 2009.

\end{thebibliography}
